University of Liège
Faculty of Applied Sciences
Department of Electrical Engineering & Computer Science

PhD dissertation

# UNDERSTANDING RANDOM FORESTS

## FROM THEORY TO PRACTICE

by Gilles Louppe

Advisor: Prof. Pierre Geurts
July 2014

# JURY MEMBERS

Louis Wehenkel, Professor at the Université de Liège (President);

Pierre Geurts, Professor at the Université de Liège (Advisor);

Bernard Boigelot, Professor at the Université de Liège;

Renaud Detry, Postdoctoral Researcher at the Université de Liège;

Gianluca Bontempi, Professor at the Université Libre de Bruxelles;

Gérard Biau, Professor at the Université Pierre et Marie Curie (France);

# ABSTRACT

Data analysis and machine learning have become an integrative part of the modern scientific methodology, offering automated procedures for the prediction of a phenomenon based on past observations, unraveling underlying patterns in data and providing insights about the problem. Yet, caution should avoid using machine learning as a black-box tool, but rather consider it as a methodology, with a rational thought process that is entirely dependent on the problem under study. In particular, the use of algorithms should ideally require a reasonable understanding of their mechanisms, properties and limitations, in order to better apprehend and interpret their results.

Accordingly, the goal of this thesis is to provide an in-depth analysis of random forests, consistently calling into question each and every part of the algorithm, in order to shed new light on its learning capabilities, inner workings and interpretability. The first part of this work studies the induction of decision trees and the construction of ensembles of randomized trees, motivating their design and purpose whenever possible. Our contributions follow with an original complexity analysis of random forests, showing their good computational performance and scalability, along with an in-depth discussion of their implementation details, as contributed within Scikit-Learn.

In the second part of this work, we analyze and discuss the interpretability of random forests in the eyes of variable importance measures. The core of our contributions rests in the theoretical characterization of the Mean Decrease of Impurity variable importance measure, from which we prove and derive some of its properties in the case of multiway totally randomized trees and in asymptotic conditions. In consequence of this work, our analysis demonstrates that variable importances as computed from non-totally randomized trees (e.g., standard Random Forest) suffer from a combination of defects, due to masking effects, misestimations of node impurity or due to the binary structure of decision trees.

Finally, the last part of this dissertation addresses limitations of random forests in the context of large datasets. Through extensive experiments, we show that subsampling both samples and features simultaneously provides on par performance while lowering at the same time the memory requirements. Overall this paradigm highlights an intriguing practical fact: there is often no need to build single models over immensely large datasets. Good performance can often be achieved by building models on (very) small random parts of the data and then combining them all in an ensemble, thereby avoiding all practical burdens of making large data fit into memory.

# ACKNOWLEDGMENTS

As the saying goes, good premises do not entail good stories. Yet, this dissertation would certainly not have come to its successful conclusion without the help, support and trust of colleagues, friends and family.

First and foremost, I would like to sincerely thank my advisor Pierre Geurts for his help, guidance and for the freedom I was granted throughout these years.

I am grateful to all members of the jury for their interest in this work and for taking the time to evaluate this dissertation.

In alphabetical order, I would also like to thank my colleagues who all contributed to create and maintain a pleasant and stimulating working environment: Antonio, Arnaud, Benjamin, Damien, Fabien, Julien, Loïc, Louis, Marie, Olivier, Raphaël, Van Anh, Vincent. Special thanks go to Antonio, Arnaud and Vincent who accepted to proofread parts of this manuscript.

I want take this opportunity to thank the Scikit-Learn team and all its contributors. This experience within the open source world really contributed to shape my vision of science and software development towards a model of rigor, pragmatism and openness. Thanks go to Gaël, Olivier, Lars, Mathieu, Andreas, Alexandre and Peter.

Special thanks go to the rowing team of the RCAE, for their friendship and good mood in all circumstances. Guys, I thank you all.

Even if I never succeeded to fully explain my research topics, I would finally like to warmly thank my dear friend Jérôme and my family for their help in moments of doubt.

Last but not least, Laura, I am forever grateful for your unconditional support and love.

# CONTENTS

# 1 INTRODUCTION

In various fields of science, technology and humanities, as in biology, meteorology, medicine or finance to cite a few, experts aim at predicting a phenomenon based on past observations or measurements. For instance, meteorologists try to forecast the weather for the next days from the climatic conditions of the previous days. In medicine, practitioners collect measurements and information such as blood pressure, age or history for diagnosing the condition of incoming patients. Similarly, in chemistry, compounds are analyzed using mass spectrometry measurements in order to determine whether they contain a given type of molecules or atoms. In all of these cases, the goal is the prediction of a response variable based on a set of observed predictor variables.

For centuries, scientists have addressed such problems by deriving theoretical frameworks from first principles or have accumulated knowledge in order to model, analyze and understand the phenomenon under study. For example, practitioners know from past experience that elderly heart attack patients with low blood pressure are generally high risk. Similarly, meteorologists know from elementary climate models that one hot, high pollution day is likely to be followed by another. For an increasing number of problems however, standard approaches start showing their limits. For example, identifying the genetic risk factors for heart disease, where knowledge is still very sparse, is nearly impractical for the cognitive abilities of humans given the high complexity and intricacy of interactions that exist between genes. Likewise, for very fine-grained meteorological forecasts, a large number of variables need to be taken into account, which quickly goes beyond the capabilities of experts to put them all into a system of equations. To break this cognitive barrier and further advance science, machines of increasing speed and capacity have been built and designed since the mid-twentieth century to assist humans in their calculations. Amazingly however, alongside this progress in terms of hardware, developments in theoretical computer science, artificial intelligence and statistics have made machines to become more than calculators. Recent advances have made them experts of their own kind, capable to learn from data and to uncover by themselves the predictive structure of problems. Techniques and algorithms that have stemmed from the field of *machine learning* have indeed now become a powerful tool for the analysis of complex and large data, successfully assisting scientists in numerous breakthroughs of various fields of science and technology. Public and famous examples include the use of boosted decision trees in the statistical analysis that led to the detection of the Higgs boson at CERN [Chatrchyan et al., 2012], the use of random forests for human pose detection in the Microsoft Kinect [Criminisi and Shotton, 2013] or the implementation of various machine learning techniques for building the IBM Watson system [Ferrucci et al., 2010], capable to compete at the human champion level on the American TV quiz show Jeopardy.

Formally, machine learning can be defined as the study of systems that can learn from data without being explicitly programmed. According to Mitchell [1997], a computer program is said to learn from data, with respect to some class of tasks and performance measure if its performance at those tasks improves with data. In particular, machine learning provides algorithms that are able to solve classification or regression tasks, hence bringing now automated procedures for the prediction of a phenomenon based on past observations. However, the goal of machine learning is not only to produce algorithms making accurate predictions, it is also to provide insights on the predictive structure of the data [Breiman et al., 1984]. If we are aiming at the latter, then our goal is to understand what variables or interactions of variables drive the phenomenon. For practitioners, which are not experts in machine learning, interpretability is indeed often as important as prediction accuracy. It allows for a better understanding of the phenomenon under study, a finer exploration of the data and an easier self-appropriation of the results. By contrast, when an algorithm is used as a black box, yielding results seemingly out of nowhere, it may indeed be difficult to trust or accept if it cannot be understood how and why the procedure came to them. Unfortunately, the current state-of-the-art in machine learning often makes it difficult for non-experts to understand and interpret the results of an algorithm. While considerable efforts have been put to improve their prediction accuracy, it is still not clearly understood what makes machine learning algorithms truly work, and under what assumptions. Likewise, few of them actually provide clear and insightful explanations about the results they generate.

In this context, the goal of this thesis is to provide a comprehensive and self-contained analysis of a class of algorithms known as decision trees [Breiman et al., 1984] and random forests [Breiman, 2001]. While these methods have proven to be a robust, accurate and successful tool for solving countless of machine learning tasks, including classification, regression, density estimation, manifold learning or semi-supervised learning [Criminisi and Shotton, 2013], there remain many gray areas in their understanding:

(a) First, the theoretical properties and statistical mechanisms that drive the algorithm are still not clearly and entirely understood. Random forests indeed evolved from empirical successes rather than from a sound theory. As such, various parts of the algo-rithm remain heuristic rather than theoretically motivated. For example, preliminary results have proven the consistency of simplified to very close variants of random forests, but consistency of the original algorithm remains unproven in a general setting.

(b) Second, while the construction process of a single decision tree can easily be described within half a page, implementing this algorithm properly and efficiently remains a challenging task involving issues that are easily overlooked. Unfortunately, implementation details are often omitted in the scientific literature and can often only be found by diving into (unequally documented) existing software implementations. As far as we know, there is indeed no comprehensive survey covering the implementation details of random forests, nor with their respective effects in terms of runtime and space complexity or learning ability.

(c) Third, interpreting the resulting model remains a difficult task, for which even machine learning experts still fail at finely analyzing and uncovering the precise predictive structure learned by the procedure. In particular, despite their extensive use in a wide range of applications, little is still known regarding variable importance measures computed by random forests. Empirical evidence suggests that they are appropriate for identifying relevant variables, but their statistical mechanisms and properties are still far from being understood.

All throughout this dissertation, our objective is therefore to call into question each and every part of the random forests methodology, both from a theoretical and practical point of view. Accordingly, this work aims at revisiting decision trees and random forests to hopefully shed new light on their learning capabilities, inner workings and interpretability.

## 1.1 OUTLINE AND CONTRIBUTIONS

Part I of this manuscript is first dedicated to a thorough treatment of decision trees and forests of randomized trees. We begin in Chapter 2 by outlining fundamental concepts of machine learning, and then proceed in Chapters 3 and 4 with a comprehensive review of the algorithms at the core of decision trees and random forests. We discuss the learning capabilities of these models and carefully study all parts of the algorithm and their complementary effects. In particular, Chapter 4 includes original contributions on the bias-variance analysis of ensemble methods, highlighting how randomization can help improve performance. Chapter 5 concludes this first part with an original space and time complexity analysis of random forests (and their variants), along with an in-depth discussion of implementation details, as contributed within the open source Scikit-Learn library. Overall, Part I therefore presents a comprehensive review of previous work on random forests, including some original contributions both from a theoretical and practical point of view.

Part II analyzes and discusses the interpretability of random forests. In Chapter 6, we study variable importances as computed with a forest of randomized trees and study how these scores can be interpreted in order to reveal the underlying predictive structure learned from the data. In particular, we derive a theoretical framework from which we prove theoretical and practical properties of variable importances. In Chapter 7, we then exploit this framework to further study variable importances as derived from actual random forests and present successful applications of variable importance measures. Part II constitutes the main contributions of this dissertation.

Finally, Part III addresses limitations of random forests in the context of large datasets. Through extensive experiments, we show in Chapter 8 that subsampling strategies provides on par performance while simultaneously lowering the memory requirements. This chapter presents original work.

## 1.2 PUBLICATIONS

This dissertation summarizes several contributions to random forests algorithms. Publications that have directly stemmed from this work include:

- [Geurts and Louppe, 2011] *Learning to rank with extremely randomized trees*, Geurts Pierre and Louppe Gilles. In JMLR: Workshop and Conference Proceedings, volume 14, 2011.

- [Louppe and Geurts, 2012] *Ensembles on random patches*, Louppe Gilles and Geurts Pierre. In Machine Learning and Knowledge Discovery in Databases, pages 346–361. Springer, 2012.

- [Louppe et al., 2013] *Understanding variable importances in forests of randomized trees*, Louppe Gilles, Wehenkel Louis, Sutera Antonio and Geurts Pierre. In Advances in Neural Information Processing Systems, pages 431–439, 2013.

- [Buitinck et al., 2013] *API design for machine learning software: experiences from the scikit-learn project*, Buitinck Lars, Louppe Gilles, Blondel Mathieu et al.. In ECML-PKDD 2013 Workshop: Languages for Data Mining and Machine Learning, 2013.

- [Botta et al., 2014] *Exploiting SNP Correlations within Random Forest for Genome-Wide Association Studies*, Botta Vincent, Louppe Gilles, Geurts Pierre and Wehenkel Louis. PloS one, 9(4):e93379, 2014.

During the course of this thesis, several fruitful collaborations have also led to the following publications. These are not discussed within this dissertation.

- [Louppe and Geurts, 2010] *A zealous parallel gradient descent algorithm*, Louppe Gilles and Geurts Pierre. In Learning on Cores, Clusters and Clouds workshop, NIPS, 2010

- [Marée et al., 2014] *A hybrid human-computer approach for large-scale image-based measurements using web services and machine learning*, Marée Raphaël, Rollus Loic, Stevens Benjamin et al. Proceedings IEEE International Symposium on Biomedical Imaging, 2014.

- [McGovern et al., 2014] *Solar Energy Prediction: An International Contest to Initiate Interdisciplinary Research on Compelling Meteorological Problems*, Amy McGovern, David John Gagne II, Lucas Eustaquio et al., 2014. *Submitted.*

- [Sutera et al., 2014] *Simple connectome inference from partial correlation statistics in calcium imaging*, Antonio Sutera, Arnaud Joly, Vincent Francois-Lavet et al., 2014. *Submitted.*

# Part I

# GROWING DECISION TREES

# 2 BACKGROUND

**OUTLINE**

In this chapter, we introduce supervised learning and fundamental concepts on which this work builds upon. In Section 2.1, we first describe classification and regression learning tasks and then formally define the notion of model. In Section 2.2, we proceed with a discussion on performance evaluation and then describe, in Section 2.3, procedures for selecting the best possible model. Finally, we conclude in Section 2.4 with a brief overview of some of the main methods in supervised learning.

## 2.1 LEARNING FROM DATA

In the examples introduced in Chapter 1, the objective which is sought is to find a systematic way of predicting a phenomenon given a set of measurements. In machine learning terms, this goal is formulated as the *supervised learning* task of inferring from collected data a model that predicts the value of an output variable based on the observed values of input variables. As such, finding an appropriate model is based on the assumption that the output variable does not take its value at random and that there exists a relation between the inputs and the output. In medicine for instance, the goal is to find a decision rule (i.e., a model) from a set of past cases (i.e., the collected data) for predicting the condition of an incoming patient (i.e., the output value) given a set of measurements such as age, sex, blood pressure or history (i.e., the input values).

To give a more precise formulation, let us assume as set of *cases* or *objects* taken from a universe $\Omega$. Let us arrange the set of measurements on a case in a pre-assigned order, i.e., take the input values to be $x_1, x_2, ..., x_p$, where $x_j \in \mathcal{X}_j$ (for $j = 1, ..., p$) corresponds to the value of the input variable $X_j$. Together, the input values $(x_1, x_2, ..., x_p)$ form a $p$-dimensional input vector $\mathbf{x}$ taking its values in $\mathcal{X}_1 \times ... \times \mathcal{X}_p = \mathcal{X}$, where $\mathcal{X}$ is defined as the input space. Similarly, let us define as $y \in \mathcal{Y}$ the value of the output variable $Y$, where $\mathcal{Y}$ is defined as the output space[1]. By definition, both the input and the output spaces are assumed to respectively contain all possible input vectors and all possible output values. Note that input variables are some-

1 Unless stated otherwise, supervised learning is reduced to the prediction of a *single* output variable. More generally however, this framework can be defined as the prediction of one or several output variables.

times known as *features*, input vectors as *instances* or *samples* and the output variable as *target*.

Among variables that define the problem, we distinguish between two general types. The former correspond to quantitative variables whose values are integer or real numbers, such as age or blood pressure. The latter correspond to qualitative variables whose values are symbolic, such as gender or condition. Formally, we define them as follows:

**Definition 2.1.** *A variable $X_j$ is* ordered *if $\mathcal{X}_j$ is a totally ordered set. In particular, $X_j$ is said to be* numerical *if $\mathcal{X}_j = \mathbb{R}$.*

**Definition 2.2.** *A variable $X_j$ is* categorical *if $\mathcal{X}_j$ is a finite set of values, without any natural order.*

In a typical supervised learning task, past observations are summarized by a dataset called *learning set*. It consists in a set of observed input vectors together with their actual output value and formally defined as follows:

**Definition 2.3.** *A* learning set *$\mathcal{L}$ is a set of $N$ pairs of input vectors and output values $(\mathbf{x}_1, y_1), ..., (\mathbf{x}_N, y_N)$, where $\mathbf{x}_i \in \mathcal{X}$ and $y_i \in \mathcal{Y}$.*

Equivalently, a set of $p$-input vectors $\mathbf{x}_i$ (for $i = 1, ..., N$) can be denoted by a $N \times p$ matrix $\mathbf{X}$, whose rows $i = 1, ..., N$ correspond to input vectors $\mathbf{x}_i$ and columns $j = 1, ..., p$ to input variables $X_j$. Similarly, the corresponding output values can be written as a vector $\mathbf{y} = (y_1, ..., y_N)$.

### Data representation

For optimal implementations of machine learning algorithms, data needs to be represented using structures which allow for high-performance numerical computation. In this work, code snippets are described under the assumption that data is represented using a data structure similar to a NumPy array [Van der Walt et al., 2011].

A NumPy array is basically a multidimensional uniform collection of values, all of the same type and organized in a given shape. For instance, a matrix $\mathbf{X}$ can be represented as a 2-dimensional NumPy array of shape $N \times p$ that contains numbers (e.g., floating point values or integers). This structure allows for random access in constant time, vectorized high-level operations and efficient memory usage.

Additionally, using a data representation which is close to the matrix formulation, like NumPy arrays, makes it possible to write implementations that are close to their original textbook formulation, thereby making them easier to code, understand and maintain.

In this framework, the supervised learning task can be stated as learning a function $\varphi : \mathcal{X} \mapsto \mathcal{Y}$ from a learning set $\mathcal{L} = (\mathbf{X}, \mathbf{y})$. The objective is to find a model such that its predictions $\varphi(\mathbf{x})$, also denoted by the variable $\hat{Y}$, are as good as possible. If $Y$ is a categorical variable then the learning task is a classification problem. If $Y$ is numerical variable, then learning task is a regression problem. Without loss of generality, the resulting models can be defined as follows:

**Definition 2.4.** *A* classifier *or* classification rule *is a function* $\varphi : \mathcal{X} \mapsto \mathcal{Y}$, *where* $\mathcal{Y}$ *is a finite set of classes (or labels) denoted* $\{c_1, c_2, ..., c_J\}$.

**Definition 2.5.** *A* regressor *is a function* $\varphi : \mathcal{X} \mapsto \mathcal{Y}$, *where* $\mathcal{Y} = \mathbb{R}$.

### Estimator interface

We follow in this work the API conventions proposed by Buitinck et al. [2013]. Learning algorithms are described as *estimator* objects implementing the following interface:

- Hyper-parameters of an algorithm are all passed to the constructor of the estimator. The constructor does not see any actual data. All it does is to attach hyper-parameters values as public attributes to the estimator object.
- Learning is performed in a `fit` method. This method is called with a learning set (e.g., supplied as two arrays `X_train` and `y_train`). Its task is to run a learning algorithm and to determine model-specific parameters from the data. The `fit` method always returns the estimator object it was called on, which now serves as a model and can be used to make predictions.
- Predictions are performed through a `predict` method, taking as input an array `X_test` and producing as output the predictions for `X_test` based on the learned parameters. In the case of classification, this method returns labels from $\mathcal{Y} = \{c_1, c_2, ..., c_J\}$. In the case of regression, it returns numerical values from $\mathcal{Y} = \mathbb{R}$.

With this API, a typical supervised learning task is performed as follows:

```
# Instantiate and set hyper-parameters
clf = DecisionTreeClassifier(max_depth=5)
# Learn a model from data
clf.fit(X_train, y_train)
# Make predictions on new data
y_pred = clf.predict(X_test)
```

## 2.2 Performance evaluation

In the statistical sense, input and output variables $X_1, ..., X_p$ and $Y$ are *random variables* taking jointly their values from $\mathcal{X} \times \mathcal{Y}$ with respect to the joint probability distribution $P(X, Y)$, where $X$ denotes the random vector $(X_1, ..., X_p)$. That is, $P(X = \mathbf{x}, Y = y)$ is the probability that random variables X and Y take values $\mathbf{x}$ and $y$ from $\mathcal{X}$ and $\mathcal{Y}$ when drawing an object uniformly at random from the universe $\Omega$.

Accordingly, using an algorithm $\mathcal{A}$ for learning a model[2] $\varphi_{\mathcal{L}}$ whose predictions are as good as possible can be stated as finding a model which minimizes its expected prediction error, defined as follows:

**Definition 2.6.** *The* expected prediction error, *also known as* generalization error *or* test error, *of the model* $\varphi_{\mathcal{L}}$ *is*[3]

$$\mathrm{Err}(\varphi_{\mathcal{L}}) = \mathbb{E}_{X,Y}\{L(Y, \varphi_{\mathcal{L}}(X))\}, \tag{2.1}$$

*where* $\mathcal{L}$ *is the learning set used to build* $\varphi_{\mathcal{L}}$ *and* L *is a loss function measuring the discrepancy between its two arguments [Geurts, 2002].*

Equation 2.1 basically measures the prediction error of $\varphi_{\mathcal{L}}$ over all possible objects in $\Omega$ (each represented by a couple $(\mathbf{x}, y) \in \mathcal{X} \times \mathcal{Y}$), including the observed couples from the learning set $\mathcal{L}$ but also all the *unseen* ones from $\mathcal{X} \times \mathcal{Y} \setminus \mathcal{L}$. Indeed, the goal is not in fact to make the very-most accurate predictions over the subset $\mathcal{L}$ of known data, but rather to learn a model which is correct and reliable on all possible data.

For classification, the most common loss function is the *zero-one* loss function[4] $L(Y, \varphi_{\mathcal{L}}(X)) = 1(Y \neq \varphi_{\mathcal{L}}(X))$, where all misclassifications are equally penalized. In this case, the generalization error of $\varphi_{\mathcal{L}}$ becomes the probability of misclassification of the model:

$$\mathrm{Err}(\varphi_{\mathcal{L}}) = \mathbb{E}_{X,Y}\{1(Y \neq \varphi_{\mathcal{L}}(X))\} = P(Y \neq \varphi_{\mathcal{L}}(X)) \tag{2.2}$$

Similarly, for regression, the most used loss function is the *squared error* loss $L(Y, \varphi_{\mathcal{L}}(X)) = (Y - \varphi_{\mathcal{L}}(X))^2$, where large differences between the true values and the predicted values are penalized more heavily than small ones. With this loss, the generalization error of the model becomes:

$$\mathrm{Err}(\varphi_{\mathcal{L}}) = \mathbb{E}_{X,Y}\{(Y - \varphi_{\mathcal{L}}(X))^2\} \tag{2.3}$$

---

2 Unless it is clear from the context, models are now denoted $\varphi_{\mathcal{L}}$ to emphasize that they are built from the learning set $\mathcal{L}$.

3 $\mathbb{E}_X\{f(X)\}$ denotes the expected value of $f(x)$ (for $x \in \mathcal{X}$) with respect to the probability distribution of the random variable X. It is defined as

$$\mathbb{E}_X\{f(X)\} = \sum_{x \in \mathcal{X}} P(X = x) f(x).$$

4 $1(\textit{condition})$ denotes the unit function. It is defined as

$$1(\textit{condition}) = \begin{cases} 1 & \text{if } \textit{condition} \text{ is true} \\ 0 & \text{if } \textit{condition} \text{ is false} \end{cases}.$$

### 2.2.1 *Estimating* $\mathrm{Err}(\varphi_{\mathcal{L}})$

In practice, the probability distribution $P(X, Y)$ is usually unknown, making the direct evaluation of $\mathrm{Err}(\varphi_{\mathcal{L}})$ infeasible. Equivalently, it is often not possible to draw additional data, thereby making infeasible the empirical estimation of $\mathrm{Err}(\varphi_{\mathcal{L}})$ on a (virtually infinite) set $\mathcal{L}'$ drawn independently from $\mathcal{L}$. In most problems, $\mathcal{L}$ constitutes the only data available, on which both the model needs to be learned and its generalization error estimated. As reviewed and described on multiple occasions by several authors [Toussaint, 1974; Stone, 1978; Breiman et al., 1984; Kohavi et al., 1995; Nadeau and Bengio, 2003; Hastie et al., 2005; Arlot and Celisse, 2010], the generalization error in Equation 2.1 can however be estimated in several ways.

To make notations clearer, let us first define $\overline{E}(\varphi_{\mathcal{L}}, \mathcal{L}')$ as the average prediction error of the model $\varphi_{\mathcal{L}}$ over the set $\mathcal{L}'$ (possibly different from the learning set $\mathcal{L}$ used to produce $\varphi_{\mathcal{L}}$), that is:

$$\overline{E}(\varphi_{\mathcal{L}}, \mathcal{L}') = \frac{1}{N'} \sum_{(\mathbf{x}_i, y_i) \in \mathcal{L}'} L(y_i, \varphi_{\mathcal{L}}(\mathbf{x}_i)) \tag{2.4}$$

where $N'$ is the size of the set $\mathcal{L}'$.

The first and simplest estimate of the generalization error is the *resubstitution estimate* or *training sample estimate*. It consists in empirically estimating $\mathrm{Err}(\varphi_{\mathcal{L}})$ on the same data as the learning set $\mathcal{L}$ used to build $\varphi_{\mathcal{L}}$, that is:

$$\widehat{\mathrm{Err}}^{\text{train}}(\varphi_{\mathcal{L}}) = \overline{E}(\varphi_{\mathcal{L}}, \mathcal{L}) \tag{2.5}$$

In general, the resubstitution error is a poor estimate of $\mathrm{Err}(\varphi_{\mathcal{L}})$. In particular, since most machine learning algorithms aim at precisely minimizing Equation 2.5 (either directly or indirectly), it typically results in an overly optimistic estimate of the generalization error, which accounts for all couples $(\mathbf{x}, y)$, i.e., not only those from $\mathcal{L}$.

The second approach is the *test sample estimate*. It consists in dividing the learning set $\mathcal{L}$ into two disjoint sets $\mathcal{L}_{\text{train}}$ and $\mathcal{L}_{\text{test}}$, called *training set* and *test set*, and then to use each part respectively for learning a model and estimating its generalization error. The test sample estimate of the generalization error of the model $\varphi_{\mathcal{L}}$ that would be obtained from $\mathcal{L}$ is then given as the average prediction error over $\mathcal{L}_{\text{test}}$ of the model $\varphi_{\mathcal{L}_{\text{train}}}$ built on $\mathcal{L}_{\text{train}}$:

$$\widehat{\mathrm{Err}}^{\text{test}}(\varphi_{\mathcal{L}}) = \overline{E}(\varphi_{\mathcal{L}_{\text{train}}}, \mathcal{L}_{\text{test}}) \tag{2.6}$$

As a rule-of-thumb, $\mathcal{L}_{\text{train}}$ is usually taken as 70% of the samples in $\mathcal{L}$ and $\mathcal{L}_{\text{test}}$ as the remaining 30%, though theoretical work [Guyon, 1997] suggests to progressively reduce the size of test set as the size of $\mathcal{L}$ increases. In any case, care must be taken when splitting $\mathcal{L}$ into two subsets, so that samples from $\mathcal{L}_{\text{train}}$ can be considered independent from those in $\mathcal{L}_{test}$ and drawn from the same distribution. This is however usually guaranteed by drawing $\mathcal{L}_{train}$ and $\mathcal{L}_{test}$ simply at random from $\mathcal{L}$. While being an unbiased estimate of $\text{Err}(\varphi_\mathcal{L})$, the test sample estimate has the drawback that it reduces the effective sample size on which the model $\varphi_{\mathcal{L}_{train}}$ is learned. If $\mathcal{L}$ is large, then this is usually not an issue, but if $\mathcal{L}$ only contains a few dozens of samples, then this strategy might not correctly approximate the true generalization error of the model that would have been learned on the entire learning set.

When $\mathcal{L}$ is small, the $K$-*fold cross-validation estimate* is usually preferred over the test sample estimate. It consists in randomly dividing the learning set $\mathcal{L}$ into $K$ disjoint subsets, $\mathcal{L}_1, ..., \mathcal{L}_K$, and then to estimate the generalization error as the average prediction error over the folds $\mathcal{L}_k$ of the models $\varphi_{\mathcal{L}\setminus\mathcal{L}_k}$ learned on the remaining data:

$$\widehat{\text{Err}}^{CV}(\varphi_\mathcal{L}) = \frac{1}{K}\sum_{k=1}^{K} \bar{E}(\varphi_{\mathcal{L}\setminus\mathcal{L}_k}, \mathcal{L}_k) \tag{2.7}$$

The assumption behind this approach is that since each model $\varphi_{\mathcal{L}\setminus\mathcal{L}_k}$ is built using almost all $\mathcal{L}$, they should all be close to the model $\varphi_\mathcal{L}$ learned on the entire set. As a result the unbiased estimates $\bar{E}(\varphi_{\mathcal{L}\setminus\mathcal{L}_k}, \mathcal{L}_k)$ should also all be close to $\text{Err}(\varphi_\mathcal{L})$. While more computationally intensive, the K-fold cross-validation estimate has the advantage that every couple $(\mathbf{x}, y) \in \mathcal{L}$ is used for estimating the generalization error of $\varphi_\mathcal{L}$. In a typical setting, $K$ is usually fixed to $10$, a value often yielding stable and reliable estimates [Kohavi et al., 1995].

**Expected generalization error**

As we have shown, our goal is to estimate the generalization error $\text{Err}(\varphi_\mathcal{L})$ conditional on the learning set $\mathcal{L}$. A related quantity is the *expected* generalization error

$$\mathbb{E}_\mathcal{L}\{\text{Err}(\varphi_\mathcal{L})\}, \tag{2.8}$$

averaging over everything which is random, including the randomness in the learning set $\mathcal{L}$ used to produce $\varphi_\mathcal{L}$. As discussed by Hastie et al. [2005], this quantity is close, yet different from $\text{Err}(\varphi_\mathcal{L})$. The authors point out that most estimates, including K-fold cross-validation, effectively estimate Equation 2.8 rather than Equation 2.1.

### 2.2.2 *Bayes model and residual error*

In theory, when the probability distribution $P(X, Y)$ is known, the best possible model, i.e., the model $\varphi_B$ which minimizes the generalization error of Equation 2.1, can be derived analytically and indepen-dently of any learning set $\mathcal{L}$. By conditioning on X, the generalization error of this model can be rewritten as:

$$\mathbb{E}_{X,Y}\{L(Y, \varphi_B(X))\} = \mathbb{E}_X\{\mathbb{E}_{Y|X}\{L(Y, \varphi_B(X))\}\} \tag{2.9}$$

In this latter form, the model which minimizes Equation 2.9 is a model which minimizes the inner expectation point-wise, that is:

$$\varphi_B(\mathbf{x}) = \arg\min_{y \in \mathcal{Y}} \mathbb{E}_{Y|X=\mathbf{x}}\{L(Y, y)\} \tag{2.10}$$

In the literature, $\varphi_B$ is known as the *Bayes model* and its generalization error $\text{Err}(\varphi_B)$ as the *residual error*. It represents the minimal error that any supervised learning algorithm can possibly attain, that is the irreducible error purely due to random deviations in the data.

**Definition 2.7.** *A model $\varphi_B$ is a* Bayes model *if, for any model $\varphi$ built from any learning set $\mathcal{L}$, $\text{Err}(\varphi_B) \leqslant \text{Err}(\varphi_\mathcal{L})$.*

In classification, when L is the zero-one loss, the Bayes model is:

$$\begin{aligned}\varphi_B(\mathbf{x}) &= \arg\min_{y \in \mathcal{Y}} \mathbb{E}_{Y|X=\mathbf{x}}\{1(Y, y)\} \\ &= \arg\min_{y \in \mathcal{Y}} P(Y \neq y | X = \mathbf{x}) \\ &= \arg\max_{y \in \mathcal{Y}} P(Y = y | X = \mathbf{x})\end{aligned} \tag{2.11}$$

Put otherwise, the best possible classifier consists in systematically predicting the most likely class $y \in \{c_1, c_2, ..., c_J\}$ given $X = \mathbf{x}$.

Similarly, for regression with the squared error loss, we have:

$$\begin{aligned}\varphi_B(\mathbf{x}) &= \arg\min_{y \in \mathcal{Y}} \mathbb{E}_{Y|X=\mathbf{x}}\{(Y - y)^2\} \\ &= \mathbb{E}_{Y|X=\mathbf{x}}\{Y\}\end{aligned} \tag{2.12}$$

In other words, the best possible regressor consists in systematically predicting the average value of Y at $X = \mathbf{x}$.

For practical problems, $P(X, Y)$ is unknown and the Bayes model cannot be derived analytically. In this context, the effectiveness of a model $\varphi_\mathcal{L}$ may be difficult to evaluate since (estimates of) $\text{Err}(\varphi_\mathcal{L})$ may not be very indicative of the goodness of $\varphi_\mathcal{L}$ if the lowest attainable error $\text{Err}(\varphi_B)$ is unknown. On simulated data however, where the distribution is known, deriving $\varphi_B$ and $\text{Err}(\varphi_B)$ is feasible. This is beneficial as $\text{Err}(\varphi_B)$ can now be used for comparing the test set error of $\varphi_\mathcal{L}$ and thereby evaluate the actual effectiveness of $\varphi_\mathcal{L}$ with respect to the best model one can possibly build.

From a theoretical point of view, the concepts of Bayes model and residual error are also useful to study the learning capabilities of an algorithm. In particular, as $\mathcal{L}$ gets arbitrarily large, a fundamental question is to know whether it is possible to produce a model $\varphi_\mathcal{L}$ which is *consistent*, that is such that its generalization error gets arbitrarily close to the lowest possible generalization error $\text{Err}(\varphi_B)$. Formally, consistency is defined as follows [Devroye et al., 1996]:

**Definition 2.8.** *A learning algorithm $\mathcal{A}$ is said to be* weakly consistent *for a certain distribution $P(X, Y)$ if $\mathbb{E}_{\mathcal{L}}\{\text{Err}(\varphi_{\mathcal{L}})\} \to \text{Err}(\varphi_B)$ as the size $N$ of the learning set $\mathcal{L}$ used to build $\varphi_{\mathcal{L}}$ using $\mathcal{A}$ tends to infinity.*

**Definition 2.9.** *A learning algorithm $\mathcal{A}$ is said to be* strongly consistent *for a certain distribution $P(X, Y)$ if $\text{Err}(\varphi_{\mathcal{L}}) \to \text{Err}(\varphi_B)$ almost surely as the size $N$ of the learning set $\mathcal{L}$ used to build $\varphi_{\mathcal{L}}$ using $\mathcal{A}$ tends to infinity.*

Note that the definition of consistency depends on the distribution $P(X, Y)$. In general, a learning algorithm $\mathcal{A}$ can be proven to be consistent for some classes of distributions, but not for others. If consistency can be proven for any distribution $P(X, Y)$, then $\mathcal{A}$ is said to be *universally* (strongly) consistent.

## 2.3 MODEL SELECTION

From the previous discussion in Section 2.2.2, it appears that to solve the supervised learning problem it would be sufficient to estimate $P(Y|X)$ from the learning sample $\mathcal{L}$ and then to define a model accordingly using either Equation 2.11 or Equation 2.12. Unfortunately, this approach is infeasible in practice because it requires $\mathcal{L}$ to grow exponentially with the number $p$ of input variables in order to compute accurate estimates of $P(Y|X)$ [Geurts, 2002].

### 2.3.1 *Selecting the (approximately) best model*

To make supervised learning work in high-dimensional input spaces with learning sets of moderate sizes, simplifying assumptions must be made on the structure of the best model $\varphi_B$. More specifically, a supervised learning algorithm assumes that $\varphi_B$ – or at least a good enough approximation – lives in a family $\mathcal{H}$ of candidate models, also known as *hypotheses* in statistical learning theory, of restricted structure. In this setting, the *model selection* problem is then defined as finding the best model among $\mathcal{H}$ on the basis of the learning set $\mathcal{L}$.

**Approximation error**

Depending on restrictions made on the structure of the problem, the Bayes model usually does not belong to $\mathcal{H}$, but there may be models $\varphi \in \mathcal{H}$ that are sufficiently close to it. As such, the *approximation error* [Bottou and Bousquet, 2011] measures how closely the models in $\mathcal{H}$ can approximate the optimal model $\varphi_B$:

$$\text{Err}(\mathcal{H}) = \min_{\varphi \in \mathcal{H}} \{\text{Err}(\varphi)\} - \text{Err}(\varphi_B) \tag{2.13}$$

To be more specific, let $\theta$ be the vector of hyper-parameters values controlling the execution of a learning algorithm $\mathcal{A}$. The application of $\mathcal{A}$ with hyper-parameters $\theta$ on the learning set $\mathcal{L}$ is a deterministic[5] process yielding a model $\mathcal{A}(\theta, \mathcal{L}) = \varphi_{\mathcal{L}} \in \mathcal{H}$. As such, our goal is to find the vector of hyper-parameters values yielding to the best model possibly learnable in $\mathcal{H}$ from $\mathcal{L}$:

$$\theta^* = \arg\min_{\theta} \text{Err}(\mathcal{A}(\theta, \mathcal{L})) \tag{2.14}$$

Again, this problem cannot (usually) be solved exactly in practice since it requires the true generalization error of a model to be computable. However approximations $\widehat{\theta}^*$ can be obtained in several ways.

When $\mathcal{L}$ is large, the easiest way to find $\widehat{\theta}^*$ is to use test sample estimates (as defined by Equation 2.6) to guide the search of the hyper-parameter values, that is:

$$\widehat{\theta}^* = \arg\min_{\theta} \widehat{\text{Err}}^{\text{test}}(\mathcal{A}(\theta, \mathcal{L})) \tag{2.15}$$

$$= \arg\min_{\theta} \overline{E}(\mathcal{A}(\theta, \mathcal{L}_{\text{train}}), \mathcal{L}_{\text{test}}) \tag{2.16}$$

In practice, solving this later equation is also a difficult task but approximations can be obtained in several ways, e.g., using either manual tuning of $\theta$, exhaustive exploration of the parameter space using grid search, or dedicated optimization procedures (e.g., using random search [Bergstra and Bengio, 2012]). Similarly, when $\mathcal{L}$ is scarce, the same procedure can be carried out in the exact same way but using K-fold cross-validation estimates (as defined by Equation 2.7) instead of test sample estimates. In any case, once $\widehat{\theta}^*$ is identified, the learning algorithm is run once again on the entire learning set $\mathcal{L}$, finally yielding the approximately optimal model $\mathcal{A}(\widehat{\theta}^*, \mathcal{L})$.

When optimizing $\theta$, special care must be taken so that the resulting model is neither too simple nor too complex. In the former case, the model is indeed said to *underfit* the data, i.e., to be not flexible enough the capture the structure between X and Y. In the later case, the model is said to *overfit* the data, i.e., to be too flexible and to capture isolated structures (i.e., noise) that are specific to the learning set.

As Figure 2.1 illustrates, this phenomenon can be observed by examining the respective training and test estimates of the model with respect to its complexity[6]. When the model is too simple, both the training and test estimates are large because of underfitting. As complexity increases, the model gets more accurate and both the training and test estimates decrease. However, when the model becomes too complex, specific elements from the training set get captured, reducing the corresponding training estimates down to 0, as if the model

5 If $\mathcal{A}$ makes use of a pseudo-random generator to mimic a stochastic process, then we assume that the corresponding random seed is part of $\theta$.

6 Unless mentioned otherwise, *complexity* here refers to the complexity of the model in function space. *Computational complexity* is studied later in Section 5.1 were perfect. At the same time, the test estimates become worse because the structure learned from the training set is actually too specific and does not generalize. The model is overfitting. The best hyper-parameter value $\theta$ is therefore the one making the appropriate trade-off and producing a model which is neither too simple nor to complex, as shown by the gray line on the figure.

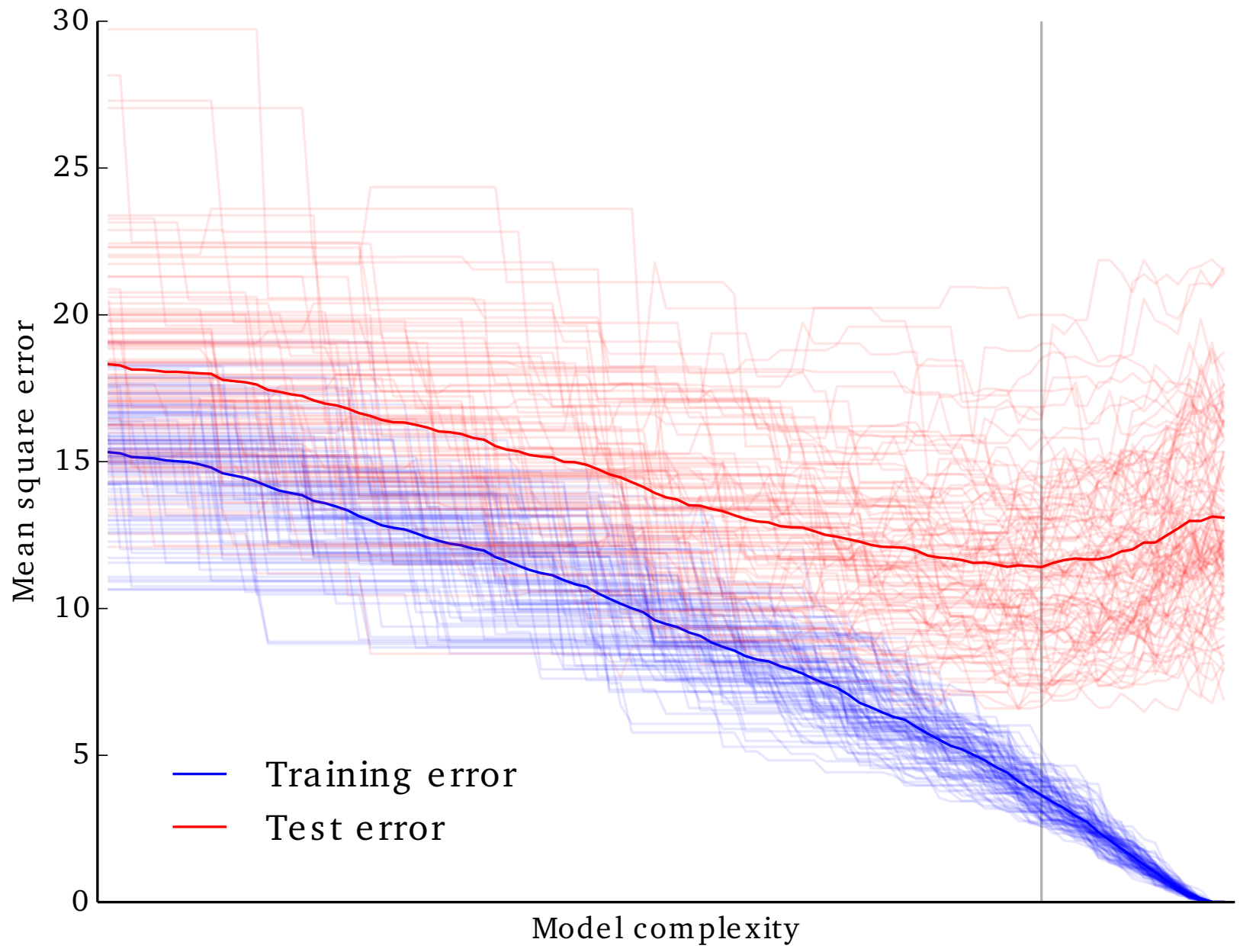

Figure 2.1: Training and test error with respect to the complexity of a model. The light blue curves show the training error over $\mathcal{L}_{\text{train}}$ while the light red curves show the test error estimated over $\mathcal{L}_{\text{test}}$ for 100 pairs of training and test sets $\mathcal{L}_{\text{train}}$ and $\mathcal{L}_{\text{test}}$ drawn at random from a known distribution. The thick blue curve is the average training error while the thick red curve is the average test error. (Figure inspired from [Hastie et al., 2005].)

As we will see later in Chapter 4, overfitting can also be explained by decomposing the generalization error in terms of bias and variance. A model which is too simple usually has high bias but low variance, while a model which is too complex usually has low bias but high variance. In those terms, finding the best model amounts to make the appropriate *bias-variance trade-off*.

### 2.3.2 *Selecting and evaluating simultaneously*

On real applications, and in particular when reporting results, it usually happens that one wants to do both model selection and model assessment. That is, having chosen a final model $\mathcal{A}(\widehat{\theta}^*, \mathcal{L})$, one wants to also estimate its generalization error.

A naive assessment of the generalization error of the selected model might be to simply use the test sample estimate $\widehat{\text{Err}}^{\text{test}}(\mathcal{A}(\widehat{\theta}^*, \mathcal{L}))$ (i.e., $\overline{E}(\mathcal{A}(\widehat{\theta}^*, \mathcal{L}_{\text{train}}), \mathcal{L}_{\text{test}})$) that was minimized during model selection. The issue with this estimate is that the learned model is not independent from $\mathcal{L}_{\text{test}}$ since its repeated construction was precisely guided by the minimization of the prediction error over $\mathcal{L}_{\text{test}}$. As a result, the minimized test sample error is in fact a biased, optimistic, estimate of the true generalization error, sometimes leading to substantial underestimations. For the same reasons, using the K-fold cross-validation estimate does not provide a better estimate since model selection was similarly guided by the minimization of this quantity.

To guarantee an unbiased estimate, the test set on which the generalization error is evaluated should ideally be kept out of the entire model selection procedure and only be used once the final model is selected. Algorithm 2.1 details such a protocol. Similarly, per-fold estimates of the generalization error should be kept out of the model selection procedure, for example using nested cross-validation within each fold, as explicited in Algorithm 2.2.

**Algorithm 2.1.** *Train-Valid-Test set protocol for both model selection and evaluation.*

(a) *Divide the learning set $\mathcal{L}$ into three parts $\mathcal{L}_{train}$, $\mathcal{L}_{valid}$ and $\mathcal{L}_{test}$;*

(b) *Perform model selection on $\mathcal{L}_{train} \cup \mathcal{L}_{valid}$ using test sample estimates, i.e., find:*

$$\widehat{\theta}^* = \arg\min_{\theta} \widehat{\text{Err}}^{test}(\mathcal{A}(\theta, \mathcal{L}_{train} \cup \mathcal{L}_{valid})) \quad (2.17)$$

$$= \arg\min_{\theta} \overline{E}(\mathcal{A}(\theta, \mathcal{L}_{train}), \mathcal{L}_{valid}); \quad (2.18)$$

(c) *Evaluate the (unbiased) generalization error of the final model as*

$$\overline{E}(\mathcal{A}(\widehat{\theta}^*, \mathcal{L}_{train} \cup \mathcal{L}_{valid}), \mathcal{L}_{test}); \quad (2.19)$$

(d) *Learn the final model $\mathcal{A}(\widehat{\theta}^*, \mathcal{L})$ on the entire learning set.*

**Algorithm 2.2.** *Nested K-fold cross-validation protocol for both model selection and evaluation.*

(a) *Divide the learning set $\mathcal{L}$ into K folds $\mathcal{L}_1, ..., \mathcal{L}_K$;*

(b) *For each fold $k = 1, ..., K$:*

  i. *Divide $\mathcal{L} \setminus \mathcal{L}_k = \mathcal{L}^{-k}$ into K folds $\mathcal{L}_1^{-k}, ..., \mathcal{L}_K^{-k}$;*

  ii. *Perform model selection on the subset $\mathcal{L}^{-k}$ using nested K-fold estimates, i.e., find:*

$$\widehat{\theta}_k^* = \arg\min_{\theta} \widehat{\text{Err}}^{CV}(\mathcal{A}(\theta, \mathcal{L}^{-k})) \quad (2.20)$$

$$= \arg\min_{\theta} \frac{1}{K} \sum_{l=1}^{K} \overline{E}(\mathcal{A}(\theta, \mathcal{L}^{-k} \setminus \mathcal{L}_l^{-k}), \mathcal{L}_l^{-k}); \quad (2.21)$$

iii. *Evaluate the generalization error of the selected sub-model as*

$$\overline{E}(\mathcal{A}(\widehat{\theta}^*_k, \mathcal{L} \setminus \mathcal{L}_k), \mathcal{L}_k); \tag{2.22}$$

(c) *Evaluate the (unbiased) generalization error of the selected model as the average generalization estimate of the sub-models selected over the folds:*

$$\frac{1}{K}\sum_{k=1}^{K} \overline{E}(\mathcal{A}(\widehat{\theta}^*_k, \mathcal{L} \setminus \mathcal{L}_k), \mathcal{L}_k); \tag{2.23}$$

(d) *Perform model selection on the entire learning set* $\mathcal{L}$ *using* K*-fold cross-validation estimates, i.e., find:*

$$\widehat{\theta}^* = \arg\min_{\theta} \widehat{Err}^{CV}(\mathcal{A}(\theta, \mathcal{L})) \tag{2.24}$$

$$= \arg\min_{\theta} \frac{1}{K}\sum_{k=1}^{K} \overline{E}(\mathcal{A}(\theta, \mathcal{L} \setminus \mathcal{L}_k), \mathcal{L}_k); \tag{2.25}$$

(e) *Learn the final model* $\mathcal{A}(\widehat{\theta}^*, \mathcal{L})$ *on the entire learning set.*

## 2.4 CLASSES OF LEARNING ALGORITHMS

Before proceeding in the next chapters, and for the rest of this work, to an in-depth analysis of the class of tree-based methods, we briefly review in this section some of the other learning algorithms that have matured in the field, including linear methods, support vector machines, neural networks and nearest neighbor methods.

### 2.4.1 *Linear methods*

One of the oldest class of supervised learning algorithms is the class of *linear methods* from the field of statistics. In these methods, the central assumption made on $\mathcal{H}$ is that the output variable $Y$ can be described as a linear combination of the input variables $X_1, ..., X_p$ (i.e., as an hyperplane), or at least that the linear model is a reasonable approximation. For regression, $\mathcal{H}$ includes all models $\varphi$ of the form:

$$\varphi(\mathbf{x}) = b + \sum_{j=1}^{p} x_j w_j \tag{2.26}$$

For binary classification (i.e., when $\mathcal{Y} = \{c_1, c_2\}$), $\mathcal{H}$ includes all models $\varphi$ of the form:

$$\varphi(\mathbf{x}) = \begin{cases} c_1 & \text{if } b + \sum_{j=1}^{p} x_j w_j > 0 \\ c_2 & \text{otherwise} \end{cases} \tag{2.27}$$

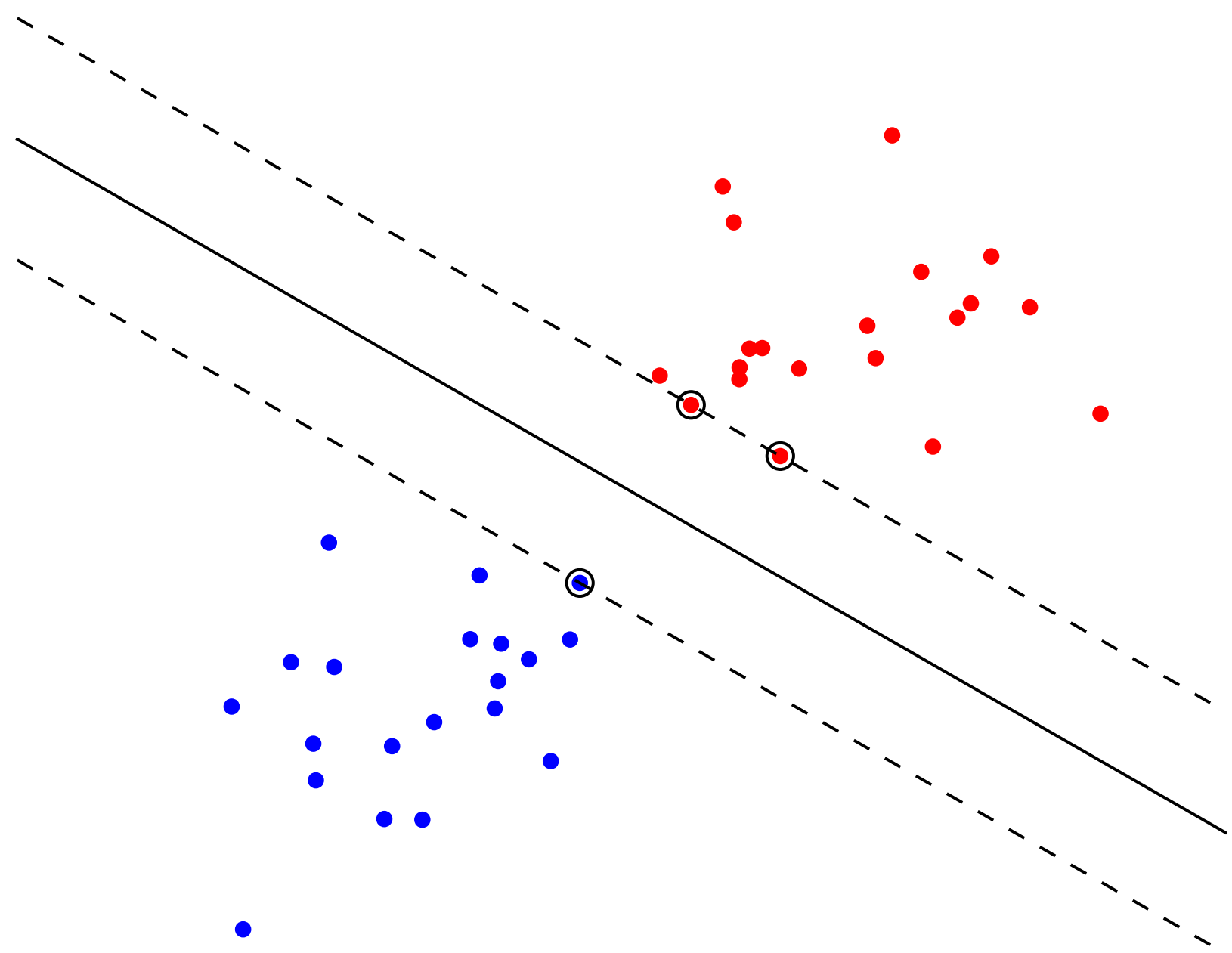

Figure 2.2: A good separating hyperplane is an hyperplane that maximizes the distance to the nearest training data points.

Linear methods come in many flavors but mostly differ from each other in the way they estimate the coefficients $b$ and $w_j$ (for $j = 1, \dots, p$), usually using some specific optimization procedure to minimize a specific criterion. Among all of them, the most famous is the method of *least squares*. It consists in finding the $b$ and $w_j$ coefficients that minimize the resubstitution estimate (Equation 2.5) using the squared error loss.

Despite an apparent simplicity, linear methods often provide reliable predictions and an interpretable description of how the input variables affect the output. Contrary to their name, linear methods can also be used to model non-linear relations between $X$ and $Y$, for example by applying these methods on (non-linear) transformations of the input variables. For more details on (generalized) linear methods, see the reviews of MacCullagh and Nelder [1989], Hastie et al. [2005], Bishop and Nasrabadi [2006] or Duda et al. [2012].

### 2.4.2 *Support vector machines*

When data points from the learning set are linearly separable, there exist several hyperplanes (i.e., several linear models) that are in fact equally as good when evaluated in resubstitution. In generalization however, these hyperplanes are usually not equivalent. As illustrated in Figure 2.2, a good separation is intuitively achieved when the distance (also known as the *margin*) to the nearest training data points is as large as possible, since in general the larger the margin the lower the generalization error of the model.

Mathematically, support vector machines [Boser et al., 1992; Cortes and Vapnik, 1995] are maximum-margin linear models of the form of Equation 2.27. Assuming without loss of generality that $\mathcal{Y} = \{-1, 1\}$ and that $b = 0$, support vector machines are learned by solving the following primal optimization problem:

$$\min_{\mathbf{w},\xi} \left\{ \frac{1}{2}\|\mathbf{w}\|^2 + C\sum_{i=1}^{N} \xi_i \right\} \tag{2.28}$$

subject to

$$y_i(\mathbf{w} \cdot \mathbf{x_i}) \geqslant 1 - \xi_i, \quad \xi_i \geqslant 0. \tag{2.29}$$

In its dual the form, the optimization problem is

$$\max_{\alpha} \left\{ \sum_{i=1}^{N} \alpha_i - \frac{1}{2} \sum_{i,j} \alpha_i \alpha_j y_i y_j \mathbf{x}_i \cdot \mathbf{x}_j \right\} \tag{2.30}$$

subject to

$$0 \leqslant \alpha_i \leqslant C, \tag{2.31}$$

where C is an hyper-parameter that controls the degree of misclassification of the model, in case classes are not linearly separable. From the solution of dual problem, we have

$$\mathbf{w} = \sum_{i=1}^{N} \alpha_i y_i \mathbf{x}_i, \tag{2.32}$$

from which the final linear model can finally be expressed.

Support vector machines extend to non-linear classification by projecting the original input space into a high-dimensional space (the so-called *kernel trick*), where a separating hyperplane can hopefully be found. Interestingly, the dual optimization problem is exactly the same, except that the dot product $\mathbf{x}_i \cdot \mathbf{x}_j$ is replaced by a *kernel* $K(\mathbf{x}_i, \mathbf{x}_j)$, which corresponds the dot product of $\mathbf{x}_i$ and $\mathbf{x}_j$ in the new space.

### 2.4.3 *Neural networks*

The family of *neural networks* methods finds its origins in attempts to identify mathematical representations of information processing in biological systems. While this objective is still far from being reached, (artificial) neural networks have nonetheless proven all their worth from a statistical point of view, and have actually grown into one of the most effective methods in machine learning.

A neural network is usually made of several units, also known as *neurons*, of the form

$$h_j(\mathbf{x}) = \sigma(w_j + \sum_{i=1}^{n} w_{ij} x_i), \tag{2.33}$$

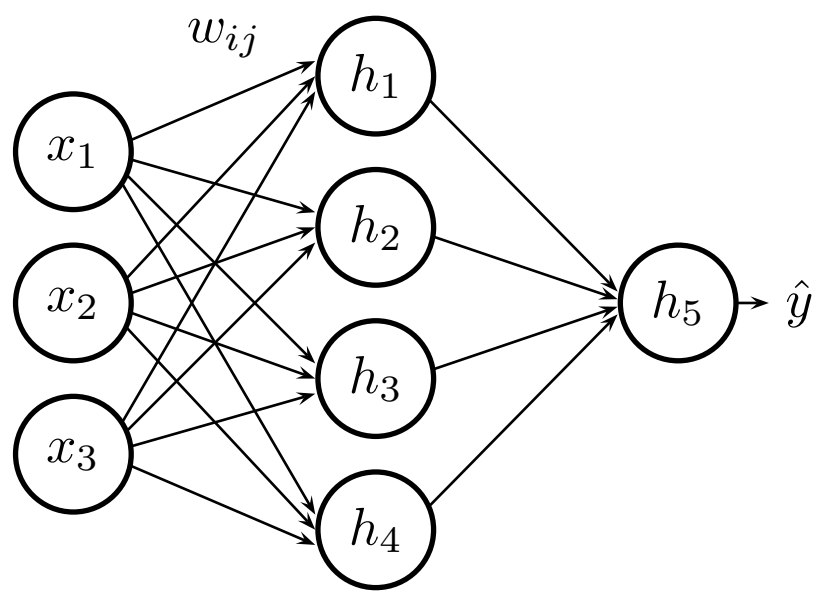

Figure 2.3: An artificial neural network.

where $\sigma$ is a non-linear activation function, such as the sign function, the sigmoid function or the softmax activation. In most cases, these units are structured into successive layers, where the outputs of a layer are directed through weighted connections, also known as *synapses*, to the inputs of the next layer. As an example, Figure 2.3 illustrates a three layered neural network. The first layer is the input layer, which transmits the input values $\mathbf{x} = (x_1, ..., x_p)$ to the second layer. The second layer is made of activation units $h_j$, taking as inputs the weighted values of the input layer and producing non-linear transformations as outputs. The third layer is made of a single activation unit, taking as inputs the weighted outputs of the second layer and producing the predicted value $\hat{y}$. Assuming that this structure is fixed and that all units from the network make use of the same activation function $\sigma$, the hypothesis space $\mathcal{H}$ therefore includes all models $\varphi$ of the form

$$\varphi(\mathbf{x}) = \sigma(w_5 + \sum_{j=1}^{4} w_{j5}\sigma(w_j + \sum_{i=1}^{p} w_{ij}x_i)). \tag{2.34}$$

As for linear methods, learning a neural network amounts to estimate the weights $w_{ij}$ that minimize some specific loss function, using some specific optimization procedure. Among all of them, the most famous is the *backpropagation algorithm* [Bryson, 1975].

Recent advances on neural networks, now themed as *deep learning*, have recently shown the capability of these models to autonomously learn high-level and very effective representations of data. On various difficult tasks, such as image classification or speech recognition, neural networks have indeed achieved outstanding performance, outperforming both human operators and state-of-the-art methods. From a theoretical point of view however, little is still known about what makes them truly work. In particular, picking the appropriate combination of number of units, layers and types of activation functions remains a delicate task, which unfortunately makes (deep) neural networks still difficult to use for non-experts. Reviews on recent developments include the works of Hinton [2007], Arel et al. [2010] or Bengio et al. [2013].

### 2.4.4 *Nearest neighbor methods*

*Nearest neighbor methods* belong to a class of non-parametric algorithms known as *prototype methods* [Hastie et al., 2005]. They distinguish from other learning algorithms in the sense that they are memory-based and require no model to be fit.

The principle behind nearest neighbor methods is to find a number of training samples closest in distance to the new sample, and then infer from these the value of the output variable. For regression, the k-nearest neighbor algorithm [Fix and Hodges, 1951] averages the output values from the k closest training samples, that is:

$$\varphi(\mathbf{x}) = \frac{1}{k} \sum_{(\mathbf{x}_i, y_i) \in NN(\mathbf{x}, \mathcal{L}, k)} y_i, \tag{2.35}$$

where $NN(\mathbf{x}, \mathcal{L}, k)$ denotes the k nearest neighbors of $\mathbf{x}$ in $\mathcal{L}$. For classification, the procedure is the same except that the predicted output value is computed as the majority class among the k nearest neighbors:

$$\varphi(\mathbf{x}) = \arg\max_{c \in \mathcal{Y}} \sum_{(\mathbf{x}_i, y_i) \in NN(\mathbf{x}, \mathcal{L}, k)} 1(y_i = c). \tag{2.36}$$

In general, the distance function used to identify the k nearest neighbors can be any metric, but the standard Euclidean distance is the most common choice. A notable variant is the *radius-based neighbor algorithm*, in which the predicted output value is computed from the training samples within a radius r of the new sample. In cases where data is not uniformly sampled, this latter algorithm can be a better choice than the k-nearest neighbor method.

Despite their simplicity, nearest neighbor methods usually yield decent results in practice. They are are often successful in classification situations where the decision boundary is very irregular. From a theoretical point of view, an important result due to Cover and Hart [1967] is the proof of consistency of the method. When the size of $\mathcal{L}$ tends to infinity and k is appropriately adjusted to $\mathcal{L}$, the generalization error of the model produced by the k-nearest neighbor method is shown to converge towards the generalization error of the Bayes model.

# 3 CLASSIFICATION AND REGRESSION TREES

**Outline**

In this chapter, we present a unified framework in which we detail (single) decision trees methods. In Section 3.1, we first give an overview of the context in which these algorithms have been developed. In Section 3.2, we proceed with a mathematical presentation, introducing all necessary concepts and notations. The general learning algorithm is then presented in Section 3.3 while specific parts of the algorithm are discussed in finer details in Sections 3.4, 3.5, 3.6. As such, specific decision tree algorithms (e.g., CART, ID3 or C4.5) are described as specializations of the general framework presented here.

## 3.1 INTRODUCTION

Since always, artificial intelligence has been driven by the ambition to understand and uncover complex relations in data. That is, to find models that can not only produce accurate predictions, but also be used to extract knowledge in an intelligible way. Guided with this twofold objective, research in machine learning has given rise to extensive bodies of works in a myriad of directions. Among all of them however, tree-based methods stand as one of the most effective and useful method, capable to produce both reliable and understandable results, on mostly any kind of data.

Historically, the appearance of *decision trees* is due to Morgan and Sonquist [1963], who first proposed a tree-based method called *automatic interaction detector* (AID) for handling multi-variate non-additive effects in the context of survey data. Building upon AID, methodological improvements and computer programs for exploratory analysis were then proposed in the following years by several authors [Sonquist, 1970; Messenger and Mandell, 1972; Gillo, 1972; Sonquist et al., 1974]. Without contest however, the principal investigators that have driven research on the modern methodological principles are Breiman [1978a,b], Friedman [1977, 1979] and Quinlan [1979, 1986] who simultaneously and independently proposed very close algorithms for the induction of tree-based models. Most notably, the unifying work of Breiman et al. [1984], later complemented with the work of Quinlan [1993], have set decision trees into a simple and consistent methodological framework, which largely contributed in making them easy to understand and easy to use by a large audience.

As we will explore in further details all throughout this work, the success of decision trees (and by extension, of all tree-based methods) is explained by several factors that make them quite attractive in practice:

- Decision trees are non-parametric. They can model arbitrarily complex relations between inputs and outputs, without any a priori assumption;
- Decision trees handle heterogeneous data (ordered or categorical variables, or a mix of both);
- Decision trees intrinsically implement feature selection, making them robust to irrelevant or noisy variables (at least to some extent);
- Decision trees are robust to outliers or errors in labels;
- Decision trees are easily interpretable, even for non-statistically oriented users.

Most importantly, decision trees are at the foundation of many modern and state-of-the-art algorithms, including forests of randomized trees (on which this work is about, see Chapter 4) or boosting methods [Freund and Schapire, 1995; Friedman, 2001], where they are used as building blocks for composing larger models. Understanding all algorithmic details of single decision trees is therefore a mandatory prerequisite for an in-depth analysis of these methods.

## 3.2 TREE STRUCTURED MODELS

When the output space is a finite set of values, like in classification where $\mathcal{Y} = \{c_1, c_2, ..., c_J\}$, another way of looking at a supervised learning problem is to notice that $Y$ defines a partition over the universe $\Omega$, that is

$$\Omega = \Omega_{c_1} \cup \Omega_{c_2} \cup \dots \cup \Omega_{c_J}, \tag{3.1}$$

where $\Omega_{c_k}$ is the of set objects for which $Y$ has value $c_k$. Similarly, a classifier $\varphi$ can also be regarded as a partition of the universe $\Omega$ since it defines an approximation $\widehat{Y}$ of $Y$. This partition however is defined on the input space $\mathcal{X}$ rather that directly on $\Omega$, that is

$$\mathcal{X} = \mathcal{X}_{c_1}^{\varphi} \cup \mathcal{X}_{c_2}^{\varphi} \cup ... \cup \mathcal{X}_{c_J}^{\varphi}, \tag{3.2}$$

where $\mathcal{X}_{c_k}^{\varphi}$ is the set of description vectors $\mathbf{x} \in \mathcal{X}$ such that $\varphi(\mathbf{x}) = c_k$. Accordingly, learning a classifier can be restated as learning a partition of $\mathcal{X}$ matching as closely as possible the best possible partition, i.e., the one engendered by the Bayes model $\varphi_B$ over $\mathcal{X}$:

$$\mathcal{X} = \mathcal{X}_{c_1}^{\varphi_B} \cup \mathcal{X}_{c_2}^{\varphi_B} \cup ... \cup \mathcal{X}_{c_J}^{\varphi_B}. \tag{3.3}$$

**Partitioning with noise**

Notice that when $Y$ cannot be univocally determined given $X = \mathbf{x}$, e.g., when there is noise on $Y$, then there may exist two distinct objects from the universe $\Omega$ such that their representations $\mathbf{x}_1$ and $\mathbf{x}_2$ in the input space are equal, yet such that the corresponding output values $y_1$ and $y_2$ are different. In other words, the subsets

$$\mathcal{X}^{\Omega}_{c_k} = \{\mathbf{x}_i | i \in \Omega, Y = c_k\} \tag{3.4}$$

may not be disjoint. By contrast, since $\varphi$ defines a function from $\mathcal{X}$ to $\mathcal{Y}$, any input $\mathbf{x} \in \mathcal{X}$ is mapped to exactly one output value $y \in \mathcal{Y}$ and the subsets $\mathcal{X}^{\varphi}_{c_k}$ are therefore necessarily disjoint, which means that no model will ever perfectly predict the true output value in all cases. As discussed in Section 2.2.2, this limitation is unavoidable and can in fact be viewed as the cause of the residual error.

From a geometrical point of view, the principle of tree structured models is beautifully simple. It consists in approximating the partition of the Bayes model by recursively partitioning the input space $\mathcal{X}$ into subspaces and then assign constant prediction values $\widehat{y} \in \mathcal{Y}$ to all objects $\mathbf{x}$ within each terminal subspace. To make things clearer, let us first define the following concepts:

**Definition 3.1.** *A* tree *is a graph* $G = (V, E)$ *in which any two vertices (or* nodes*) are connected by exactly one path.*

**Definition 3.2.** *A* rooted tree *is a tree in which one of the nodes has been designated as the* root*. In our case, we additionally assume that a rooted tree is a* directed *graph, where all edges are directed away from the root.*

**Definition 3.3.** *If there exists an edge from* $t_1$ *to* $t_2$ *(i.e., if* $(t_1, t_2) \in E$*) then node* $t_1$ *is said to be the* parent *of node* $t_2$ *while node* $t_2$ *is said to be a* child *of node* $t_1$*.*

**Definition 3.4.** *In a rooted tree, a node is said to be* internal *if it has one or more children and* terminal *if it has no children. Terminal nodes are also known as* leaves*.*

**Definition 3.5.** *A* binary tree *is a rooted tree where all internal nodes have exactly two children.*

In those terms, a *tree-structured model* (or *decision tree*) can be defined as a model $\varphi : \mathcal{X} \mapsto \mathcal{Y}$ represented by a rooted tree (often binary, but not necessarily), where any node $t$ represents a subspace $\mathcal{X}_t \subseteq \mathcal{X}$ of the input space, with the root node $t_0$ corresponding to $\mathcal{X}$ itself. Internal nodes $t$ are labeled with a *split* $s_t$ taken from a set of questions $\mathcal{Q}$. It divides the space $\mathcal{X}_t$ that node $t$ represents into disjoint subspaces respectively corresponding to each of its children. For instance, the set of all binary splits is the set $\mathcal{Q}$ of questions $s$ of the form *"Does*

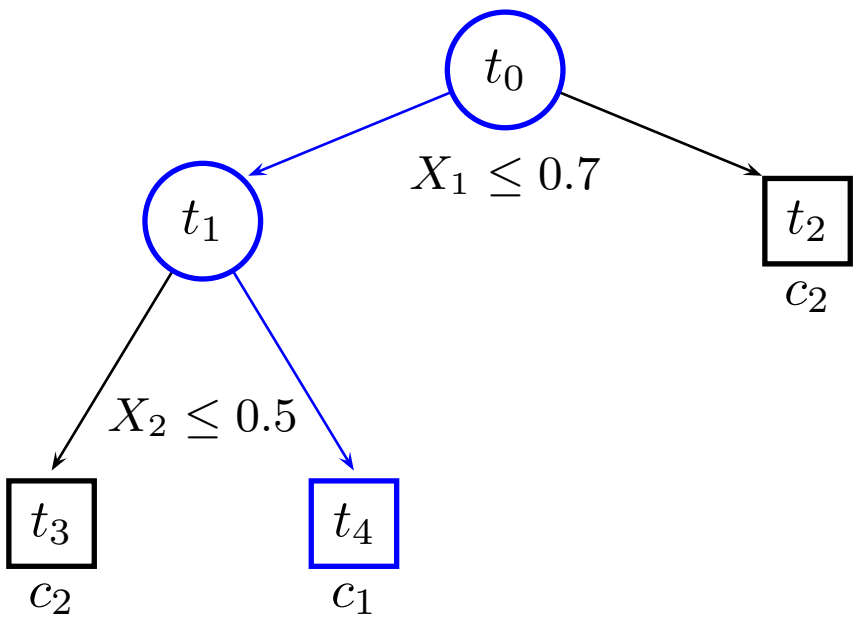

Figure 3.1: A decision tree $\varphi$ built for a binary classification problem from an input space $\mathcal{X} = [0, 1] \times [0, 1]$. (Figure inspired from Breiman et al. [1984].)

$\mathbf{x} \in \mathcal{X}_A$?", where $\mathcal{X}_A \subset \mathcal{X}$ is some subset of the input space. Any split $s$ of this form divides $\mathcal{X}_t$ into two subspaces respectively corresponding to $\mathcal{X}_t \cap \mathcal{X}_A$ for the left child of $t$ and to $\mathcal{X}_t \cap (\mathcal{X} \setminus \mathcal{X}_A)$ for the right child of $t$. Terminal nodes are labeled with a best guess value $\widehat{y}_t \in \mathcal{Y}$ of the output variable. If $\varphi$ is a classification tree, then $\widehat{y}_t \in \{c_1, ..., c_J\}$ while if $\varphi$ is a regression tree, then $\widehat{y}_t \in \mathbb{R}$. As such, the predicted output value $\varphi(\mathbf{x})$ is the label of the leaf reached by the instance $\mathbf{x}$ when it is propagated through the tree by following the splits $s_t$ (see Algorithm 3.1).

**Algorithm 3.1.** *Prediction of the output value $\widehat{y} = \varphi(\mathbf{x})$ in a decision tree.*

```
1: function Predict(φ, x)
2:     t = t_0
3:     while t is not a terminal node do
4:         t = the child node t' of t such that x ∈ X_t'
5:     end while
6:     return ŷ_t
7: end function
```

As an example, Figure 3.1 illustrates a decision tree $\varphi$ made of five nodes and partitioning the input space $\mathcal{X} = \mathcal{X}_1 \times \mathcal{X}_2 = [0; 1] \times [0; 1]$ for a binary classification problem (where $\mathcal{Y} = \{c_1, c_2\}$). Node $t_0$ is the root node and corresponds to the whole input space $\mathcal{X}_{t_0} = \mathcal{X}$. It is labeled with the binary split $X_1 \leqslant 0.7$ (i.e., the question *"Is $X_1 \leqslant 0.7$?"*) which divides $\mathcal{X}_{t_0}$ into two disjoint subsets $\mathcal{X}_{t_1} \cup \mathcal{X}_{t_2}$. The first set corresponds to its left child $t_1$ and represents the set of all input vectors $\mathbf{x} \in \mathcal{X}_0$ such that $x_1 \leqslant 0.7$. Similarly, the second set corresponds to its right child $t_2$ and represents the set of all input vectors $\mathbf{x} \in \mathcal{X}_{t_0}$ such that $x_1 > 0.7$. Likewise, $t_1$ is labeled with the split $X_2 \leqslant 0.5$ which further divides $\mathcal{X}_{t_1}$ into two disjoint subsets $\mathcal{X}_{t_3} \cup \mathcal{X}_{t_4}$ respectively corresponding to the sets of all input vectors $\mathbf{x} \in \mathcal{X}_{t_1}$ such that $x_2 \leqslant 0.5$ and $x_2 > 0.5$. Terminal nodes $t_2$, $t_3$ and $t_4$ are represented by squares labeled with an output value $\widehat{y}_t$. They form together a partition (as defined by Equation 3.2) of $\mathcal{X}$, where each set $\mathcal{X}^{\varphi}_{c_k}$ is obtained from the union of the subspaces $\mathcal{X}_t$ of all terminal nodes $t$ such that $\widehat{y}_t = c_k$. In this case, $\mathcal{X}^{\varphi}_{c_1} = \mathcal{X}_{t_4}$ while $\mathcal{X}^{\varphi}_{c_2} = \mathcal{X}_{t_2} \cup \mathcal{X}_{t_3}$. As shown in Figure 3.2, the partition induced by $\varphi$ on $\mathcal{X}$ divides the input space into subspaces that are more and more class homogeneous, starting from $\mathcal{X}$ at the root node, then $\mathcal{X}_{t_1} \cup \mathcal{X}_{t_2}$ at the second level of tree and finally $(\mathcal{X}_{t_3} \cup \mathcal{X}_{t_4}) \cup \mathcal{X}_{t_2}$ at the leaves. As we will explore in Section 3.6, the partition is in this case made of rectangles because of the nature of the splits $s_t \in \mathcal{Q}$ dividing the nodes. Predictions are made by propagating instances through the tree and using as output value the value labeling the terminal nodes in which they fall into. For example, as shown in blue in Figure 3.1, an object $\mathbf{x} = (x_1 = 0.2, x_2 = 0.7)$ falls into $t_4$ and therefore $\varphi(\mathbf{x}) = \widehat{y}_{t_4} = c_1$.

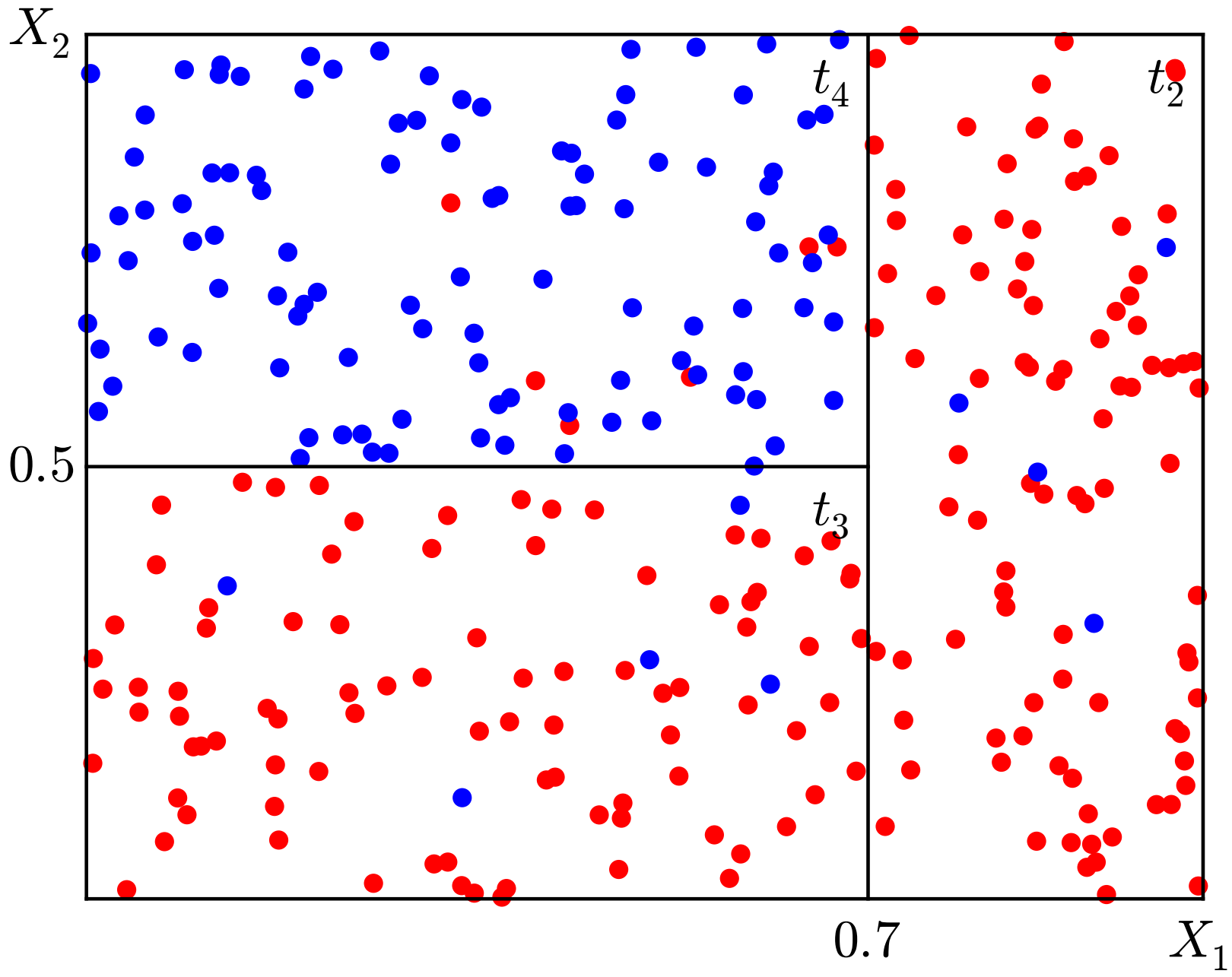

Figure 3.2: Partition of $\mathcal{X}$ induced by the decision tree $\varphi$ into homogeneous subspaces. Blue dots correspond to objects of class $c_1$ while red dots correspond to objects of class $c_2$. (Figure inspired from Breiman et al. [1984].)

Finally, let us note that from a graph theory point of view, decision trees belong to a larger family of methods known as *induction graphs* [Zighed and Rakotomalala, 2000]. In this more general framework, the structure is a directed acyclic graph, which allows for nodes to be both divided and recombined.

## 3.3 INDUCTION OF DECISION TREES

Learning a decision tree ideally amounts to determine the tree structure producing the partition which is closest to the partition engendered by $Y$ over $\mathcal{X}$. Since it is unknown, the construction of a decision tree is usually driven instead with the objective of finding a model which partitions the learning set $\mathcal{L}$ as well as possible. Among all decision trees $\varphi \in \mathcal{H}$ however, there may exist several of them that explain $\mathcal{L}$ equally best. Following Occam's Razor principles [Blumer et al., 1987] of preferring the explanation which makes as few assumptions as possible, that is to favor the simplest solution that fits the data, learning a decision tree from $\mathcal{L}$ is therefore usually restated as finding the smallest tree $\varphi^*$ (in terms of internal nodes) minimizing its resubstitution estimate $\overline{E}(\varphi^*, \mathcal{L})$. While this assumption makes sense from a generalization point of view, it also makes sense regarding interpretability. A decision tree which is small is easier to understand than a large and complex tree.

As shown by Hyafil and Rivest [1976], finding the smallest tree $\varphi^*$ that minimizes its resubstitution estimate is an NP-complete problem. As a consequence, under the assumption that $P \neq NP$, there exists no efficient algorithm for finding $\varphi^*$, thereby suggesting that finding efficient heuristics for constructing near-optimal decision trees is the best solution to keep computation requirements within realistic boundaries.

Following the framework of Breiman et al. [1984], let us broadly define an *impurity* measure $i(t)$ as a function that evaluates the goodness of any node $t$ (we delay to Section 3.6 for a more precise definition). Let assume that the smaller $i(t)$, the *purer* the node and the better the predictions $\widehat{y}_t(\mathbf{x})$ for all $\mathbf{x} \in \mathcal{L}_t$, where $\mathcal{L}_t$ is the subset of learning samples falling into $t$, that is all $(\mathbf{x}, y) \in \mathcal{L}$ such that $\mathbf{x} \in \mathcal{X}_t$. Starting from a single node representing the whole learning set $\mathcal{L}$, near-optimal decision trees can then be grown greedily by iteratively dividing nodes into purer nodes. That is, by iteratively dividing into smaller subsets the subsets of $\mathcal{L}$ represented by the nodes, until all terminal nodes cannot be made purer, hence guaranteeing near-optimal predictions over $\mathcal{L}$. The greedy assumption to make the resulting tree as small as possible, thereby seeking for good generalization, is then to divide each node $t$ using the split $s^*$ that locally maximizes the decrease of impurity of the resulting child nodes.

Formally, the decrease of impurity of a binary split $s$ is defined as follows:

**Definition 3.6.** *The* impurity decrease *of a binary split $s \in \mathcal{Q}$ dividing node $t$ into a left node $t_L$ and a right node $t_R$ is*

$$\Delta i(s, t) = i(t) - p_L i(t_L) - p_R i(t_R) \tag{3.5}$$

*where $p_L$ (resp., $p_R$) is the proportion $\frac{N_{t_L}}{N_t}$ (resp., $\frac{N_{t_R}}{N_t}$) of learning samples from $\mathcal{L}_t$ going to $t_L$ (resp., to $t_R$) and where $N_t$ is the size of the subset $\mathcal{L}_t$.*

On the basis of this concept, a general greedy procedure for the induction of decision trees can now be described formally as outlined in Algorithm 3.2. Note that for the sake of clarity, Algorithm 3.2 is formulated in terms of binary splits. However, as we will explore later in Section 3.6, it generalizes naturally to $n$-ary multiway splits.

**Algorithm 3.2.** *Greedy induction of a binary decision tree.*

1: **function** BUILDDECISIONTREE($\mathcal{L}$)
2: Create a decision tree $\varphi$ with root node $t_0$
3: Create an empty stack $S$ of *open* nodes $(t, \mathcal{L}_t)$
4: $S$.PUSH($(t_0, \mathcal{L})$)
5: **while** $S$ is not empty **do**
6: $t, \mathcal{L}_t = S$.POP( )
7: **if** the stopping criterion is met for $t$ **then**
8: $\widehat{y}_t =$ some constant value
9: **else**
10: Find the split on $\mathcal{L}_t$ that maximizes impurity decrease

$$s^* = \arg\max_{s \in \Omega} \Delta i(s, t)$$

11: Partition $\mathcal{L}_t$ into $\mathcal{L}_{t_L} \cup \mathcal{L}_{t_R}$ according to $s^*$
12: Create the left child node $t_L$ of $t$
13: Create the right child node $t_R$ of $t$
14: $S$.PUSH($(t_R, \mathcal{L}_{t_R})$)
15: $S$.PUSH($(t_L, \mathcal{L}_{t_L})$)
16: **end if**
17: **end while**
18: **return** $\varphi$
19: **end function**

The rest of this chapter is now dedicated to a detailed discussion of specific parts of Algorithm 3.2. Section 3.4 discusses assignment rules for terminal nodes (Line 8 of Algorithm 3.2) while Section 3.5 outlines stopping criteria for deciding when a node becomes terminal (Line 7 of Algorithm 3.2). Section 3.6 then presents families $\Omega$ of splitting rules, impurity criteria $i(t)$ to evaluate the goodness of splits and strategies for finding the best split $s^* \in \Omega$ (Line 10 of Algorithm 3.2). As we will see, the crux of the problem is in finding good splits and in knowing when to stop splitting.

### LOOKAHEAD SEARCH

Obviously, the greedy assumption on which Algorithm 3.2 relies may produce trees that are suboptimal. A natural improvement of the greedy strategy is lookahead search, which consists in evaluating the goodness of a split, but also of those deeper in the tree, assuming this former split was effectively performed. As empirically investigated in [Murthy and Salzberg, 1995], such an approach is however not only more computationally intensive, it is also not significantly better than greedily induced decision trees. For this reason, more elaborate variants of Algorithm 3.2 are not considered within this work. Let us note however that trees grown with lookahead search are usually shorter, which may be a strong advantage when interpretability matters.

## 3.4 ASSIGNMENT RULES

Let us assume that node t has been declared terminal given some stopping criterion (See next Section 3.5). The next step (Line 8 of Algorithm 3.2) in the induction procedure is to label t with a constant value $\widehat{y}_t$ to be used as a prediction of the output variable Y. As such, node t can be regarded as a simplistic model defined locally on $\mathcal{X}_t \times \mathcal{Y}$ and producing the same output value $\widehat{y}_t$ for all possible input vectors falling into t.

Let us first notice that, for a tree $\varphi$ of *fixed* structure, minimizing the global generalization error is strictly equivalent to minimizing the local generalization error of each simplistic model in the terminal nodes. Indeed,

$$\begin{aligned} Err(\varphi) &= \mathbb{E}_{X,Y}\{L(Y, \varphi(X))\} \\ &= \sum_{t \in \widetilde{\varphi}} P(X \in \mathcal{X}_t) \mathbb{E}_{X,Y|t}\{L(Y, \widehat{y}_t)\} \end{aligned} \tag{3.6}$$

where $\widetilde{\varphi}$ denotes the set of terminal nodes in $\varphi$ and where the inner expectation[1] is the local generalization error of the model at node t. In this later form, a model which minimizes $Err(\varphi)$ is a model which minimizes the inner expectation leaf-wise. Learning the best possible decision tree (of fixed structure) therefore simply amounts to find the best constants $\widehat{y}_t$ at each terminal node.

### 3.4.1 *Classification*

When L is the zero-one loss, the inner expectation in Equation 3.6 is minimized by the plurality rule:

$$\begin{aligned} \widehat{y}_t^* &= \arg\min_{c \in \mathcal{Y}} \mathbb{E}_{X,Y|t}\{1(Y, c)\} \\ &= \arg\min_{c \in \mathcal{Y}} P(Y \neq c | X \in \mathcal{X}_t) \\ &= \arg\max_{c \in \mathcal{Y}} P(Y = c | X \in \mathcal{X}_t) \end{aligned} \tag{3.7}$$

Put otherwise, the generalization error of t is minimized by predicting the class which is the most likely for the samples in the subspace of t. Note that if the maximum is achieved by two or more different classes, then $\widehat{y}_t^*$ is assigned arbitrarily as any one of the maximizing classes.

Equation 3.7 cannot be solved without the probability distribution $P(X, Y)$. However, its solution can be approximated by using estimates of the local generalization error. Let $N_t$ denotes the number of objects in $\mathcal{L}_t$ and let $N_{ct}$ denotes the number of objects of class c in $\mathcal{L}_t$. Then, the proportion $\frac{N_{ct}}{N_t}$ can be interpreted as the estimated probability[2]

1 The joint expectation of X and Y is taken over all objects $i \in \Omega$ such that $\mathbf{x}_i \in \mathcal{X}_t$.

2 Lower case p denotes an estimated probability while upper case P denotes a theoretical probability.

$p(Y = c|X \in \mathcal{X}_t)$ (shortly denoted $p(c|t)$) of class $c$ in $t$ and therefore be used to solve Equation 3.7:

$$\begin{aligned}\widehat{y}_t &= \arg\min_{c \in \mathcal{Y}} 1 - p(c|t) \\ &= \arg\max_{c \in \mathcal{Y}} p(c|t) \qquad (3.8)\end{aligned}$$

Similarly, let us also define the proportion $\frac{N_t}{N}$ as the estimated probability $p(X \in \mathcal{X}_t)$ (shortly denoted $p(t)$). Plugging this estimate into Equation 3.6 and approximating the local generalization error with $1 - p(\widehat{y}_t|t)$ as done above, it follows:

$$\begin{aligned}\widehat{Err}(\varphi) &= \sum_{t \in \widetilde{\varphi}} p(t)(1 - p(\widehat{y}_t|t)) \\ &= \sum_{t \in \widetilde{\varphi}} \frac{N_t}{N}(1 - \frac{N_{\widehat{y}_t t}}{N_t}) \\ &= \frac{1}{N} \sum_{t \in \widetilde{\varphi}} N_t - N_{\widehat{y}_t t} \\ &= \frac{1}{N} \sum_{t \in \widetilde{\varphi}} \sum_{\mathbf{x}, y \in \mathcal{L}_t} 1(y \neq \widehat{y}_t) \\ &= \frac{1}{N} \sum_{\mathbf{x}, y \in \mathcal{L}} 1(y \neq \varphi(\mathbf{x})) \\ &= \widehat{Err}^{\text{train}}(\varphi) \qquad (3.9)\end{aligned}$$

Accordingly, approximating Equation 3.6 through local probability estimates computed from class proportions in $\mathcal{L}_t$ reduces to the resubstitution estimate of $\varphi$ (Equation 2.5). In other words, assignment rule 3.8 in fact minimizes the resubstitution estimate rather than the true generalization error.

An important property of assignment rule 3.8 is that the more one splits a terminal node *in any way*, the smaller the resubstitution estimate $\widehat{Err}^{\text{train}}(\varphi)$ becomes.

**Proposition 3.1.** *For any non-empty split of a terminal node $t \in \widetilde{\varphi}$ into $t_L$ and $t_R$, resulting in a new tree $\varphi'$ where $\widehat{y}_{t_L}$ and $\widehat{y}_{t_R}$ are assigned with rule 3.8,*

$$\widehat{Err}^{train}(\varphi) \geqslant \widehat{Err}^{train}(\varphi')$$

*with equality if $\widehat{y}_t = \widehat{y}_{t_L} = \widehat{y}_{t_R}$.*

*Proof.*

$$\begin{aligned}\widehat{Err}^{\text{train}}(\varphi) &\geqslant \widehat{Err}^{\text{train}}(\varphi') \\ \sum_{t \in \widetilde{\varphi}} p(t)(1 - p(\widehat{y}_t|t)) &\geqslant \sum_{t \in \widetilde{\varphi'}} p(t)(1 - p(\widehat{y}_t|t)) \\ p(t)(1 - p(\widehat{y}_t|t)) &\geqslant p(t_L)(1 - p(\widehat{y}_{t_L}|t_L)) + p(t_R)(1 - p(\widehat{y}_{t_R}|t_R))\end{aligned}$$

$$\frac{N_t}{N}(1 - \max_{c \in \mathcal{Y}} \frac{N_{ct}}{N_t}) \geqslant \frac{N_{t_L}}{N}(1 - \max_{c \in \mathcal{Y}} \frac{N_{ct_L}}{N_{t_L}}) + \frac{N_{t_R}}{N}(1 - \max_{c \in \mathcal{Y}} \frac{N_{ct_R}}{N_{t_R}})$$
$$N_t - \max_{c \in \mathcal{Y}} N_{ct} \geqslant N_{t_L} - \max_{c \in \mathcal{Y}} N_{ct_L} + N_{t_R} - \max_{c \in \mathcal{Y}} N_{ct_R}$$
$$\max_{c \in \mathcal{Y}} N_{ct} \leqslant \max_{c \in \mathcal{Y}} N_{ct_L} + \max_{c \in \mathcal{Y}} N_{ct_R}$$
$$\max_{c \in \mathcal{Y}}(N_{ct_L} + N_{ct_R}) \leqslant \max_{c \in \mathcal{Y}} N_{ct_L} + \max_{c \in \mathcal{Y}} N_{ct_R}$$

which is true since $\max_{c \in \mathcal{Y}} N_{ct_L}$ (resp., $t_R$) is necessarily greater or equal to the left term $N_{ct_L}$ (resp., to right term $N_{ct_R}$) in the left-hand side of the equation. Equality holds if the majority classes are the same in $t$, $t_L$ and $t_R$. □

As a corollary of Proposition 3.1, the resubstitution estimate is minimal when terminal nodes can no longer be divided. In particular, it is equal to zero if the tree can be *fully developed*, that is if terminal nodes can be divided until they all contain exactly one object from $\mathcal{L}$. Based on this result, we will show however in Section 3.6.2 why using the resubstitution estimate to evaluate the goodness of a split $s$ has serious defects for classification.

### 3.4.2 *Regression*

When $L$ is the squared error loss, the inner expectation in Equation 3.6 is minimized by the expectation of $Y$ in $t$:

$$\begin{aligned} \widehat{y}_t^* &= \arg\min_{\widehat{y} \in \mathcal{Y}} \mathbb{E}_{X,Y|t}\{(Y - \widehat{y})^2\} \\ &= \mathbb{E}_{X,Y|t}\{Y\} \end{aligned} \tag{3.10}$$

Again, Equation 3.10 cannot be solved without the probability distribution $P(X,Y)$. However, its solution can approximated using estimates of the local generalization error:

$$\begin{aligned} \widehat{y}_t &= \arg\min_{\widehat{y} \in \mathcal{L}} \frac{1}{N_t} \sum_{x,y \in \mathcal{L}_t} (y - \widehat{y})^2 \\ &= \frac{1}{N_t} \sum_{x,y \in \mathcal{L}_t} y \end{aligned} \tag{3.11}$$

As in classification, using $p(t)$ as estimate of $P(X \in \mathcal{X}_t)$ and approximating the local generalization error with $\frac{1}{N_t} \sum_{x,y \in \mathcal{L}_t} (y - \widehat{y}_t)^2$ as done above, one can show that Equation 3.6 reduces to the resubstitution estimate of $\varphi$, thereby indicating that assignment rule 3.11 also minimizes the training error rather the true generalization error.

Most importantly, one can also show that assignment rule 3.11 behaves in the same way as 3.8. The more one splits a terminal node in any way, the smaller the resubstitution estimate $\widehat{Err}^{train}(\varphi)$ becomes.

**Proposition 3.2.** *For any non-empty split of a terminal node $t \in \widetilde{\varphi}$ into $t_L$ and $t_R$, resulting in a new tree $\varphi'$ where $\widehat{y}_{t_L}$ and $\widehat{y}_{t_R}$ are assigned with rule 3.11,*

$$\widehat{Err}^{train}(\varphi) \geqslant \widehat{Err}^{train}(\varphi')$$

*with equality if $\widehat{y}_{t_L} = \widehat{y}_{t_R} = \widehat{y}_t$.*

*Proof.*

$$\widehat{Err}^{train}(\varphi) \geqslant \widehat{Err}^{train}(\varphi')$$

$$\sum_{t\in\widetilde{\varphi}} p(t)(\frac{1}{N_t}\sum_{x,y\in\mathcal{L}_t}(y-\widehat{y}_t)^2) \geqslant \sum_{t\in\widetilde{\varphi'}} p(t)(\frac{1}{N_t}\sum_{x,y\in\mathcal{L}_t}(y-\widehat{y}_t)^2)$$

$$\sum_{x,y\in\mathcal{L}_t}(y-\widehat{y}_t)^2 \geqslant \sum_{x,y\in\mathcal{L}_{t_L}}(y-\widehat{y}_{t_L})^2 + \sum_{x,y\in\mathcal{L}_{t_R}}(y-\widehat{y}_{t_R})^2$$

$$N_t\widehat{y}_t^2 \leqslant N_{t_L}\widehat{y}_{t_L}^2 + N_{t_R}\widehat{y}_{t_R}^2$$

$$\frac{1}{N_t}(\sum_{x,y\in\mathcal{L}_t} y)^2 \leqslant \frac{1}{N_{t_L}}(\sum_{x,y\in\mathcal{L}_{t_L}} y)^2 + \frac{1}{N_{t_R}}(\sum_{x,y\in\mathcal{L}_{t_R}} y)^2$$

For simplifying, let us denote $s(t) = \sum_{x,y\in\mathcal{L}_t} y = N_t\widehat{y}_t$ the sum of the output values in t. Remark that $s(t) = s(t_L) + s(t_R)$. It comes:

$$\frac{s(t)^2}{N_t} \leqslant \frac{s(t_L)^2}{N_{t_L}} + \frac{s(t_R)^2}{N_{t_R}}$$

$$\frac{(s(t_L)+s(t_R))^2}{N_{t_L}+N_{t_R}} \leqslant \frac{s(t_L)^2}{N_{t_L}} + \frac{s(t_R)^2}{N_{t_R}}$$

$$\frac{(s(t_L)N_{t_R} - s(t_R)N_{t_L})^2}{N_{t_L}N_{t_R}(N_{t_L}+N_{t_R})} \geqslant 0$$

Which is necessary true since the numerator is non-negative and the denominator is strictly positive. Equality holds if $s(t_L)N_{t_R} = s(t_R)N_{t_L}$, that is if $\widehat{y}_{t_L} = \widehat{y}_{t_R}$. □

As for Proposition 3.1, a corollary of Proposition 3.2 is that the resubstitution estimate is minimal when terminal nodes can no longer be divided.

## 3.5 STOPPING CRITERIA

As we have shown through Propositions 3.1 and 3.2, the deeper a decision tree, the smaller its training estimate, and the better we think it is. As discussed in Section 2.3 however, increasing model complexity, in this case by enlarging the set of terminal nodes, is likely to eventually capture noise in the learning set and to cause overfitting. In other words, it is not because the training estimate can be proved to go down to zero that the test estimate converges accordingly. On the contrary, it is very likely to diverge as complexity increases. To prevent this phenomenon it is therefore necessary to find the right trade-off between a tree which is not too shallow nor too deep. The problem is in knowing when to stop splitting, i.e., in how to carefully formulate Line 7 of Algorithm 3.2.

Let us first consider the stopping criteria that are inherent to the iterative partition procedure, regardless of overfitting. Node $t$ is inevitably set as a terminal node when $\mathcal{L}_t$ can no longer be split, which happens in the following cases:

(a) When the output values of the samples in $\mathcal{L}_t$ are homogeneous. That is, if $y = y'$ for all $(\mathbf{x}, y), (\mathbf{x}', y') \in \mathcal{L}_t$. In particular, this is necessarily the case when $N_t = 1$.

(b) When the input variables $X_j$ are each locally constant in $\mathcal{L}_t$. That is, if $x_j = x'_j$ for all $(\mathbf{x}, y), (\mathbf{x}', y') \in \mathcal{L}_t$, for each input variable $X_j$. In this situation, it is indeed not possible to divide $\mathcal{L}_t$ into two (or more) non-empty subsets.

To prevent overfitting, the stopping criterion is then usually complemented with heuristics halting the recursive partition if $\mathcal{L}_t$ has become too small or if no sufficiently good split can be found. The most common approaches are formulated as follows:

(c) Set $t$ as a terminal node if it contains less than $N_{min}$ samples. ($N_{min}$ is also known as `min_samples_split`.)

(d) Set $t$ as a terminal node if its depth $d_t$ is greater or equal to a threshold $d_{max}$. ($d_{max}$ is also known as `max_depth`.)

(e) Set $t$ as a terminal node if the total decrease in impurity is less than a fixed threshold $\beta$. That is, if $p(t)\Delta i(s^*, t) < \beta$.

(f) Set $t$ as a terminal node if there is no split such that $t_L$ and $t_R$ both count a least $N_{leaf}$ samples. ($N_{leaf}$ is also known as `min_samples_leaf`.)

In all of the above, stopping criteria are defined in terms of user-defined hyper-parameters ($N_{min}$, $d_{max}$, $\beta$ or $N_{leaf}$) that have to be tuned in order to find the right trade-off. Ideally, they need to be such that they are neither too strict nor too loose for the tree to be neither too shallow nor too deep. Too large a tree will have a higher generalization error than the right sized tree. Likewise, too small a tree will not use some of the information in $\mathcal{L}$, again resulting in a higher generalization error than the right sized tree. As described in Section 2.3, choosing the appropriate parameter values is usually performed using a dedicated model selection procedure, which can be computationally expensive, yet usually imperative for reaching good generalization performance.

**Pre-pruning and post-pruning**

While the stopping criteria presented above may give good results in practice, the strategy of stopping early the induction of the tree is in general unsatisfactory. There may be nodes $t$ for which the stopping criterion is met but whose descendants $t_L$ and $t_R$ may have splits that would have in fact reduced the generalization error of the tree. By declaring $t$ as a terminal node, the good splits on $t_L$ and $t_R$ are never exploited.

Another way of looking at the problem of finding the right sized tree consists in fully developing all nodes and then to *prune* instead of stopping. That is, to sequentially remove the nodes that degrade the generalization error (estimated on an independent validation set) until the optimal tree is found. Since nodes are pruned after the induction of the tree, this framework is also known as *post-pruning*. By opposition, early stopping as described earlier is also referred to as *pre-pruning*.

In the context of single decision trees, post-pruning usually yields better results than pre-pruning. From an interpretability point of view, it is also a very effective framework for simplifying decision trees and better understand the structure in the data. However, in the context of ensemble of decision trees, we will see in Chapter 4 that (pre- or post-)pruning is no longer required to achieve good generalization performance. For this reason, we refer to the literature for more detailed discussion on the topic (e.g., [Breiman et al., 1984; Mingers, 1989a; Zighed and Rakotomalala, 2000]).

## 3.6 SPLITTING RULES

Assuming that the stopping criterion is not met, we now focus on the problem of finding the split $s^* \in \Omega$ of $t$ that maximizes the impurity decrease $\Delta i(s^*, t)$ (Line 10 of Algorithm 3.2).

### 3.6.1 *Families $\Omega$ of splitting rules*

As introduced in Section 3.2, a split $s$ of node $t$ is a question that divides the space $\mathcal{X}_t$ into disjoint subspaces respectively corresponding to each of the children of $t$. Formally, a split is defined as a partition:

**Definition 3.7.** *A split $s$ of node $t$ is a partition of $\mathcal{X}_t$, that is a set of non-empty subsets of $\mathcal{X}_t$ such that every element $\mathbf{x} \in \mathcal{X}_t$ is in exactly one of these subsets (i.e., $\mathcal{X}_t$ is a disjoint union of the subsets).*

If $s$ divides $\mathcal{X}_t$ into two subsets, then $s$ is said to be a *binary* split and its left and right children are denoted $t_L$ and $t_R$. In the general case however, $s$ may divide $\mathcal{X}_t$ into more than two subsets, resulting in as many child nodes $t_{i_1}, ..., t_{i_N}$ as subsets (See Figure 3.3).

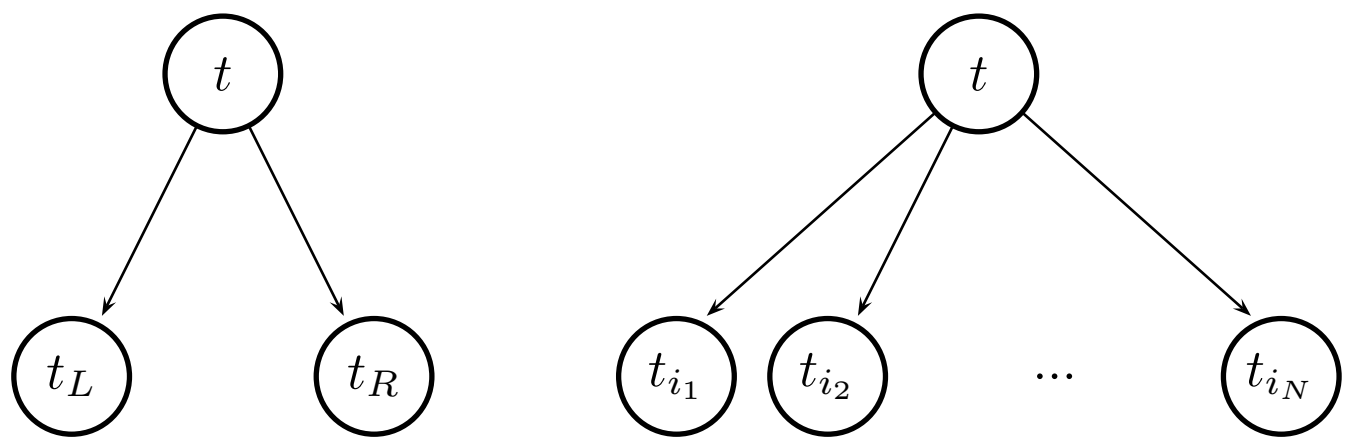

Figure 3.3: Binary split of t (left) and N-ary multiway split of t (right).

The set $\mathcal{S}$ of all possible splits s is the set of all partitions of $\mathcal{X}_t$. It is infinitely large as soon as one the input variables can take infinitely many values. For instance, if $\mathcal{X}_t = \mathbb{R}$, then there exists infinitely many ways of partitioning $\mathcal{X}_t$ into two (or more) subsets. Among all of them however, only those that yield a partition of $\mathcal{L}_t$ into non-empty subsets are worth considering. For example, in the binary case, splits resulting in one of the children being empty are not considered because their goodness cannot be evaluated on the learning set. As such, the optimization problem of finding the best split $s^*$ of $\mathcal{X}_t$ can be roughly restated as finding the best partition of the node samples $\mathcal{L}_t$.

Assuming distinct input values for all $N_t$ node samples, the number of partitions of $\mathcal{L}_t$ into k non-empty subsets is given by the Stirling number of the second kind [Knuth, 1992]

$$S(N_t, k) = \frac{1}{k!} \sum_{j=0}^{k} (-1)^{k-j} \binom{k}{j} j^{N_t}, \tag{3.12}$$

which reduces to $2^{N_t - 1} - 1$ for binary partitions. Given the exponential growth in the number of partitions, the naive strategy of enumerating all partitions and picking the best of them is often computationally intractable. For this reason, simplifying assumptions must be made on the best split $s^*$. More specifically, the induction algorithm usually assumes that $s^*$ – or at least a good enough approximation – lives in a family $\mathcal{Q} \subseteq \mathcal{S}$ of candidate splits of restricted structure.

The usual family $\mathcal{Q}$ of splits is the set of binary splits defined on a single variable and resulting in non-empty subsets of $\mathcal{L}_t$:

$$\mathcal{Q} = \{s | s \in \bigcup_{j=1}^{p} \mathcal{Q}(X_j), \mathcal{L}_{t_L} \neq \phi, \mathcal{L}_{t_R} \neq \phi\} \tag{3.13}$$

- If $X_j$ is an ordered variable taking values in $\mathcal{X}_j$, then the set of binary splits on $X_j$ is the set of all binary non-crossing partitions $s_j^v$ of $\mathcal{X}_j$:

$$\mathcal{Q}(X_j) = \{(\{\mathbf{x} | x_j \leqslant v\}, \{\mathbf{x} | x_j > v\}) | v \in \mathcal{X}_j\} \tag{3.14}$$

From a geometrical point of view, splits of this form partition the input space $\mathcal{X}$ with axis-parallel hyperplanes, as previously illustrated on Figure 3.2. Accordingly, the decision threshold $v$ thus corresponds to the distance of the separating hyperplane from the origin.

- If $X_j$ is a categorical variable taking values in $\mathcal{X}_j = \{b_1, \dots, b_L\}$, then the set of binary splits on $X_j$ is the set of all binary non-empty partitions of $\mathcal{X}_j$:

$$\mathcal{Q}(X_j) = \{(\{\mathbf{x}|x_j \in \mathcal{B}\}, \{\mathbf{x}|x_j \in \overline{\mathcal{B}}\})|\mathcal{B} \subset \{b_1, \dots, b_L\}\} \tag{3.15}$$

where $\overline{\mathcal{B}} = \{b_1, \dots, b_L\} \setminus \mathcal{B}$ is the complementary set of $\mathcal{B}$.

Given decomposition 3.13 into subsets $\mathcal{Q}(X_j)$, the split $s^* \in \mathcal{Q}$ is the best of the best splits defined on each input variable. That is,

$$s^* = \underset{\substack{s_j^* \\ j=1,\dots,p}}{\arg\max}\, \Delta i(s_j^*, t) \tag{3.16}$$

$$s_j^* = \underset{\substack{s \in Q(X_j) \\ \mathcal{L}_{t_L}, \mathcal{L}_{t_R} \neq \phi}}{\arg\max}\, \Delta i(s, t) \tag{3.17}$$

In this framework, the crux of the problem therefore simply reduces to the implementation of Equation 3.17.

While partitions of form 3.14 and 3.15 certainly constitute the most common types of splits, alternatives proposed over the years in the literature are worth mentioning. In ID3 and C4.5, Quinlan [1986, 1993] replaces binary splits on categorical variables with multiway splits. That is, if $\mathcal{X}_j$ counts L different values $b_1, \dots, b_L$, then splitting on $X_j$ divides $\mathcal{X}_t$ into L child nodes – one for each value of the variable. In our framework, $Q(X_j)$ therefore reduces to the singleton

$$Q(X_j) = \{(\{\mathbf{x}|x_j = b_l\}|b_l \in \mathcal{X}_j)\}. \tag{3.18}$$

In oblique decision trees, Heath et al. [1993] propose to replace axis-parallel cutting hyperplanes with hyperplanes of any orientation, resulting into smaller, yet as accurate, decision trees. In this setting however, decomposition 3.16 no longer holds, hence making the search of the best split often more computationally intensive. More recently, several authors (e.g., [Gama, 2004; Criminisi and Shotton, 2013; Botta, 2013]) outlined more general frameworks in which splits are defined as multi-variate functions, thereby revisiting separating hyperplanes like in oblique trees or investigating more advanced models like quadratic surfaces or decision trees inside decision trees. Finally, in another direction, several authors [Adamo, 1980; Yuan and Shaw, 1995; Olaru and Wehenkel, 2003] propose in fuzzy decision trees to redefine the notion of split as overlapping fuzzy sets (instead of disjoint subsets), hence allowing samples close to the decision threshold $v$ to propagate into both child nodes.

| $i$ | $X_1$ | $X_2$ | $X_3$ | $Y$ |
|---|---|---|---|---|
| 0 | 0 | 0 | 0 | $c_1$ |
| 1 | 0 | 0 | 1 | $c_1$ |
| 2 | 0 | 1 | 0 | $c_2$ |
| 3 | 0 | 1 | 1 | $c_2$ |
| 4 | 0 | 1 | 1 | $c_2$ |
| 5 | 1 | 0 | 0 | $c_2$ |
| 6 | 1 | 0 | 0 | $c_2$ |
| 7 | 1 | 0 | 0 | $c_2$ |
| 8 | 1 | 0 | 0 | $c_2$ |
| 9 | 1 | 1 | 1 | $c_2$ |

Table 3.1: Toy binary classification problem

### 3.6.2 *Goodness of split*

In this section, we describe impurity measures $i(t)$ for evaluating the goodness of splitting rules. As they are the heart of our contributions regarding interpretability of tree-based methods (See Chapter 6), the necessary mathematical foundations that have driven their methodological development are studied here in details. Section 3.6.2.1 reviews impurity functions for classification while Section 3.6.2.2 discusses criteria for regression.

#### 3.6.2.1 *Classification*

Let us consider a binary classification problem defined on three binary categorical input variables $X_1$, $X_2$ and $X_3$. Let us assume that we have a 10-sample learning set $\mathcal{L}$ as listed in Table 3.1. At the root node $t_0$, $\mathcal{L}_{t_0}$ contains all learning samples and we are looking for the best binary split defined on one of the input variables. Since each of them is binary and categorical, $\mathcal{Q}(X_j)$ (for $j = 1, 2, 3$) counts exactly one single split per input variable, each partitioning $\mathcal{L}_{t_0}$ into two subsets

$$\mathcal{L}_{t_L} = \{(\mathbf{x}, y) | (\mathbf{x}, y) \in \mathcal{L}_{t_0}, x_j = 0\},$$
$$\mathcal{L}_{t_R} = \{(\mathbf{x}, y) | (\mathbf{x}, y) \in \mathcal{L}_{t_0}, x_j = 1\}.$$

As introduced in Section 3.3, our objective is to partition $t_0$ using the split $s^*$ that maximizes the impurity decrease

$$\Delta i(s, t) = i(t) - p_L i(t_L) - p_R i(t_R),$$

where the impurity function $i(t)$ evaluates the goodness of node $t$. Since our goal is to build the decision tree that minimizes generalization error, it seems natural to take as proxy $i(t)$ the local resubstitution estimate at node $t$:

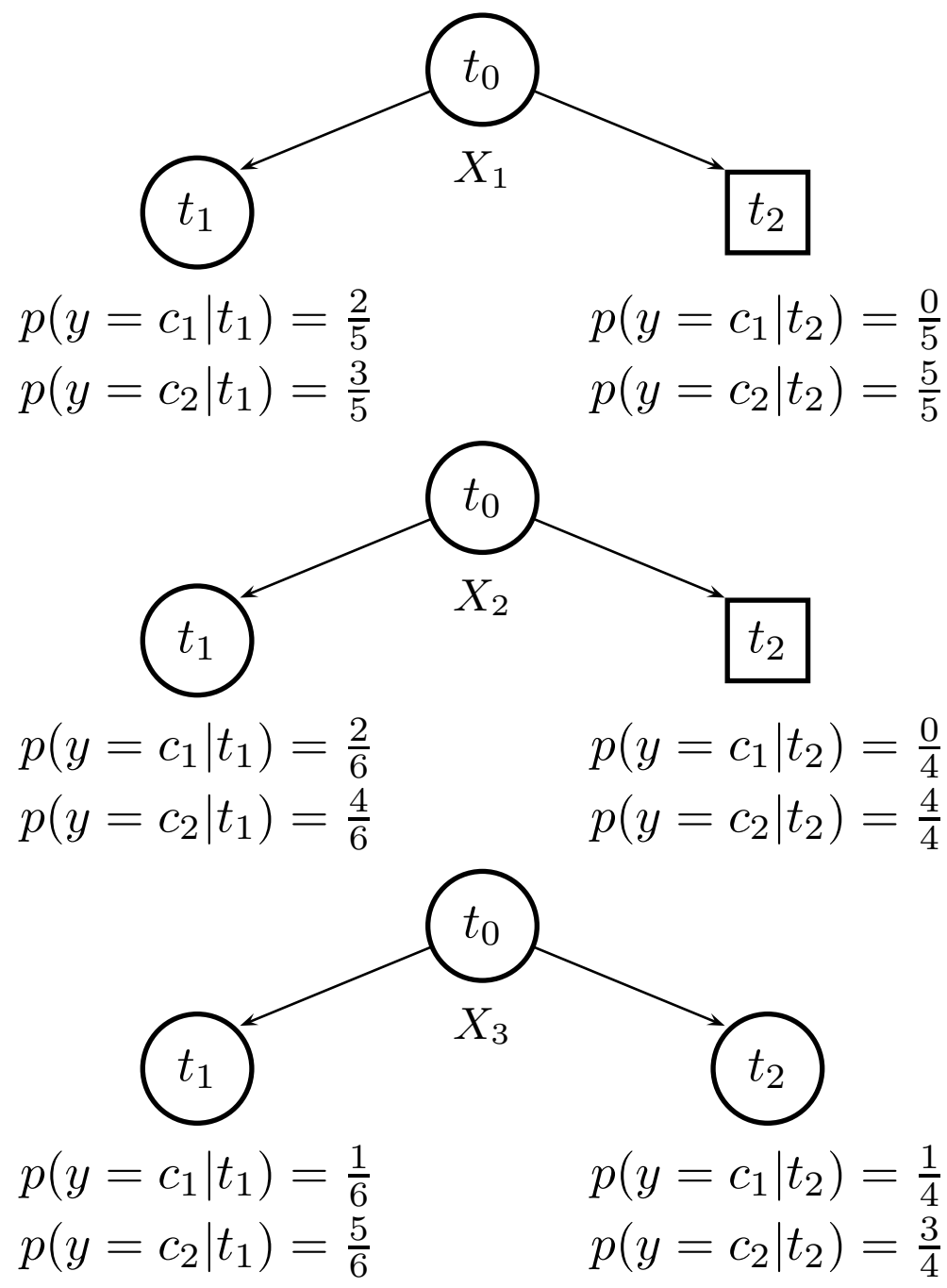

Figure 3.4: Splitting on $X_1$ versus splitting on either $X_2$ or $X_3$.

**Definition 3.8.** *In classification, the impurity function* $i_R(t)$ *based on the local resubstitution estimate defined on the zero-one loss is:*

$$i_R(t) = 1 - p(\widehat{y}_t|t) = 1 - \max_{c \in \mathcal{Y}} p(c|t) \tag{3.19}$$

With this criterion, selecting the best split would amount to pick the split that most decreases training error. In spite of its natural attractiveness however, $i_R(t)$ has two serious defects:

(a) As a corollary of Proposition 3.1, $\Delta i(s,t)$ is zero for all splits if the majority class remains the same in both child nodes, that is if $\widehat{y}_t = \widehat{y}_{t_L} = \widehat{y}_{t_R}$.

In our toy problem, splitting either on $X_1$, $X_2$ or $X_3$ produces child nodes in which $c_2$ remains the majority class. As a result, $\Delta i(s, t_0) = 0$ for all three possible splits, as if they were equally bad, or as if no further improvement could be achieved.

(b) It does not sensitively account for changes in the a posteriori class distributions $p(y|t_L)$ and $p(y|t_R)$.

In our example, none of the three splits improves training error. However, splitting on $X_1$ is likely to yield a tree which is simpler than if the root node had been split on either $X_2$ or $X_3$. As shown in Figure 3.4, splitting on $X_1$ indeed produces a right child node that is already terminal and for which no further work is required. Splitting on $X_2$ also results in the right child node to be pure. However, the left child node is in this case larger, suggesting that more work is further required. Finally, splitting on $X_3$ results in child nodes that have to be both divided further. Splitting on $X_1$ should therefore be preferred, even if it does not reduce immediately the training error.

In fact, these issues are largely due to the fact that the tree growing algorithm 3.2 is based on a one-step optimization procedure. A good impurity criterion should ideally also take into account the possibility of further improvement deeper in the tree. To avoid the undesirable properties of $i_R(t)$, the choice of $i(t)$ should therefore be guided by ensuring that $i(t)$ gets progressively smaller when $t$ gets more homogeneous towards one (or some) of the classes (while not being necessarily better in terms of misclassification error) and larger when $t$ is more heterogeneous. In CART, Breiman et al. [1984] identify a class of impurity function $i(t)$ meeting these requirements:

**Theorem 3.3.** *Let $\Phi(p_1, \dots, p_J)$ be a strictly concave $J$-ary function defined on $0 \leqslant p_k \leqslant 1$, for $k = 1, \dots, J$, $\sum_{k=1}^{J} p_k = 1$ and such that*

- $\Phi(1, \dots, 0) = \Phi(0, 1, \dots, 0) = \dots = \Phi(0, \dots, 1)$ *is minimal;*
- $\Phi(\frac{1}{J}, \dots, \frac{1}{J})$ *is maximal.*

*Then, for $i(t) = \Phi(p(c_1|t), \dots, p(c_J|t))$ and any split $s$,*

$$\Delta i(s, t) \geqslant 0,$$

*with equality if and only if $p(c_k|t_L) = p(c_k|t_R) = p(c_k|t)$ for $k = 1, \dots, J$.*

*Proof.* Let us first remark that

$$\begin{aligned} p(c_k|t) &= \frac{N_{c_k t}}{N_t} \\ &= \frac{N_{c_k t_L} + N_{c_k t_R}}{N_t} \\ &= \frac{N_{t_L}}{N_t} \frac{N_{c_k t_L}}{N_{t_L}} + \frac{N_{t_R}}{N_t} \frac{N_{c_k t_R}}{N_{t_R}} \\ &= p_L p(c_k|t_L) + p_R p(c_k|t_R) \end{aligned}$$

By strict concavity, it comes

$$\begin{aligned} i(t) &= \Phi(p(c_1|t), \dots, p(c_J|t)) \\ &= \Phi(p_L p(c_1|t_L) + p_R p(c_1|t_R), \dots, p_L p(c_J|t_L) + p_R p(c_J|t_R)) \\ &\geqslant p_L \Phi(p(c_1|t_L), \dots, p(c_J|t_L)) + p_R \Phi(p(c_1|t_R), \dots, p(c_J|t_R)) \\ &= p_L i(t_L) + p_R i(t_R) \end{aligned}$$

with equality if and only if $p(c_k|t_L) = p(c_k|t_R)$ for $k = 1, \dots, J$. □

To better understand Theorem 3.3, let us consider a strictly concave function $\Phi$ that satisfies the requirements and let us evaluate on our toy problem the goodness of the split defined on $X_1$. As illustrated in the first plot of Figure 3.5, $i(t)$ is maximum when the uncertainty on the output value is the largest (i.e., at $p(c_1|t) = p(c_2|t) = \frac{1}{2}$) and then gets progressively smaller when certainty grows towards one of the classes (i.e., as $p(c_1|t)$ gets close to 0 or 1). As shown by Theorem 3.3, $i(t)$ is also necessarily larger than the weighted sum (illustrated by the gray dot) of impurities of the child nodes. Visually, the further away $p(c_1|t_L)$ and $p(c_1|t_R)$ from $p(c_1|t)$, the larger $\Delta i(s,t)$ (illustrated by the red line) and the better the split $s$. In particular, even if $i(t_L)$ is larger than $i(t)$ on the first plot of Figure 3.5, impurity is in fact globally smaller. By contrast, when $s$ does not significantly change the a posteriori class distributions, as illustrated in the second plot of Figure 3.5 when splitting $t_0$ on $X_3$, then the closer $p(c_1|t_L)$ and $p(c_1|t_R)$ from $p(c_1|t)$ and the smaller $\Delta i(s,t)$. Yet, as long as the a posteriori class distributions do not exactly coincide with the a priori class distribution, then even the slightest changes get captured by $\Delta i(s,t)$ because of strict concavity. It is only when they exactly coincide (i.e., when $p(c_1|t_L) = p(c_1|t_R) = p(c_1|t)$) that the three red points overlap and that $\Delta i(s,t) = 0$.

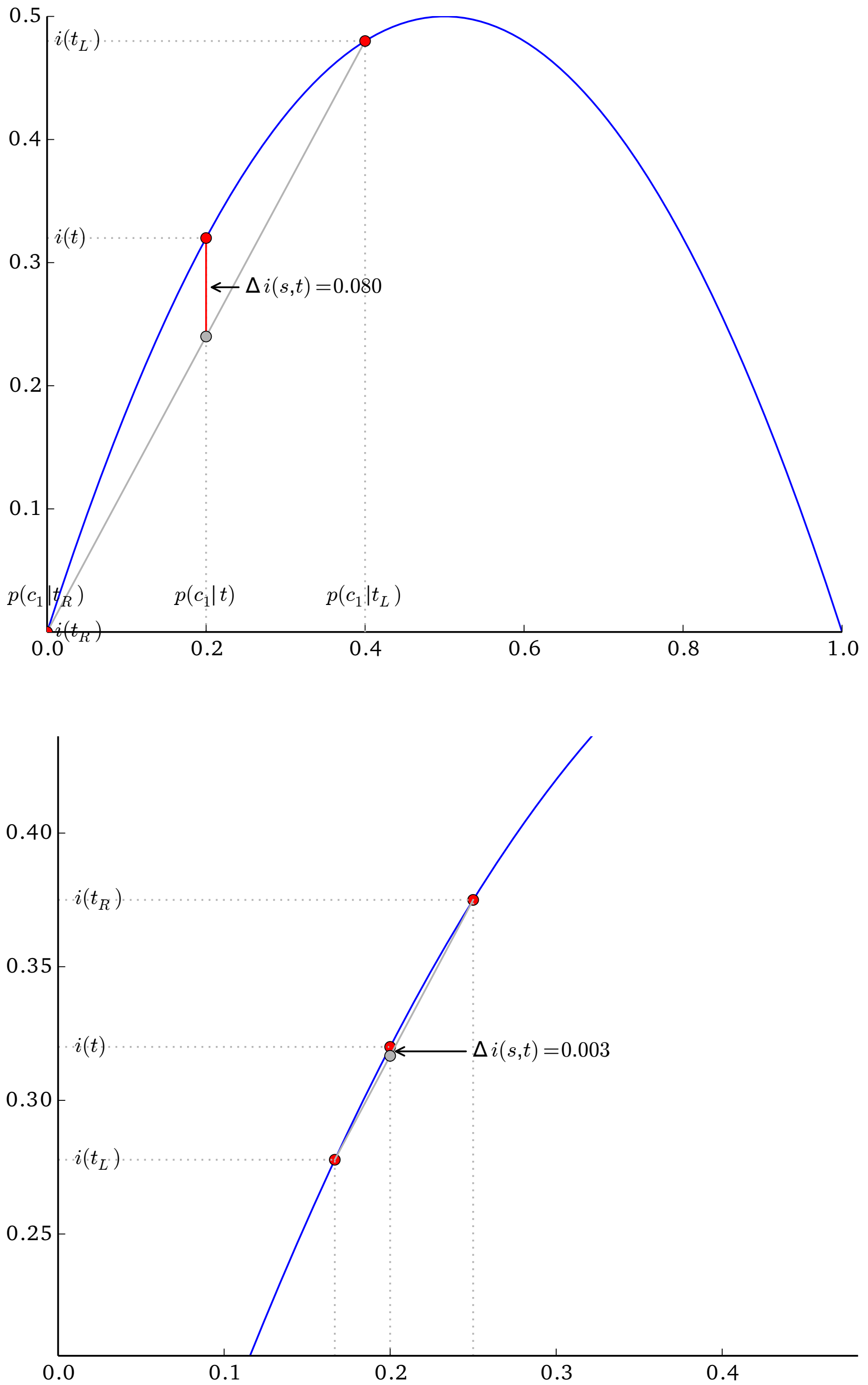

Figure 3.5: Splitting $t_0$ on $X_1$ (above) versus splitting $t_0$ on $X_3$ (below), as evaluated by a strictly concave impurity function. The figures show $i$ with respect to the local probability $p(c_1)$ of the first class.

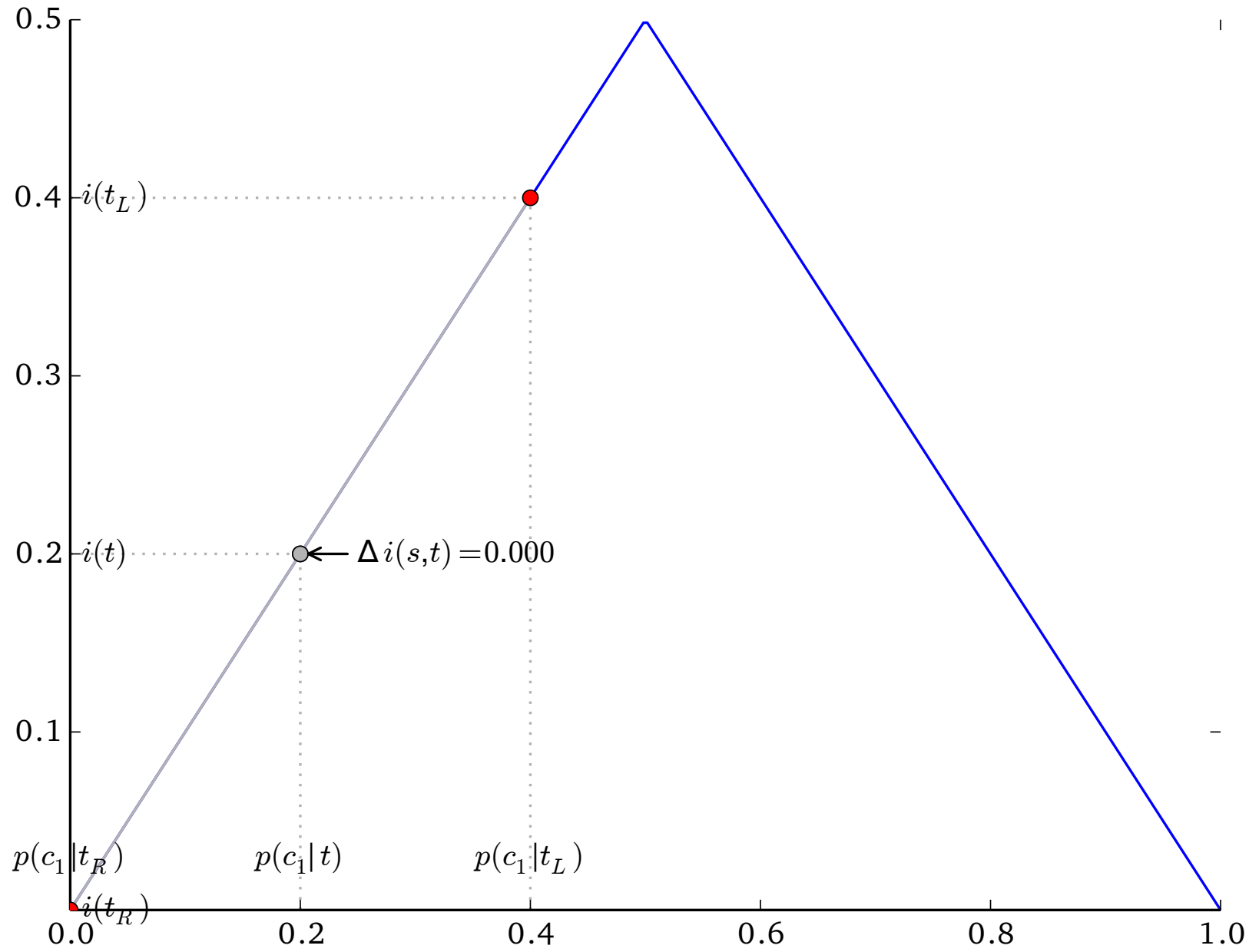

Figure 3.6: Splitting $t_0$ on $X_1$, as evaluated by the resubstitution estimate $i_R(t)$. The figure shows $i_R$ with respect to the local probability $p(c_1)$ of the first class.

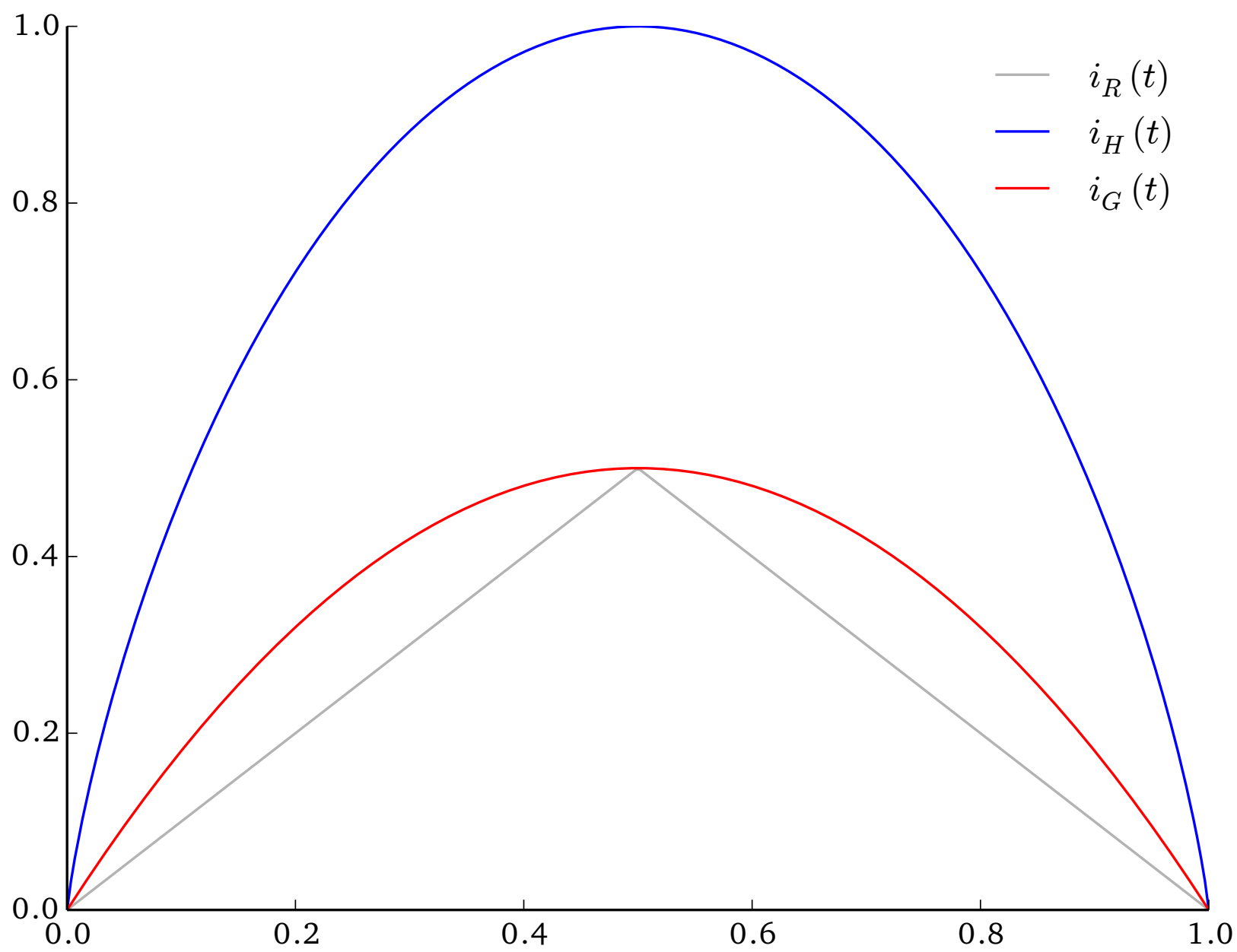

Figure 3.7: Comparison of $i_R(t)$, $i_H(t)$ and $i_G(t)$ with respect to $p(c_1|t)$.

Graphically, it also becomes obvious when comparing Figure 3.5 with Figure 3.6 why the resubstitution estimate $i_R(t)$ is not an appropriate impurity function. As long as the majority class within the child nodes $t_L$ and $t_R$ is the same as in $t$, the three red dots remain aligned in a perfect line and the gray dot overlap with $i(t)$. As a result, $\Delta i(s,t)$ is necessarily null no matter the a posteriori class distributions induced by the split, be them close or far from $p(c_1|t)$.

The most common impurity criteria used for classification trees are the Shannon entropy and the Gini index:

**Definition 3.9.** *The impurity function $i_H(t)$ based on the Shannon entropy [Shannon and Weaver, 1949] is:*

$$i_H(t) = -\sum_{k=1}^{J} p(c_k|t) \log_2(p(c_k|t)) \tag{3.20}$$

**Definition 3.10.** *The impurity function $i_G(t)$ based on the Gini index [Gini, 1912] is:*

$$i_G(t) = \sum_{k=1}^{J} p(c_k|t)(1 - p(c_k|t)) \tag{3.21}$$

As illustrated on Figure 3.7, both $i_H(t)$ and $i_G(t)$ satisfy the requirements of Theorem 3.3 and exhibit the properties that we are looking for. The entropy-based impurity $i_H(t)$ quantifies the uncertainty of $Y$ within node $t$. With this criterion, the impurity decrease $\Delta i_H(s,t)$, also known as the *information gain*, represents the information learned about $Y$ by splitting $t$ into $t_L$ and $t_R$. The Gini-based impurity $i_G(t)$ measures how often a randomly chosen object $\mathbf{x} \in \mathcal{L}_t$ would be incorrectly classified if it were randomly labeled by a class $c \in \mathcal{Y}$ according to the distribution $p(y|t)$.

While the Shannon entropy 3.20 and the Gini index 3.21 are robust and reliable impurity functions usually producing accurate decision trees, they are not exempt of defects. One of the most well-known issues is end-cut preference [Morgan and Messenger, 1973; Breiman et al., 1984], that is the tendency to favor unbalanced splits in which $p_L$ is close to zero or one, often resulting in deep and uninterpretable decision trees. Another defect is the propensity of preferring splits based on input variables with many outcomes rather than selecting input variables without bias with respect to their cardinality [Quinlan, 1986; Strobl et al., 2007a]. Intuitively, if $|\mathcal{X}_1| > |\mathcal{X}_2|$ and that both $X_1$ and $X_2$ are independent of $Y$, then $X_1$ is indeed more likely to be selected than $X_2$ because of chance. To overcome these issues, numerous variants of impurity measures have been proposed in the literature, including normalization techniques, distance-based measures or causality-based impurity functions (see [Wehenkel, 1996; Zighed and Rakotomalala, 2000; Maimon and Rokach, 2005] for reviews on the topic). From an empirical point of view however, several studies [Mingers, 1989b; Miyakawa, 1989; De Mántaras, 1991] have shown that, while the impurity function may have a significant impact on the structure of the decision tree, improvements in terms of classification accuracy are usually not that significant.

#### 3.6.2.2 *Regression*

When the output variable $Y$ is quantitative, Proposition 3.2 shows that splitting $t$ in any way reduces the squared error loss on the training set. In particular, the reduction is positive as long as the values assigned to the child nodes are different from the value at $t$. Contrary to classification, this only happens in rare circumstances (i.e., when the mean output values at the child nodes coincide with the mean output value in $t$), hence making the regression resubstitution estimate sensitive to changes in (the means of) the a posteriori distributions, even if these are only slight. As a result, the local resubstitution estimate does not exhibit the undesirable properties that we had in classification and in fact constitutes a good criterion for regression.

**Definition 3.11.** *In regression, the impurity function $i_R(t)$ based on the local resubstitution estimate defined on the squared error loss is:*

$$i_R(t) = \frac{1}{N_t} \sum_{x,y \in \mathcal{L}_t} (y - \widehat{y}_t)^2 \tag{3.22}$$

Another way of looking at criterion 3.22 is to notice that it corresponds to the within node variance of the output value in $t$. Accordingly, $s^*$ is the split that maximizes the reduction of variance $\Delta i(s,t)$ in the child nodes.

**Equivalence between classification and regression trees**

When $\mathcal{Y} = \{0, 1\}$, the learning problem can either be stated as a binary classification task where $c_1 = 0$ and $c_2 = 1$ or as a regression task where the goal is to predict the most accurate predictions $\widehat{y} \in \mathcal{Y} \subseteq \mathbb{R}$. Interestingly, regression trees built in this setting are in fact strictly equivalent to classification trees built using the Gini index $i_G(t)$. Indeed since $\widehat{y}_t = \frac{1}{N_t} \sum_{x,y \in \mathcal{L}_t} y = p(c_2|t) = 1 - p(c_1|t)$, it comes:

$$
\begin{aligned}
i_R(t) &= \frac{1}{N_t} \sum_{x,y \in \mathcal{L}_t} (y - \widehat{y}_t)^2 \\
&= \frac{1}{N_t} \sum_{x,y \in \mathcal{L}_t} y - 2y\widehat{y}_t + \widehat{y}_t^2 \\
&= \widehat{y}_t - 2\widehat{y}_t^2 + \widehat{y}_t^2 \\
&= \widehat{y}_t(1 - \widehat{y}_t) \\
&= p(c_2|t)(1 - p(c_2|t)) \\
&= \frac{1}{2} i_G(t) \qquad (3.23)
\end{aligned}
$$

From a practical point of view, this means that both classification and regression trees could be implemented using the single criterion 3.22.

### 3.6.3 *Finding the best binary split*

Now that we have described families $\mathcal{Q}$ of splitting rules and impurity criteria to evaluate their respective goodness, the last thing which is missing to have a complete specification of the induction algorithm 3.2 is an efficient optimization procedure for finding the best split $s^* \in \mathcal{Q}$. Assuming that $\mathcal{Q}$ is the set of binary univariate splits, we showed in Section 3.6.1 that $s^*$ is the best of the best binary splits $s_j^*$ defined on each input variable. This leads to the following procedure:

**Algorithm 3.3.** *Find the best split $s^*$ that partitions $\mathcal{L}_t$.*

```
1:  function FindBestSplit(L_t)
2:      Δ = −∞
3:      for j = 1, . . . , p do
4:          Find the best binary split s_j^* defined on X_j
5:          if Δi(s_j^*, t) > Δ then
6:              Δ = Δi(s_j^*, t)
7:              s^* = s_j^*
8:          end if
9:      end for
10:     return s^*
11: end function
```

Let us now discuss line 4 for ordered and categorical variables.

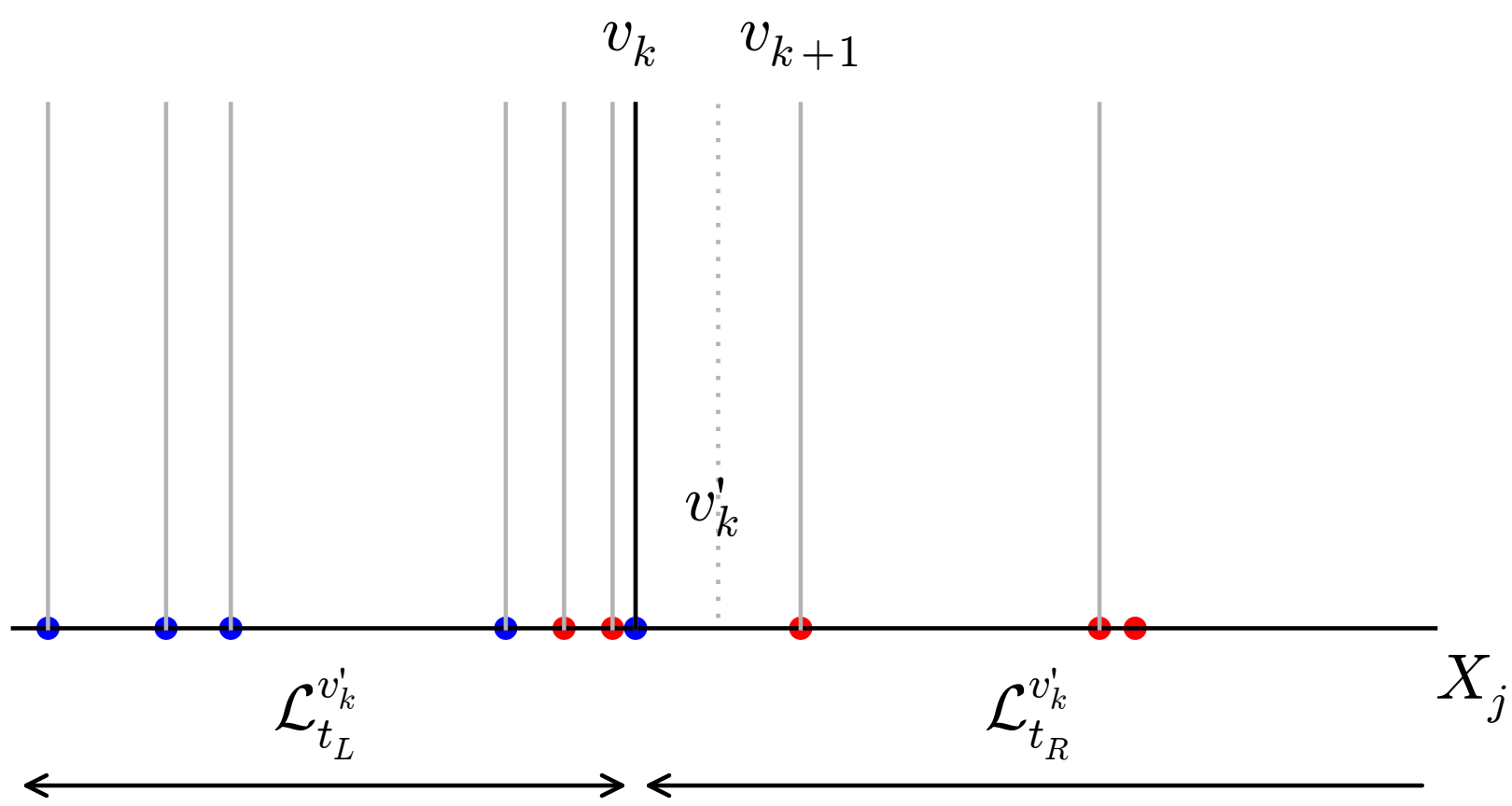

Figure 3.8: Binary partitions of $\mathcal{L}_t$ on the ordered variable $X_j$. Setting the decision threshold $v$ to any value in $[v_k; v_{k+1}[$ yields identical partitions of $\mathcal{L}_t$, but not of $\mathcal{X}_t$.

#### 3.6.3.1 *On an ordered variable*

Let $X_j$ be an ordered variable and let $\mathcal{Q}(X_j)$ be the set of all binary non-crossing partitions of $\mathcal{X}_j$, as defined in Equation 3.14. Let also $\mathcal{X}_{j|\mathcal{L}_t} = \{x_j | \mathbf{x}, y \in \mathcal{L}_t\}$ denotes the set of unique values of $X_j$ within the node samples in t. The best split $s_j^v \in \mathcal{Q}(X_j)$ on $X_j$ is the best binary partition of $\mathcal{L}_t$ into two non-empty subsets:

$$\mathcal{L}^v_{t_L} = \{(\mathbf{x}, y) | (\mathbf{x}, y) \in \mathcal{L}_t, x_j \leqslant v\}$$
$$\mathcal{L}^v_{t_R} = \{(\mathbf{x}, y) | (\mathbf{x}, y) \in \mathcal{L}_t, x_j > v\}$$

where $v$ is the decision threshold of the split. As illustrated in Figure 3.8, there exist $|\mathcal{X}_{j|\mathcal{L}_t}| - 1$ partitions of $\mathcal{L}_t$ into two such non-empty subsets, i.e., one for each value $v_k \in \mathcal{X}_{j|\mathcal{L}_t}$, except for the last one which leads to an invalid partition. In particular, if $X_j$ is locally constant in t (i.e., if all dots overlap in Figure 3.8), then $|\mathcal{X}_{j|\mathcal{L}_t}| = 1$ and $X_j$ cannot be used to partition t. More importantly, there usually exist several thresholds $v$ producing the same partition of the node samples. If $v_k$ and $v_{k+1}$ are two immediately consecutive values in $\mathcal{X}_{j|\mathcal{L}_t}$, then all splits $s_j^v$ for $v \in [v^k, v^{k+1}[$ indeed produce the same partition of $\mathcal{L}_t$ as $s_j^{v_k}$ does. In terms of impurity decrease $\Delta i$, all of them are thus equivalent when evaluated on $\mathcal{L}_t$. In generalization however, these splits may not be strictly the same since they do not produce the same partition of $\mathcal{X}_t$. As a compromise, the decision thresholds that are thus usually chosen are the mid-cut-points $v'_k = \frac{v_k + v_{k+1}}{2}$ between consecutive values of the variable, as shown by the dotted line in Figure 3.8. In practice, this is indeed a good-enough heuristic on most problems.

In this framework, the best split $s_j^*$ is the split $s_j^{v'_k}$ that maximizes impurity decrease. Computationally, the exhaustive evaluation of all these splits can be carried out efficiently when considering decision thresholds $v'_k$ in sorted order, by remarking that $\Delta i(s_j^{v'_{k+1}}, t)$ can be computed from $\Delta i(s_j^{v'_k}, t)$ in a number of operations linearly proportional the number of samples going from the right child to left child. As such, the exhaustive evaluation of all intermediate splits can be performed in linear time with respect to $N_t$, hence guaranteeing good performance as further shown in Chapter 5.

- In classification, evaluating either $\Delta i_H(s,t)$ or $\Delta i_G(s,t)$ boils down to the computation of $p(c|t)$, $p(c|t_L)$ and $p(c|t_R)$ for all classes $c \in \mathcal{Y}$. The necessary statistics for computing these probabilities are the numbers $N_{ct}$, $N_{ct_L}$ and $N_{ct_R}$ of samples of class $c$ in $t$, $t_L$ and $t_R$, along with the total numbers $N_t$, $N_{t_L}$ and $N_{t_R}$ of samples in $t$, $t_L$ and $t_R$. $N_{ct}$ and $N_t$ can be computed once for all splits. The four other statistics can be computed iteratively, from $s_j^{v'_k}$ to $s_j^{v'_{k+1}}$:

$$N_{t_L}^{v'_{k+1}} = N_{t_L}^{v'_k} + |\{(\mathbf{x},y)|(\mathbf{x},y) \in \mathcal{L}_{t_L}^{v'_{k+1}} \cap \mathcal{L}_{t_R}^{v'_k}\}| \tag{3.24}$$

$$N_{ct_L}^{v'_{k+1}} = N_{ct_L}^{v'_k} + |\{(\mathbf{x},y)|(\mathbf{x},y) \in \mathcal{L}_{t_L}^{v'_{k+1}} \cap \mathcal{L}_{t_R}^{v'_k}, y = c\}| \tag{3.25}$$

$$N_{t_R}^{v'_{k+1}} = N_{t_R}^{v'_k} - |\{(\mathbf{x},y)|(\mathbf{x},y) \in \mathcal{L}_{t_L}^{v'_{k+1}} \cap \mathcal{L}_{t_R}^{v'_k}\}| \tag{3.26}$$

$$N_{ct_R}^{v'_{k+1}} = N_{ct_R}^{v'_k} - |\{(\mathbf{x},y)|(\mathbf{x},y) \in \mathcal{L}_{t_L}^{v'_{k+1}} \cap \mathcal{L}_{t_R}^{v'_k}, y = c\}| \tag{3.27}$$

  where $\mathcal{L}_{t_L}^{v'_{k+1}} \cap \mathcal{L}_{t_R}^{v'_k}$ are the node samples going from the right child node to the left child node when switching from $s_j^{v'_k}$ to $s_j^{v'_{k+1}}$.

- In regression, $i_R(t)$ can be re-expressed as the difference between the mean of the squared output values and the square of the mean output value, that is

$$i_R(t) = \frac{\Sigma_{t^2}}{N_t} - (\frac{\Sigma_t}{N_t})^2 \tag{3.28}$$

  where $\Sigma_t = \sum_{x,y \in \mathcal{L}_t} y$ and $\Sigma_{t^2} = \sum_{x,y \in \mathcal{L}_t} y^2$. In this form, the necessary statistics for computing $\Delta i_R(s,t)$ are $\Sigma_t$, $\Sigma_{t^2}$, $\Sigma_{t_L}$, $\Sigma_{t_L^2}$, $\Sigma_{t_R}$ and $\Sigma_{t_R^2}$, along with the total numbers $N_t$, $N_{t_L}$ and $N_{t_R}$ of samples in $t$, $t_L$ and $t_R$. As in classification, $\Sigma_t$, $\Sigma_{t^2}$ and $N_t$ can be computed once for all splits, while $N_{t_L}$ and $N_{t_R}$ can be computed from one split to another using Equations 3.24 and 3.26. The four other statistics can be computed iteratively, from $s_j^{v'_k}$ to $s_j^{v'_{k+1}}$:

$$\Sigma_{t_L}^{v'_{k+1}} = \Sigma_{t_L}^{v'_k} + \sum_{(\mathbf{x},y) \in \mathcal{L}_{t_L}^{v'_{k+1}} \cap \mathcal{L}_{t_R}^{v'_k}} y \tag{3.29}$$

$$\Sigma_{t_L^2}^{v'_{k+1}} = \Sigma_{t_L^2}^{v'_k} + \sum_{(\mathbf{x},y) \in \mathcal{L}_{t_L}^{v'_{k+1}} \cap \mathcal{L}_{t_R}^{v'_k}} y^2 \tag{3.30}$$

$$\Sigma_{t_R}^{v'_{k+1}} = \Sigma_{t_R}^{v'_k} - \sum_{(\mathbf{x},y) \in \mathcal{L}_{t_L}^{v'_{k+1}} \cap \mathcal{L}_{t_R}^{v'_k}} y \tag{3.31}$$

$$\Sigma_{t_R^2}^{v'_{k+1}} = \Sigma_{t_R^2}^{v'_k} - \sum_{(\mathbf{x},y) \in \mathcal{L}_{t_L}^{v'_{k+1}} \cap \mathcal{L}_{t_R}^{v'_k}} y^2 \tag{3.32}$$

Starting from the initial partition $v'_0 = -\infty$, that is where $t_L$ is empty and $t_R$ corresponds to $t$, and using the above update equations to switch from $v'_k$ to $v'_{k+1}$, the search of best split $s_j^*$ on $X_j$ can finally be implemented as described in Algorithm 3.4 and further illustrated in Figure 3.9.

**Algorithm 3.4.** *Find the best split $s_j^*$ on $X_j$ that partitions $\mathcal{L}_t$.*

1: **function** FindBestSplit($\mathcal{L}_t$, $X_j$)
2: $\quad \Delta = 0$
3: $\quad k = 0$
4: $\quad v'_k = -\infty$
5: $\quad$ Compute the necessary statistics for $i(t)$
6: $\quad$ Initialize the statistics for $t_L$ to $0$
7: $\quad$ Initialize the statistics for $t_R$ to those of $i(t)$
8: $\quad$ Sort the node samples $\mathcal{L}_t$ such that $x_{1,j} \leqslant x_{2,j} \leqslant \cdots \leqslant x_{N_t,j}$
9: $\quad i = 1$
10: $\quad$ **while** $i \leqslant N_t$ **do**
11: $\quad\quad$ **while** $i+1 \leqslant N_t$ and $x_{i+1,j} = x_{i,j}$ **do**
12: $\quad\quad\quad i = i + 1$
13: $\quad\quad$ **end while**
14: $\quad\quad i = i + 1$
15: $\quad\quad$ **if** $i \leqslant N_t$ **then**
16: $\quad\quad\quad v'_{k+1} = \frac{x_{i,j} + x_{i-1,j}}{2}$
17: $\quad\quad\quad$ Update the necessary statistics from $v'_k$ to $v'_{k+1}$
18: $\quad\quad\quad$ **if** $\Delta i(s_j^{v'_{k+1}}, t) > \Delta$ **then**
19: $\quad\quad\quad\quad \Delta = \Delta i(s_j^{v'_{k+1}}, t)$
20: $\quad\quad\quad\quad s_j^* = s_j^{v'_{k+1}}$
21: $\quad\quad\quad$ **end if**
22: $\quad\quad\quad k = k + 1$
23: $\quad\quad$ **end if**
24: $\quad$ **end while**
25: $\quad$ **return** $s_j^*$
26: **end function**

As we will see later in Chapter 5, the algorithm complexity is upper bounded by the complexity of the sorting operation. In the context of randomized decision trees however, where finding the very best split $s_j^*$ is not as crucial, we will show that approximation techniques based on sub-sampling of the node samples or on the discretization of the variable $X_j$ can help reduce the computational complexity of procedure without significant impact on accuracy.

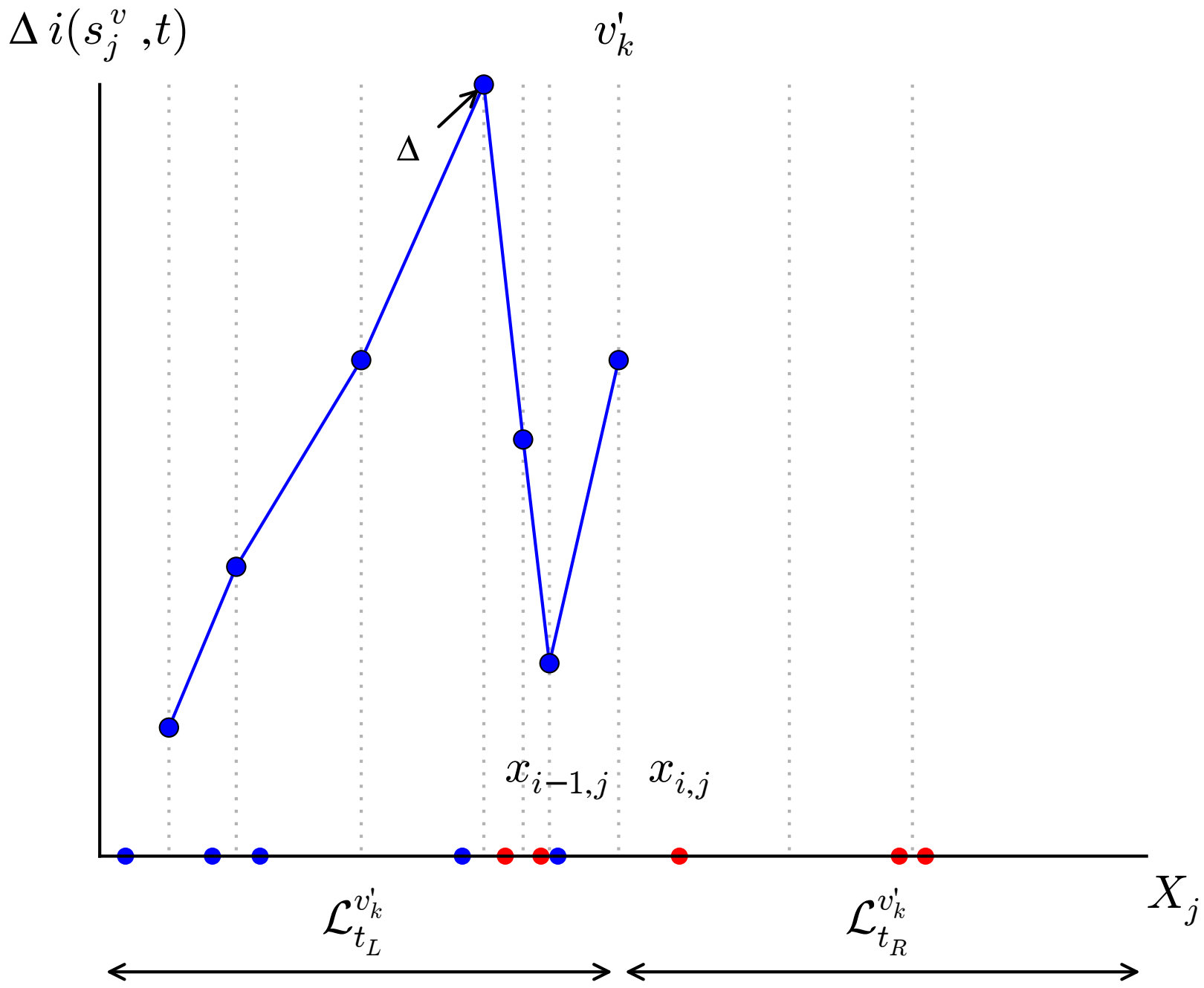

Figure 3.9: Loop invariant of Algorithm 3.4. At the end of each iteration, $\Delta i(s_j^{v'_k}, t)$ has been computed from the statistics of the previous split at $v'_{k-1}$ and compared to the best reduction in impurity $\Delta$ found so far within the splits at $v'_0 = -\infty$ to $v'_{k-1}$.

#### 3.6.3.2 *On a categorical variable*

Let $X_j$ be a categorical variable and let $\mathcal{Q}(X_j)$ be the set of all binary partitions of $\mathcal{X}_j = \{b_1, \dots, b_L\}$, as defined in Equation 3.15. The number of non-empty binary partitions in this set is $2^{L-1} - 1$, which may quickly become highly prohibitive to explore exhaustively when $X_j$ counts a high number $L$ of categories.

Fortunately, in binary classification, this exponential complexity can be reduced from looking at $2^{L-1} - 1$ partitions to $L - 1$ partitions, thanks a theoretical result due to Fisher [1958] and Breiman et al. [1984]:

**Theorem 3.4.** *Let reorder the categories of* $X_j$ *such that*

$$p(c_1|t, X_j = b_{l_1}) \leqslant p(c_1|t, X_j = b_{l_2}) \leqslant \dots \leqslant p(c_1|t, X_j = b_{l_L}).$$

*If the impurity function* $i(t)$ *satisfies the requirements of Theorem 3.3, then one of the* $L - 1$ *subsets* $\mathcal{B} = \{b_{l_1}, \dots, b_{l_h}\}, h = 1, \dots, L-1$ *defines a binary partition of the node samples into*

$$\mathcal{L}_{t_L}^{\mathcal{B}} = \{(\mathbf{x}, y) | (\mathbf{x}, y) \in \mathcal{L}_t, x_j \in \mathcal{B}\}$$
$$\mathcal{L}_{t_R}^{\mathcal{B}} = \{(\mathbf{x}, y) | (\mathbf{x}, y) \in \mathcal{L}_t, x_j \in \overline{\mathcal{B}}\}$$

*which maximizes* $\Delta i(s, t)$, *for* $s \in \mathcal{Q}(X_j)$.

Intuitively, this result indicates that all those categories $b_l$ leading to high probabilities of being of class $c_1$ should be put together into one node and the categories leading to lower probabilities into the other. In practice, this also means that the search of the best binary split can be performed as if $X_j$ was an ordered variable, by virtually replacing categorical values $b_l$ with $p(c_1|t, X_j = b_l)$ and using Algorithm 3.4 to find the best binary split on these ordered values.

Unfortunately, Theorem 3.4 does not extend to multi-class classification (i.e., when $J > 2$). In this case, an exhaustive exploration of all $2^{L-1} - 1$ splits must be performed in order to find the best split. When the number $L$ of categories is small, this can usually be done within reasonable time. However, as $L$ gets larger, this quickly becomes infeasible in practice. As an approximation, the usual strategy [Liaw and Wiener, 2002] consists in exhaustively evaluating all splits when $L \leqslant 10$ and then switch to random sampling of the subsets $\mathcal{B} \subset \{b_1, \dots, b_L\}$ when $L > 10$. In that case, $2^8$ random subsets $\mathcal{B}$ are typically drawn and evaluated.

In regression, given the equivalence of $i_R(t)$ with the Gini index, conditions of Theorem 3.4 are met and the same algorithmic trick can be applied as in binary classification to compute the optimal split. In this case, $X_j$ is transformed into an ordered variable by virtually replacing categorical values $b_l$ with the mean output value at $b_l$, i.e., $\frac{1}{N_{lt}} \sum_{\mathbf{x},y \in \mathcal{L}_t, x_j = b_l} y$, on which Algorithm 3.4 is then applied to find the best binary split on these ordered values.

## 3.7 MULTI-OUTPUT DECISION TREES

Up to now, we have assumed that the response was modeled as a single target variable $Y$. In some contexts however, the goal may be to predict several output values $\mathbf{y} = (y_1, \dots, y_q) \in \mathcal{Y}_1 \times \dots \times \mathcal{Y}_q$ for the same input $\mathbf{x}$. A famous example of such multi-objective task is multi-label classification (see [Tsoumakas and Katakis, 2007]), in which a sample $\mathbf{x}$ can be assigned to multiple binary labels $Y_k$ (for $k = 1, \dots, q$). Another example is image annotation, in which the goal is to annotate (i.e., classify) every pixel of an image in order to identify regions or objects (see [Zhang et al., 2012]).

The most straightforward strategy to solve a multi-objective task is to build an independent and separate model for each of the $q$ output variables $Y_k$. Distinctively, decision trees offer an alternative strategy which consists in learning a single *multi-output* model, capable of predicting all output variables at once. Algorithmically, *multi-output decision trees* closely relate to predictive clustering trees [Blockeel et al., 2000], output kernel trees [Geurts et al., 2006b] or multi-objective decision trees [Kocev et al., 2007], and can easily be implemented on the basis of the induction procedure developed in this chapter, provided the following two changes:

- Leaves are labeled with output vectors $\mathbf{y}_t = (y_{t,1}, \dots, y_{t,q})$, where each $y_{t,k}$ (for $k = 1, \dots, q$) is computed using assign-ment rule 3.7 in classification (resp. assignment rule 3.10 in regression), as previously done in single output decision trees.

- The impurity decrease of a split is computed as the average impurity decrease over the q output variables. Accordingly, splits are optimized with respect to all output variables, thereby exploiting correlations between $Y_1, \dots, Y_q$ whenever possible.

The main advantage of multi-output decision trees is that dependencies between output variables can be taken into account, whereas building q individual models cannot exploit such correlations, which may affect the overall accuracy of the model. A second advantage is that building a single model is often less computationally expensive than building q different models, both from a time and space complexity point of view.

# 4 RANDOM FORESTS

**OUTLINE**

In this chapter, we present the well-known family of *random forests* methods. In Section 4.1, we first describe the bias-variance decomposition of the prediction error and then present, in Section 4.2, how aggregating randomized models through ensembles reduces the prediction error by decreasing the variance term in this decomposition. In Section 4.3, we revisit random forests and its variants and study how randomness introduced into the decision trees reduces prediction errors by decorrelating the decision trees in the ensemble. Properties and features of random forests are then outlined in Section 4.4 while their consistency is finally explored in Section 4.5.

## 4.1 BIAS-VARIANCE DECOMPOSITION

In section 2.2, we defined the generalization error of a model $\varphi_{\mathcal{L}}$ as its expected prediction error according to some loss function L

$$\text{Err}(\varphi_{\mathcal{L}}) = \mathbb{E}_{X,Y}\{L(Y, \varphi_{\mathcal{L}}(X))\}. \tag{4.1}$$

Similarly, the expected prediction error of $\varphi_{\mathcal{L}}$ at $X = \mathbf{x}$ can be expressed as

$$\text{Err}(\varphi_{\mathcal{L}}(\mathbf{x})) = \mathbb{E}_{Y|X=\mathbf{x}}\{L(Y, \varphi_{\mathcal{L}}(\mathbf{x}))\}. \tag{4.2}$$

In regression, for the squared error loss, this latter form of the expected prediction error additively decomposes into bias and variance terms which together constitute a very useful framework for diagnosing the prediction error of a model. In classification, for the zero-one loss, a similar decomposition is more difficult to obtain. Yet, the concepts of bias and variance can be transposed in several ways to classification, thereby providing comparable frameworks for studying the prediction error of classifiers.

### 4.1.1 *Regression*

In regression, assuming that L is the squared error loss, the expected prediction error of a model $\varphi_{\mathcal{L}}$ at a given point $X = \mathbf{x}$ can be rewritten with respect to the Bayes model $\varphi_B$:

$$\text{Err}(\varphi_{\mathcal{L}}(\mathbf{x}))$$

$$
\begin{aligned}
&= \mathbb{E}_{Y|X=\mathbf{x}}\{(Y - \varphi_{\mathcal{L}}(\mathbf{x}))^2\} \\
&= \mathbb{E}_{Y|X=\mathbf{x}}\{(Y - \varphi_B(\mathbf{x}) + \varphi_B(\mathbf{x}) - \varphi_{\mathcal{L}}(\mathbf{x}))^2\} \\
&= \mathbb{E}_{Y|X=\mathbf{x}}\{(Y - \varphi_B(\mathbf{x}))^2\} + \mathbb{E}_{Y|X=\mathbf{x}}\{(\varphi_B(\mathbf{x}) - \varphi_{\mathcal{L}}(\mathbf{x}))^2\} \\
&\hookrightarrow + \mathbb{E}_{Y|X=\mathbf{x}}\{2(Y - \varphi_B(\mathbf{x}))(\varphi_B(\mathbf{x}) - \varphi_{\mathcal{L}}(\mathbf{x}))\} \\
&= \mathbb{E}_{Y|X=\mathbf{x}}\{(Y - \varphi_B(\mathbf{x}))^2\} + \mathbb{E}_{Y|X=\mathbf{x}}\{(\varphi_B(\mathbf{x}) - \varphi_{\mathcal{L}}(\mathbf{x}))^2\} \\
&= \text{Err}(\varphi_B(\mathbf{x})) + (\varphi_B(\mathbf{x}) - \varphi_{\mathcal{L}}(\mathbf{x}))^2
\end{aligned} \tag{4.3}
$$

since $\mathbb{E}_{Y|X=\mathbf{x}}\{Y - \varphi_B(\mathbf{x})\} = \mathbb{E}_{Y|X=\mathbf{x}}\{Y\} - \varphi_B(\mathbf{x}) = 0$ by definition of the Bayes model in regression. In this form, the first term in the last expression of Equation 4.3 corresponds to the (irreducible) residual error at $X = \mathbf{x}$ while the second term represents the discrepancy of $\varphi_{\mathcal{L}}$ from the Bayes model. The farther from the Bayes model, the more sub-optimal the model and the larger the error.

If we further assume that the learning set $\mathcal{L}$ is itself a random variable (sampled from the population $\Omega$) and that the learning algorithm is deterministic, then the expected discrepancy over $\mathcal{L}$ with the Bayes model can further be re-expressed in terms of the average prediction $\mathbb{E}_{\mathcal{L}}\{\varphi_{\mathcal{L}}(\mathbf{x})\}$ over the models learned from all possible learning sets of size $N$:

$$
\begin{aligned}
&\mathbb{E}_{\mathcal{L}}\{(\varphi_B(\mathbf{x}) - \varphi_{\mathcal{L}}(\mathbf{x}))^2\} \\
&= \mathbb{E}_{\mathcal{L}}\{(\varphi_B(\mathbf{x}) - \mathbb{E}_{\mathcal{L}}\{\varphi_{\mathcal{L}}(\mathbf{x})\} + \mathbb{E}_{\mathcal{L}}\{\varphi_{\mathcal{L}}(\mathbf{x})\} - \varphi_{\mathcal{L}}(\mathbf{x}))^2\} \\
&= \mathbb{E}_{\mathcal{L}}\{(\varphi_B(\mathbf{x}) - \mathbb{E}_{\mathcal{L}}\{\varphi_{\mathcal{L}}(\mathbf{x})\})^2\} + \mathbb{E}_{\mathcal{L}}\{(\mathbb{E}_{\mathcal{L}}\{\varphi_{\mathcal{L}}(\mathbf{x})\} - \varphi_{\mathcal{L}}(\mathbf{x}))^2\} \\
&\hookrightarrow + \mathbb{E}_{\mathcal{L}}\{2(\varphi_B(\mathbf{x}) - \mathbb{E}_{\mathcal{L}}\{\varphi_{\mathcal{L}}(\mathbf{x})\})(\mathbb{E}_{\mathcal{L}}\{\varphi_{\mathcal{L}}(\mathbf{x})\} - \varphi_{\mathcal{L}}(\mathbf{x}))\} \\
&= \mathbb{E}_{\mathcal{L}}\{(\varphi_B(\mathbf{x}) - \mathbb{E}_{\mathcal{L}}\{\varphi_{\mathcal{L}}(\mathbf{x})\})^2\} + \mathbb{E}_{\mathcal{L}}\{(\mathbb{E}_{\mathcal{L}}\{\varphi_{\mathcal{L}}(\mathbf{x})\} - \varphi_{\mathcal{L}}(\mathbf{x}))^2\} \\
&= (\varphi_B(\mathbf{x}) - \mathbb{E}_{\mathcal{L}}\{\varphi_{\mathcal{L}}(\mathbf{x})\})^2 + \mathbb{E}_{\mathcal{L}}\{(\mathbb{E}_{\mathcal{L}}\{\varphi_{\mathcal{L}}(\mathbf{x})\} - \varphi_{\mathcal{L}}(\mathbf{x}))^2\}
\end{aligned} \tag{4.4}
$$

since $\mathbb{E}_{\mathcal{L}}\{\mathbb{E}_{\mathcal{L}}\{\varphi_{\mathcal{L}}(\mathbf{x})\} - \varphi_{\mathcal{L}}(\mathbf{x})\} = \mathbb{E}_{\mathcal{L}}\{\varphi_{\mathcal{L}}(\mathbf{x})\} - \mathbb{E}_{\mathcal{L}}\{\varphi_{\mathcal{L}}(\mathbf{x})\} = 0$. In summary, the expected generalization error additively decomposes as formulated in Theorem 4.1.

**Theorem 4.1.** *For the squared error loss, the bias-variance decomposition of the expected generalization error* $\mathbb{E}_{\mathcal{L}}\{\text{Err}(\varphi_{\mathcal{L}}(\mathbf{x}))\}$ *at* $X = \mathbf{x}$ *is*

$$
\mathbb{E}_{\mathcal{L}}\{\text{Err}(\varphi_{\mathcal{L}}(\mathbf{x}))\} = \textit{noise}(\mathbf{x}) + \textit{bias}^2(\mathbf{x}) + \textit{var}(\mathbf{x}), \tag{4.5}
$$

*where*

$$
\begin{aligned}
\textit{noise}(\mathbf{x}) &= \text{Err}(\varphi_B(\mathbf{x})), \\
\textit{bias}^2(\mathbf{x}) &= (\varphi_B(\mathbf{x}) - \mathbb{E}_{\mathcal{L}}\{\varphi_{\mathcal{L}}(\mathbf{x})\})^2, \\
\textit{var}(\mathbf{x}) &= \mathbb{E}_{\mathcal{L}}\{(\mathbb{E}_{\mathcal{L}}\{\varphi_{\mathcal{L}}(\mathbf{x})\} - \varphi_{\mathcal{L}}(\mathbf{x}))^2\}.
\end{aligned}
$$

This bias-variance decomposition of the generalization error is due to Geman et al. [1992] and was first proposed in the context of neural networks. The first term, noise$(\mathbf{x})$, is the residual error. It is entirely independent of both the learning algorithm and the learning set and provides for any model a theoretical lower bound on its generalization error. The second term, $\text{bias}^2(\mathbf{x})$, measures the discrepancy between the average prediction and the prediction of the Bayes model. Finally, the third term, $\text{var}(\mathbf{x})$, measures the variability of the predictions at $X = \mathbf{x}$ over the models learned from all possible learning sets. All three terms are illustrated in Figure 4.1 for a toy and artificial regression problem. Both $\text{noise}(\mathbf{x})$ and $\text{var}(\mathbf{x})$ measures the spread of the two densities while $\text{bias}^2(\mathbf{x})$ is the distance between their means.

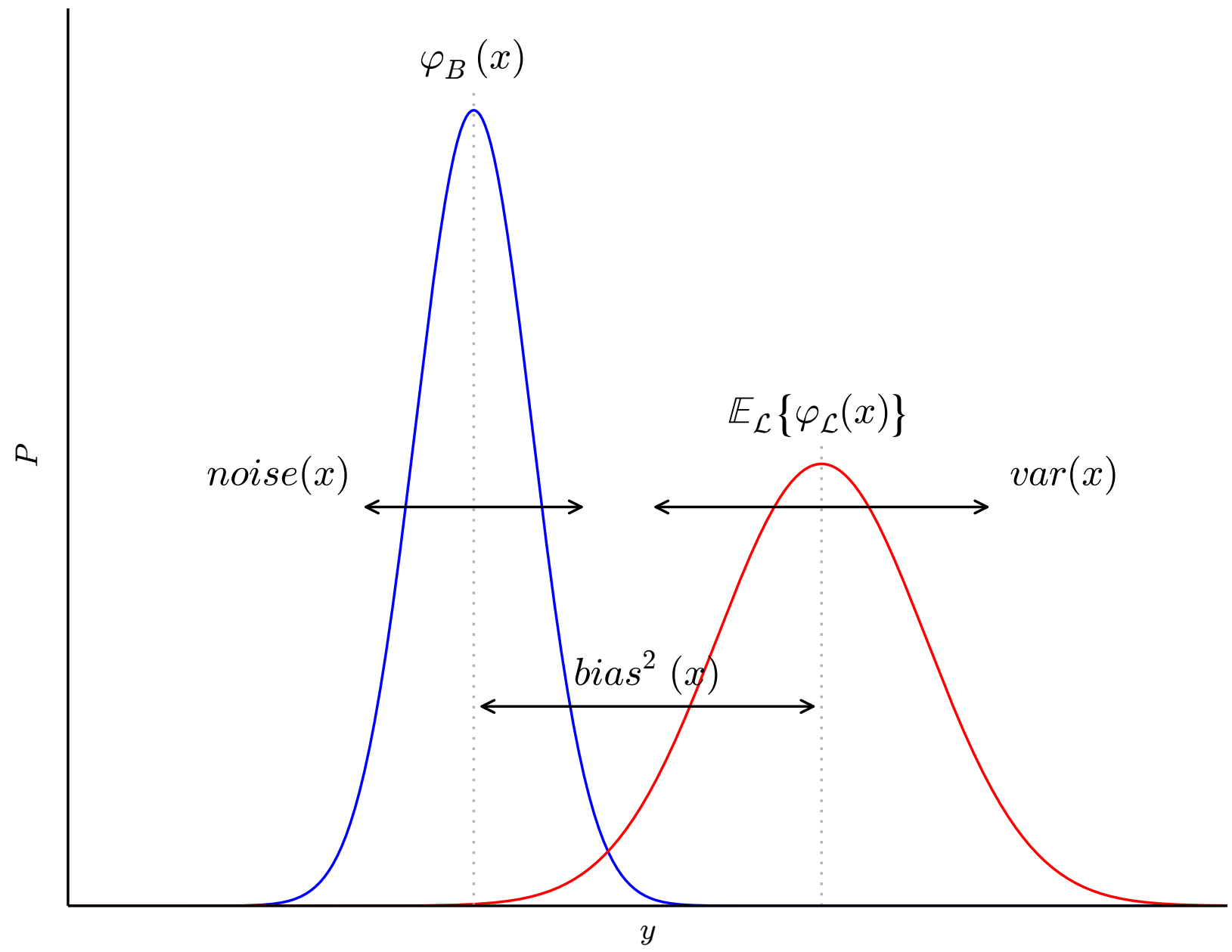

Figure 4.1: Residual error, bias and variance at $X = \mathbf{x}$. (Figure inspired from [Geurts, 2002].)

As a typical example, the bias-variance decomposition framework can be used as a tool for diagnosing underfitting and overfitting (as previously introduced in Section 2.3). The upper plots in Figure 4.2 illustrate in light red predictions $\varphi_{\mathcal{L}}(\mathbf{x})$ for polynomials of degree 1, 5 and 15 learned over random learning sets $\mathcal{L}$ sampled from a noisy cosine function. Predictions $\mathbb{E}_{\mathcal{L}}\{\varphi_{\mathcal{L}}(\mathbf{x})\}$ of the average model are represented by the thick red lines. Predictions for the model learned over the learning set, represented by the blue dots, are represented in gray. Predictions of the Bayes model are shown by blue lines and coincide with the unnoised cosine function that defines the regression problem. The lower plots in the figure illustrate the bias-variance decomposition of the expected generalization error of the polynomials.

Clearly, polynomials of degree 1 (left) suffer from underfitting. In terms of bias and variance, this translates into low variance but high bias as shown in the lower left plot of Figure 4.2. Indeed, due to the low degree of the polynomials (i.e., due to the low model complexity), the resulting models are almost all identical and the variability of the predictions from one model to another is therefore quite low.

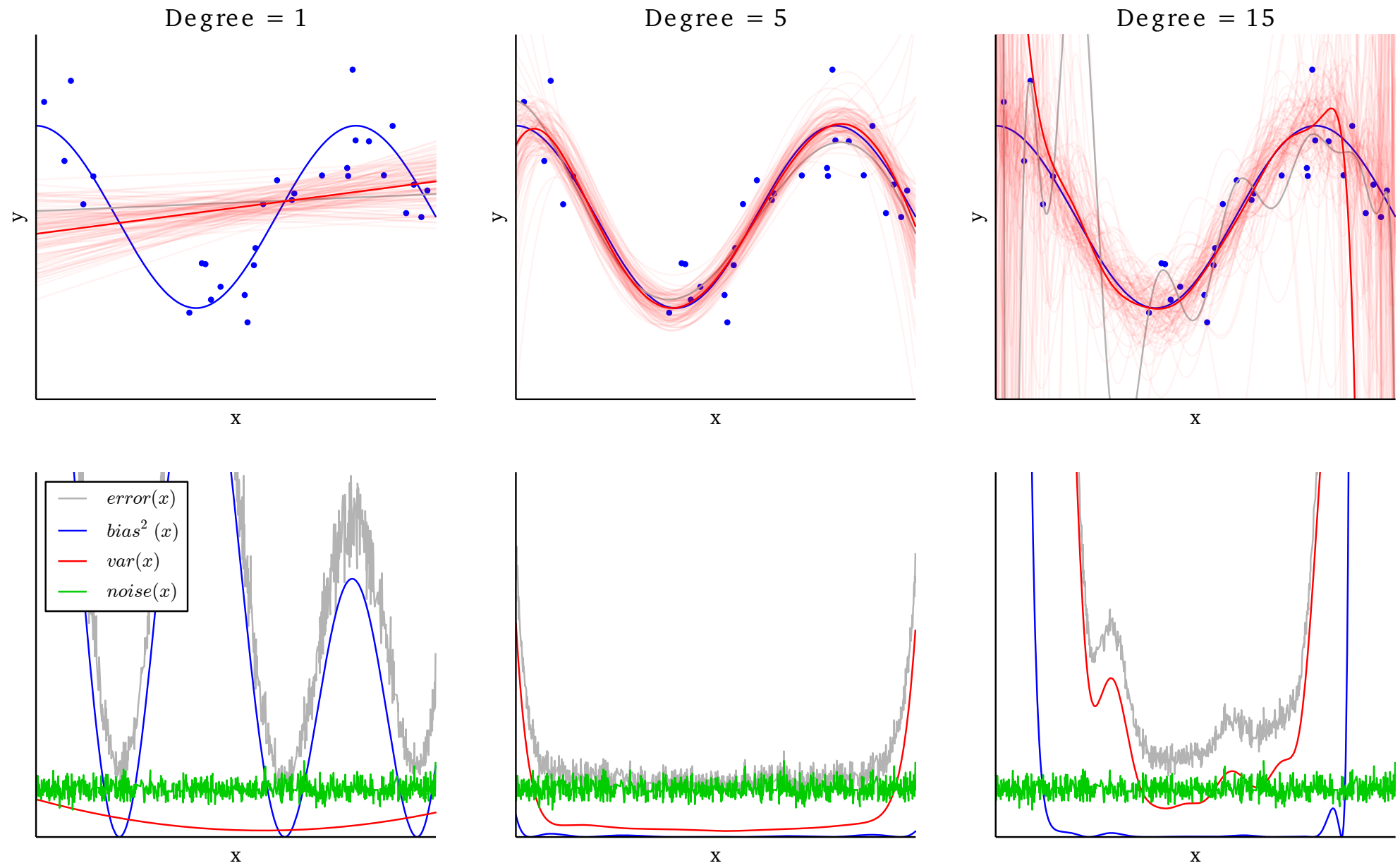

Figure 4.2: Bias-variance decomposition of the expected generalization error for polynomials of degree 1, 5 and 15.

Also, because of low complexity, none of them really fits the trend of the training points, even approximately, which implies that the average model is far from approximating the Bayes model. This results in high bias. On the other hand, polynomials of degree 15 (right) suffer from overfitting. In terms of bias and variance, the situation is the opposite. Predictions have low bias but high variance, as shown in the lower right plot of Figure 4.2. The variability of the predictions is large because the high degree of the polynomials (i.e., the high model complexity) captures noise in the learning set. Indeed, compare the gray line with the blue dots – they almost all intersect. Put otherwise, small changes in the learning set result in large changes in the obtained model and therefore in its predictions. By contrast, the average model is now quite close from the Bayes model, which results in low bias[1]. Finally, polynomials of degree 5 (middle) are neither too simple nor too complex. In terms of bias and variance, the trade-off is well-balanced between the two extreme situations. Bias and variance are neither too low nor too large.

### 4.1.2 *Classification*

In direct analogy with the bias-variance decomposition for the squared error loss, similar decompositions have been proposed in the literature for the expected generalization error based on the zero-one loss, i.e., for $\mathbb{E}_{\mathcal{L}}\{\mathbb{E}_{Y|X=\mathbf{x}}\{1(\varphi_{\mathcal{L}}(x) \neq Y)\}\} = P_{\mathcal{L},Y|X=\mathbf{x}}(\varphi_{\mathcal{L}}(\mathbf{x}) \neq Y)$. Most no-

1 Note however the Gibbs-like phenomenon resulting in both high variance and high bias at the boundaries of $\mathcal{X}$.

tably, Dietterich and Kong [1995], Breiman [1996], Kohavi et al. [1996], Tibshirani [1996] and Domingos [2000] have all developed additive decompositions similar to Theorem 4.1 by redefining the concepts of bias and variance in the case of classification. While these efforts have all provided useful insight into the nature of classification error, none of them really have provided a seductively as simple and satisfactory framework as in regression (for reviews, see [Friedman, 1997; James, 2003; Geurts, 2005]).

An interesting connection with Theorem 4.1 however is to remark that classification algorithms usually work by computing estimates

$$\widehat{p}_{\mathcal{L}}(Y = c | X = \mathbf{x}) \tag{4.6}$$

of the conditional class probability (e.g., $\widehat{p}_{\mathcal{L}}(Y = c|X = \mathbf{x}) = p(c|t)$ in decision trees, as defined in Section 3.4) and then deriving a classification rule by predicting the class that maximizes this estimate, that is:

$$\varphi_{\mathcal{L}}(\mathbf{x}) = \arg\max_{c \in \mathcal{Y}} \widehat{p}_{\mathcal{L}}(Y = c|X = \mathbf{x}) \tag{4.7}$$

As such, a direction for studying classification models is to relate the bias-variance decomposition of these numerical estimates to the expected misclassification error of classification rule 4.7.

We now reproduce the results of Friedman [1997] who made this connection explicit for the case of binary classification. Let us first decompose the expected classification error into an irreducible part associated with the random nature of the output $Y$ and a reducible part that depends on $\varphi_{\mathcal{L}}(\mathbf{x})$, in analogy with Equation 4.3 for the squared error loss. (Note that, to simplify notations, we assume that all probabilities based on the random variable $Y$ is with respect to the distribution of $Y$ at $X = \mathbf{x}$.)

$$\begin{aligned}
&\mathbb{E}_{\mathcal{L}}\{\mathbb{E}_{Y|X=\mathbf{x}}\{1(\varphi_{\mathcal{L}}(\mathbf{x}) \neq Y)\}\} \\
&= P_{\mathcal{L}}(\varphi_{\mathcal{L}}(\mathbf{x}) \neq Y) \\
&= 1 - P_{\mathcal{L}}(\varphi_{\mathcal{L}}(\mathbf{x}) = Y) \\
&= 1 - P_{\mathcal{L}}(\varphi_{\mathcal{L}}(\mathbf{x}) = \varphi_B(\mathbf{x}))P(\varphi_B(\mathbf{x}) = Y) \\
&\quad\; - P_{\mathcal{L}}(\varphi_{\mathcal{L}}(\mathbf{x}) \neq \varphi_B(\mathbf{x}))P(\varphi_B(\mathbf{x}) \neq Y) \\
&= P(\varphi_B(\mathbf{x}) \neq Y) + P_{\mathcal{L}}(\varphi_{\mathcal{L}}(\mathbf{x}) \neq \varphi_B(\mathbf{x})) \\
&\quad\; - 2P_{\mathcal{L}}(\varphi_{\mathcal{L}}(\mathbf{x}) \neq \varphi_B(\mathbf{x}))P(\varphi_B(\mathbf{x}) \neq Y) \\
&= P(\varphi_B(\mathbf{x}) \neq Y) + P_{\mathcal{L}}(\varphi_{\mathcal{L}}(\mathbf{x}) \neq \varphi_B(\mathbf{x}))(2P(\varphi_B(\mathbf{x}) = Y) - 1)
\end{aligned} \tag{4.8}$$

In this form, the first term is the irreducible error of the Bayes model. The second term is the increased error due to the misestimation of the optimal decision boundary. The probability $P_{\mathcal{L}}(\varphi_{\mathcal{L}}(\mathbf{x}) \neq \varphi_B(\mathbf{x}))$ is the probability for the model of making a decision which is different from the decision of the Bayes model. This happens when the estimate $\widehat{p}_{\mathcal{L}}(Y = \varphi_B(\mathbf{x}))$ is lower than 0.5, that is:

$$P_{\mathcal{L}}(\varphi_{\mathcal{L}}(\mathbf{x}) \neq \varphi_B(\mathbf{x})) = P_{\mathcal{L}}(\widehat{p}_{\mathcal{L}}(Y = \varphi_B(\mathbf{x})) < 0.5) \tag{4.9}$$

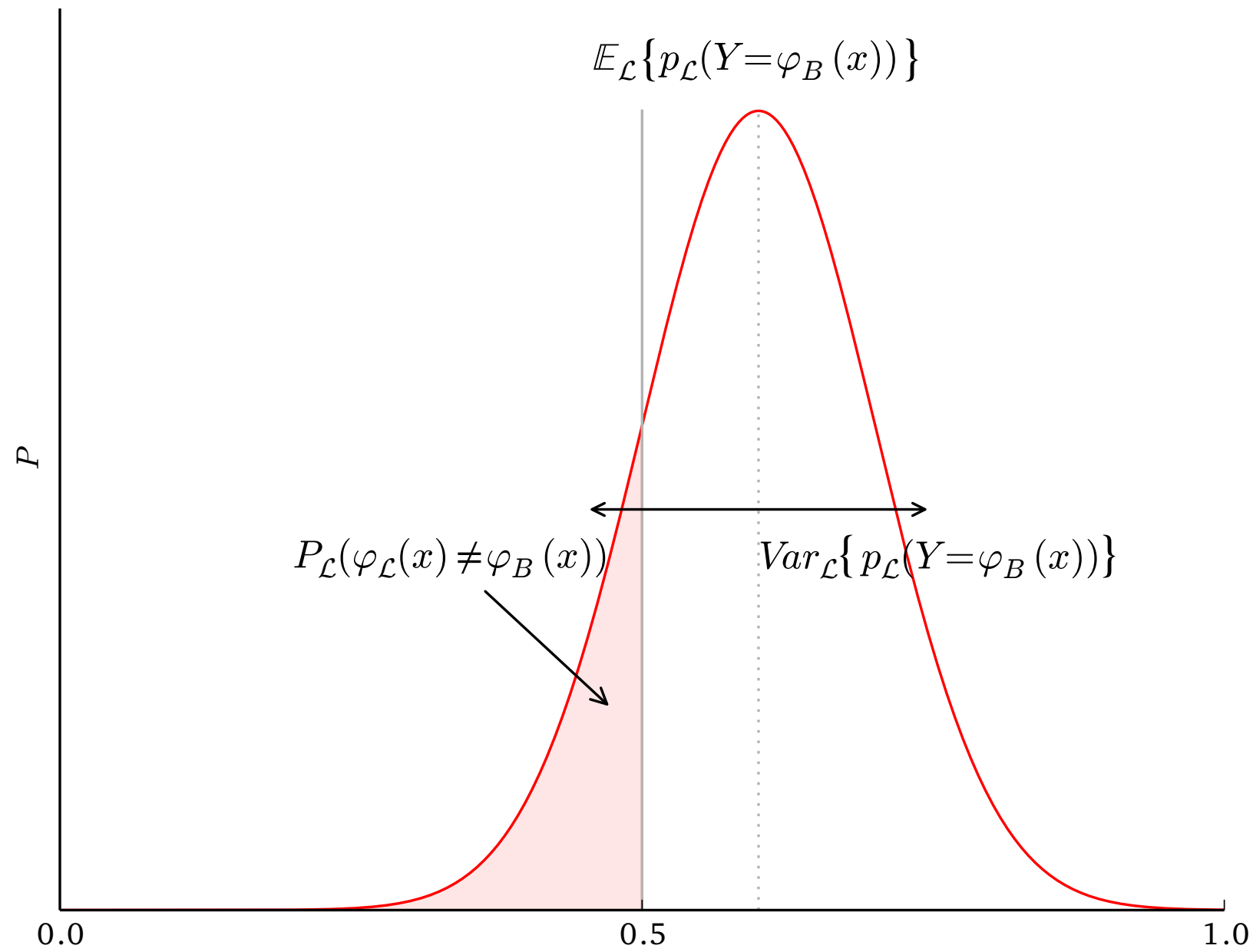

Figure 4.3: Probability distribution of the estimate $\widehat{p}_{\mathcal{L}}(Y = \varphi_B(\mathbf{x}))$.

As Figure 4.3 illustrates, probability 4.9 in fact corresponds to the tail area on the left side of the decision threshold (at 0.5) of the distribution of the estimate.

If we now further assume[2] that the estimate $\widehat{p}_{\mathcal{L}}(Y = \varphi_B(\mathbf{x}))$ is normally distributed, then probability 4.9 can be computed explicitly from its mean and variance:

$$P_{\mathcal{L}}(\widehat{p}_{\mathcal{L}}(Y = \varphi_B(\mathbf{x})) < 0.5) = \Phi(\frac{0.5 - \mathbb{E}_{\mathcal{L}}\{\widehat{p}_{\mathcal{L}}(Y = \varphi_B(\mathbf{x}))\}}{\sqrt{\mathbb{V}_{\mathcal{L}}\{\widehat{p}_{\mathcal{L}}(Y = \varphi_B(\mathbf{x}))\}}}) \tag{4.10}$$

where $\Phi(x) = \frac{1}{\sqrt{2\pi}} \int_{-\infty}^{x} \exp(-\frac{t^2}{2}) dt$ is the cumulative distribution function of the standard normal distribution. In summary, the expected generalization error additively decomposes as formulated in Theorem 4.2.

**Theorem 4.2.** *For the zero-one loss and binary classification, the expected generalization error* $\mathbb{E}_{\mathcal{L}}\{\text{Err}(\varphi_{\mathcal{L}}(\mathbf{x}))\}$ *at* $X = \mathbf{x}$ *decomposes as follows:*

$$\begin{aligned} \mathbb{E}_{\mathcal{L}}\{\text{Err}(\varphi_{\mathcal{L}}(\mathbf{x}))\} &= P(\varphi_B(\mathbf{x}) \neq Y) \\ &+ \Phi(\frac{0.5 - \mathbb{E}_{\mathcal{L}}\{\widehat{p}_{\mathcal{L}}(Y = \varphi_B(\mathbf{x}))\}}{\sqrt{\mathbb{V}_{\mathcal{L}}\{\widehat{p}_{\mathcal{L}}(Y = \varphi_B(\mathbf{x}))\}}})(2P(\varphi_B(\mathbf{x}) = Y) - 1) \end{aligned} \tag{4.11}$$

2 For single decision trees, the normal assumption is certainly not satisfied in all cases, but the qualitative conclusions are still generally valid. When the computations of the estimates involve some averaging process, e.g., as further developed in the case of ensemble of randomized trees, this approximation is however fairly reasonable.

As a result, Theorem 4.2 establishes a direct connection between the regression variance of the estimates and the classification error of the resulting model. In practice, this decomposition has important consequences:

- When the expected probability estimate $\mathbb{E}_{\mathcal{L}}\{\widehat{p}_{\mathcal{L}}(Y = \varphi_B(\mathbf{x})\}$ for the true majority class is greater than 0.5, a reduction of variance of the estimate results in a decrease of the total misclassification error. If $\mathbb{V}_{\mathcal{L}}\{\widehat{p}_{\mathcal{L}}(Y = \varphi_B(\mathbf{x}))\} \to 0$, then $\Phi \to 0$ and the expected generalization error tends to the error of the Bayes model. In particular, the generalization error can be driven to its minimum value whatever the regression bias of the estimate (at least as long as $\mathbb{E}_{\mathcal{L}}\{\widehat{p}_{\mathcal{L}}(Y = \varphi_B(\mathbf{x})\} > 0.5$).
- Conversely, when $\mathbb{E}_{\mathcal{L}}\{\widehat{p}_{\mathcal{L}}(Y = \varphi_B(\mathbf{x})\} < 0.5$, a decrease of variance might actually increase the total misclassification error. If $\mathbb{V}_{\mathcal{L}}\{\widehat{p}_{\mathcal{L}}(Y = \varphi_B(\mathbf{x}))\} \to 0$, then $\Phi \to 1$ and the error is maximal.

## 4.2 ENSEMBLE METHODS BASED ON RANDOMIZATION

Both theorems 4.1 and 4.2 reveal the role of variance in the expected generalization error of a model. In light of these results, a sensible approach for reducing generalization error would therefore consist in driving down the prediction variance, provided the respective bias can be kept the same or not be increased too much.

As it happens, *ensemble methods* constitute a beautifully simple way to do just that. Specifically, the core principle of ensemble methods based on randomization is to introduce random perturbations into the learning procedure in order to produce several different models from a single learning set $\mathcal{L}$ and then to combine the predictions of those models to form the prediction of the ensemble. How predictions are combined and why does it help is formally studied in the next sections.

### 4.2.1 *Randomized models*

Given a learning set $\mathcal{L}$, a learning algorithm $\mathcal{A}$ deterministically produces a model $\mathcal{A}(\theta, \mathcal{L})$, denoted $\varphi_{\mathcal{L},\theta}$, where $\theta$ are hyper-parameters controlling the execution of $\mathcal{A}$. Let us assume that $\theta$ includes a random seed parameter for mimicking some stochastic behavior in $\mathcal{A}$, hence producing (pseudo-)randomized models that are more or less different from one random seed to another. (We defer the discussion on specific random perturbations in the case of decision trees to Section 4.3.)

In this context, the bias-variance decomposition can be extended to account for everything that is random, hence considering both $\mathcal{L}$ and $\theta$ as random variables[3]. Accordingly, theorems 4.1 and 4.2 naturally extend to the expected generalization error $\mathbb{E}_{\mathcal{L},\theta}\{Err(\varphi_{\mathcal{L},\theta}(\mathbf{x}))\}$ of the randomized model $\varphi_{\mathcal{L},\theta}$ by replacing expectations $\mathbb{E}_{\mathcal{L}}\{.\}$ and variances $\mathbb{V}_{\mathcal{L}}\{.\}$ with their respective counterparts $\mathbb{E}_{\mathcal{L},\theta}\{.\}$ and $\mathbb{V}_{\mathcal{L},\theta}\{.\}$ computed over the joint distribution of $\mathcal{L}$ and $\theta$. In regression, the bias-variance decomposition of the squared error loss thus becomes:

$$\mathbb{E}_{\mathcal{L},\theta}\{Err(\varphi_{\mathcal{L},\theta}(\mathbf{x}))\} = \text{noise}(\mathbf{x}) + \text{bias}^2(\mathbf{x}) + \text{var}(\mathbf{x}), \tag{4.12}$$

where

$$\text{noise}(\mathbf{x}) = Err(\varphi_B(\mathbf{x})), \tag{4.13}$$

$$\text{bias}^2(\mathbf{x}) = (\varphi_B(\mathbf{x}) - \mathbb{E}_{\mathcal{L},\theta}\{\varphi_{\mathcal{L},\theta}(\mathbf{x})\})^2, \tag{4.14}$$

$$\text{var}(\mathbf{x}) = \mathbb{E}_{\mathcal{L},\theta}\{(\mathbb{E}_{\mathcal{L},\theta}\{\varphi_{\mathcal{L},\theta}(\mathbf{x})\} - \varphi_{\mathcal{L},\theta}(\mathbf{x}))^2\}. \tag{4.15}$$

In this form, variance now accounts for both the prediction variability due to the randomness of the learning set $\mathcal{L}$ and the variability due to the randomness of the learning algorithm itself. As such, the variance of a randomized algorithm is typically larger than the variance of its deterministic counterpart. Depending on the strength of randomization, bias also usually increases, but often to a smaller extent than variance.

While randomizing an algorithm might seem counter-intuitive, since it increases both variance and bias, we will show in Section 4.2.3 that combining several such randomized models might actually achieve better performance than a single non-randomized model.

### 4.2.2 *Combining randomized models*

Let us assume a set of $M$ randomized models $\{\varphi_{\mathcal{L},\theta_m} | m = 1, \dots, M\}$, all learned on the same data $\mathcal{L}$ but each built from an independent random seed $\theta_m$. Ensemble methods work by combining the predictions of these models into a new *ensemble* model, denoted $\psi_{\mathcal{L},\theta_1,\dots,\theta_M}$, such that the expected generalization error of the ensemble is (hopefully) smaller than the expected generalization error of the individual randomized models.

In regression, for the squared error loss, the most common way to combine the randomized models into an ensemble is to average their predictions to form the final prediction:

$$\psi_{\mathcal{L},\theta_1,\dots,\theta_M}(\mathbf{x}) = \frac{1}{M}\sum_{m=1}^{M} \varphi_{\mathcal{L},\theta_m}(\mathbf{x}) \tag{4.16}$$

The rationale is that the average prediction is the prediction that minimizes the average squared error with respect to the individual predictions of the models. In that sense, the average prediction is the closest prediction with respect to all individual predictions.

3 From now on, and without loss of generality, we assume that the random variable $\theta$ only controls the randomness of the learning algorithm.

**AMBIGUITY DECOMPOSITION**

For prediction averaging, as defined in Equation 4.16, the *ambiguity decomposition* [Krogh et al., 1995] guarantees the generalization error of the ensemble to be lower than the average generalization error of its constituents. Formally, the ambiguity decomposition states that

$$\text{Err}(\psi_{\mathcal{L},\theta_1,\dots,\theta_M}) = \overline{E} - \overline{A} \tag{4.17}$$

where

$$\overline{E} = \frac{1}{M}\sum_{m=1}^{M} \text{Err}(\varphi_{\mathcal{L},\theta_m}), \tag{4.18}$$

$$\overline{A} = \mathbb{E}_X\{\frac{1}{M}\sum_{m=1}^{M} (\varphi_{\mathcal{L},\theta_m}(X) - \psi_{\mathcal{L},\theta_1,\dots,\theta_M}(X))^2\}. \tag{4.19}$$

The first term is the average generalization error of the individual models. The second term is the ensemble ambiguity and corresponds to the variance of the individual predictions around the prediction of the ensemble. Since $\overline{A}$ is non-negative, the generalization error of the ensemble is therefore smaller than the average generalization error of its constituents.

In classification, for the zero-one loss, predictions are usually aggregated by considering the models in the ensemble as a committee and then resorting to *majority voting* to form the final prediction:

$$\psi_{\mathcal{L},\theta_1,\dots,\theta_M}(\mathbf{x}) = \arg\max_{c \in \mathcal{Y}} \sum_{m=1}^{M} 1(\varphi_{\mathcal{L},\theta_m}(\mathbf{x}) = c) \tag{4.20}$$

Similarly, the rationale is that the majority prediction is the prediction that minimizes the average zero-one error with respect to the individual predictions. Alternatively, when individual models provide class probability estimates $\widehat{p}_{\mathcal{L},\theta m}(Y = c|X = \mathbf{x})$, *soft voting* [Zhou, 2012] consists in averaging the class probability estimates and then predict the class which is the most likely:

$$\psi_{\mathcal{L},\theta_1,\dots,\theta_M}(\mathbf{x}) = \arg\max_{c \in \mathcal{Y}} \frac{1}{M}\sum_{m=1}^{M} \widehat{p}_{\mathcal{L},\theta m}(Y = c|X = \mathbf{x}) \tag{4.21}$$

As empirically investigated by Breiman [1994], both approaches yield results that are nearly identical[4]. From a practical point of view however, Equation 4.21 has the advantage of providing smoother class probability estimates for the ensemble, which may prove to be useful in critical applications, e.g., when (estimates of) the certainty about

4 In the case of ensembles of fully developed decision trees that perfectly classify all samples from $\mathcal{L}$, majority voting and soft voting are exactly equivalent.

predictions is as important as the predictions themselves. Additionally, combining predictions in this way makes it easy to study the expected generalization error of the ensemble – it suffices to plug the averaged estimates into Theorem 4.2. For these reasons, and for the rest of this work, predictions in classification are now assumed to be combined with soft voting (see Equation 4.21) unless mentioned otherwise.

**Condorcet's jury theorem**

Majority voting, as defined in Equation 4.20, finds its origins in the *Condorcet's jury theorem* from the field of political science. Let consider a group of $M$ voters that wishes to reach a decision by majority vote. The theorem states that if each voter has an independent probability $p > \frac{1}{2}$ of voting for the correct decision, then adding more voters increases the probability of the majority decision to be correct. When $M \to \infty$, the probability that the decision taken by the group is correct approaches 1. Conversely, if $p < \frac{1}{2}$, then each voter is more likely to vote incorrectly and increasing $M$ makes things worse.

### 4.2.3 *Bias-variance decomposition of an ensemble*

Let us now study the bias-variance decomposition of the expected generalization error of an ensemble $\psi_{\mathcal{L},\theta_1,\dots,\theta_M}$, first in the case in case of regression and then for classification.

To simplify notations in the analysis below, let us denote the mean prediction at $X = \mathbf{x}$ of a single randomized model $\varphi_{\mathcal{L},\theta_m}$ and its respective prediction variance as:

$$\mu_{\mathcal{L},\theta_m}(\mathbf{x}) = \mathbb{E}_{\mathcal{L},\theta_m}\{\varphi_{\mathcal{L},\theta_m}(\mathbf{x})\} \tag{4.22}$$

$$\sigma^2_{\mathcal{L},\theta_m}(\mathbf{x}) = \mathbb{V}_{\mathcal{L},\theta_m}\{\varphi_{\mathcal{L},\theta_m}(\mathbf{x})\} \tag{4.23}$$

#### 4.2.3.1 *Regression*

From Theorem 4.1, the expected generalization error of an ensemble $\psi_{\mathcal{L},\theta_1,\dots,\theta_M}$ made of $M$ randomized models decomposes into a sum of noise$(\mathbf{x})$, bias$^2(\mathbf{x})$ and var$(\mathbf{x})$ terms.

The noise term only depends on the intrinsic randomness of $Y$. Its value stays therefore the same, no matter the learning algorithm:

$$\text{noise}(\mathbf{x}) = \mathbb{E}_{Y|X=\mathbf{x}}\{(Y - \varphi_B(\mathbf{x}))^2\} \tag{4.24}$$

The (squared) bias term is the (squared) difference between the prediction of the Bayes model and the average prediction of the model. For an ensemble, the average prediction is in fact the same as the av-erage prediction of the corresponding randomized individual model. Indeed,

$$\begin{aligned}\mathbb{E}_{\mathcal{L},\theta_1,\dots,\theta_M}\{\psi_{\mathcal{L},\theta_1,\dots,\theta_M}(\mathbf{x})\} &= \mathbb{E}_{\mathcal{L},\theta_1,\dots,\theta_M}\{\frac{1}{M}\sum_{m=1}^{M}\varphi_{\mathcal{L},\theta_m}(\mathbf{x})\} \\ &= \frac{1}{M}\sum_{m=1}^{M}\mathbb{E}_{\mathcal{L},\theta_m}\{\varphi_{\mathcal{L},\theta_m}(\mathbf{x})\} \\ &= \mu_{\mathcal{L},\theta}(\mathbf{x}) \end{aligned} \tag{4.25}$$

since random variables $\theta_m$ are independent and all follow the same distribution. As a result,

$$\text{bias}^2(\mathbf{x}) = (\varphi_B(\mathbf{x}) - \mu_{\mathcal{L},\theta}(\mathbf{x}))^2, \tag{4.26}$$

which indicates that the bias of an ensemble of randomized models is the same as the bias of any of the randomized models. Put otherwise, combining randomized models has no effect on the bias.

On variance on the other hand, ensemble methods show all their raison d'etre, virtually reducing the variability of predictions to almost nothing and thereby improving the accuracy of the ensemble. Before considering the variance of $\psi_{\mathcal{L},\theta_1,\dots,\theta_M}(\mathbf{x})$ however, let us first derive the correlation coefficient $\rho(\mathbf{x})$ between the predictions of two randomized models built on the same learning set, but grown from two independent random seeds $\theta'$ and $\theta''$. From the definition of the Pearson's correlation coefficient, it comes:

$$\begin{aligned}\rho(\mathbf{x}) &= \frac{\mathbb{E}_{\mathcal{L},\theta',\theta''}\{(\varphi_{\mathcal{L},\theta'}(\mathbf{x}) - \mu_{\mathcal{L},\theta'}(\mathbf{x}))(\varphi_{\mathcal{L},\theta''}(\mathbf{x}) - \mu_{\mathcal{L},\theta''}(\mathbf{x}))\}}{\sigma_{\mathcal{L},\theta'}(\mathbf{x})\sigma_{\mathcal{L},\theta''}(\mathbf{x})} \\ &= \frac{\mathbb{E}_{\mathcal{L},\theta',\theta''}\{\varphi_{\mathcal{L},\theta'}(\mathbf{x})\varphi_{\mathcal{L},\theta''}(\mathbf{x}) - \varphi_{\mathcal{L},\theta'}(\mathbf{x})\mu_{\mathcal{L},\theta''}(\mathbf{x}) - \varphi_{\mathcal{L},\theta''}(\mathbf{x})\mu_{\mathcal{L},\theta'}(\mathbf{x}) + \mu_{\mathcal{L},\theta'}(\mathbf{x})\mu_{\mathcal{L},\theta''}(\mathbf{x})\}}{\sigma^2_{\mathcal{L},\theta}(\mathbf{x})} \\ &= \frac{\mathbb{E}_{\mathcal{L},\theta',\theta''}\{\varphi_{\mathcal{L},\theta'}(\mathbf{x})\varphi_{\mathcal{L},\theta''}(\mathbf{x})\} - \mu^2_{\mathcal{L},\theta}(\mathbf{x})}{\sigma^2_{\mathcal{L},\theta}(\mathbf{x})} \end{aligned} \tag{4.27}$$

by linearity of the expectation and exploiting the fact that random variables $\theta'$ and $\theta''$ follow the same distribution. Intuitively, $\rho(\mathbf{x})$ represents the strength of the random perturbations introduced in the learning algorithm. When it is close to 1, predictions of two randomized models are highly correlated, suggesting that randomization has no sensible effect on the predictions. By contrast, when it is close to 0, predictions of the randomized models are decorrelated, hence indicating that randomization has a strong effect on the predictions. At the limit, when $\rho(\mathbf{x}) = 0$, predictions of two models built on the same learning set $\mathcal{L}$ are independent, which happens when they are perfectly random. As proved later with Equation 4.31, let us finally also remark that the correlation term $\rho(\mathbf{x})$ is non-negative, which confirms that randomization has a decorrelation effect only.

From Equation 4.27, the variance of $\psi_{\mathcal{L},\theta_1,\dots,\theta_M}(\mathbf{x})$ can now be derived as follows:

$$\begin{aligned}
\text{var}(\mathbf{x}) &= \mathbb{V}_{\mathcal{L},\theta_1,\dots,\theta_M}\{\frac{1}{M}\sum_{m=1}^{M}\varphi_{\mathcal{L},\theta_m}(\mathbf{x})\} \\
&= \frac{1}{M^2}\left[\mathbb{E}_{\mathcal{L},\theta_1,\dots,\theta_M}\{(\sum_{m=1}^{M}\varphi_{\mathcal{L},\theta_m}(\mathbf{x}))^2\} - \mathbb{E}_{\mathcal{L},\theta_1,\dots,\theta_M}\{\sum_{m=1}^{M}\varphi_{\mathcal{L},\theta_m}(\mathbf{x})\}^2\right]
\end{aligned}$$

by exploiting the facts that $\mathbb{V}\{aX\} = a^2\mathbb{V}\{X\}$, $\mathbb{V}\{X\} = \mathbb{E}\{X^2\} - \mathbb{E}\{X\}^2$ and the linearity of expectation. By rewriting the square of the sum of the $\varphi_{\mathcal{L},\theta_m}(\mathbf{x})$ terms as a sum over all pairwise products $\varphi_{\mathcal{L},\theta_i}(\mathbf{x})\varphi_{\mathcal{L},\theta_j}(\mathbf{x})$, the variance can further be rewritten as:

$$\begin{aligned}
&= \frac{1}{M^2}\left[\mathbb{E}_{\mathcal{L},\theta_1,\dots,\theta_M}\{\sum_{i,j}\varphi_{\mathcal{L},\theta_i}(\mathbf{x})\varphi_{\mathcal{L},\theta_j}(\mathbf{x})\} - (M\mu_{\mathcal{L},\theta}(\mathbf{x}))^2\right] \\
&= \frac{1}{M^2}\left[\sum_{i,j}\mathbb{E}_{\mathcal{L},\theta_i,\theta_j}\{\varphi_{\mathcal{L},\theta_i}(\mathbf{x})\varphi_{\mathcal{L},\theta_j}(\mathbf{x})\} - M^2\mu^2_{\mathcal{L},\theta}(\mathbf{x})\right] \\
&= \frac{1}{M^2}\Big[M\mathbb{E}_{\mathcal{L},\theta}\{\varphi_{\mathcal{L},\theta}(\mathbf{x})^2\} \\
&\quad \hookrightarrow + (M^2 - M)\mathbb{E}_{\mathcal{L},\theta',\theta''}\{\varphi_{\mathcal{L},\theta'}(\mathbf{x})\varphi_{\mathcal{L},\theta''}(\mathbf{x})\} - M^2\mu^2_{\mathcal{L},\theta}(\mathbf{x})\Big] \\
&= \frac{1}{M^2}\Big[M(\sigma^2_{\mathcal{L},\theta}(\mathbf{x}) + \mu^2_{\mathcal{L},\theta}(\mathbf{x})) \\
&\quad \hookrightarrow + (M^2 - M)(\rho(\mathbf{x})\sigma^2_{\mathcal{L},\theta}(\mathbf{x}) + \mu^2_{\mathcal{L},\theta}(\mathbf{x})) - M^2\mu^2_{\mathcal{L},\theta}(\mathbf{x})\Big] \\
&= \frac{\sigma^2_{\mathcal{L},\theta}(\mathbf{x})}{M} + \rho(\mathbf{x})\sigma^2_{\mathcal{L},\theta}(\mathbf{x}) - \rho(\mathbf{x})\frac{\sigma^2_{\mathcal{L},\theta}(\mathbf{x})}{M} \\
&= \rho(\mathbf{x})\sigma^2_{\mathcal{L},\theta}(\mathbf{x}) + \frac{1-\rho(\mathbf{x})}{M}\sigma^2_{\mathcal{L},\theta}(\mathbf{x}) \qquad (4.28)
\end{aligned}$$

As the size of the ensemble gets arbitrarily large, i.e., as $M \to \infty$, the variance of the ensemble reduces to $\rho(\mathbf{x})\sigma^2_{\mathcal{L},\theta}(\mathbf{x})$. Under the assumption that randomization has some effect on the predictions of randomized models, i.e., assuming $\rho(\mathbf{x}) < 1$, the variance of an ensemble is therefore strictly smaller than the variance of an individual model. As a result, the expected generalization error of an ensemble is strictly smaller than the expected error of a randomized model. As such, improvements in predictions are solely the result of variance reduction, since both noise$(\mathbf{x})$ and bias$^2(\mathbf{x})$ remain unchanged. Additionally, when random effects are strong, i.e., when $\rho(\mathbf{x}) \to 0$, variance reduces to $\frac{\sigma^2_{\mathcal{L},\theta}(\mathbf{x})}{M}$, which can further be driven to 0 by increasing the size of the ensemble. On the other hand, when random effects are weak, i.e., when $\rho(\mathbf{x}) \to 1$, then variance reduces to $\sigma^2_{\mathcal{L},\theta}(\mathbf{x})$ and building an ensemble brings no benefit. Put otherwise, the stronger the random effects, the larger the reduction of variance due to ensembling, and vice-versa.

In summary, the expected generalization error of an ensemble additively decomposes as stated in Theorem 4.3.

**Theorem 4.3.** *For the squared error loss, the bias-variance decomposition of the expected generalization error* $\mathbb{E}_{\mathcal{L}}\{\text{Err}(\psi_{\mathcal{L},\theta_1,\ldots,\theta_M}(\mathbf{x}))\}$ *at* $X = \mathbf{x}$ *of an ensemble of* $M$ *randomized models* $\varphi_{\mathcal{L},\theta_m}$ *is*

$$\mathbb{E}_{\mathcal{L}}\{\text{Err}(\psi_{\mathcal{L},\theta_1,\ldots,\theta_M}(\mathbf{x}))\} = \textit{noise}(\mathbf{x}) + \textit{bias}^2(\mathbf{x}) + \textit{var}(\mathbf{x}), \tag{4.29}$$

*where*

$$\begin{aligned}
\textit{noise}(\mathbf{x}) &= \text{Err}(\varphi_B(\mathbf{x})),\\
\textit{bias}^2(\mathbf{x}) &= (\varphi_B(\mathbf{x}) - \mathbb{E}_{\mathcal{L},\theta}\{\varphi_{\mathcal{L},\theta}(\mathbf{x})\})^2,\\
\textit{var}(\mathbf{x}) &= \rho(\mathbf{x})\sigma^2_{\mathcal{L},\theta}(\mathbf{x}) + \frac{1-\rho(\mathbf{x})}{M}\sigma^2_{\mathcal{L},\theta}(\mathbf{x}).
\end{aligned}$$

In light of Theorem 4.3, the core principle of ensemble methods is thus to introduce random perturbations in order to decorrelate as much as possible the predictions of the individual models, thereby maximizing variance reduction. However, random perturbations need to be carefully chosen so as to increase bias as little as possible. The crux of the problem is to find the right trade-off between randomness and bias.

### Alternative variance decomposition

Geurts [2002] (Chapter 4, Equation 4.31) alternatively decomposes the ensemble variance as

$$\text{var}(\mathbf{x}) = \mathbb{V}_{\mathcal{L}}\{\mathbb{E}_{\theta|\mathcal{L}}\{\varphi_{\mathcal{L},\theta}(\mathbf{x})\}\} + \frac{1}{M}\mathbb{E}_{\mathcal{L}}\{\mathbb{V}_{\theta|\mathcal{L}}\{\varphi_{\mathcal{L},\theta}(\mathbf{x})\}\}. \tag{4.30}$$

The first term of this decomposition is the variance due to the randomness of the learning set $\mathcal{L}$, averaged over the random perturbations due to $\theta$. It measures the dependence of the model on the learning set, independently of $\theta$. The second term is the expectation over all learning sets of the variance with respect to $\theta$. It measures the strength of the random effects. As the decomposition shows, only this last part of the variance can be reduced as a result of averaging, which is consistent with our previous conclusions. The stronger the random effects, the larger the variance with respect to $\theta$, and hence the larger of reduction of variance due to ensembling.

**Proposition 4.4.** *Decompositions 4.28 and 4.30 of the prediction variance for an ensemble are equivalent.*

*Proof.* From Equations 4.28 and 4.30, equivalence holds if Equation 4.27 is equivalent to

$$\rho(\mathbf{x}) = \frac{\mathbb{V}_{\mathcal{L}}\{\mathbb{E}_{\theta|\mathcal{L}}\{\varphi_{\mathcal{L},\theta}(\mathbf{x})\}\}}{\mathbb{V}_{\mathcal{L}}\{\mathbb{E}_{\theta|\mathcal{L}}\{\varphi_{\mathcal{L},\theta}(\mathbf{x})\}\} + \mathbb{E}_{\mathcal{L}}\{\mathbb{V}_{\theta|\mathcal{L}}\{\varphi_{\mathcal{L},\theta}(\mathbf{x})\}\}}. \tag{4.31}$$

From the law of total variance, the denominator of Equation 4.27 expands to the denominator of Equation 4.31:

$$\sigma^2_{\mathcal{L},\theta}(\mathbf{x}) = \mathbb{V}_{\mathcal{L}}\{\mathbb{E}_{\theta|\mathcal{L}}\{\varphi_{\mathcal{L},\theta}(\mathbf{x})\}\} + \mathbb{E}_{\mathcal{L}}\{\mathbb{V}_{\theta|\mathcal{L}}\{\varphi_{\mathcal{L},\theta}(\mathbf{x})\}\} \tag{4.32}$$

Similarly, the numerator of Equation 4.31 can be reexpressed as the numerator of Equation 4.27, thereby proving the equivalence:

$$\begin{aligned}
&\mathbb{V}_{\mathcal{L}}\{\mathbb{E}_{\theta|\mathcal{L}}\{\varphi_{\mathcal{L},\theta}(\mathbf{x})\}\} \\
&= \mathbb{E}_{\mathcal{L}}\{(\mathbb{E}_{\theta|\mathcal{L}}\{\varphi_{\mathcal{L},\theta}(\mathbf{x})\} - \mathbb{E}_{\mathcal{L}}\{\mathbb{E}_{\theta|\mathcal{L}}\{\varphi_{\mathcal{L},\theta}(\mathbf{x})\}\})^2\} \\
&= \mathbb{E}_{\mathcal{L}}\{(\mathbb{E}_{\theta|\mathcal{L}}\{\varphi_{\mathcal{L},\theta}(\mathbf{x})\} - \mu_{\mathcal{L},\theta}(\mathbf{x}))^2\} \\
&= \mathbb{E}_{\mathcal{L}}\{\mathbb{E}_{\theta|\mathcal{L}}\{\varphi_{\mathcal{L},\theta}(\mathbf{x})\}^2\} - \mu^2_{\mathcal{L},\theta}(\mathbf{x}) \\
&= \mathbb{E}_{\mathcal{L}}\{\mathbb{E}_{\theta'|\mathcal{L}}\{\varphi_{\mathcal{L},\theta'}(\mathbf{x})\}\mathbb{E}_{\theta''|\mathcal{L}}\{\varphi_{\mathcal{L},\theta''}(\mathbf{x})\}\} - \mu^2_{\mathcal{L},\theta}(\mathbf{x}) \\
&= \mathbb{E}_{\mathcal{L},\theta',\theta''}\{\varphi_{\mathcal{L},\theta'}(\mathbf{x})\varphi_{\mathcal{L},\theta''}(\mathbf{x})\} - \mu^2_{\mathcal{L},\theta}(\mathbf{x}).
\end{aligned} \tag{4.33}$$

□

In this form, $\rho(\mathbf{x})$ is interpreted as the ratio between the variance due to the learning set and the total variance, accounting for random effects due to both the learning set and the random perturbations. It is close to 1 when variance is mostly due to the learning set, hence yielding correlated predictions. Conversely, it is close to 0 when variance is mostly due to the random perturbations induced by $\theta$, hence decorrelating the predictions.

#### 4.2.3.2 *Classification*

The decomposition of the expected generalization error of an ensemble in classification directly follows from theorems 4.2 and 4.3. Building an ensemble always reduces the variance of the class probability estimate $\mathbb{E}_{\mathcal{L},\theta}\{\widehat{p}_{\mathcal{L},\theta}(Y = \varphi_B(\mathbf{x})\}$ (as shown in Equation 4.28), which results in a decrease of the misclassification error if the expected estimate remains strictly greater than 0.5 in a randomized model.

**Shortcomings addressed by ensembles**

In complement with the formal bias-variance analysis carried out in the previous section, Dietterich [2000b] identifies three fundamental reasons intuitively explaining why ensembles often work better than single models.

The first reason is statistical. When the learning set is too small, a learning algorithm can typically find several models in the hypothesis space $\mathcal{H}$ that all give the same performance on the training data. Provided their predictions are uncorrelated, averaging several models reduces the risk of choosing the wrong hypothesis.

The second reason is computational. Many learning algorithms rely on some greedy assumption or local search that may get stuck in local optima. As such, an ensemble made of individual models built from many different starting points may provide a better approximation of the true unknown function that any of the single models.

Finally, the third reason is representational. In most cases, for a learning set of finite size, the true function cannot be represented by any of the candidate models in $\mathcal{H}$. By combining several models in an ensemble, it may be possible to expand the space of representable functions and to better model the true function.

## 4.3 RANDOM FORESTS

Random forests[5] form a family of methods that consist in building an ensemble (or *forest*) of decision trees grown from a randomized variant of the tree induction algorithm (as described in Chapter 3). Decision trees are indeed ideal candidates for ensemble methods since they usually have low bias and high variance, making them very likely to benefit from the averaging process. As we will review in this section, random forests methods mostly differ from each other in the way they introduce random perturbations into the induction procedure. As highlighted in the previous section, the difficulty is to inject randomness while minimizing $\rho(\mathbf{x})$ and simultaneously maintaining a low bias in the randomized decision trees.

### 4.3.1 *Randomized induction algorithms*

Kwok and Carter [1990]:
: Historically, the earliest mention of ensemble of decision trees is due to Kwok and Carter [1990]. In this work, the authors empirically observe that averaging multiple decision trees with different structure consistently produces better results than any of the constituents of the ensemble. This approach however was not based on randomization nor was fully automatic: decision trees were generated by first manually selecting at the top of the tree splits that were almost as good as the optimal splits, and then expanded using the classical ID3 induction procedure.

Breiman [1994]:
: In a now famous technical report, Breiman [1994] was one of the earliest to show, both theoretically and empirically, that ag-

5 The term *random forests*, without capitals, is used to denote any ensemble of decision trees. Specific variants are referred to using their original designation, denoted with capitals. In particular, the ensemble method due to Breiman [2001] is denoted as *Random Forests*, with capitals.

gregating multiple versions of an estimator into an ensemble can give substantial gains in accuracy. He notes and shows that the average model $\mathbb{E}_{\mathcal{L}}\{\varphi_{\mathcal{L}}\}$ has a lower expected generalization error than $\varphi_{\mathcal{L}}$. As such, *Bagging* consists in approximating $\mathbb{E}_{\mathcal{L}}\{\varphi_{\mathcal{L}}\}$ by combining models built from *bootstrap samples* [Efron, 1979] $\mathcal{L}^m$ (for $m = 1, \dots, M$) of the learning set $\mathcal{L}$. The $\{\mathcal{L}^m\}$ form replicates of $\mathcal{L}$, each consisting of N cases $(\mathbf{x}, y)$, drawn at random but *with replacement* from $\mathcal{L}$.

Note that even though $|\mathcal{L}| = |\mathcal{L}^m| = N$, 37% of the couples $(\mathbf{x}, y)$ from $\mathcal{L}$ are on average missing in the bootstrap replicates. Indeed, after N draws with replacement, the probability of never have been selected is

$$(1 - \frac{1}{N})^N \approx \frac{1}{e} \approx 0.368. \tag{4.34}$$

When the learning algorithm is unstable (i.e., when small changes in the learning set can cause large changes in the learned models), Bagging generates individual models that are different from one bootstrap sample to another, hence making them likely to benefit from the averaging process. In some cases however, when the learning set $\mathcal{L}$ is small, subsampling 67% of the objects might lead to an increase of bias (e.g., because of a decrease in model complexity) which is too large to be compensated by a decrease of variance, hence resulting in overall poorer performance. Despite this defect, Bagging has proven to be an effective method in numerous applications, one of its strengths being that it can be used to combine any kind of models – i.e., not only decision trees.

Dietterich and Kong [1995]:
Building upon [Kwok and Carter, 1990], Dietterich and Kong [1995] propose to randomize the choice of the best split at a given node by selecting uniformly at random one of the 20 best splits of node $t$. The authors empirically show in [Dietterich and Kong, 1995; Dietterich, 2000a] that randomizing in this way gives results that are slightly better than Bagging in low noise settings. Experiments show however that when noise is important, Bagging usually yield better results. From a bias-variance point of view, this method virtually does not change bias but increases variance due to randomization.

Amit et al. [1997]:
In the context of handwritten character recognition, where the number $p$ of input variables is typically very large, Amit et al. [1997] propose a randomized variant of the tree induction algorithm that consists in searching for the best split at each node over a random subsample of the variables.

Denoting $K \leqslant p$ (also known as `mtry` or `max_features`) the number of variables effectively considered at each node, this variant replaces Algorithm 3.3 with the following randomized alternative:

**Algorithm 4.1.** *Find the best split* $s^*$ *that partitions* $\mathcal{L}_t$*, among a random subset of* $K \leqslant p$ *input variables.*

```
function FindBestSplitRandom(L_t, K)
    Δ = −∞
    Draw K random indices j_k from 1, ..., p
    for k = 1, ..., K do
        Find the best binary split s*_{j_k} defined on X_{j_k}
        if Δi(s*_{j_k}, t) > Δ then
            Δ = Δi(s*_{j_k}, t)
            s* = s*_{j_k}
        end if
    end for
    return s*
end function
```

When the output Y can be explained in several ways, this randomized algorithm generates trees that are each structurally different, yet individually good. As a result, bias usually increases only slightly, while the increased variance due to randomization can be cancelled out through averaging. The optimal trade-off between these quantities can otherwise be adjusted by tuning the value of K. As $K \to 1$, the larger the bias but the larger the variance of the individual models and hence the more effective the averaging process. Conversely, as $K \to p$, the smaller the bias but also the smaller the variance of the individual models and therefore the less beneficial the ensemble.

Ho [1998]:

Inspired from the principles of Bagging [Breiman, 1994] and random subsets of variables [Amit et al., 1997], Ho [1998] proposes with the *Random Subspace* (RS) method to build a *decision forest* whose trees are grown on random subsets of the input variables – drawn once, prior to the construction of each tree – rather than on all p variables. As empirically evaluated at several occasions [Ho, 1998; Panov and Džeroski, 2007; Louppe and Geurts, 2012], the Random Subspace method is a powerful generic ensemble method that can achieve near state-of-the-art performance on many problems. Again, the optimal trade-off between the variance due to randomization and the increase of bias can be controlled by tuning the size of the random subset.

Breiman [2001]:

In his seminal *Random Forests* (RF) paper, Breiman [2001] com-

bines Bagging [Breiman, 1994] with random variable selection at each node [Amit et al., 1997]. Injecting randomness simultaneously with both strategies yields one the most effective off-the-shelf methods in machine learning, working surprisingly well for almost any kind of problems. The author empirically shows that Random Forests give results that are competitive with boosting [Freund and Schapire, 1995] and arcing algorithms [Breiman, 1996], which both are designed to reduce bias while forests focus on variance reduction.

While the original principles are due to several authors (as discussed above), Breiman is often cited as the father of forests of randomized trees. Parts of this recognition are certainly due to the pioneer theoretical analysis that has always complemented his empirical analysis of algorithms. In contrast with other authors, another reason might also be his efficient software implementation [Breiman, 2002] that was made freely available, allowing users outside of the machine learning community to quickly and easily apply Random Forests to their problems.

Cutler and Zhao [2001]:

With *Perfect Random Tree Ensembles* (PERT), Cutler and Zhao [2001] propose to grow a forest of perfectly fit decision trees in which both the (ordered) variable to split on and the discretization threshold are chosen at random. More specifically, given a node t, the split variable $X_j$ is drawn at random using Algorithm 4.1 with $K = 1$ while the cut-point $v$ is set midway between two randomly drawn samples using the following procedure (instead of Algorithm 3.4):

**Algorithm 4.2.** *Draw a random split on* $X_j$ *that partitions* $\mathcal{L}_t$*.*

```
1: function FindRandomSplit-PERT(L_t, X_j)
2:     Draw (x_1, y_1), (x_2, y_2) ∈ L_t such that y_1 ≠ y_2
3:     Draw α uniformly at random from [0, 1]
4:     v = α x_{1,j} + (1 − α) x_{2,j}
5:     return s_j^v
6: end function
```

The induction of the tree proceeds using such random splits until all nodes become pure or until it is no longer possible to draw samples of different output values.

From a practical point of view, PERT is an easily coded and a very efficient ensemble method since there is no impurity criterion to evaluate when splitting the nodes of a tree. Regarding accuracy, experimental comparisons in [Cutler and Zhao, 2001] show that PERT is often nearly as good as Random Forests [Breiman, 2001], while resulting however in random trees that are typically larger than decision trees grown with less randomization. The simplicity of the method also allows for an amenable theoretical analysis of forests of randomized trees, as carried out in [Zhao, 2000].

Geurts et al. [2006a]:
As investigated in [Wehenkel, 1997; Geurts and Wehenkel, 2000], the notoriously high variance of decision trees partly finds its origins from the high dependence of the splits with the random nature of the learning set. The authors empirically show that the variance of the optimal cut-point $v$ (in the case of ordered input variables) may indeed be very high, even for large sample sizes. In particular, Geurts [2002] shows that cut-point variance appears to be responsible for a significant part of the generalization error of decision trees. As a way to smoothen the decision boundary, Geurts et al. [2006a] hence propose in *Extremely Randomized Trees* (ETs) to combine random variable selection [Amit et al., 1997] with random discretization thresholds. As a drop-in replacement of Algorithm 3.4, the authors propose the following simplistic but effective procedure for drawing splits at random:

**Algorithm 4.3.** *Draw a random split on* $X_j$ *that partitions* $\mathcal{L}_t$.

1: **function** FindRandomSplit-ETs($\mathcal{L}_t$, $X_j$)
2: $\quad$ $\min_j = \min(\{x_{i,j} | (\mathbf{x}_i, y_i) \in \mathcal{L}_t\})$
3: $\quad$ $\max_j = \max(\{x_{i,j} | (\mathbf{x}_i, y_i) \in \mathcal{L}_t\})$
4: $\quad$ Draw $v$ uniformly at random from $[\min_j, \max_j[$
5: $\quad$ **return** $s_j^v$
6: **end function**

With respect to decomposition 4.30 of variance, extremely randomized trees can therefore be seen as a way to transfer cut-point variance from the variance term due to the learning set to the (reducible) variance term that is due to random effects.

In the special case where $K = 1$, Extremely Randomized Trees reduce to *Totally Randomized Trees*, in which both a single variable $X_j$ and a discretization threshold $v$ are drawn at random at each node. As a result, the structure of such trees can be learned in an unsupervised way, independently of the output variable $Y$. In this setting, Totally Randomized Trees are very close to Perfect Random Tree Ensembles [Cutler and Zhao, 2001] since both draw $X_j$ and $v$ at random. These methods are however not strictly equivalent since they do not draw the random discretization thresholds $v$ with respect to the same probability distribution.

Rodriguez et al. [2006]:
In a different direction, Rodriguez et al. [2006] propose in *Ro-*

*tation Forests* to generate randomized decision trees based on feature extraction. As in Bagging [Breiman, 1994], individual decision trees are built on bootstrap replicates $\mathcal{L}^m$ of the learning set $\mathcal{L}$. To further enhance diversity (i.e., to further decorrelate the predictions of the constituents of the ensemble), for each of the $M$ bootstrap replicates, input variables are randomly partitioned into $q$ subsets of $\frac{p}{q}$ variables, principal component analysis (PCA) is run separately on each subset, and a new set $\widetilde{\mathcal{L}^m}$ of $p$ extracted input variables is constructed by pooling all principal components from the $q$ projections. In this way, bootstrap replicates $\mathcal{L}^m$ are each independently transformed linearly into a new input space using $q$ axis rotations. In this framework, decision trees are particularly suited because they are sensitive to changes in the input space and still can be very accurate. As reported in [Rodriguez et al., 2006; Kuncheva and Rodríguez, 2007], Rotation Forests favorably compare with other tree-based ensemble methods, yielding results that are often as good, sometimes better, than Random Forests [Breiman, 2001]. In terms of complexity however, the computational overhead due to the $q$ axis rotations should not be overlooked when resources are limited.

### 4.3.2 *Illustration*

As a summary and illustrative example, let us consider a simulated toy regression problem such that

$$Y = \sum_{j=1}^{5} X_j, \tag{4.35}$$

where all input variables $X_1, \dots, X_5$ are independent random Gaussian variables of zero mean and unit variance. To simulate noise in the data, 5 additional random Gaussian input variables $X_6, \dots, X_{10}$, all independent from $Y$, are further appended to the learning set.

Let us compare for this problem the bias-variance decomposition of the expected generalization error of a Random Forest [Breiman, 2001] (RF) and of Extremely Randomized Trees [Geurts et al., 2006a] (ETs). For both methods, the error is estimated as derived from Theorem 4.3, using 100 randomly drawn learning sets $\mathcal{L}$ of 50 samples, on which ensembles of 10 trees are built. Their generalization error is estimated on an independent test set of 1000 samples.

As the left plot in Figure 4.4 shows, the expected generalization error additively decomposes into (squared) bias and variance terms. For small values of $K$, random effects are strong, leading to high bias and low variance. By contrast, for larger values of $K$, random effects are less important, reducing bias but increasing variance. For both methods, too small a value of $K$ appears however to lead to a too large increase of bias, for which averaging is not able to compensate. For RF, the optimal trade-off is at $K = 8$, which indicates that bagging and random variable selection are here complementary with regard to the level of randomness injected into the trees. Indeed, using bagging only (at $K = 10$) for randomizing the construction of the trees yield results that are slightly worse than when both techniques are combined. For ETs, the optimal trade-off is at $K = 10$, suggesting that random thresholds $v$ provide by themselves enough randomization on this problem.

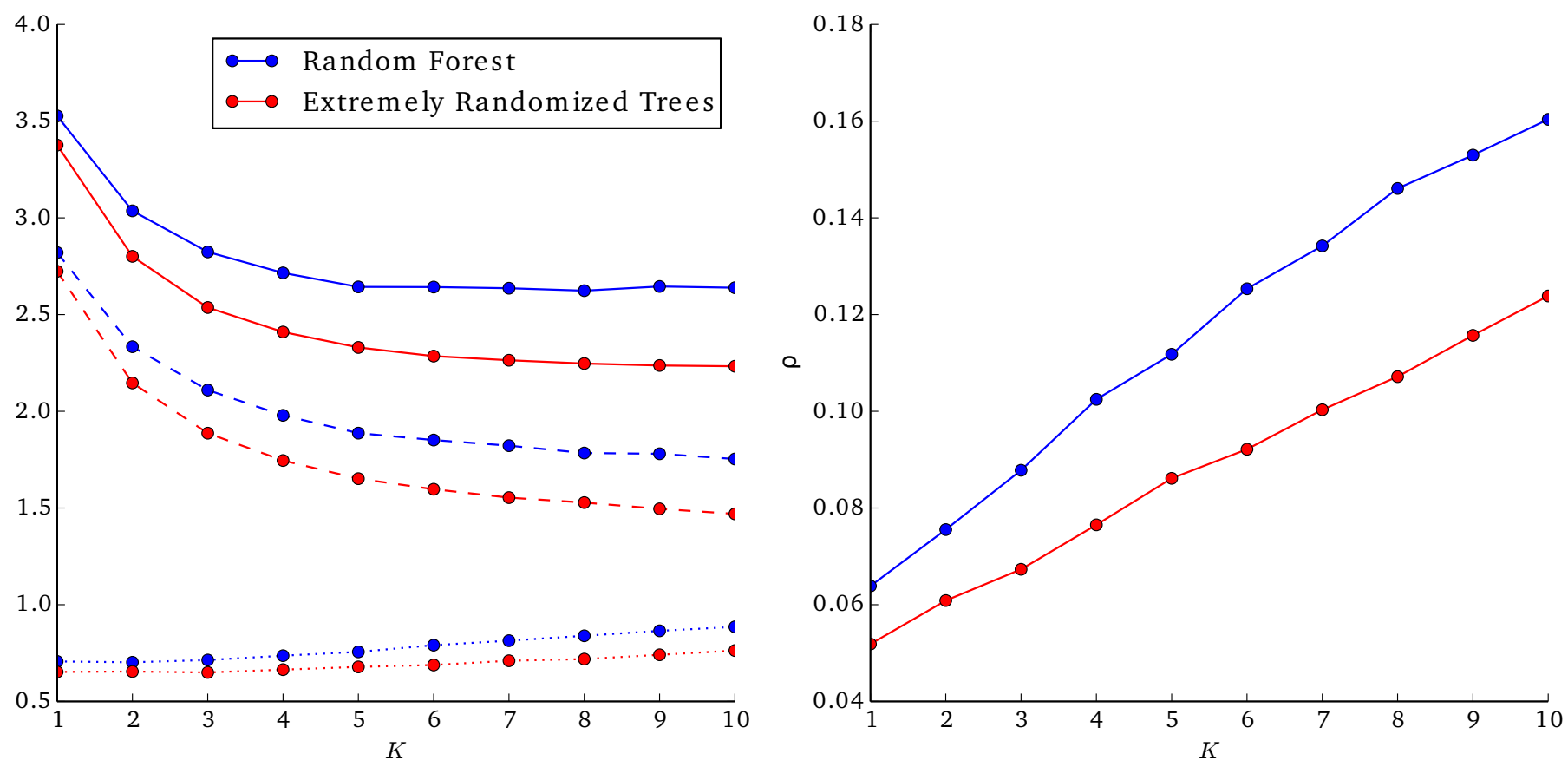

Figure 4.4: (Left) Bias-variance decomposition of the generalization error with respect to the hyper-parameter K. The total error is shown by the plain lines. Bias and variance terms are respectively shown by the dashed and dotted lines. (Right) Average correlation coefficient $\rho(\mathbf{x})$ over the predictions of two randomized trees grown from the same learning set but with different random seeds.

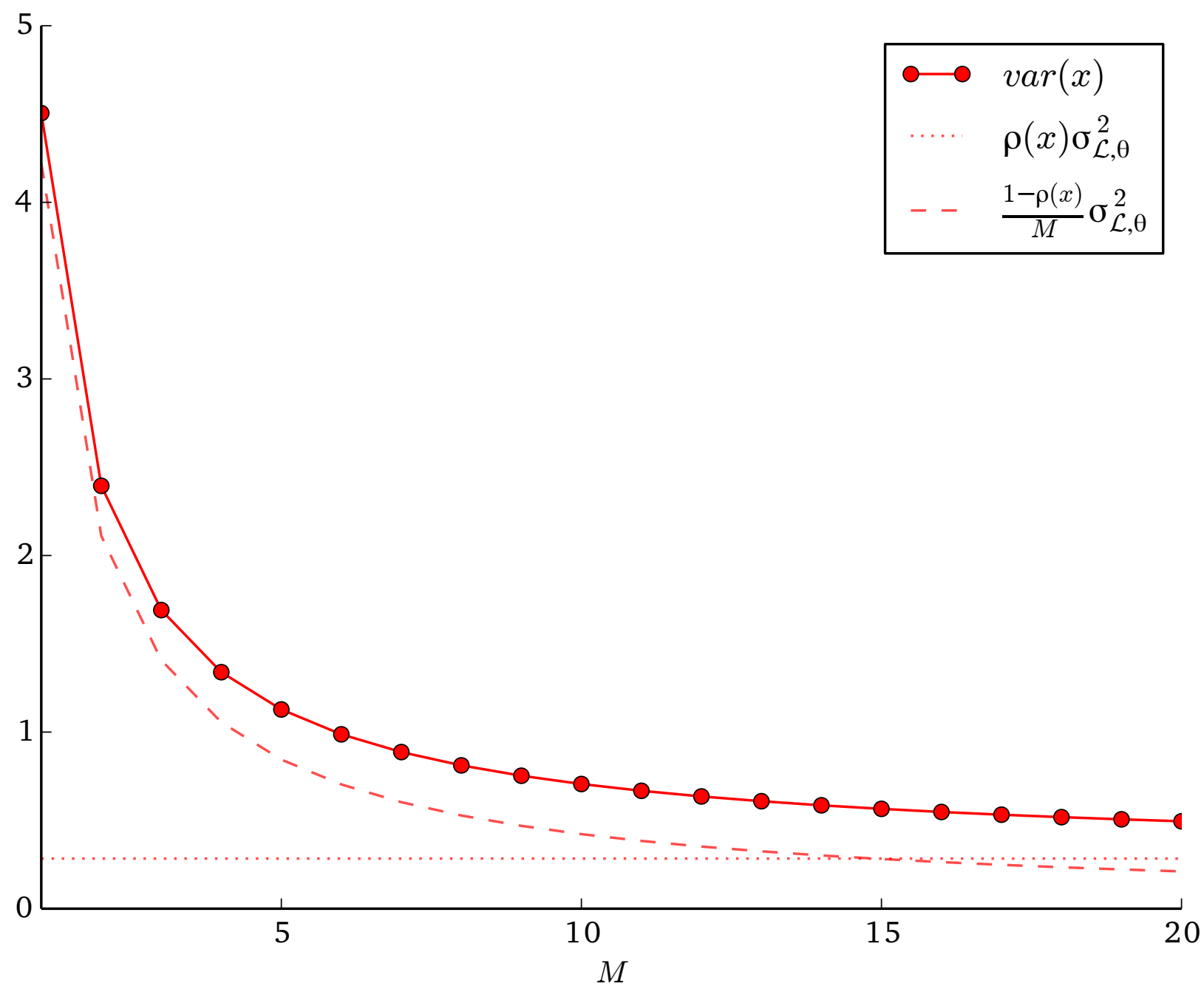

Figure 4.5: Decomposition of the variance term $\text{var}(\mathbf{x})$ with respect to the number M of trees in the ensemble.

The right plot of Figure 4.4 illustrates the (averaged) correlation coefficient $\rho(\mathbf{x})$ between the predictions of two randomized trees grown on the same learning set. As expected, the smaller K, the stronger the random effects, therefore the less correlated the predictions and the more variance can be reduced from averaging. The plot also confirms that ETs are inherently less correlated than trees built in RF, which is not surprising given the fact that the former method randomizes the choice of the discretization threshold while the latter does not. In cases where such a randomization does not induce too large an increase of bias, as in this problem, ETs are therefore expected to yield better results than RF. (The choice of the optimal randomization strategy is however highly dependent on the problem and no general conclusion should be drawn from this toy regression problem.)

As confirmed by Figure 4.5 for RF, variance also additively decomposes into

$$\text{var}(\mathbf{x}) = \rho(\mathbf{x})\sigma^2_{\mathcal{L},\theta}(\mathbf{x}) + \frac{1-\rho(\mathbf{x})}{M}\sigma^2_{\mathcal{L},\theta}(\mathbf{x}). \tag{4.36}$$

The first term is the variance due to the learning set $\mathcal{L}$ and remains constant as the number $M$ of trees increases. The second term is the variance due to random effects and decreases as $M$ increases. Of particular interest is variance at $M = 1$, which corresponds to the variance of a single decision tree. As the figure clearly shows, averaging several decision trees into an ensemble allows to significantly reduce this quantity. At the limit, when $M \to \infty$, variance tends to $\rho(\mathbf{x})\sigma^2_{\mathcal{L},\theta}(\mathbf{x})$, as shown by the dotted line and as expected from Theorem 4.3.

## 4.4 PROPERTIES AND FEATURES

### 4.4.1 *Out-of-bag estimates*

An interesting feature of ensemble methods that construct models on bootstrap samples, like Bagging or Random Forests, is the built-in possibility of using the left-out samples $\mathcal{L} \setminus \mathcal{L}^m$ to form estimates of important statistics. In the case of generalization error, the *out-of-bag* estimate at $(\mathbf{x}_i, y_i)$ consists in evaluating the prediction of the ensemble using only the individual models $\varphi_{\mathcal{L}^m}$ whose bootstrap samples $\mathcal{L}^m$ did not include $(\mathbf{x}_i, y_i)$. That is, in regression,

$$\widehat{\text{Err}}^{OOB}(\psi_{\mathcal{L}}) = \frac{1}{N} \sum_{(\mathbf{x}_i, y_i) \in \mathcal{L}} L(\frac{1}{M^{-i}} \sum_{l=1}^{M^{-i}} \varphi_{\mathcal{L}^{m_{k_l}}}(\mathbf{x}_i), y_i), \tag{4.37}$$

where $m_{k_1}, \dots, m_{k_{M^{-i}}}$ denote the indices of $M^{-i}$ the trees that have been built from bootstrap replicates that do not include $(\mathbf{x}_i, y_i)$. For classification, the out-of-bag estimate of the generalization error is similar to Equation 4.37, except that the out-of-bag average prediction is replaced with the class which is the most likely, as computed from the out-of-bag class probability estimates.

Out-of-bag estimates provide accurate estimates of the generalization error of the ensemble, often yielding statistics that are as good or even more precise than K-fold cross-validation estimates [Wolpert and Macready, 1999]. In practice, out-of-bag estimates also constitute a computationally efficient alternative to K-fold cross-validation, reducing to M the number of invocations of the learning algorithm, instead of otherwise having to build $K \times M$ base models.

While out-of-bag estimates constitute an helpful tool, their benefits should however be put in balance with the potential decrease of accuracy that the use of bootstrap replicates may induce. As shown experimentally in [Louppe and Geurts, 2012], bootstrapping is in fact rarely crucial for random forests to obtain good accuracy. On the contrary, not using bootstrap samples usually yield better results.

### 4.4.2 *Variable importances*

In most machine learning tasks, the goal is not only to find the most accurate model of the response but also to identify which of the input variables are the most important to make the predictions, e.g., in order to lead to a deeper understanding of the problem under study.

In this context, random forests offer several mechanisms for assessing the *importance* of an input variable, and therefore enhance the interpretability of the model. These are the object of Chapter 6, in which we study variable importance measures and develop original contributions to further improve their understanding.

### 4.4.3 *Proximity measures*

Another helpful built-in feature of tree-based ensemble methods is the *proximity measure* [Breiman, 2002] between two sample points. Formally, the proximity between $(\mathbf{x}_1, y_1)$ and $(\mathbf{x}_2, y_2)$ is defined as the number of times both samples reach the same leaf $t$ within each de-cision tree, normalized by the number of trees in the forest. That is,

$$\text{proximity}(\mathbf{x}_1, \mathbf{x}_2) = \frac{1}{M} \sum_{m=1}^{M} \sum_{t \in \widetilde{\varphi}_{\mathcal{L},\theta_m}} 1(\mathbf{x}_1, \mathbf{x}_2 \in \mathcal{X}_t) \quad (4.38)$$

where $\widetilde{\varphi}_{\mathcal{L},\theta_m}$ denotes the set of terminal nodes in $\varphi_{\mathcal{L},\theta_m}$. The idea is that the proximity measure gives an indication of how close the samples are in the eyes of the random forest [Hastie et al., 2005], even if the data is high-dimensional or involves mixed input variables. When proximity is close to 1, samples propagate into the same leaves and are therefore similar according to the forest. On the other hand, when it is close to 0, samples reach different leaves, suggesting that they are structurally different from each other. The proximity measure depends on both the depth and the number of trees in the forest. When trees are shallow, samples are more likely to end up in the same leaves than when trees are grown more deeply, thereby impacting on the spread of the measure. Likewise, the more trees in the forest, the smoother the measure since the larger the number $M + 1$ of values the proximity measure can take.

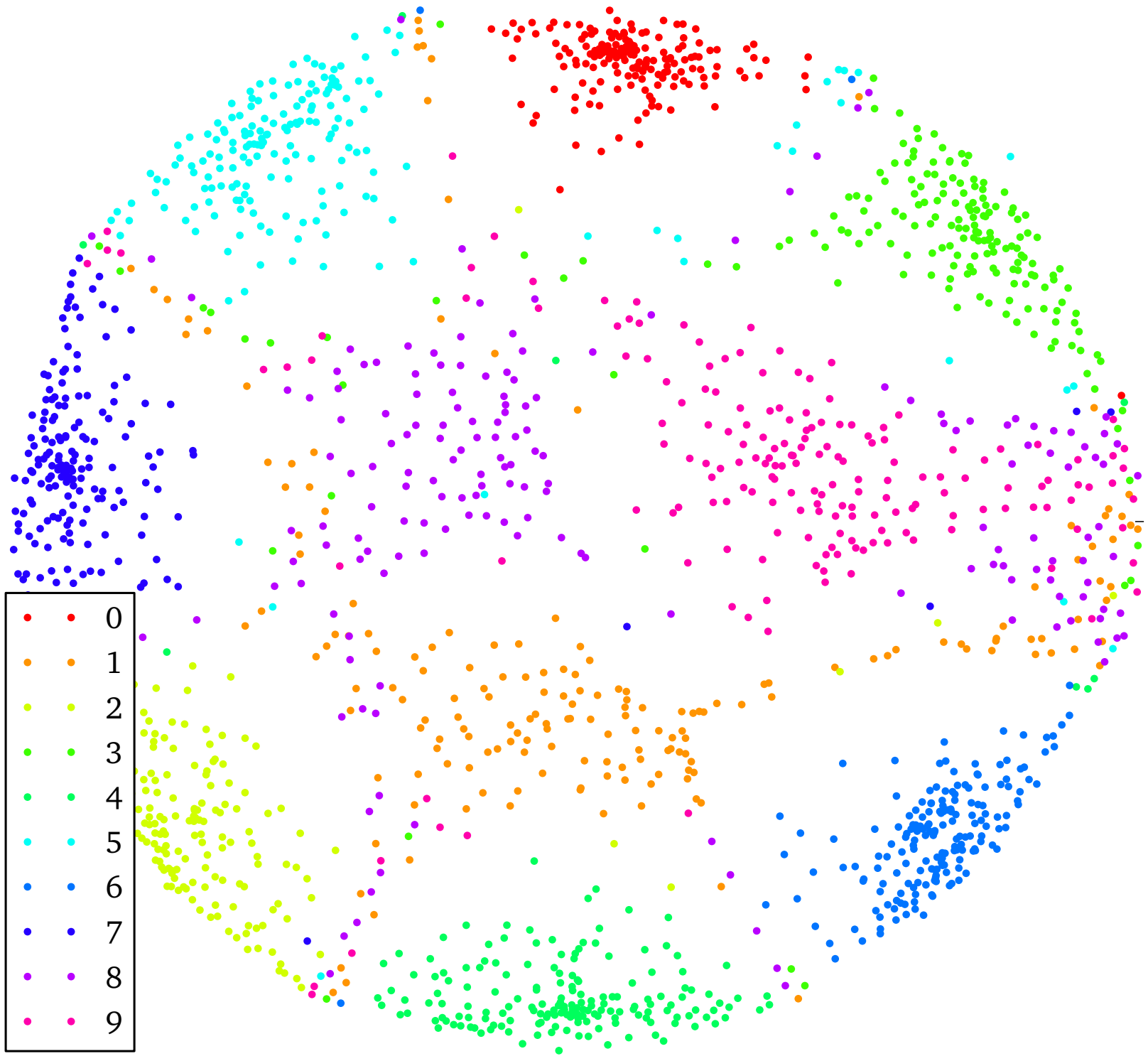

Figure 4.6: Proximity plot for a 10-class handwritten digit classification task.

For exploratory purposes, the $N \times N$ proximity matrix $P$ such that

$$P_{ij} = \text{proximity}(\mathbf{x}_i, \mathbf{x}_j), \quad (4.39)$$

can be used to visually represent how samples are close together with respect to the random forest. Using $1 - P$ as a dissimilarity matrix, the level of similarity between individual samples can be visualized e.g., by projecting them on a d-dimensional space (e.g., on a plane) such that the distances between any pair of samples in that space correspond as best as possible to the dissimilarities in $1 - P$. As an illustrative example, Figure 4.6 represents the proximity matrix learned for a 10-class handwritten digit classification task, as projected on a plane using Multidimensional Scaling [Kruskal, 1964]. Samples from a same class form identifiable clusters, which suggests that they share similar structure (since they end up in the same leaves). The figure also highlights classes for which the random forest makes errors. In this case, digits 1 and 8 are the more dispersed, suggesting high within-class variance, but also overlap the most with samples of other classes, indicating that the random forest fails to identify the true class for these samples.

Additionally, proximity measures can be used for identifying outliers within a learning set $\mathcal{L}$. A sample $\mathbf{x}$ can be considered as an outlier if its average proximity with respect to all other samples is small, which indeed indicates that $\mathbf{x}$ is structurally different from the other samples (since they do not end up in the same leaves). In the same way, in classification, within-class outliers can be identified by computing the average proximity of a sample with respect to all other samples of the same class. Conversely, proximity measures can be used for identifying class prototypes, by considering samples whose proximity with all other samples of the same class is the largest.

An alternative formulation of proximity measures is to consider a random forest as a mapping $\phi : \mathcal{X} \mapsto \mathcal{X}'$ which transforms a sample $\mathbf{x}$ into the (one-hot-encoded) indices of the leaves it ends up in. That is, $\mathbf{x}'_t$ is 1 for all leaves $t$ of the forest in which $\mathbf{x}$ falls in, and 0 for all the others:

$$\phi(\mathbf{x}) = \Big(1(\mathbf{x} \in \mathcal{X}_{t_1}), \dots, 1(\mathbf{x} \in \mathcal{X}_{t_L})\Big) \tag{4.40}$$

where $t_1, \dots, t_L \in \widetilde{\psi}$ denote the leafs of all $M$ trees in the forest $\psi$. In this formalism, the proximity between $\mathbf{x}_1$ and $\mathbf{x}_2$ corresponds to a *kernel* [Scholkopf and Smola, 2001], that can be defined as the normalized dot product of the samples, as represented in $\mathcal{X}'$:

$$\text{proximity}(\mathbf{x}_1, \mathbf{x}_2) = \frac{1}{M} \phi(\mathbf{x}_1) \cdot \phi(\mathbf{x}_2) \tag{4.41}$$

Interestingly, $\phi$ provides a non-linear transformation to a sparse very high-dimensional space, taking somehow into account the structure of the problem. If two samples are structurally similar, then they will end up in the same leafs and their representations in the projected space will be close, even if they may in fact appear quite dissimilar in the original space.

In close connection, [Geurts et al., 2006a] show that a regression tree $\varphi$ can be expressed as a kernel-based model by formulating the prediction $\varphi(\mathbf{x})$ as a scalar product over the input space defined by the normalized characteristic function $\phi'$ of the leaf nodes. That is,

$$\varphi(\mathbf{x}) = \sum_{(\mathbf{x}_i, y_i) \in \mathcal{L}} y_i K_\varphi(\mathbf{x}_i, \mathbf{x}) \tag{4.42}$$

where

$$K_\varphi(\mathbf{x}_i, \mathbf{x}) = \phi'(\mathbf{x}_i) \cdot \phi'(\mathbf{x}), \tag{4.43}$$

$$\phi'(\mathbf{x}) = \Big(\frac{1(\mathbf{x} \in \mathcal{X}_1)}{\sqrt{N_1}}, \dots, \frac{1(\mathbf{x} \in \mathcal{X}_{t_L})}{\sqrt{N_{t_L}}}\Big), \tag{4.44}$$

and where $t_1, \dots, t_L \in \widetilde{\varphi}$ denote the leafs of $\varphi$. Likewise, the formulation can be extended to an ensemble $\psi$ of $M$ decision trees by defining $K_\psi$ as the average kernel over $K_{\varphi_m}$ (for $m = 1, \dots, M$).

From a more practical point of view, such forest-based transforms $\phi$ and $\phi'$ have proven to be helpful and efficient embeddings, e.g., when combined with linear methods or support vector machines [Moosmann et al., 2006; Marée et al., 2013]. In particular, they find useful applications in computer vision for transforming the raw pixel input space into a new feature space hopefully capturing structures and patterns in images.

### 4.4.4 *Missing data*

Because of practical limitations, physical constraints or for privacy reasons, real-world data are often imperfect, erroneous or incomplete. In particular, most machine learning algorithms are often not applicable on data containing missing values because they implicitly assume an ideal scenario in which all values are known for all input variables. Fortunately, random forests offer several mechanisms for dealing with this issue.

*Ternary decision trees.*
: The simplest strategy is to explicitly model missing values in the structure of decision trees by considering ternary splits instead of binary splits. That is, partition $t$ into $t_L$, $t_R$ and $t_M$, such that $t_L$ and $t_R$ are defined from a binary split $s_j^v : X_j \leqslant v$, for all node samples where the value of $X_j$ is known, and such that $t_M$ contain all node samples for which the value of $X_j$ is missing.

*Propagate in both child nodes*
: An alternative strategy is to propagate samples for which the value of the split variable is missing into both the left and right child nodes. Accordingly, samples going into both child nodes should be re-weighted by half their sample weight (see Chapter 5), so that they are not unfairly taken into account more than the other samples.

*Imputation.*

Finally, random forests also offer several mechanisms for imputing missing values. A simple approach, due to [Breiman, 2002], consists first in filling missing values with a rough and inaccurate approximation (e.g., the median). Then build a random forest on the completed data and update the missing values of each sample by the weighted mean value over the samples that are the closest (as defined by the proximity measure). The procedure is then repeated until convergence, typically after 4 to 6 iterations.

Alternatively, missing data imputation can be considered as a supervised learning problem in itself, where the response variable is the input variable for which values are missing. As such, the MissForest algorithm [Stekhoven and Bühlmann, 2012] consists in iteratively building a random forest on the observed parts of the data in order to predict the missing values for a given variable.

## 4.5 Consistency

Despite their extensive use in practice, excellent performance and relative algorithmic simplicity, the mathematical mechanisms that drive random forests are still not well understood. More specifically, the fundamental theoretical question of the consistency (see definitions 2.8 and 2.9) of random forests, i.e., whether convergence towards an optimal model is guaranteed provided an infinitely large learning set, remains an open and difficult problem. In this section, we review theoretical works that have investigated simplified versions of the algorithm, for which the construction procedure is often made data-independent, hence making the theoretical analysis typically more tractable. With the hope that results obtained for these simplified models will provide insights on the mathematical properties of the actual algorithm, the long-term objective of this line of research is usually to prove the consistency of the original Random Forest algorithm [Breiman, 2001], hence bridging the gap between theory and practice.

Breiman et al. [1984]:

Single decision trees are proved to be consistent, both in regression and classification.

Note that these results do not extend to the Random Forest algorithm for the following reasons:

- In single decision trees, the number of samples in terminal nodes is let to become large, while trees in random forests are usually fully developed;

- Single decision trees do not make use of bootstrap sampling;
- The splitting strategy in single decision trees consists in selecting the split that maximizes the Gini criterion. By contrast, in random forests, the splitting strategy is randomized.

Zhao [2000]:

One of the earliest works studying the consistency of ensembles of randomized trees is due to Zhao [2000]. In classification, the author conjectures that PERT is (weakly) consistent, but establishes its strong consistency (Theorem 4.4.2) provided the construction of the trees stops early. More specifically, strong consistency is guaranteed provided:

(a) Trees are grown infinitely deeply while forming leaves with infinitely many node samples, hence making empirical class proportions in terminal nodes converge towards their theoretical counterparts. This is guaranteed, e.g., by stopping the construction when $N_t < N_{min}^{(N)}$, such that $N_{min}^{(N)} \to 0$ and $N \times N_{min}^{(N)} \to \infty$ as $N \to \infty$ (where $N_{min}^{(N)}$ is the value of the $N_{min}$ parameter for a forest grown on a learning set of size N);

(b) The posterior class probabilities induced by the ensemble are all continuous in the input space $\mathcal{X}$.

Given the close formulations of the methods, results from [Zhao, 2000] regarding PERT extend to Extremely Randomized Trees provided the posterior class probabilities are continuous. In Appendix F of [Geurts et al., 2006a], Extremely Randomizes Trees (for $K = 1$ and $M \to \infty$) are shown to be a continuous piecewise multilinear function of its arguments, which should therefore suffice to establish the strong consistency of the method when trees are built totally at random.

Breiman [2004]:

In this work, consistency is studied for a simplified variant of the Random Forest algorithm, assuming (i) no bootstrap sampling, (ii) that variables are selected as split variables with probability $p(m)$ (for $m = 1, \dots, p$), (iii) that splits on relevant variables are set at the midpoint of the values of the selected variable, (iv) that splits on irrelevant variables are set at random points along the values of the selected variable and (v) that trees are balanced. Under these assumptions, it can be shown that this variant reduces to an (adaptive) nearest neighbor algorithm [Lin and Jeon, 2002], for which (weak) consistency conditions are met (both in regression and classification). Additionally, this work studies the bias-variance decomposition of this simplified method and shows that the rate of convergence towards the Bayes error only depends on the number $r$ of relevant variables, hence explaining why random forests work well even with many noise variables.

Biau et al. [2008]:

In binary classification, Biau et al. [2008] proves that if the randomized base models in an ensemble are consistent, then the corresponding majority or soft voting ensembles are also consistent. (This result was later expanded both for multi-class classification [Denil et al., 2013b] and regression [Denil et al., 2013a].)

From this proposition, the consistency of Purely Random Forests [Breiman, 2000] is established. Let us assume that the input space $\mathcal{X}$ is supported on $[0,1]^p$ and that terminal nodes represent hyper rectangles of $[0,1]^p$, called cells, and forming together a partition of $[0,1]^p$. At each step, one of the current terminal nodes $t$ and one the $p$ input variables are chosen uniformly at random. The selected node $t$ is then split along the chosen variable at a random location, along the length of the chosen side in $t$. This procedure is repeated $k \geqslant 1$ times, which amounts to developing trees in random order. In this setting, and similarly to PERT, (strong) consistency of Purely Random Forests is guaranteed whenever $k^{(N)} \to \infty$ and $\frac{k^{(N)}}{N} \to 0$ as $N \to \infty$ (where $k^{(N)}$ is the value of $k$ for a forest grown on learning set of size $N$) – which is equivalent to letting the number of points in terminal nodes grow to infinity.

In Purely Random Forests, let us remark that trees are built in a data-independent manner, without even looking at the samples in $\mathcal{L}$. In this same work, and assuming no bootstrap sampling, the authors show that consistency is however also guaranteed when the position of the cut is chosen in a data-dependent manner, by selecting a random gap between consecutive node samples (ordered along the chosen variable) and then sampling uniformly within the gap.

Biau [2012]:

In this work, the authors more closely approaches the consistency of the actual Random Forest algorithm and prove the consistency of the following variant.

Again, let us assume that the input space $\mathcal{X}$ is supported on $[0,1]^p$ and that terminal nodes represent hyper rectangles of $[0,1]^p$, forming together a partition of the input space. At each step, all current terminal nodes are independently split using one of the $p$ input variables $X_j$, drawn with probability $p_j^{(N)}$, and using as threshold the mid-point of the chosen side in $t$. This procedure is repeated $\lceil \log_2 k \rceil$ times, which amounts to developing trees in breadth-first order until depth $\lceil \log_2 k \rceil$. (Strong) Consistency is then guaranteed whenever $p_j^{(N)} k^{(N)} \to \infty$ (for $j = 1, \dots, p$) and $\frac{k^{(N)}}{N} \to 0$ as $N \to \infty$.

In particular, by properly defining the probabilities $p_j$, this result can be shown to include the situation where, at each node, randomness is introduced by selecting at random a subset of $K$ variables and splitting along the one that maximizes some impurity criterion, like in Random Forest. (Note however that best cut-points coincide with the mid-points only for some probability distributions.) Assuming no bootstrap sampling and provided that two independent datasets are used for evaluating the goodness of the splits and fitting the prediction values at leafs, consistency of the method is then also established.

Interestingly, and corroborating results of [Breiman, 2004], this work also highlights the fact that performance of random forests only depends on the number $r$ of relevant variables, and not on $p$, making the method robust to overfitting.

Denil et al. [2013a]:

Building upon [Biau, 2012], Denil et al. [2013a] narrowed the gap between theory and practice by proving the consistency of the following variant.

For each tree in the forest, the learning set is partitioned randomly into structure points (used for determining splits) and estimation points (used for fitting values at leafs). Again, let us assume that the input space $\mathcal{X}$ is supported on $[0,1]^p$ and that terminal nodes represent hyper rectangles of $[0,1]^p$, forming together a partition of the input space. At each step, current terminal nodes are expanded by drawing at random $\min(1 + \text{Poisson}(\lambda), p)$ variables and then looking for the cut-point that maximizes the impurity criterion, as computed over $m$ randomly drawn structure points. The construction halts when no split leading to child nodes with less than $k$ node samples can be found.

In regression, assuming not bootstrap sampling, (strong) consistency of this variant is guaranteed whenever $k^{(N)} \to \infty$ and $\frac{k^{(N)}}{N} \to 0$ as $N \to \infty$ (where $k^{(N)}$ is the value of $k$ for a forest grown on learning set of size $N$).

Scornet et al. [2014]:

This work establishes the first consistency result for the original Random Forest algorithm. In particular, (strong) consistency is obtained in the context of regression additive models, assuming subsampling without replacement (instead of bootstrap sampling). This work is the first result establishing consistency when (i) splits are chosen in a data-dependent manner and (ii) leafs are not let to grow to an infinite number of node samples.

In conclusions, despite the difficulty of the mathematical analysis of the method, these theoretical works provide together converging arguments all confirming why random forests – including the Random Forest algorithm but also Extremely Randomized Trees – appear to work so well in practice.

Finally, let us complete this review by mentioning consistency results in the case of domain-specific adaptations of random forests, including quantile regression [Meinshausen, 2006], survival analysis [Ishwaran and Kogalur, 2010] and online forest construction [Denil et al., 2013b].

# 5 COMPLEXITY ANALYSIS AND IMPLEMENTATION

**Outline**

In this chapter, we study the computational efficiency of tree-based ensemble methods. In sections 5.1 and 5.2, we derive and discuss the time complexity of random forests, first for building them from data and then for making predictions. In Section 5.3, we discuss technical details that are critical for efficiently implementing random forests. Finally, we conclude in Section 5.4 with benchmarks of the random forest implementation developed within this work and compare our solution with alternative implementations.

## 5.1 COMPLEXITY OF THE INDUCTION PROCEDURE

The dominant objective of most machine learning methods is to find models that maximize accuracy. For models of equivalent performance, a secondary objective is usually to minimize complexity, both functional, as studied in the previous chapters, and computational. With regards to the later, a first and immediate aspect of computational complexity in decision trees and random forests is the *time complexity* for learning models, that is the number of operations required for building models from data.

Given the exponential growth in the number of possible partitions of $N$ samples, we chose in Section 3.6 to restrict splitting rules to binary splits defined on single variables. Not only this is sufficient for reaching good accuracy (as discussed in the previous chapter), it also allows for time complexity to effectively remain within reasonable bounds.

Formally, let $T(N)$ denotes the time complexity for building a decision tree from a learning set $\mathcal{L}$ of $N$ samples. From Algorithm 3.2, $T(N)$ corresponds to the number of operations required for splitting a node of $N$ samples and then for recursively building two sub-trees respectively from $N_{t_L}$ and $N_{t_R}$ samples. Without loss of generality, let us assume that learning samples all have distinct input values, such that it is possible to build a fully developed decision tree where each leaf contains exactly one sample. Under this assumption, determining time complexity therefore boils down to solve the recurrence equation

$$\begin{cases} T(1) = c_1 \\ T(N) = C(N) + T(N_{t_L}) + T(N_{t_R}), \end{cases} \tag{5.1}$$

where $c_1$ is the constant cost for making a leaf node out of a single sample and $C(N)$ is the runtime complexity for finding a split $s$ and then partitioning the $N$ node samples into $t_L$ and $t_R$. In particular, this later operation requires at least to iterate over all $N$ samples, which sets a linear lower bound on the time complexity within a node, i.e., $C(N) = \Omega(N)$. As for finding the split $s$, this can be achieved in several ways, as outlined in the previous chapters for several randomized variants of the induction procedure. As we will see, not only this has an impact on the accuracy of the resulting model, it also drives the time complexity of the induction procedure.

**Big O notations**

Time complexity analyzes the asymptotic behavior of an algorithm with respect to the size $N$ of its input and its hyper-parameters. In this way, big O notations are used to formally express an asymptotic upper bound on the growth rate of the number $f(N)$ of steps in the algorithm. Formally, we write that

$$f(N) = O(g(N)) \text{ if } \exists c > 0, N_0 > 0, \forall N > N_0, f(N) \leqslant cg(N) \quad (5.2)$$

to express that $f(N)$ is asymptotically upper bounded by $g(N)$, up to some negligible constant factor $c$. Similarly, big $\Omega$ notations are used to express an asymptotic lower bound on the growth rate of the number of steps in the algorithm. Formally, we write that

$$f(N) = \Omega(g(N)) \text{ if } \exists c > 0, N_0 > 0, \forall N > N_0, cg(N) \leqslant f(N) \quad (5.3)$$

to express that $f(N)$ is asymptotically lower bounded by $g(N)$. Consequently, if $f(N)$ is both $O(g(N))$ and $\Omega(g(N))$ then $f(N)$ is both lower bounded and upper bounded asymptotically by $g(N)$ (possibly for different constants), which we write using big $\Theta$ notations:

$$f(N) = \Theta(g(N)) \text{ if } f(N) = O(g(N)) = \Omega(g(N)). \quad (5.4)$$

In the original CART induction procedure [Breiman et al., 1984] (Algorithms 3.3 and 3.4), finding a split $s$ requires, for each of the $p$ variables $X_j$ (for $j = 1, \ldots, p$), to sort the values $x_{i,j}$ of all $N$ node samples (for $i = 1, \ldots, N$) and then to iterate over these in order to find the best threshold $v$. Assuming that the value of the impurity criterion can be computed iteratively for consecutive thresholds, the most costly operation is sorting, whose time complexity is at worst $O(N \log N)$ using an optimal algorithm. As a result, the overall within-node complexity is $C(N) = O(pN \log N)$. In a randomized tree as built with the Random Forest algorithm [Breiman, 2001] (RF, Algorithm 4.1), the search of the best split $s$ is carried out in the same way, but only on $K \leqslant p$ of the input variables, resulting in a within-node complexity $C(N) = O(KN \log N)$. In Extremely Randomized Trees [Geurts et al., 2006a] (ETs, Algorithm 4.3), discretization thresholds are drawn at random within the minimum and maximum node sample values of $X_j$, making sort no longer necessary. As such, the within-node complexity reduces to the time required for finding these lower and upper bounds, which can be done in linear time, hence $C(N) = O(KN)$. Finally, in Perfect Random Tree Ensembles [Cutler and Zhao, 2001] (PERT, Algorithm 4.2), cut-points $v$ are set midway between two randomly drawn samples, which can be done in constant time, independently of the number $N$ of node samples. Yet, the within-node complexity is lower bounded by the time required for partitioning the node samples into $t_L$ and $t_R$, hence $C(N) = O(N)$.

Since both CART and PERT can respectively be considered as special cases of RF and ETs with regards to time complexity (i.e., for $K = p$ in CART, for $K = 1$ in PERT), let us consider the overall complexity $T(N)$ when either $C(N) = O(KN \log N)$ or $C(N) = O(KN)$ (for $K = 1, \dots, p$). To make our analysis easier, let us further assume that it exists $c_2, c_3 > 0$ such that $c_2 KN \log N \leqslant C(N) \leqslant c_3 KN \log N$ for all $N \geqslant 1$ (resp. $c_2 KN \leqslant C(N) \leqslant c_3 KN$ for all $N \geqslant 1$). Results presented below extend however to the case where $C(N) = \Theta(KN)$ (resp. $\Theta(KN \log N)$), at the price of a more complicated mathematical analysis. Time complexity is studied in three cases:

*Best case.*
: The induction procedure is the most efficient when node samples can always be partitioned into two balanced subsets of $\frac{N}{2}$ samples.

*Worst case.*
: By contrast, the induction procedure is the least efficient when splits are unbalanced. In the worst case, this results in splits that move a single sample in the first sub-tree and put all $N-1$ others in the second sub-tree.

*Average case.*
: The average case corresponds to the average time complexity, as taken over all possible learning sets $\mathcal{L}$ (of a given size and for a given distribution) and all random seeds $\theta$. Since the analysis of this case is hardly feasible without strong assumptions on the structure of the problem, we consider instead the case where the sizes of the possible partitions of a node are all equiprobable. We believe this is a good enough approximation, that should at least help us derive the order of magnitude of the complexity (instead of deriving it exactly).

Since the induction algorithm looks for splits that most decrease impurity (in particular, as $K \to p$), we assume that balanced partitions of the nodes samples should be more likely than partitions that are unbalanced (such a claim is however highly data dependent). Under this assumption, random forests should therefore approach the best case rather than the average case (for which partitions are equally likely). However, as randomization increases (e.g., as $K \to 1$), splits yield partitions that tend to be more equiprobable, hence approaching the average case. As such, we believe the complexity for the true average case should lie in-between the complexity of the best case and the complexity of the average case as defined above. As for the worst case, it should only arise in exceptional circumstances.

Let us mention that due to the end-cut preference of classical impurity criteria, worst splits are in fact likely to be selected (see [Breiman et al., 1984], Theorem 11.1) if none of the $K$ selected variables are informative with respect to the output. (Note that this effect is less pronounced in ETs because thresholds are drawn at random rather than being optimized locally, as in RF.) In the analysis carried below, intermingling effects due to good splits at the top of trees and such degenerated splits due to end-cut preference when input variables become independent of the output are not explicitly taken into account. Yet, the simplifying assumptions defining our average case appear to constitute a good-enough approximation, as further confirmed in empirical benchmarks in Section 5.4.

MASTER THEOREM

Some of the results outlined in the remaining of this section make use of the Master Theorem [Bentley et al., 1980; Goodrich and Tamassia, 2006], which is recalled below to be self-contained.

**Theorem 5.1.** *Let consider the recurrence equation*

$$\begin{cases} T(1) = c & \text{if} \quad n < d \\ T(n) = aT(\frac{n}{b}) + f(n) & \text{if} \quad n \geqslant d \end{cases} \tag{5.5}$$

*where $d \geqslant 1$ is an integer constant, $a > 0$, $c > 0$ and $b > 1$ are real constants, and $f(n)$ is a function that is positive for $n \geqslant d$.*

*(a) If there is a small constant $\epsilon > 0$, such that $f(n)$ is $O(n^{\log_b a - \epsilon})$, then $T(n)$ is $\Theta(n^{\log_b a})$.*

*(b) If there is a constant $k \geqslant 0$, such that $f(n)$ is $\Theta(n^{\log_b a} \log^k n)$, then $T(n)$ is $\Theta(n^{\log_b a} \log^{k+1} n)$.*

*(c) If there are small constants $\epsilon > 0$ and $\delta < 1$, such that $f(n)$ is $\Omega(n^{\log_b a + \epsilon})$ and $af(\frac{n}{b}) \leqslant \delta f(n)$, for $n \geqslant d$, then $T(n)$ is $\Theta(f(n))$.*

**Theorem 5.2.** *For* $c_2 KN \log N \leqslant C(N) \leqslant c_3 KN \log N$ *(for all* $N \geqslant 1$*), the time complexity for building a decision tree in the best case is* $\Theta(KN \log^2 N)$.

*Proof.* Let us rewrite $T(N)$ as $KT'(N)$ where, in the best case,

$$\begin{cases} T'(1) = c'_1 \\ T'(N) = C'(N) + 2T'(\frac{N}{2}), \end{cases} \tag{5.6}$$

with $c'_1 = \frac{c_1}{K}$ and $C'(N) = \frac{C(N)}{K}$. In this form, the second case of the Master Theorem applies for $f(N) = C'(N) = \Theta(N \log N)$, $a = 2$, $b = 2$ and $k = 1$. Accordingly, $T'(N) = \Theta(N \log^{k+1} N) = \Theta(N \log^2 N)$ and $T(N) = \Theta(KN \log^2 N)$. □

**Theorem 5.3.** *For* $c_2 KN \leqslant C(N) \leqslant c_3 KN$ *(for all* $N \geqslant 1$*), the time complexity for building a decision tree in the best case is* $\Theta(KN \log N)$.

*Proof.* In the best case, $T(N) = KT'(N)$, where

$$\begin{cases} T'(1) = c'_1 \\ T'(N) = C'(N) + 2T'(\frac{N}{2}). \end{cases} \tag{5.7}$$

Again, the second case of the Master Theorem applies, this time for $f(N) = C'(N) = \Theta(N)$, $a = 2$, $b = 2$ and $k = 0$. Accordingly, $T'(N) = \Theta(N \log^{k+1} N) = \Theta(N \log N)$ and $T(N) = \Theta(KN \log N)$. □

**Theorem 5.4.** *For* $c_2 KN \log N \leqslant C(N) \leqslant c_3 KN \log N$ *(for all* $N \geqslant 1$*), the time complexity for building a decision tree in the worst case is* $\Omega(KN^2)$ *and* $O(KN^2 \log N)$.

*Proof.* In the worst case, $T(N) = KT'(N)$, where

$$\begin{cases} T'(1) = c'_1 \\ T'(N) = C'(N) + T'(1) + T'(N-1). \end{cases} \tag{5.8}$$

Let us first consider an upper bound on $T'(N)$. By expanding the recurrence, we have

$$\begin{aligned} T'(N) &\leqslant c_3 N \log N + c'_1 + T'(N-1) \\ &= \sum_{i=1}^{N} c'_1 + c_3 i \log i \\ &= c'_1 N + \sum_{i=1}^{N} c_3 i \log i \\ &\leqslant c'_1 N + c_3 \log N \sum_{i=1}^{N} i \\ &= c'_1 N + c_3 \log N \frac{N(N+1)}{2} \end{aligned}$$

$$= O(N^2 \log N). \tag{5.9}$$

Similarly, a lower bound on $T'(N)$ can be derived in the following way:

$$\begin{aligned}
T'(N) &\geqslant c_2 N \log N + c_1' + T'(N-1) \\
&= \sum_{i=1}^{N} c_1' + c_2 i \log i \\
&= c_1' N + \sum_{i=1}^{N} c_2 i \log i \\
&\geqslant c_1' N + c_2 \sum_{i=2}^{N} i \\
&= c_1' N + c_2 \frac{N(N+1)}{2} - c_2 \\
&= \Omega(N^2).
\end{aligned} \tag{5.10}$$

As such, $T(N) = \Omega(KN^2)$ and $O(KN^2 \log N)$ in the worst case. □

Let us remark that the lower and upper bounds derived for $T(N)$ in Theorem 5.4 are not tight. Nevertheless, given the fact that

$$\sum_{i=1}^{N} i \log i = \log(H(N)), \tag{5.11}$$

where $H(N)$ is the hyperfactorial function, complexity could be re-expressed tightly as $T(N) = \Theta(K \log(H(N)))$.

**Theorem 5.5.** *For $c_2 KN \leqslant C(N) \leqslant c_3 KN$ (for all $N \geqslant 1$), the time complexity for building a decision tree in the worst case is $\Theta(KN^2)$.*

*Proof.* In the worst case, $T(N) = KT'(N)$, where

$$\begin{cases} T'(1) = c_1' \\ T'(N) = C'(N) + T'(1) + T'(N-1). \end{cases} \tag{5.12}$$

By expanding the recurrence, we have

$$\begin{aligned}
T'(N) &\leqslant c_3 N + c_1' + T'(N-1) \\
&= c_1' + \sum_{i=2}^{N} (c_1' + c_3 i) \\
&= c_1' N + \frac{c_3}{2}(N^2 + 3N - 4) \\
&= O(N^2).
\end{aligned} \tag{5.13}$$

In the exact same way, $T'(N)$ can be lower bounded by $c_2 N + c_1' + T'(N-1)$ and shown to be $\Omega(N^2)$. As a result, $T(N) = \Theta(KN^2)$ in the worst case. □

**Theorem 5.6.** *For* $c_2 KN \log N \leqslant C(N) \leqslant c_3 KN \log N$ *(for all* $N \geqslant 1$*), the time complexity for building a decision tree in the average case is* $\Theta(KN \log^2 N)$.

*Proof.* In the average case, $T(N) = KT'(N)$, where

$$\begin{cases} T'(1) = c'_1 \\ T'(N) = C'(N) + \frac{1}{N-1} \sum_{i=1}^{N-1} (T'(i) + T'(N-i)). \end{cases} \tag{5.14}$$

Let us first consider an upper bound on $T'(N)$ by assuming that $C'(N) = c_3 N \log N$. By symmetry and by multiplying both sides of the last equation by $(N-1)$, it comes

$$(N-1)T'(N) = (N-1)(c_3 N \log N) + 2 \sum_{i=1}^{N-1} T'(i). \tag{5.15}$$

For $N \geqslant 3$, substituting $N$ with $N-1$ similarly yields

$$(N-2)T'(N-1) = (N-2)(c_3(N-1)\log(N-1)) + 2 \sum_{i=1}^{N-2} T'(i). \tag{5.16}$$

Subtracting Equation 5.16 from 5.15, it comes after simplifications and division of both sides by $N(N-1)$:

$$\frac{T'(N)}{N} = \frac{T'(N-1)}{N-1} + c_3 \frac{2}{N} \log(N-1) + c_3 \log \frac{N}{N-1}. \tag{5.17}$$

Let us now introduce $S(N) = \frac{T'(N)}{N}$. From the last equation, it comes

$$\begin{aligned} S(N) &= S(N-1) + c_3 \frac{2}{N} \log(N-1) + c_3 \log \frac{N}{N-1} \\ &= c'_1 + c_3 \sum_{i=2}^{N} \frac{2}{i} \log(i-1) + \log \frac{i}{i-1} \\ &= c'_1 + c_3 \log N + 2c_3 \sum_{i=2}^{N} \frac{1}{i} \log(i-1) \\ &\leqslant c'_1 + c_3 \log N + 2c_3 \log N \sum_{i=2}^{N} \frac{1}{i} \\ &= c'_1 + c_3 \log N + 2c_3 \log N (H_N - 1) \\ &= O(H_N \log N) \end{aligned} \tag{5.18}$$

where $H_N$ is the N-th harmonic number. Using

$$H_N \approx \log N + \gamma + \frac{1}{2N} + O(\frac{1}{N^2}) \tag{5.19}$$

as approximation (where $\gamma$ is the Euler-Mascheroni constant), we have

$$T'(N) = O(NH_N \log N)$$

$$= O(N(\log N + \gamma + \frac{1}{2N} + \frac{1}{N^2}) \log N)$$
$$= O(N \log^2 N).$$

Given Theorem 5.2, we also know that $T'(N)$ cannot be faster than $\Omega(N \log^2 N)$, thereby setting a lower bound on the complexity of the average case.

As a result, $T(N) = \Theta(KN \log^2 N)$ in the average case. □

**Theorem 5.7.** *For $c_2KN \leqslant C(N) \leqslant c_3KN$ (for all $N \geqslant 1$), the time complexity for building a decision tree in the average case is $\Theta(KN \log N)$.*

*Proof.* In the average case, $T(N) = KT'(N)$, where

$$\begin{cases} T'(1) = c'_1 \\ T'(N) = C'(N) + \frac{1}{N-1} \sum_{i=1}^{N-1} T'(i) + T'(N-i). \end{cases} \tag{5.20}$$

Let us first consider an upper bound on $T'(N)$ by assuming that $C'(N) = c_3N$. By symmetry and by multiplying both sides of the last equation by $(N-1)$, it comes

$$(N-1)T'(N) = c_3N(N-1) + 2\sum_{i=1}^{N-1} T'(i). \tag{5.21}$$

For $N \geqslant 3$, substituting $N$ with $N-1$ similarly yields

$$(N-2)T'(N-1) = c_3(N-1)(N-2) + 2\sum_{i=1}^{N-2} T'(i). \tag{5.22}$$

Subtracting Equation 5.22 from 5.21, it comes after simplifications and division of both sides by $N(N-1)$:

$$\frac{T'(N)}{N} = \frac{T'(N-1)}{N-1} + \frac{2c_3}{N}. \tag{5.23}$$

Let us now introduce $S(N) = \frac{T'(N)}{N}$. From the last equation, it comes

$$\begin{aligned} S(N) &= S(N-1) + \frac{2c_3}{N} \\ &= c'_1 + 2c_3 \sum_{i=2}^{N} \frac{1}{i} \\ &= O(2H_N) \end{aligned} \tag{5.24}$$

where $H_N$ is the N-th harmonic number. Using

$$H_N \approx \log N + \gamma + \frac{1}{2N} + O(\frac{1}{N^2}) \tag{5.25}$$

as approximation (where $\gamma$ is the Euler-Mascheroni constant), we have

$$T'(N) = O(2H_N N)$$

$$= O(2N \log N + 2N\gamma + 1 + \frac{2}{N})$$
$$= O(N \log N). \quad (5.26)$$

In the exact same way, when assuming that $C'(N) = c_2 N$, $T'(N)$ can be shown to be lower bounded by $\Omega(N \log N)$. As a result, $T(N) = \Theta(KN \log N)$ in the average case. □

From Theorems 5.2-5.7, the total time complexity for building an ensemble of $M$ randomized trees can finally be derived. As summarized in Table 5.1, complexity remains within polynomial time for all variants studied in this work. In the best case, complexity is linear with the number of split variables considered at each node (i.e., $O(K)$) and linearithmic or quasilinear with the number of (unique) samples effectively used to build each tree (i.e., $O(N \log N)$ or $O(N \log^2 N)$). In the very worst case however, this later dependency becomes quadratic (i.e., $O(N^2)$ or $O(N^2 \log N)$), which might greatly affect performance in the context of large datasets. Reassuringly, the analysis of the average case shows that pathological cases are not dominant and that, on average, for all methods, complexity behaves as in the best case.

As expected, the complexity of Bagging [Breiman, 1994] is about $M$ times as large as the complexity of building a single decision tree. Yet, by taking into account the fact that bagged decision trees are built on bootstrap replicates $\mathcal{L}^m$ that contain about 63.2% of unique samples, complexity can be expressed with respect to $\widetilde{N} = 0.632N$ instead of $N$. From an implementation point of view, repeated occurrences of the same input vector can indeed be accounted for using sample weights for the evaluation of the impurity criterion, thereby simulating a learning set of $N$ samples from $\widetilde{N}$ unique samples while reducing the effective size of the learning set. Assuming close constant factors, building a single bagged decision tree is therefore asymptotically

$$\lim_{N \to \infty} \frac{N \log^2 N}{0.632N \log^2 0.632N} = 1.582 \quad (5.27)$$

times faster than building a regular decision tree. The complexity of RF is identical to Bagging, except that the dependency to $p$ becomes a dependency to $K$, resulting in an average speedup of $\frac{p}{K}$. In other words, not only choosing $K \leqslant p$ improves accuracy, it also significantly decreases the time required for building trees in a forest. In ETs, due to the fact that samples are no longer required to be sorted at each node, the dependency to $N$ becomes linearithmic instead of quasilinear. With respect to RF, speedup is asymptotically $O(\log N)$ in the best and average cases. Finally, PERT shows to be the fastest method of all, which is due to the fact that only a single split variable is considered at each split. For $K = 1$ however, ETs and PERT have identical time complexity. Note that the analysis presented here is only valid asymptotically. In practice, constant factors might lead to different observed results, though they should not significantly deviate from our conclusions if algorithms are all implemented from a common code-base.

| | *Best case* | *Worst case* | *Average case* |
|---|---|---|---|
| CART | $\Theta(pN\log^2 N)$ | $O(pN^2\log N)$ | $\Theta(pN\log^2 N)$ |
| Bagging | $\Theta(Mp\widetilde{N}\log^2 \widetilde{N})$ | $O(Mp\widetilde{N}^2\log \widetilde{N})$ | $\Theta(Mp\widetilde{N}\log^2 \widetilde{N})$ |
| RF | $\Theta(MK\widetilde{N}\log^2 \widetilde{N})$ | $O(MK\widetilde{N}^2\log \widetilde{N})$ | $\Theta(MK\widetilde{N}\log^2 \widetilde{N})$ |
| ETs | $\Theta(MKN\log N)$ | $\Theta(MKN^2)$ | $\Theta(MKN\log N)$ |
| PERT | $\Theta(MN\log N)$ | $\Theta(MN^2)$ | $\Theta(MN\log N)$ |

Table 5.1: Time complexity for building forests of $M$ randomized trees. $N$ denotes the number of samples in $\mathcal{L}$, $p$ the number of input variables and $K$ the number of variables randomly drawn at each node. $\widetilde{N} = 0.632N$, due to the fact that bootstrap samples draw, on average, 63.2% of unique samples.

## 5.2 COMPLEXITY OF THE PREDICTION PROCEDURE

A second facet of computational complexity in random forests is the time required for making predictions. For a single test point, this represents the number of operations for traversing each of the $M$ trees, from the root to one of the leaves. As such, computational complexity is directly related to the average depth of the leaves reached by the test samples. Assuming that both the learning set and the test set are drawn from the same probability distribution, the complexity analysis of the prediction procedure therefore reduces to the analysis of the average depth of the leaves induced by a learning set $\mathcal{L}$.

As in Section 5.1, let us assume a fully developed decision tree, where each leaf contains exactly one sample from $\mathcal{L}$. Likewise let us derive the average depth $D(N)$ of the leaves in the best, worst and average cases.

**Theorem 5.8.** *In the best case, the average depth of the leaves is* $\Theta(\log N)$.

*Proof.* For a decision tree which is perfectly balanced, node samples are recursively split into two subsets of $\frac{N_t}{2}$ samples at each node, which leads to the following recurrence equation in the best case:

$$\begin{cases} D(1) = 1 \\ D(N) = 1 + \frac{1}{2}D(\frac{N}{2}) + \frac{1}{2}D(\frac{N}{2}) = 1 + D(\frac{N}{2}). \end{cases} \tag{5.28}$$

With this equation, the second case of the Master Theorem applies for $f(N) = 1$, $a = 1$, $b = 2$ and $k = 0$. Accordingly, $D(N) = \Theta(\log^{k+1} N) = \Theta(\log N)$. □

**Theorem 5.9.** *In the worst case, the average depth of the leaves is* $\Theta(N)$.

*Proof.* In the worst case, the recurrence equation of the average depth of the leaves is:

$$\begin{cases} D(1) = 1 \\ D(N) = 1 + \frac{1}{N}D(1) + \frac{N-1}{N}D(N-1). \end{cases} \tag{5.29}$$

Let us now introduce $S(N) = ND(N)$. From the previous equation, it comes

$$\begin{aligned} S(N) &= N + 1 + S(N-1) \\ &= 1 + \sum_{i=2}^{N}(i+1) \\ &= \frac{N^2}{2} + \frac{3N}{2} - 1. \end{aligned} \tag{5.30}$$

Since $D(N) = \frac{S(N)}{N}$, we have $D(N) = \frac{1}{2}(N - \frac{2}{N} + 3) = \Theta(N)$. □

**Theorem 5.10.** *In the average case, the average depth of the leaves is* $\Theta(\log N)$.

*Proof.* In the average case, the recurrence equation of the average depth of the leaves is:

$$\begin{cases} D(1) = 1 \\ D(N) = 1 + \frac{1}{N-1}\sum_{i=1}^{N-1}(\frac{i}{N}D(i) + \frac{N-i}{N}D(N-i)). \end{cases} \tag{5.31}$$

By symmetry and by multiplying both sides by $N(N-1)$, it comes

$$N(N-1)D(N) = N(N-1) + 2\sum_{i=1}^{N-1} iD(i). \tag{5.32}$$

For $N \geqslant 3$, substituting $N$ with $N-1$ similarly yields

$$(N-1)(N-2)D(N-1) = (N-1)(N-2) + 2\sum_{i=1}^{N-2} iD(i). \tag{5.33}$$

Subtracting Equation 5.33 from 5.32, it comes after simplifications and division of both sides by $N(N-1)$

$$\begin{aligned} D(N) &= D(N-1) + \frac{2}{N} \\ &= 1 + 2\sum_{i=2}^{N}\frac{1}{i} \\ &= 2H_N - 1 \end{aligned} \tag{5.34}$$

where $H_N$ is the N-th harmonic number. Using

$$H_N \approx \log N + \gamma + \frac{1}{2N} + O(\frac{1}{N^2}) \tag{5.35}$$

as approximation (where $\gamma$ is the Euler-Mascheroni constant), we have $D(N) = \Theta(\log N)$. □

From Theorems 5.8-5.10, the total time complexity for computing predictions out of a forest of $M$ trees is therefore $\Theta(M \log N)$ in the best case and $\Theta(MN)$ is the very worst case. In accordance with previous results, the analysis of the average case shows however that pathological cases are not dominant and that, on average, complexity behaves once again as in the best case.

**Randomized trees are not equiprobable**

When trees are built totally at random, that is when all sizes of partitions are equiprobable, the analysis of the average case shows that the average depth of the leaves remains logarithmic with size of the learning set. Assuming unique input values for all samples and considering the special case $p = 1$, the random construction of a decision tree is in fact equivalent to the insertion in random order of $N$ unique keys into a binary search tree. As shown exactly in [Sedgewick and Flajolet, 2013], the average depth of a binary search tree built from equiprobable permutations of $N$ unique keys is $O(\log N)$, which corroborates our results. By contrast, if trees were drawn uniformly at random from the set of all possible trees (i.e., in the case of *Catalan* trees), then the average depth can be shown to be $O(\sqrt{N})$. The fundamental difference is that binary search trees built from random permutations are *not* all equiprobable. There may exist several permutations mapping to the same binary search tree. In particular, short trees are more likely to occur than degeneratedly deep trees. Accordingly, decision trees, even built totally at random, are also not all equiprobable. Because of the recursive construction procedure, deep degenerated decision trees are less likely than shorter decision trees.

## 5.3 IMPLEMENTATION

Implementing decision trees and random forests involves many issues that are easily overlooked if not considered with care. In this section, we describe the random forest implementation that has been developed in this work and deployed within the Scikit-Learn machine learning library [Pedregosa et al., 2011]. *The first part of this section is based on previous work published in [Buitinck et al., 2013].*

### 5.3.1 *Scikit-Learn*

The Scikit-Learn library provides a comprehensive suite of machine learning tools for Python. It extends this general-purpose programming language with machine learning operations: learning algorithms, preprocessing tools, model selection procedures and a composition mechanism to produce complex machine learning work-flows. The ambition of the library is to provide efficient implementations of well-established algorithms within a programming environment that is accessible to non-experts and reusable in various scientific areas. The library is distributed under the simplified BSD license to encourage its use in both academia and industry.

Scikit-Learn is designed to tie in with the set of numeric and scientific packages centered around the NumPy [Oliphant, 2007] and SciPy [Van der Walt et al., 2011] libraries. NumPy augments Python with a contiguous numeric array datatype and fast array computing primitives, while SciPy extends it further with common numerical operations, either by implementing these in Python/NumPy or by wrapping existing C/C++/Fortran code. The name "scikit" derives from "SciPy toolkit" and is shared with *scikit-image*. IPython [Perez and Granger, 2007] and Matplotlib [Hunter, 2007] complement SciPy to provide a MATLAB-like working environment suited for scientific use.

Since 2007, Scikit-Learn has been developed by more than a dozen core developers, mostly researchers in fields ranging from neuroscience to astrophysics. It also benefits from occasional contributors proposing small bug-fixes or improvements. Development proceeds on GitHub[1], a platform which greatly facilitates this kind of collaboration. Because of the large number of developers, emphasis is put on keeping the project maintainable. In particular, code must follow specific quality guidelines, such as style consistency and unit-test coverage. Documentation and examples are required for all features, and major changes must pass code review by at least two other developers. The popularity of the project can be gauged from various indicators such as the hundreds of citations in scientific publications, successes in various machine learning challenges (e.g., Kaggle), and statistics derived from our repositories and mailing list. At the time of writing[2], the project is watched by 3,445 people and forked 1,867 times on GitHub; the mailing list receives more than 300 mails per month; version control logs show more than 200 unique contributors to the code-base and the online documentation receives 37,000 unique visitors and 295,000 pageviews per month.

Our implementation guidelines emphasize writing efficient but readable code. In particular, we focus on making the code-base maintainable and understandable in order to favor external contributions. Whenever practical, algorithms implemented in Scikit-Learn are written in Python, using NumPy vector operations for numerical work. This allows the code to remain concise, readable and efficient. For critical algorithms that cannot be easily and efficiently expressed as NumPy operations, we rely on Cython [Behnel et al., 2011] to achieve competitive performance and scalability. Cython is a compiled programming language that extends Python with static typing. It pro-

1 `https://github.com/scikit-learn`
2 July 2014.

duces efficient C extension modules that are importable from the Python run-time system. Examples of Cython code in Scikit-Learn are stochastic gradient descent for linear models, some clustering algorithms, and decision trees.

#### 5.3.1.1 *Data*

Machine learning revolves around data, so good data structures are paramount to designing software for it. In most tasks, data is modeled by a set of $p$ numerical variables, so that a single *sample* is a vector $\mathbf{x} \in \mathbb{R}^p$. A collection of such samples is naturally regarded as the rows in a matrix $\mathbf{X}$ of size $N \times p$. In the common case of supervised learning (classification, regression), we have an additional vector $\mathbf{y}$ of length $N$ at training time and want an algorithm to produce such a $\mathbf{y}$ for new data.

Scikit-Learn's data representation is kept as close as possible to this matrix formulation: datasets are encoded as two-dimensional NumPy arrays or SciPy sparse matrices [Van der Walt et al., 2011], while target vectors are flat arrays of numbers or strings. While these may seem rather unsophisticated when compared to more object-oriented constructs, such as the ones used by Weka [Hall et al., 2009], they allow us to rely on efficient NumPy and SciPy vector operations while keeping the code close to the textbook. Given the pervasiveness of NumPy and SciPy in many other scientific Python packages, many scientific users of Python will already be familiar with these data structures, and a collection of available data loading and conversion tools facilitate interoperability. For tasks where the inputs are text files or semi-structured objects, we provide *vectorizer* objects that efficiently convert such data to the NumPy or SciPy formats.

The public interface is oriented towards processing a batch of samples, rather than a single sample, in each API call. While classification and regression algorithms can make predictions for single samples, Scikit-Learn objects are not optimized for this use case. (The online learning algorithms in the library are intended to take mini-batches.) Batch processing makes optimal use of NumPy and SciPy by preventing the overhead inherent to Python function calls or due to per-element dynamic type checking. Although this might seem to be an artifact of the Python language, and therefore an implementation detail that leaks into the API, we argue that APIs should be designed so as not to tie a library to a suboptimal implementation strategy. Batch processing also enables fast implementations in lower-level languages, where memory hierarchy effects and the possibility of internal parallelization come into play.

#### 5.3.1.2 *Estimators*

The *estimator* interface is at the core of the library. It defines instantiation mechanisms of objects and exposes a `fit` method for learning a model from training data. All supervised and unsupervised learning algorithms (e.g., for classification, regression or clustering) are offered as objects implementing this interface. Machine learning tasks like feature extraction and selection are also provided as estimators.

Estimator initialization and actual learning are strictly separated, in a way that is similar to partial function application: an estimator is initialized from a set of named hyper-parameter values (e.g., the number of trees in a forest) and can be considered a function that maps these values to actual learning algorithms. The constructor does not see any actual data. All it does is attach the given parameters to the object. For the sake of model inspection, hyper-parameters are set as public attributes, which is especially important in model selection settings. Default hyper-parameter values are provided for all built-in estimators. These defaults are set to be relevant in many common situations in order to make estimators effective *out-of-the-box*.

Actual learning is performed by the `fit` method. This method is called with training data (e.g., supplied as two arrays `X_train` and `y_train` in supervised learning estimators). Its task is to run a learning algorithm and to determine model-specific parameters from the training data and set these as attributes on the estimator object. As a convention, the parameters learned by an estimator are exposed as public attributes with names suffixed with a trailing underscore (e.g., `coef_` for the learned coefficients of a linear model), again to facilitate model inspection. In the partial application view, `fit` is a function from data to a model of that data. It always returns the estimator object it was called on, which now serves as a model and can be used to perform predictions or transformations of new data.

The choice to let a single object serve dual purpose as estimator and model has been driven by usability and technical considerations. Having two coupled instances (an estimator object used as a factory, and a model object produced by the estimator) makes a library harder to learn and use. From the developer point of view, decoupling estimators from models would create parallel class hierarchies and increases the overall maintenance burden. A good reason for decoupling would be to make it possible to ship a model to a new environment where the full library cannot be installed. However, our inspectable setup where model parameters are documented public attributes and prediction formulas follow standard textbooks, goes a long way in solving this problem.

To illustrate the initialize-fit sequence, let us consider a supervised learning task using a single decision tree. Given the API defined above, solving this problem is as simple as the following example.

```python
# Import the tree module
from sklearn.tree import DecisionTreeClassifier
# Instantiate and set hyper-parameters
clf = DecisionTreeClassifier(min_samples_split=5)
# Learn a model from data
clf.fit(X_train, y_train)
```

In this snippet, a `DecisionTreeClassifier` estimator is first initialized by setting the `min_samples_split` hyper-parameter to 5 (See Section 3.5). Upon calling `fit`, a model is learned from the training arrays `X_train` and `y_train`, and stored on the object for later use. Since all estimators share the same interface, using a different learning algorithm is as simple as replacing the constructor (the class name). To build a random forest on the same data, one would simply replace `DecisionTreeClassifier(min_samples_split=5)` in the snippet above by `RandomForestClassifier()`.

In Scikit-Learn, classical learning algorithms are not the only objects to be implemented as estimators. For example, preprocessing routines (e.g., scaling of features) or feature extraction techniques (e.g., vectorization of text documents) also implement the *estimator* interface. Even stateless processing steps, that do not require the `fit` method to perform useful work, implement the estimator interface. This design pattern is of prime importance for consistency, composition and model selection reasons, as further illustrated in [Buitinck et al., 2013].

#### 5.3.1.3 *Predictors*

The *predictor* interface extends the notion of an estimator by adding a `predict` method that takes an array `X_test` and produces predictions for `X_test`, based on the learned parameters of the estimator (we call the input to `predict` "`X_test`" in order to emphasize that `predict` generalizes to new data). In the case of supervised learning estimators, this method returns the predicted labels or values computed by the model:

```python
# Make predictions on new data
y_pred = clf.predict(X_test)
```

Apart from `predict`, predictors may also implement methods that quantify the confidence of predictions. In the case of linear models, the `decision_function` method returns the distance of samples to the separating hyperplane. Some predictors also provide a `predict_proba` method which returns class probability estimates.

Finally, supervised predictors provide a `score` function to assess their performance on a batch of input data. This method takes as input arrays `X_test` and `y_test` and typically computes the coefficient of determination between `y_test` and `predict(X_test)` in regression, or the accuracy in classification. The only requirement is that the `score` method return a value that quantifies the quality of its predictions (the higher, the better). An unsupervised estimator may also expose a `score` function to compute, for instance, data likelihood under its model.

#### 5.3.1.4 *API for random forests*

Scikit-Learn provides efficient implementations of decision trees and random forests, all offered as objects implementing the estimator and predictor interfaces presented above. Most of the hyper-parameters described in this work are supported.

`DecisionTreeClassifier`, `DecisionTreeRegressor`:
: Implement single decision trees [Breiman et al., 1984], as described in Chapter 3.

`BaggingClassifier`, `BaggingRegressor`:
: Implement Bagging [Breiman, 1994], Random Subspaces [Ho, 1998] and Ensembles of Random Patches [Louppe and Geurts, 2012], as described in Chapters 4 and 8.

`RandomForestClassifier`, `RandomForestRegressor`:
: Implement Random Forest [Breiman, 2001], as described in Chapter 4.

`ExtraTreesClassifier`, `ExtraTreesRegressor`:
: Implement Extremely Randomized Trees [Geurts et al., 2006a], as described in Chapter 4.

### 5.3.2 *Internal data structures*

Among all possible ways of representing a decision tree, one of the simplest and most direct representations is to adopt an object-oriented approach. In this paradigm, a decision tree is naturally represented as a hierarchy of high-level objects, where each object corresponds to a node of the tree and comprises attributes referencing its children or storing its split and value. Such a representation would make for a correct, intuitive and flexible implementation but may in fact not be the most appropriate when aiming for high-performance. One of the biggest issues indeed is that object-oriented programming usually fragments complex and nested objects into non-contiguous memory blocks, thereby requiring multiple levels of indirection for traversing the structure. In particular, this design can easily impair computing times in performance-critical applications, e.g., by not making it possible to leverage CPU cache or pre-fetching mechanisms. At the price of less abstraction and flexibility, we adopt instead in this work a compact low-level representation of decision trees, allowing us for a fine-grained and complete control over memory management and CPU mechanisms.

Let `t` $\in \{0, \dots, T-1\}$ denote unique node identifiers and let `t` $= 0$ corresponds to the identifier of the root node. The data structure that we propose for representing decision trees consists in using contiguous arrays for simultaneously storing the content of all nodes, as defined below:

`left_child` (array of $T$ integers):
: Store the node identifier `left_child[t]` $\in \{0, \dots, T-1\}$ of the left child of node `t`.

`right_child` (array of $T$ integers):
: Store the node identifier `right_child[t]` $\in \{0, \dots, T-1\}$ of the right child of node `t`.

`feature` (array of $T$ integers):
: Store the identifier `feature[t]` $\in \{1, \dots, p\}$ of the variable used for splitting node `t`.

`threshold` (array of $T$ reals):
: Store the cut-off value `threshold[t]` $\in \mathbb{R}$ used for splitting node `t`.

`impurity` (array of $T$ reals):
: Store the impurity $i(t)$ at node `t`, as computed on the learning set $\mathcal{L}$.

`n_samples` (array of $T$ integers or reals):
: Store the (weighted) number `n_samples[t]` of learning samples reaching node `t`.

`value` (array of $T \times J$ reals or $T$ reals):
: Store the class probability estimates (i.e., the number of samples of each class) or the mean regression values, as computed from the learning samples reaching node `t`.

As an example, Table 5.2 illustrates the array representation of the decision tree shown in Figure 5.2, as built from the learning set $\mathcal{L}$ of Figure 5.1. Internal nodes ($t_0$, $t_2$, $t_3$ and $t_6$) are nodes for which the corresponding values in `left_child`, `right_child`, `feature` and `threshold` are not empty, while leaves ($t_1$, $t_4$, $t_7$ and $t_8$) correspond to nodes for which these values are not defined. In the case of classification, the `value` array contains the number of samples of each class for each node. From these, class probability estimates $p_{\mathcal{L}}(Y = c|t)$ can be computed by dividing `value[t][c]` by the number of samples `n_samples[t]` in $t$. In the case of regression, `value[t]` would contain the average output value for the samples in $t$. Finally, let us note that storing node impurities in the `impurity` array is not strictly necessary, since these values are only used during construction. Yet, storing them can prove to be valuable, e.g., for later inspection of the decision tree or for computing variable importances (see Chapter 6) more readily.

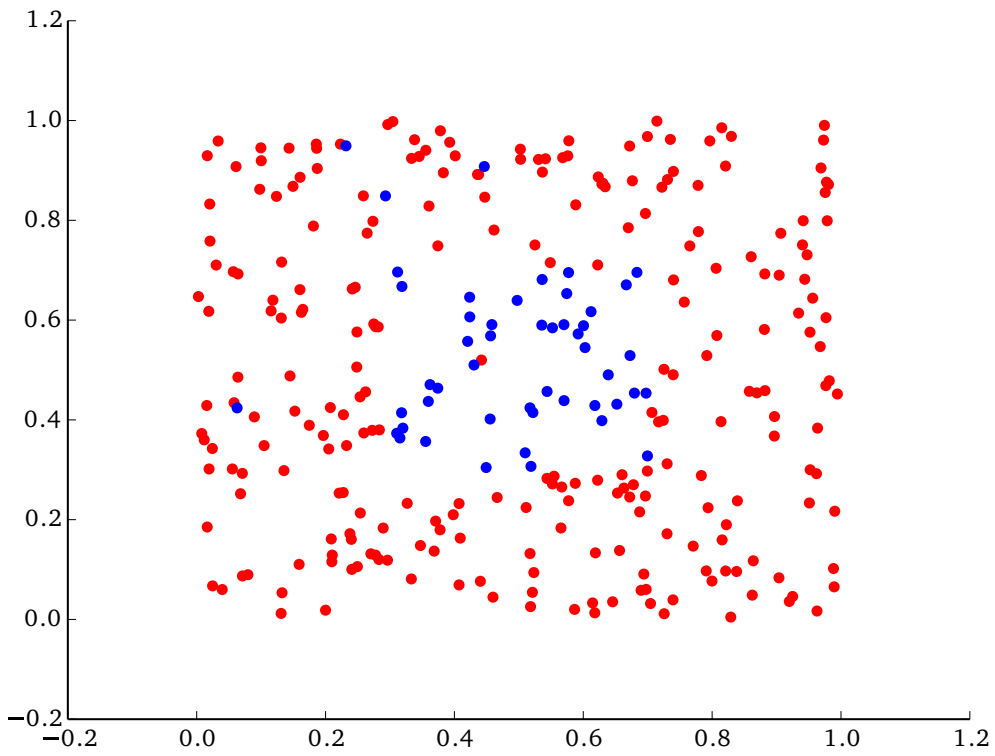

Figure 5.1: Binary classification task.

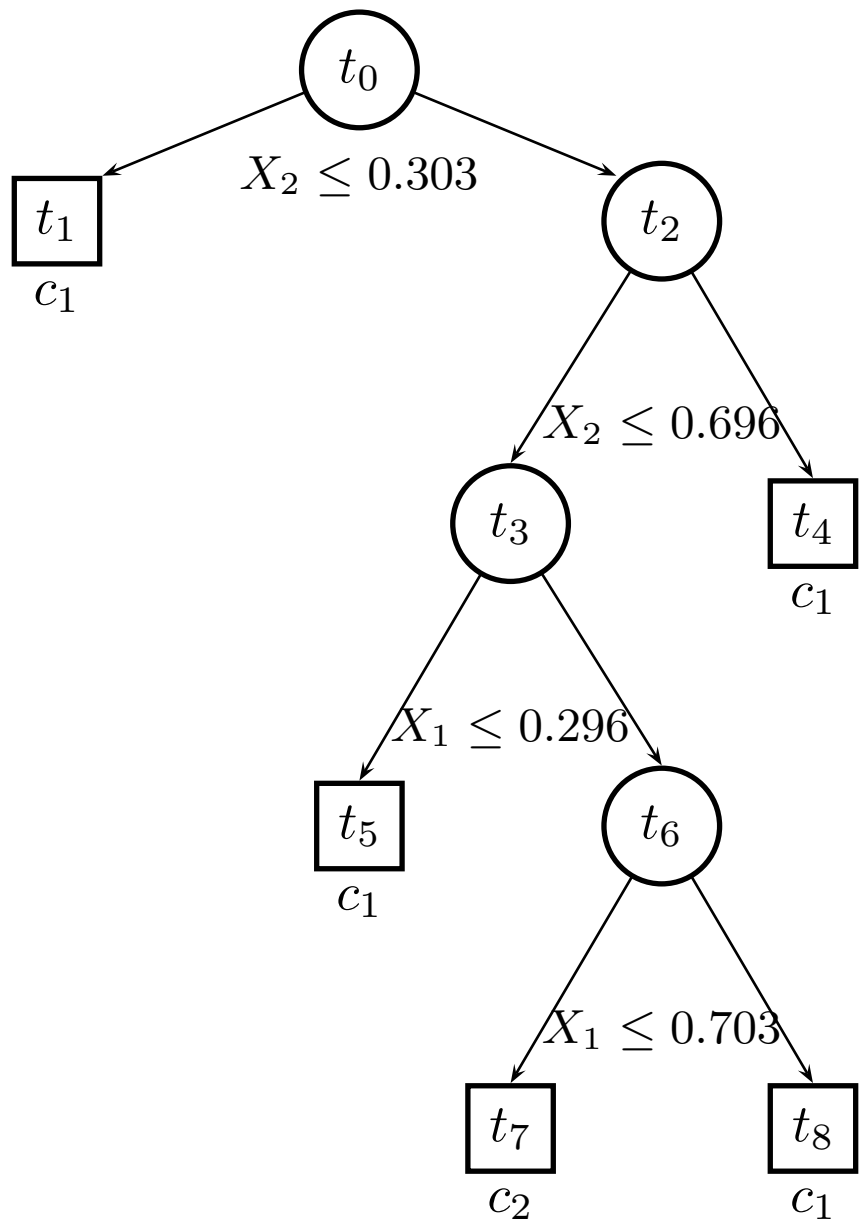

Figure 5.2: Decision tree learned from Figure 5.1.

| t | 0 | 1 | 2 | 3 | 4 | 5 | 6 | 7 | 8 |
|---|---|---|---|---|---|---|---|---|---|
| left_child | 1 | – | 3 | 5 | – | – | 7 | – | – |
| right_child | 2 | – | 4 | 6 | – | – | 8 | – | – |
| feature | 2 | – | 2 | 1 | – | – | 1 | – | – |
| threshold | 0.303 | – | 0.696 | 0.296 | – | – | 0.703 | – | – |
| impurity | 0.273 | 0. | 0.368 | 0.482 | 0.065 | 0.051 | 0.48 | 0.042 | 0. |
| n_samples | 300 | 99 | 201 | 113 | 88 | 38 | 75 | 46 | 29 |
| value | 251/49 | 99/0 | 152/49 | 67/46 | 85/3 | 37/1 | 30/45 | 1/45 | 29/0 |

Table 5.2: Array representation of the decision tree in Figure 5.2.

By implementing the array representation as dynamic tables [Cormen et al., 2001], insertion of new nodes has an $O(1)$ amortized cost, while keeping memory allocations at a minimum and thereby reducing the overhead that would otherwise occur using per-node data structures. Maintaining the representation contiguously in memory also brings the benefits of leveraging CPU caches, which might greatly improve performance when fast predictions are critical. Since nodes are accessed successively and repeatedly at test time, storing them in nearby memory locations is indeed highly recommended.

### 5.3.3 *Builders*

The implementation of the induction procedure in Scikit-Learn revolves around three nested components:

(a) The first component is a `Builder` object whose role is to effectively build the tree array representation presented above, by recursively partitioning nodes using splits found by a `Splitter` object.

(b) The second component is a `Splitter` object whose role is to find splits on internal nodes.

(c) The third component is a `Criterion` object whose role is to evaluate the goodness of a split.

The most direct `Builder` implementation is the depth-first greedy induction procedure as originally outlined in Algorithm 3.2. A critical aspect of this algorithm is how to keep account of the the input samples $(\mathbf{x}, y) \in \mathcal{L}_t$ that arrive at node $t$. A straightforward solution is to store lists of indices of the input samples at node $t$. As pointed out in [Criminisi and Shotton, 2013], building many such lists is however inefficient because of repeated memory allocations and deallocations. A better solution is to have instead all indices stored in a single static array and to use in-place reordering operations when partitioning $t$. More specifically, let assume that `samples` is the list of all indices and that `start` and `end` are bounds such that `samples[start:end]` contains the sample indices at the current node `t`. If `t` is split on feature `feature[t]` and threshold `threshold[t]`, then $\mathcal{L}_t$ can be partitioned into $\mathcal{L}_{t_L}$ and $\mathcal{L}_{t_R}$ by reorganizing `samples[start:end]` into `samples[start:pos]` and `samples[pos:end]`, such that all samples in the first part are on the left side of the split while all samples in the second part are on the right side of the split. The induction procedure then proceeds by pushing the bounds `start:pos` and `pos:end` on the stack $S$, as an efficient way to effectively represent $\mathcal{L}_{t_L}$ and $\mathcal{L}_{t_R}$.

An alternative implementation of the `Builder` interface consists in replacing the stack $S$ in Algorithm 3.2 by a priority queue, hence changing the order in which the nodes are split. In particular, if nodes $t$ are prioritized by weighted potential impurity decrease $p(t)\Delta i(s^*, t)$ (where $s^*$ is the best split on $t$), then the greedy induction procedure switches from a depth-first construction of the tree to a *best-first* induction procedure. In this way, an additional stopping criterion can be defined as the maximum number `max_leaf_nodes` of leaves in the tree. If nodes are expanded in best-first order, then the resulting tree only includes the most significant splits, thereby pre-pruning all (seemingly) irrelevant branches without wasting computational resources. Alternatives also include prioritizing nodes according to their impurity $i(t)$ or their size $N_t$.

Likewise, the stack $S$ in Algorithm 3.2 can be replaced by a deque data structure, in which case the induction procedure switches to a *breadth-first* construction of the tree. With some modifications, breadth-first construction shows to be more efficient when it is expensive to randomly access the data [Criminisi and Shotton, 2013], e.g., when data are too big to fit into memory and must be streamed from disk. Breadth-first induction also proves to be an efficient strategy when combined with parallelism, e.g., for building a whole level of the decision tree simultaneously using multiple cores [Liao et al., 2013].

### 5.3.4 *Splitters*

In Scikit-Learn, `Splitter` objects implement search procedures for finding splits on internal nodes. In the case of decision trees, they implement algorithms 3.3 and 3.4 for finding the best split over all $p$ input variables. In the case of randomized trees, they implement the search procedure 4.1 over $K \leqslant p$ randomly drawn input variables, combined with either algorithm 3.4 or 4.3 to respectively obtain Random Forests or Extremely Randomized Trees. Again, several aspects of the algorithm should be considered with care to obtain an efficient implementation:

*Data access.*

Looking for the best split on the input variable $X_j$ and partitioning $\mathcal{L}_t$ requires to repeatedly access over the node sample values $x_{i,j}$ (i.e., over values in the $j$-th column if samples are represented as a two dimensional array), which may comes at a non negligible cost if data is not ordered properly. However, due to CPU caching and pre-fetching effects, the closer the values in memory, usually the lower it takes to access them. In our case, this can be exploited in two ways to minimize the cost of data access:

- By storing the learning set $\mathcal{L}$ using column-major order (i.e., Fortran array memory layout), hence storing the values $x_{i,j}$ of $X_j$ (for all samples $i = 1, \ldots, N$) contiguously in memory.
- By pre-fetching the node sample values $x_{i,j}$ (for $i = 1, \ldots, N_t$) manually into a static and contiguous buffer, on which the search procedure is then applied.

*Sort algorithm.*

In CART and Random Forests, finding the best split on $X_j$ requires to sort the sample indices `samples[start:end]` along their respective values on $X_j$, i.e., such that

$$\texttt{X[samples[start], j]} \leqslant \cdots \leqslant \texttt{X[samples[start+i], j]}$$

$$\leqslant \cdots \leqslant \texttt{X[samples[end-1], j]},$$

for $i = 0, \ldots, N_t - 1$. As shown in Section 5.1, this operation drives the complexity of the overall induction procedure and is therefore critical in the implementation of the algorithm.

To guarantee the lowest time complexity in all cases, we rely on Introsort [Musser, 1997] which combines Quicksort and Heapsort into a hybrid sorting algorithm. Basically, sorting in Introsort begins with Quicksort and then switches to Heapsort when the depth of partitioning exceeds $\Theta(\log N)$. As such, the practical $\Theta(N \log N)$ performance of Introsort is comparable to the average performance of Quicksort, except that performance in the worst case is bounded to $\Theta(N \log N)$ due to Heapsort, instead of $\Theta(N^2)$ in the worst case for Quicksort. Additionally, Quicksort is implemented using the median-of-three as pivot [Bentley and McIlroy, 1993], hence yielding better partitions and further improving performance.

In the context of the tree induction procedure, let us finally notice that partitioning node samples $\mathcal{L}_t$ into subsets $\mathcal{L}_{t_L}$ and $\mathcal{L}_{t_R}$ not only makes them purer in terms of the output variable $Y$, it also indirectly makes node samples more and more identical in terms of the values taken by the input variables. As we get deeper into the tree, this property can be leveraged by implementing Quicksort as a 3-way recursive partition [Bentley and McIlroy, 1993], as illustrated in Figure 5.3. In this variant, elements that are identical to the pivot are all set at their final positions in a single recursive call, hence accelerating the sorting procedure. If the number of distinct values is constant, 3-way Quicksort can be shown [Sedgewick and Wayne, 2011] to reduce running times from $O(N \log N)$ to $O(NH)$, where $H$ is the Shannon entropy defined on the frequencies of the values to be sorted. As a consequence, in the case of many duplicates (i.e., when $H = O(1)$), $C(N) = O(N)$ and using 3-way Quicksort for sorting the sample indices in fact reduces the overall complexity of Random Forests to the asymptotic complexity of Extremely Randomized Trees (see Section 5.1).

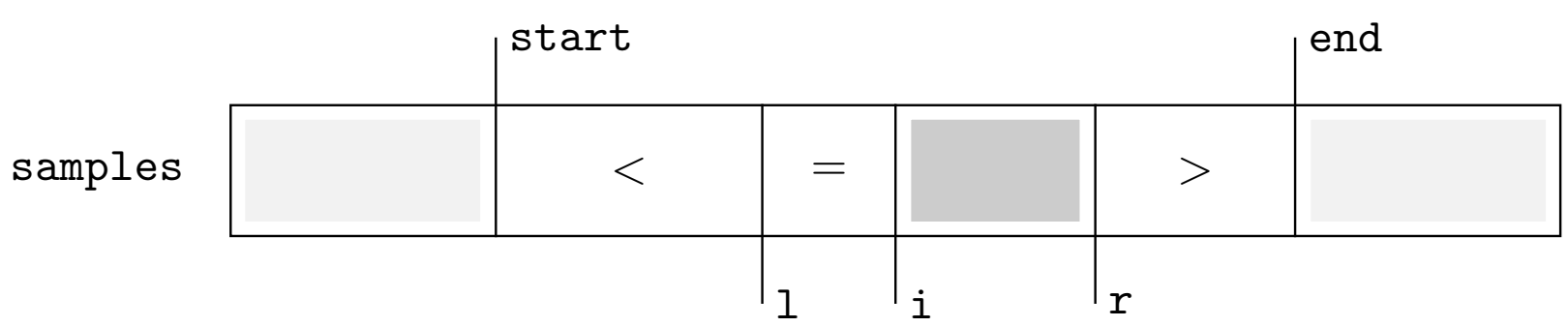

Figure 5.3: 3-way partition in Quicksort. Once `i=r`, elements that are identical to the pivot are all set at their final positions `samples[l:r]`, in a single recursive call. The sub-arrays `samples[start:l]` and `samples[r:end]` are then recursively sorted.

*Skipping locally constant input variables.*

In the extreme case, recursively partitioning node samples might result in input variables $X_j$ to become locally constant at $t$. That is, $x_j = x'_j$ for all $(\mathbf{x}, y), (\mathbf{x}', y) \in \mathcal{L}_t$. In such a case, there is no valid split on $X_j$ and trying to find one at $t$, but also at any of the descendant nodes, is a waste of computational resources. Accordingly, Algorithm 4.1 can be extended to account for variables that are known to be constant, thereby skipping non-informative variables if we know in advance that no valid split can be found. To stay as close as possible to the original algorithm, note however that constant variables are still included when drawing $K$ variables at random.

*Split approximations.*

In Chapter 4, we showed that an increase of bias due to randomization is beneficial as long as it sufficiently decorrelates the predictions of the individual trees. Given this result, a legitimate question is to wonder if finding the very best splits $s_j^*$ really is imperative in the first place to achieve good accuracy in random forests. As proved by the good results usually obtained by Extremely Randomized Trees, even splits that are drawn at random are in fact often sufficient to reach good (sometimes even better) performance. More importantly, not only this yields comparable accuracy, it is also more computationally efficient. At midway between exact best splits (as in Algorithm 3.4) and random splits (as in Algorithm 4.3), an intermediate and pragmatic solution is therefore to consider approximations of the best splits. The two most common strategies are based on *subsampling* and *binning*:

- Let $N_0$ denote an upper limit on the node sample size. If $N_t > N_0$, the subsampling approximation of the best split $s_j^*$ at $t$ on $X_j$ consists in looking for the best split using only $N_0$ randomly drawn node samples from $\mathcal{L}_t$, and all node samples otherwise. The rationale behind this strategy is that, in the first nodes of the tree, the best splits are often so markedly superior to all other splits that this clear-cut information also reflects in a subsample. For nodes further down in the tree (i.e., for $N_t \leqslant N_0$), splits are usually less marked but then all node samples in $\mathcal{L}_t$ are used, hence guaranteeing to find the same splits as in the original procedure. Historically, the subsampling strategy was first proposed in [Breiman et al., 1984] in the context of single decision trees, for datasets too large to be held in memory.
- An alternative but close strategy is to consider only a subset of the possible discretization thresholds $v'_k$ instead of exhaustively evaluating all candidate splits. In binning, this is performed by grouping together node samples that are close to each other into bins and then evaluating only the thresholds in-between these groups. The simplest strategy of all is to divide the value interval of $X_j$ at $t$ into $D$ intervals of equal length. Another strategy is to sort the node samples along their respective values of $X_j$ and to partition them into $D$ subsets of equal size. More elaborate algorithms, also known as *attribute discretization* procedures, are reviewed in [Zighed and Rakotomalala, 2000].

*Pre-sorting data.*

As shown earlier, sorting node samples for each candidate split significantly impacts the overall complexity of the induction procedure. An alternative strategy is possible [Breiman, 2002] and consists in presorting the samples for all $p$ input variables before the construction of the tree. More specifically, let $i = 1, \dots, \widetilde{N}$ denote the original indices of the unique input samples in $\mathcal{L}^m$. For each input variable $X_j$ (for $j = 1, \dots, p$), the sorted indices $\sigma^j_1, \dots, \sigma^j_{\widetilde{N}}$, such that

$$x_{\sigma^j_1, j} \leqslant \dots \leqslant x_{\sigma^j_{\widetilde{N}}, j}, \tag{5.36}$$

can be computed in $O(p\widetilde{N} \log \widetilde{N})$. Given these indices, finding the best split at a node then simply amounts to iterate, in that order, over the samples $\sigma^j_1, \dots, \sigma^j_{\widetilde{N}}$ (for all $K$ of the split variables $X_j$), which reduces complexity to $O(K\widetilde{N})$ instead of $O(K\widetilde{N} \log \widetilde{N})$. Partitioning the node samples into $t_L$ and $t_R$ then requires to partition all $p$ lists of sorted indices into $2p$ sub-lists of $\widetilde{N}_{t_L}$ and $\widetilde{N}_{t_R}$ sorted indices $\sigma^j_1, \dots, \sigma^j_{\widetilde{N}_{t_L}}$ and $\sigma^j_1, \dots, \sigma^j_{\widetilde{N}_{t_R}}$, which can be done in $O(p\widetilde{N})$. In total, the within-node complexity is

$$C(\widetilde{N}) = O(K\widetilde{N} + p\widetilde{N}) = O(p\widetilde{N}). \tag{5.37}$$

Using this alternative strategy, the time complexity of RF for the best, the worst and the average cases is respectively $O(Mp\widetilde{N} \log \widetilde{N})$, $O(Mp\widetilde{N}^2)$ and $O(Mp\widetilde{N} \log \widetilde{N})$, as derived from theorems 5.3, 5.5 and 5.7 for $K = p$. Neglecting constant factors, the ratio between the two implementations is $O(\frac{p}{K \log \widetilde{N}})$, which might not necessarily lead to faster building times depending on $K$ and the size of the problem. For this reason, pre-sorting is in fact not used within the Scikit-Learn implementation of random forests.

### 5.3.5 *Criteria*

The last components of the implementation of decision trees in Scikit-Learn are `Criterion` objects for evaluating the goodness of splits. Supported criteria are `"gini"` and `"entropy"` in classification, and the reduction of variance "mse" in regression. All of them implement the update mechanisms described in Section 3.6.3.1, such that the iterative evaluation of $N_t - 1$ splits on an ordered variable remains a $O(N_t)$ operation.

### 5.3.6 *Parallelization*

Scikit-Learn implementation of random forests supports parallelism to distribute computations on multi-core architectures. The simple, yet effective, approach that we implement is to consider the construction of a forest as an embarrassingly parallel problem, in which the $M$ decision trees are built concurrently and independently on multiple processors. Assuming no overhead due to parallelization, the theoretical speedup of this strategy is equal to the number $N_p$ of processors used for building the forest. In practice however, this upper bound is not always strictly reached due to the variance of the time required for building randomized trees. It may indeed happen that building trees take longer than building some others, thereby under-utilizing computational resources since jobs may have finished sooner than others.

More fine-grained approaches consist in building decision trees one at time, but using multiple processors. In the *node-based decomposition*, the construction of a tree is distributed by building multiple nodes concurrently. In this case, workload can be divided between processors using a breadth-first induction procedure and assigning whole sub-trees to develop once the waiting queue contains $N_p$ nodes. Once again however, sub-trees may take longer to build than others, hence under-utilizing computational resources if the resulting decision tree is not balanced. To alleviate this issue, processors may be assigned single nodes to develop (that is, one at a time) using a producer-consumer scheme to balance workload in a fair way. Unfortunately, this latter approach often proves to be dominated in practice by the overhead due to task assignment and bookkeeping activities. Alternatively, the construction of a tree can also be distributed using an *attribute-based decomposition*. In this case, the workload is divided by parallelizing the search of the $K$ best splits at a given node (see Algorithm 4.1). For detailed reviews on the topic, see [Kufrin, 1997; Srivastava et al., 2002].

Using similar or hybrid approaches, let us finally note that the parallel induction of random forests extend to other distributed architectures, including GPUs [Sharp, 2008; Liao et al., 2013], FPGA [Narayanan et al., 2007] or clusters [Mitchell et al., 2011].

## 5.4 BENCHMARKS

### 5.4.1 *Complexity on artificial data*

In this section, we study the empirical complexity of random forests on artificial data. Our goal is to investigate whether the theoretical results that have been derived in Sections 5.1 and 5.2 hold true in practice, even without our assumptions necessarily satisfied.

For our experiments on simulated data, we consider the Friedman #1 artificial regression problem [Friedman, 1991]. It consists in 10 independent input variables $X_1, \ldots, X_{10}$, each uniformly distributed over $[0, 1]$. The output $Y$ is a real-valued variable and is given by

$$Y = 10\sin(\pi X_1 X_2) + 20(X_3 - 0.48)^2 + 10X_4 + 5X_5 + \epsilon \tag{5.38}$$

where $\epsilon$ is $\mathcal{N}(0, 1)$. We study the effects of the number $M$ of trees in the forest, the number $N$ of input samples, the number $p$ of input variables (by adding new irrelevant variables), the number $K$ of variables drawn at each node and the effects of using bootstrap replicates $\mathcal{L}^m$ instead of $\mathcal{L}$. Hyper-parameters are studied independently for both (fully developed) Random Forests (RF) and Extremely Randomized Trees (ETs) built on $N = 1000$ learning samples of $p = 10$ input variables, with $M = 250$ trees and $K = \sqrt{p}$ set by default. Statistics reported below are averages over 10 runs. Figures 5.4-5.8 all illustrate the effect of one the hyper-parameters with respect to the time required for building a forest (on the left plots), the average depth of the trees (in the middle plots) and the mean squared error of the ensemble (on the right plots).

Figure 5.4 first demonstrates the effect of the number $M$ of trees. As expected, the time complexity of the induction procedure is $O(M)$ while the average depth of the leaves is constant with $M$, since the later has no effect on the structure of the individual trees. The left plot also confirms that ETs are faster to build than RF, even if trees in ETs are on average one level deeper that trees in RFs, as shown in the middle plot. Finally, the mean squared error of the model shows to be inversely proportional to the number of trees in the forest, which confirms our previous results on variance reduction from Chapter 4.

Figure 5.5 considers the effect of the number $N$ of training samples. As shown on the left plot, building times grow slightly faster than linearly, which appears to confirm the respective $O(N \log^2 N)$ and $O(N \log N)$ dependencies of RF and ETs with the size of the learning set (as derived in Section 5.1). Similarly, the middle plot confirms that the average depth of the leaves grows in $O(\log N)$ with $N$ (as shown in Section 5.2). As expected, the right plot also illustrates that the more the data, the better the resulting model. Additionally, it shows that RF and ETs tend towards the same error rate as $N$ increases, even if RF was strictly better on less data. This confirms once again that finding the very best splits is not strictly necessary to achieve good results.

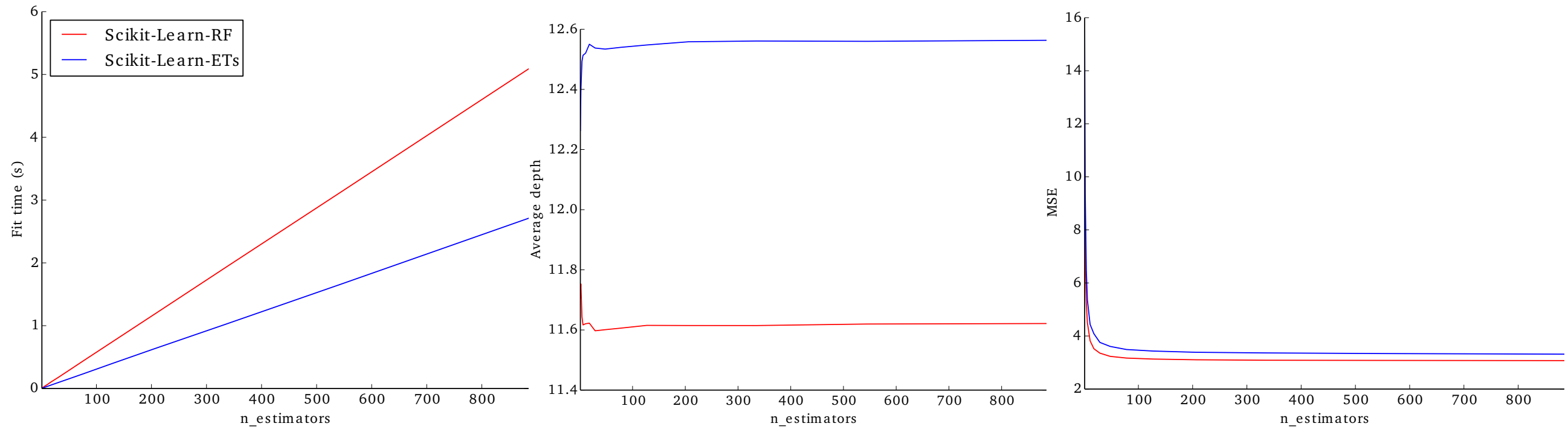

Figure 5.4: Effect of the number $M$ (= `n_estimators`) of trees on training time (left), average depth (middle) and mean squared error (right). $N = 1000$, $p = 10$, $K = \sqrt{p}$.

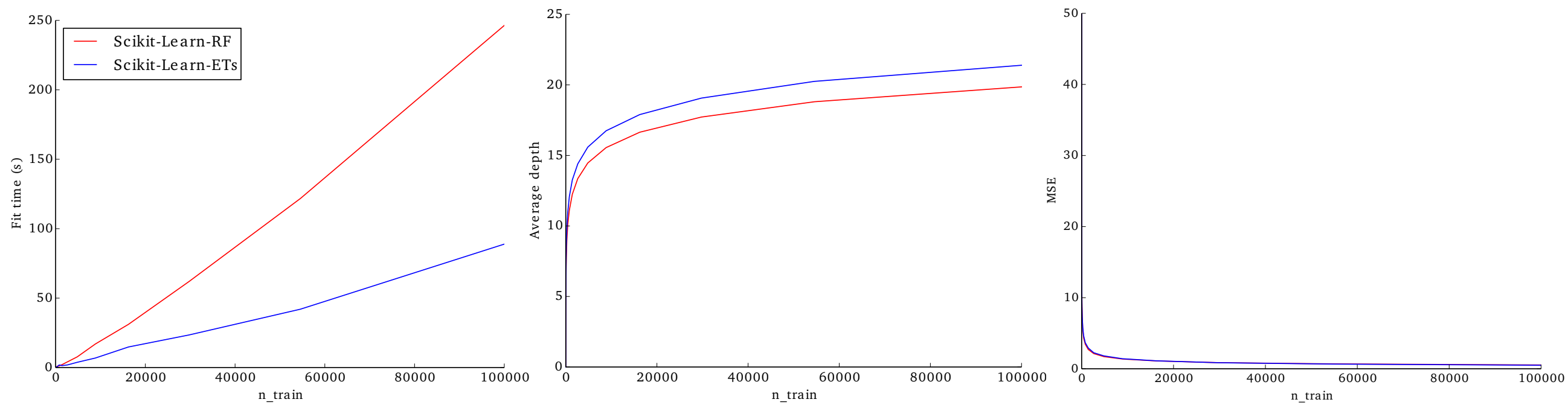

Figure 5.5: Effect of the size $N$ (= `n_train`) of the learning set on training time (left), average depth (middle) and mean squared error (right). $M = 250$, $p = 10$, $K = \sqrt{p}$.

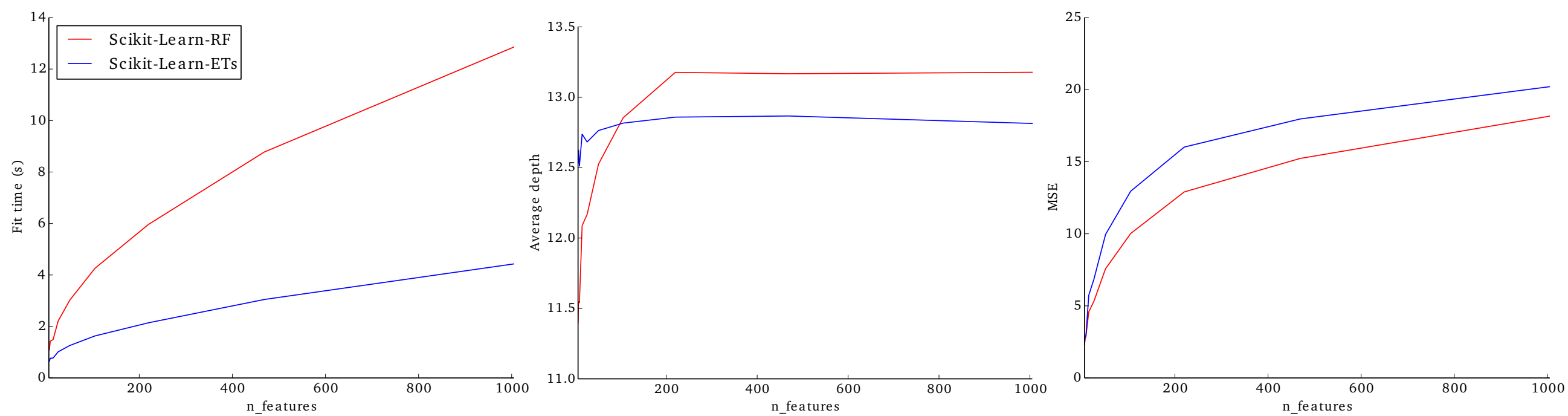

Figure 5.6: Effect of the number $p$ (= `n_features`) of input variables on training time (left), average depth (middle) and mean squared error (right). $N = 1000$, $M = 250$, $K = \sqrt{p}$.

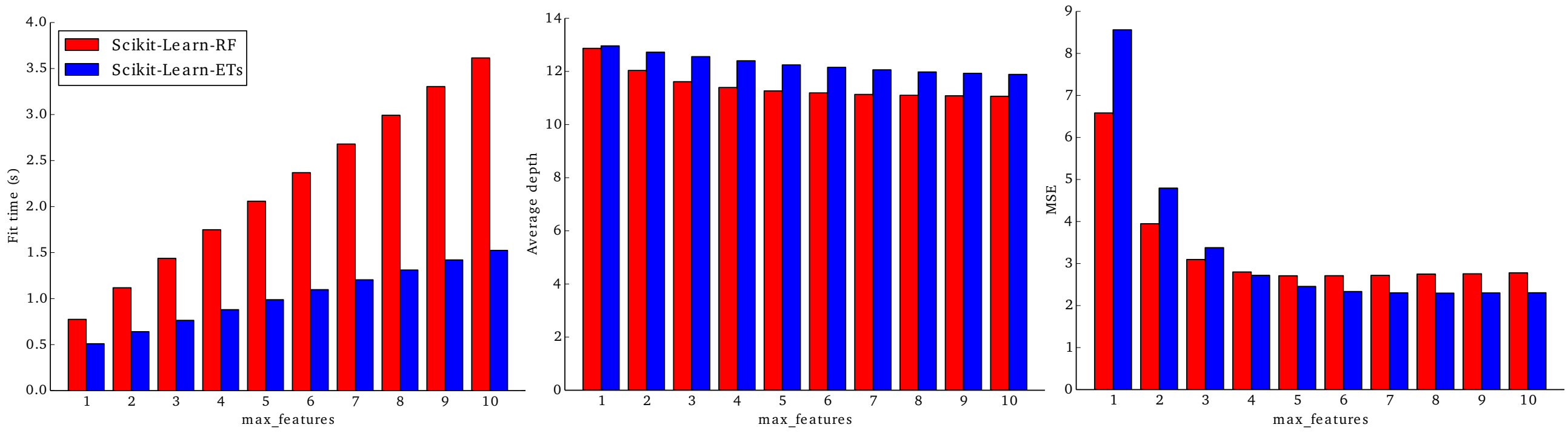

Figure 5.7: Effect of K (= `max_features`) in random variable selection on training time (left), average depth (middle) and mean squared error (right). $N = 1000$, $M = 250$, $p = 10$.

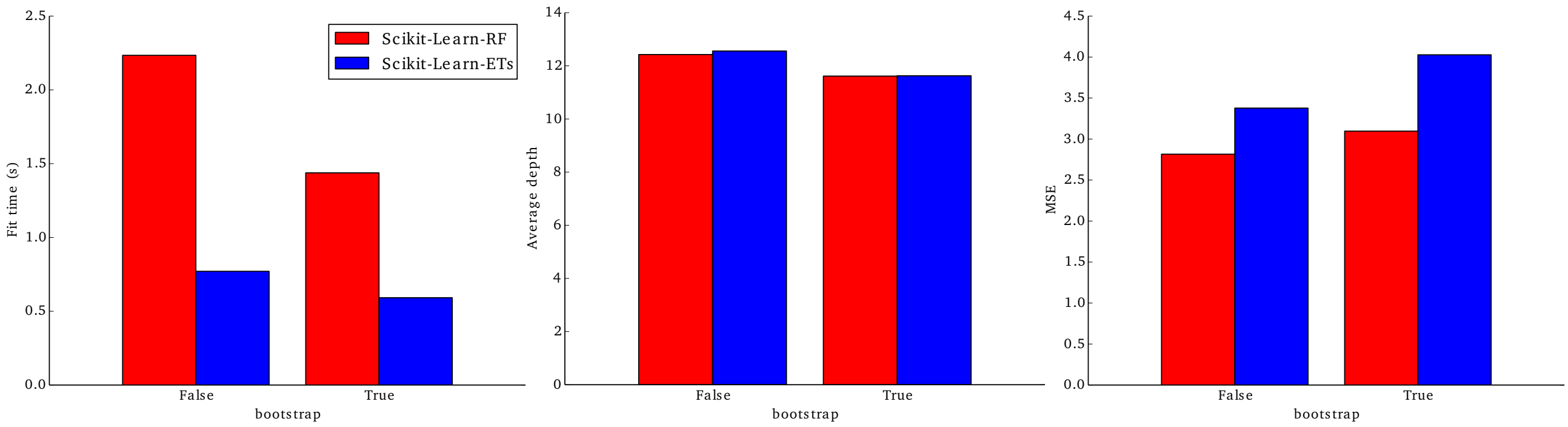

Figure 5.8: Effect of bootstrap replicates on training time (left), average depth (middle) and mean squared error (right). $N = 1000$, $M = 250$, $p = 10$, $K = \sqrt{p}$.

Next, Figure 5.6 shows the effect of irrelevant and noisy variables. The right plot first indicates that adding irrelevant variables deteriorates the accuracy of the forests. This is expected since the more the irrelevant variables, the higher the probability of splitting on one of them and therefore of fitting noise in the data. More importantly, the left and middle plots show that RF is more affected than ETs by the addition of noisy variables. This is actually a consequence of the end-cut preference of the classical impurity criteria. When splitting on noisy variables and looking for the best threshold, partitions of the node samples are more likely to be unbalanced rather than the opposite. This explains why on the middle plot the average depth of RF eventually gets larger than the depth of ETs – which suffers less from end-cut preference since thresholds are not optimized locally.

Figure 5.7 illustrates the effect of random variable selection. On the left plot, building times grow as expected in $O(K)$ with the number $K$ of variables randomly drawn at each node. The average depth of the tree shows that increasing $K$ slightly reduces the depth of the trees, which is explained by the fact that better splits are selected as more variables are evaluated. Finally, the right plot again confirms results from Chapter 4. Random perturbations negatively impact the resulting model when they are too strong (at $K = 1$) but then become beneficial as $K$ increases. For RF, the optimal trade-off is at $K = 5$ while the lowest error can be achieved in ETs for $K = 7$.

Finally, Figure 5.8 studies the effect of using bootstrap replicates. For RF, learning trees from learning sets $\mathcal{L}^m$ drawn with replacement from $\mathcal{L}$ (`bootstrap=True`) is about 1.5 faster than building trees directly on $\mathcal{L}$ (`bootstrap=False`) , which is in accordance with our theoretical results from Section 5.1. The effect for ETs is less strong, but still leads to a non-negligible speedup. The average depth for RF and ETs is roughly 0.5 lower when using bootstrap replicates. Again, this is expected since the average depth grows with $O(\log N)$ and that reducing the effective size of the learning set to $\widetilde{N} = 0.632N$ reduces its logarithm by $-\log(0.632) = 0.662$. Finally, the right plot shows that using bootstrap replicates actually impairs performance on this problem.

While not discussed here, further experiments carried out on other artificial data, for both regression and classification, lead to similar conclusions. They all corroborate the theoretical results presented earlier.

### 5.4.2 *Comparison with other implementations*

Due to their apparent simplicity, random forests have been reimplemented at multiple occasions in various machine learning libraries and programming languages. In this section, we compare the Scikit-Learn implementation developed within this work with popular alternatives available in other machine learning packages.

Table 5.3 summarizes the main open source implementations of random forests along with some of their supported features. All implement the original Random Forest algorithm [Breiman, 2001] and therefore also provide an interface for Bagging [Breiman, 1994] (since it corresponds to the special case $K = p$). Scikit-Learn, OpenCV and OK3 also offer variants like Extremely Randomized Trees [Geurts et al., 2006a] or Random Subspaces [Ho, 1998]. All support both classification and regression tasks, with the exceptions of Weka and H2O which appear to only support classification. From a parameter point of view, implementations mostly differ from each other in the impurity and stopping criteria they support. The most common impurity criterion is Gini (or, equivalently, MSE) but some packages, like Orange, include alternative measures such as entropy, gain ratio or Relief [Kira and Rendell, 1992]. The most common stopping criteria are an upper bound on the depth of the trees (e.g., `max_depth` in Scikit-Learn) and a minimum number of samples required for splitting a node (e.g., `nmin` in OK3). Some packages also allow to set an upper limit on the number of nodes in the trees or to stop the construction when a fixed accuracy is reached. Despite the embarrassing parallelism, only Scikit-Learn, Weka and H2O make it possible to build random forests using multiple cores. Most notably, H2O builds on top of Hadoop to provide a fully distributed implementation of random forests (i.e., on a cluster).

| Library | Algorithms | Tasks | Impurity | Stopping criteria | Variable importances | Multi-thread | Open source | Language | Reference |
|---|---|---|---|---|---|---|---|---|---|
| *Scikit-Learn* | Bagging, RF, ETs, RS, RP | C/R | Gini, Entropy, MSE | `max_depth`, `min_samples_split`, `min_samples_leaf`, `max_leaf_nodes` | ✓ | ✓ | BSD | Python | [Pedregosa et al., 2011] |
| *OpenCV* | Bagging, RF, ETs | C/R | Gini | `max_depth`, `min_samples_count`, `forest_accuracy` | ✓ | ✗ | BSD | C/C++ | [Bradski and Kaehler, 2008] |
| *OK3* | Bagging, RF, ETs | C/R | MSE | `varmin`, `nmin`, `maxnbsplits` | ✓ | ✗ | Source-available | C | [Geurts et al., 2006a] |
| *Weka* | Bagging, RF | C | Gini | `depth` | ✗ | ✓ | GPL | Java | [Hall et al., 2009] |
| *randomForest* | Bagging, RF | C/R | Gini | `nodesize`, `maxnodes` | ✓ | ✗ | GPL | R | [Liaw and Wiener, 2002] |
| *Orange* | Bagging, RF | C/R | GainRatio, Gini, Relief, MSE | `worst_acceptable`, `min_susbet`, `min_instances`, `max_depth`, `max_majority` | ✗ | ✗ | GPL | Python | [Demšar et al., 2013] |
| *H2O* | Bagging, RF | C | Gini, Entropy | `maxDepth` | ✗ | ✓ | Apache | Java | [Vitek, 2013] |

Table 5.3: Popular libraries for random forests.

For performance reasons, implementations listed in Table 5.3 have mostly been written in low-level programming languages like C, C++ or Java, but usually interface with higher level languages, like Python, MATLAB or R, for convenience reasons. In the case of OpenCV for example, the core implementation is written in C++ but can easily been called from Python through bindings shipped with the library. Let us finally note that Table 5.3 is by no means an exhaustive list of all implementations of random forests. Domain-specific implementations (e.g., Random Jungle [Schwarz et al., 2010] for GWAS or TMVA [Hoecker et al., 2007] in particle physics) and proprietary software (e.g., Salford Systems, which owns the *Random Forests* trademark) are not listed nor evaluated here.

In the experiments carried out below, we benchmark the Scikit-Learn implementations of RF and ETs against all libraries listed in Table 5.3. We do not compare against H2O since it requires a dedicated cloud environment. Implementations are all benchmarked on 29 well-known and publicly available classification datasets that were chosen a priori and independently of the results obtained. Overall, these datasets cover a wide range of conditions (including both artificial and real data), with a sample size $N$ ranging from 208 to 70000 and a number $p$ of variables varying from 6 to 24496. Implementations are evaluated for (fully developed) Random Forests (RF) and Extremely Randomized Trees (ETs) whenever available, using respectively 75% and 25% of the original dataset as learning and test sets, $M = 250$ trees and $K = \sqrt{p}$ set by default. Results reported below are averages over 10 runs, with the learning and test sets resampled at each fold. Tables 5.4 and 5.5 respectively report the average time required for building a random forest and the average time required making predictions, on all 29 datasets and for all implementations. Computing times are reported using the Scikit-Learn implementation of RF as a baseline. Results in bold indicate that the implementation is the fastest on the dataset. Remark that times are not reported on CIFAR10 for R and Orange because these implementations failed at building a single forest within 96 hours for CPU time.

With respect to RF, Table 5.4 shows that Scikit-Learn is the fastest on average, far before all others. OK3 is the second fastest implementation (2.63× slower on average), Weka is third (2.92× slower), OpenCV is fourth (12.37× slower), Orange is fifth (17.80× slower) and the R implementation is last (30.73× slower). In particular, the R implementation shows to be very ineffective on large datasets, when both N and p are moderately large (e.g., on MNIST, as further illustrated in Figure 5.9). A possible explanation is that the R implementation uses pre-sorting (see Section 5.3.4), which makes complexity linear with p rather than with K. Regarding Orange, results can be explained by the fact that the implementation is purely written in Python (i.e., an interpreted high-level programming language), while all others are written in optimized low-level programming languages. With respect to ETs, the Scikit-Learn implementation is again the fastest, far before all others. More interestingly, the table also shows that ETs are usually faster to build than RF, which confirms the theoretical results presented earlier. For the Scikit-Learn implementation, ETs are indeed empirically $\frac{1}{0.71} = 1.41\times$ faster than RF on average.

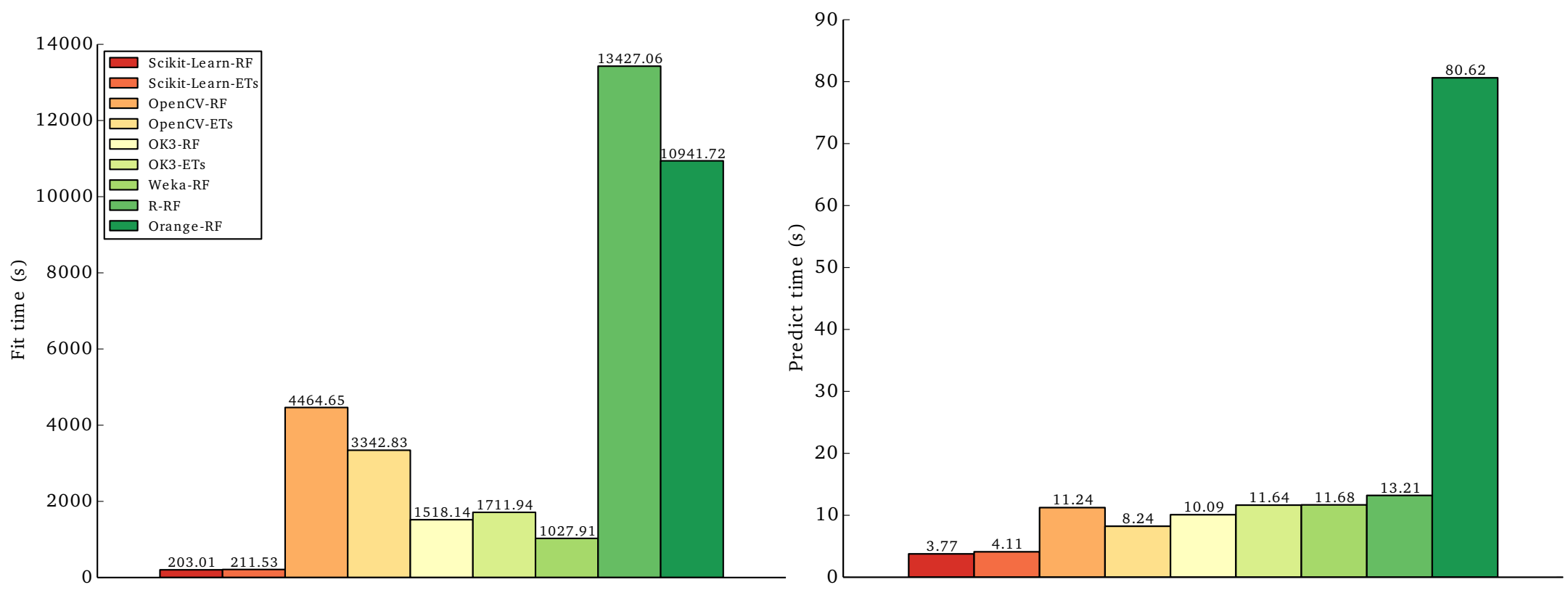

Figure 5.9: Average time required for building a forest on the MNIST dataset (left) and average time required for making predictions (right).

Table 5.5 reports results regarding the average time required for making predictions on the test set. Again, the Scikit-Learn implementation of RF is the fastest implementation on average. OpenCV is second (2.07$\times$ slower on average), OK3 is third (3.71$\times$ slower), Weka is fourth (4.31$\times$ slower), R is fifth (9.64$\times$ slower) and Orange is last (19.82$\times$ slower). By contrast with fitting times, making predictions from ETs is now slightly slower than RF, which is explained by the fact that trees in ETs are on average deeper than in RF.

In conclusions, benchmarks show that the Scikit-Learn implementation is on average significantly faster than OpenCV, OK3, Weka, R and Orange. While these results suffer from a selection bias (i.e., they depend on the 29 selected datasets), we believe that the general conclusions extend to other datasets. More importantly, the careful design and implementation of each and every component of random forests, as discussed all throughout this work, eventually shows to be highly rewarding in terms of computational performance.

| Dataset | Sklearn-RF | Sklearn-ETs | OpenCV-RF | OpenCV-ETs | OK3-RF | OK3-ETs | Weka-RF | R-RF | Orange-RF |
|---|---|---|---|---|---|---|---|---|---|
| ARCENE | 1.00 | **0.77** | 26.33 | 16.73 | 0.81 | 0.85 | 8.70 | 22.84 | 4.78 |
| BREAST2 | 1.00 | **0.41** | 35.10 | 19.03 | 1.20 | 0.99 | 3.88 | 33.41 | 11.42 |
| CIFAR10 | 1.00 | **0.52** | 22.82 | 13.76 | 3.78 | 2.30 | 3.29 | – | – |
| DIABETES | 1.00 | 0.88 | 0.73 | 0.75 | **0.59** | 0.65 | 2.25 | 1.07 | 2.42 |
| DIG44 | 1.00 | **0.39** | 1.01 | 0.71 | 2.68 | 2.03 | 1.70 | 1.60 | 10.85 |
| IONOSPHERE | 1.00 | 0.71 | 0.79 | 0.56 | 0.53 | **0.34** | 1.87 | 0.80 | 3.01 |
| ISOLET | 1.00 | **0.36** | 5.39 | 3.39 | 4.48 | 2.76 | 2.24 | 8.70 | 16.58 |
| LETTER | 1.00 | **0.94** | 2.34 | 1.82 | 9.05 | 10.57 | 3.13 | 4.43 | 21.04 |
| LIVER | 1.00 | 0.95 | 0.38 | 0.46 | **0.33** | 0.42 | 2.80 | 0.58 | 1.42 |
| MADELON | 1.00 | **0.46** | 5.82 | 4.07 | 1.38 | 0.82 | 2.06 | 6.18 | 48.48 |
| MARTI0 | 1.00 | **0.31** | 5.70 | 4.45 | 1.08 | 0.55 | 1.36 | 18.99 | 7.16 |
| MNIST | **1.00** | 1.04 | 21.99 | 16.47 | 7.48 | 8.43 | 5.06 | 66.14 | 53.90 |
| MNIST3VS8 | 1.00 | **0.95** | 18.74 | 13.90 | 3.32 | 3.04 | 3.68 | 54.89 | 27.80 |
| MNIST4VS9 | 1.00 | **0.97** | 20.21 | 14.70 | 3.46 | 3.22 | 4.09 | 57.10 | 29.42 |
| MUSK2 | 1.00 | **0.44** | 4.59 | 2.44 | 2.00 | 0.95 | 1.97 | 5.48 | 22.63 |
| PENDIGITS | 1.00 | **0.61** | 1.60 | 1.12 | 3.61 | 3.09 | 2.30 | 2.64 | 9.24 |
| REGED0 | 1.00 | **0.43** | 6.65 | 4.29 | 1.18 | 0.64 | 1.63 | 10.02 | 7.04 |
| RING-NORM | 1.00 | **0.22** | 0.63 | 0.43 | 1.13 | 0.42 | 1.11 | 1.21 | 9.92 |
| SATELLITE | 1.00 | **0.56** | 2.10 | 1.36 | 2.69 | 1.97 | 1.94 | 2.56 | 13.88 |
| SECOM | 1.00 | **0.21** | 5.59 | 2.61 | 1.54 | 0.48 | 1.00 | 5.46 | 12.88 |
| SEGMENT | 1.00 | **0.76** | 1.09 | 0.81 | 1.91 | 1.79 | 1.83 | 1.64 | 5.97 |
| SIDO0 | **1.00** | 1.55 | 133.11 | 140.63 | 9.70 | 8.62 | 9.69 | 468.85 | 49.11 |
| SONAR | 1.00 | 0.78 | 0.80 | 0.64 | 0.39 | **0.30** | 1.82 | 0.81 | 2.61 |
| SPAMBASE | **1.00** | 1.41 | 4.06 | 3.36 | 2.24 | 3.25 | 2.90 | 4.59 | 15.85 |
| TIS | **1.00** | 1.84 | 26.83 | 24.59 | 3.73 | 4.31 | 5.41 | 74.77 | 69.51 |
| TWO-NORM | 1.00 | **0.30** | 0.98 | 0.66 | 1.12 | 0.55 | 1.21 | 1.33 | 17.12 |
| VEHICLE | 1.00 | **0.74** | 1.40 | 1.08 | 1.63 | 1.46 | 2.23 | 1.69 | 6.51 |
| VOWEL | 1.00 | 0.59 | 0.67 | **0.59** | 1.87 | 1.50 | 1.99 | 1.08 | 8.18 |
| WAVEFORM | 1.00 | **0.43** | 1.20 | 0.94 | 1.49 | 0.98 | 1.49 | 1.50 | 9.79 |
| *Average* | 1.00 | **0.71** | 12.37 | 10.22 | 2.63 | 2.32 | 2.92 | 30.73 | 17.80 |
| *Median* | 1.00 | **0.61** | 4.06 | 2.44 | 1.87 | 1.46 | 2.23 | 4.51 | 11.14 |

Table 5.4: Average time required for building a random forest, relative to the Scikit-Learn implementation of the Random Forest algorithm (first column). The lower, the better.

| Dataset | Sklearn-RF | Sklearn-ETs | OpenCV-RF | OpenCV-ETs | OK3-RF | OK3-ETs | Weka-RF | R-RF | Orange-RF |
|---|---|---|---|---|---|---|---|---|---|
| ARCENE | 1.00 | 1.02 | **0.38** | 0.40 | 0.55 | 0.78 | 5.65 | 5.46 | 6.70 |
| BREAST2 | **1.00** | 1.11 | 2.00 | 1.76 | 3.63 | 4.97 | 33.92 | 33.26 | 44.73 |
| CIFAR10 | **1.00** | 1.13 | 3.17 | 2.25 | 2.82 | 3.05 | 6.23 | – | – |
| DIABETES | 1.00 | 1.25 | 1.00 | 1.78 | 1.34 | 2.47 | 1.49 | **0.99** | 11.41 |
| DIG44 | **1.00** | 1.25 | 3.60 | 4.40 | 4.62 | 5.86 | 3.25 | 1.44 | 21.86 |
| IONOSPHERE | 1.00 | 1.05 | **0.35** | 0.61 | 0.47 | 0.74 | 1.36 | 0.45 | 6.56 |
| ISOLET | **1.00** | 1.19 | 2.21 | 2.25 | 2.03 | 2.63 | 2.92 | 2.17 | 12.28 |
| LETTER | 1.00 | 1.15 | 2.25 | 2.71 | 3.18 | 3.87 | 2.69 | **0.85** | 14.91 |
| LIVER | 1.00 | 1.14 | **0.49** | 0.93 | 0.62 | 1.22 | 1.54 | 0.58 | 6.01 |
| MADELON | **1.00** | 1.29 | 2.83 | 4.01 | 7.28 | 11.03 | 3.58 | 3.01 | 33.41 |
| MARTI0 | 1.00 | 1.10 | **0.68** | 1.00 | 0.73 | 1.09 | 3.23 | 2.42 | 9.80 |
| MNIST | **1.00** | 1.09 | 2.98 | 2.18 | 2.68 | 3.09 | 3.10 | 3.50 | 21.39 |
| MNIST3VS8 | **1.00** | 1.05 | 1.74 | 1.71 | 1.90 | 2.37 | 2.55 | 3.46 | 15.76 |
| MNIST4VS9 | **1.00** | 1.21 | 2.29 | 2.24 | 2.42 | 2.92 | 3.21 | 5.21 | 19.46 |
| MUSK2 | **1.00** | 1.15 | 2.62 | 3.31 | 3.10 | 4.18 | 2.88 | 2.07 | 29.81 |
| PENDIGITS | **1.00** | 1.18 | 2.70 | 3.72 | 3.66 | 4.97 | 2.89 | 1.03 | 22.49 |
| REGED0 | 1.00 | 1.09 | **0.48** | 0.69 | 0.51 | 0.81 | 2.84 | 2.37 | 10.70 |
| RING-NORM | **1.00** | 1.48 | 4.35 | 6.24 | 5.74 | 8.29 | 3.30 | 1.59 | 28.55 |
| SATELLITE | **1.00** | 1.07 | 2.64 | 3.43 | 3.46 | 4.51 | 2.63 | 1.16 | 23.14 |
| SECOM | **1.00** | 1.07 | 1.87 | 1.98 | 2.09 | 2.22 | 3.06 | 2.74 | 17.35 |
| SEGMENT | 1.00 | 1.49 | 1.25 | 2.15 | 1.66 | 2.86 | 2.23 | **0.82** | 18.98 |
| SIDO0 | **1.00** | 1.12 | 2.02 | 2.18 | 2.52 | 2.96 | 11.25 | 184.54 | 46.13 |
| SONAR | 1.00 | 1.03 | **0.20** | 0.33 | 0.23 | 0.41 | 1.01 | 0.39 | 4.70 |
| SPAMBASE | **1.00** | 1.38 | 2.91 | 3.90 | 4.32 | 8.89 | 3.67 | 1.73 | 26.51 |
| TIS | **1.00** | 1.33 | 2.70 | 2.66 | 2.82 | 3.66 | 3.40 | 3.70 | 16.65 |
| TWO-NORM | **1.00** | 1.33 | 4.09 | 5.93 | 5.35 | 7.57 | 3.52 | 1.52 | 32.80 |
| VEHICLE | **1.00** | 1.17 | 1.74 | 2.50 | 2.26 | 3.21 | 2.57 | 1.07 | 17.61 |
| VOWEL | 1.00 | 1.18 | 1.07 | 1.60 | 1.43 | 2.12 | 1.53 | **0.76** | 10.83 |
| WAVEFORM | **1.00** | 1.35 | 3.48 | 4.95 | 4.37 | 6.34 | 3.40 | 1.52 | 24.53 |
| *Average* | **1.00** | 1.19 | 2.07 | 2.54 | 2.68 | 3.76 | 4.31 | 9.64 | 19.82 |
| *Median* | **1.00** | 1.15 | 2.21 | 2.24 | 2.52 | 3.05 | 3.06 | 1.66 | 18.29 |

Table 5.5: Average time required for making predictions, relative to the Scikit-Learn implementation of the Random Forest algorithm (first column). The lower, the better.

# Part II

# INTERPRETING RANDOM FORESTS

# 6 UNDERSTANDING VARIABLE IMPORTANCES

**Outline**

In this chapter, we study variable importance measures as computed from forests of randomized trees. In Section 6.1, we first present how random forests can be used to assess the importance of input variables. We then derive in Section 6.2 a characterization in asymptotic conditions and show that variable importances derived from totally randomized trees offer a three-level decomposition of the information jointly contained in the input variables about the output. In Section 6.3, we show that this characterization only depends on the relevant variables and then discuss these ideas in Section 6.4 in the context of variants closer to the Random Forest algorithm. Finally, we illustrate these results on an artificial problem in Section 6.5. *This chapter is based on previous work published in [Louppe et al., 2013].*

An important task in many scientific fields is the prediction of a response variable based on a set of predictor variables. In many situations though, the aim is not only to make the most accurate predictions of the response but also to identify which predictor variables are the most important to make these predictions, e.g. in order to understand the underlying process. Because of their applicability to a wide range of problems and their capability to build accurate models and, at the same time, to provide variable importance measures, random forests have become a major data analysis tool used with success in various scientific areas.

Despite their extensive use in applied research, only a couple of works have studied the theoretical properties and statistical mechanisms of these algorithms. Zhao [2000], Breiman [2004], Meinshausen [2006], Lin and Jeon [2002] and Biau et al. [2008]; Biau [2012] investigated simplified to very close variants of these algorithms and proved the consistency of those variants. Little is known however regarding the variable importances computed by random forests, and – as far as we know – the work of Ishwaran [2007] is indeed the only theoretical study of tree-based variable importance measures. In this chapter, we aim at filling this gap and present a theoretical analysis of the Mean Decrease Impurity importance derived from ensembles of randomized trees.

## 6.1 VARIABLE IMPORTANCES

### 6.1.1 *Importances in single decision trees*

In the context of single decision trees, Breiman et al. [1984] first defined the measure of importance of a variable $X_j$ as

$$\text{Imp}(X_j) = \sum_{t \in \varphi} \Delta I(\tilde{s}_t^j, t), \tag{6.1}$$

where $\tilde{s}_t^j$ is the best *surrogate split* for $s_t$, that is the closest split defined on variable $X_j$ that can mimic the actual split $s_t$ defined at node $t$. The use of surrogate splits was proposed to account for masking effects: it may indeed happen that some variable $X_{j_2}$ never occurs in any split because it leads to splits that are slightly worse, and therefore not selected, than those of some other variable $X_{j_1}$. However, if $X_{j_1}$ is removed and another tree is grown, $X_{j_2}$ may now occur prominently within the splits and the resulting tree may be nearly as good as the original tree. In such a case, a relevant measure should detect the importance of $X_{j_2}$. Accordingly, if $X_{j_2}$ is being masked at $t$ by $X_{j_1}$ (i.e., if $X_{j_1}$ is used to split $t$), but if $\tilde{s}_t^{j_2}$ is similar to $s_t$, but not quite as good, then $\Delta I(\tilde{s}_t^{j_2}, t)$ will be nearly as large as $\Delta I(s_t, t)$ and therefore the proposed measure will indeed account for the importance of $X_{j_2}$.

### 6.1.2 *Importances in forests*

Thanks to randomization, masking effects are dampened within forests of randomized trees. Even if $X_{j_2}$ is being masked by $X_{j_1}$ there is indeed still a chance for $X_{j_2}$ to be chosen as a split if $X_{j_1}$ is not selected among the $K$ variables chosen at random. Depending on the value $K$, masking effects do not disappear entirely though. The use of bootstrap samples also helps reduce masking effects, making $X_{j_1}$ or $X_{j_2}$ just slightly better than the other due to the variations in the bootstrap samples.

For this reason, Breiman [2001, 2002] proposed to evaluate the importance of a variable $X_j$ for predicting $Y$ by adding up the weighted impurity decreases $p(t)\Delta i(s_t, t)$ for all nodes $t$ where $X_j$ is used, averaged over all trees $\varphi_m$ (for $m = 1, \dots, M$) in the forest:

$$\text{Imp}(X_j) = \frac{1}{M} \sum_{m=1}^{M} \sum_{t \in \varphi_m} 1(j_t = j) \Big[ p(t) \Delta i(s_t, t) \Big], \tag{6.2}$$

where $p(t)$ is the proportion $\frac{N_t}{N}$ of samples reaching $t$ and where $j_t$ denotes the identifier of the variable used for splitting node $t$. When using the Gini index as impurity function, this measure is known as the *Gini importance* or *Mean Decrease Gini*. However, since it can be defined for any impurity measure $i(t)$, we will refer to Equation 6.2 as the *Mean Decrease Impurity* importance (MDI), no matter the impurity measure $i(t)$. We will characterize and derive results for this measure in the rest of this text.

In addition to MDI, Breiman [2001, 2002] also proposed to evaluate the importance of a variable $X_j$ by measuring the *Mean Decrease Accuracy* (MDA) or, equivalently, by measuring the *Mean Increase Error*, of the forest when the values of $X_j$ are randomly permuted in the out-of-bag samples. For that reason, this latter measure is also known as the *permutation importance*. Formally, in regression, the permutation importance of $X_j$ is given as:

$$\text{Imp}(X_j) = \mathbb{E}_{\pi_j}\left\{ \frac{1}{N} \sum_{(\mathbf{x}'_i, y_i) \in \pi_j(\mathcal{L})} L\left(\frac{1}{M^{-i}} \sum_{l=1}^{M^{-i}} \varphi_{\mathcal{L}^{m_{k_l}}}(\mathbf{x}'_i), y_i\right) \right\}$$
$$\hookrightarrow -\frac{1}{N} \sum_{(\mathbf{x}_i, y_i) \in \mathcal{L}} L\left(\frac{1}{M^{-i}} \sum_{l=1}^{M^{-i}} \varphi_{\mathcal{L}^{m_{k_l}}}(\mathbf{x}_i), y_i\right) \tag{6.3}$$

where $\pi_j(\mathcal{L})$ denotes a replicate of $\mathcal{L}$ in which in the values of $X_j$ have been randomly permuted, and where $m_{k_1}, \dots, m_{k_{M^{-i}}}$ denote the indices of the trees that have been built from bootstrap replicates that do not include $(\mathbf{x}_i, y_i)$. For classification, the permutation importance is derived similarly as in Equation 6.3, except that the out-of-bag average predictions are replaced with the class which is the most likely, as computed from the out-of-bag class probability estimates. Its rationale is that randomly permuting the input variable $X_j$ should break its association with the response $Y$. Therefore, if $X_j$ is in fact associated to $Y$, permuting its values should also result in a substantial increase of error, as here measured by the difference between the out-of-bag estimates of the generalization error. That is, the larger the increase of error, the more important the variable, and vice-versa.

Thanks to popular machine learning softwares [Breiman, 2002; Liaw and Wiener, 2002; Pedregosa et al., 2011], both of these variable importance measures have shown their practical utility in an increasing number of experimental studies. Little is known however regarding their inner workings. Strobl et al. [2007b] compare both MDI and MDA and show experimentally that the former is biased towards some predictor variables. As explained by White and Liu [1994] in case of single decision trees, this bias stems from an unfair advantage given by the usual impurity functions $i(t)$ towards predictors with a large number of values (see Section 7.2.2). Strobl et al. [2008] later showed that MDA is biased as well, and that it overestimates the importance of correlated variables – although this effect was not confirmed in a later experimental study by Genuer et al. [2010]. From a theoretical point of view, Ishwaran [2007] provides a detailed theoretical development of a simplified version of MDA, giving key insights for the understanding of the actual MDA.

## 6.2 THEORETICAL STUDY

### 6.2.1 *Background*

To be self-contained, we first recall several definitions from information theory (see [Cover and Thomas, 2012], for further properties).

We suppose that we are given a probability space $(\Omega, \mathcal{E}, \mathbb{P})$ and consider random variables defined on it taking a finite number of possible values. We use upper case letters to denote such random variables (e.g. $X, Y, Z, W \ldots$) and calligraphic letters (e.g. $\mathcal{X}, \mathcal{Y}, \mathcal{Z}, \mathcal{W} \ldots$) to denote their image sets (of finite cardinality), and lower case letters (e.g. $x, y, z, w \ldots$) to denote one of their possible values. For a (finite) set of (finite) random variables $X = \{X_1, \ldots, X_p\}$, we denote by $P_X(\mathbf{x}) = P_X(x_1, \ldots, x_p)$ the probability $\mathbb{P}(\{\omega \in \Omega \mid \forall j : 1, \ldots, p : X_j(\omega) = x_j\})$, and by $\mathcal{X} = \mathcal{X}_1 \times \cdots \times \mathcal{X}_p$ the set of joint configurations of these random variables. Given two sets of random variables, $X = \{X_1, \ldots, X_p\}$ and $Y = \{Y_1, \ldots, Y_q\}$, we denote by $P_{X|Y}(\mathbf{x} \mid \mathbf{y}) = P_{X,Y}(\mathbf{x}, \mathbf{y})/P_Y(\mathbf{y})$ the conditional density of X with respect to Y.[1]

With these notations, the joint (Shannon) entropy of a set of random variables $X = \{X_1, \ldots, X_p\}$ is thus defined by

$$H(X) = -\sum_{\mathbf{x} \in \mathcal{X}} P_X(\mathbf{x}) \log_2 P_X(\mathbf{x}), \tag{6.4}$$

while the mean conditional entropy of a set of random variables $X = \{X_1, \ldots, X_p\}$, given the values of another set of random variables $Y = \{Y_1, \ldots, Y_q\}$ is defined by

$$H(X \mid Y) = -\sum_{\mathbf{x} \in \mathcal{X}} \sum_{\mathbf{y} \in \mathcal{Y}} P_{X,Y}(\mathbf{x}, \mathbf{y}) \log_2 P_{X|Y}(\mathbf{x} \mid \mathbf{y}). \tag{6.5}$$

The mutual information among the set of random variables $X = \{X_1, \ldots, X_p\}$ and the set of random variables $Y = \{Y_1, \ldots, Y_q\}$ is defined by

$$\begin{aligned} I(X;Y) &= -\sum_{\mathbf{x} \in \mathcal{X}} \sum_{\mathbf{y} \in \mathcal{Y}} P_{X,Y}(\mathbf{x}, \mathbf{y}) \log_2 \frac{P_X(\mathbf{x}) P_Y(\mathbf{y})}{P_{X,Y}(\mathbf{x}, \mathbf{y})} \\ &= H(X) - H(X \mid Y) \\ &= H(Y) - H(Y \mid X) \end{aligned} \tag{6.6}$$

The mean conditional mutual information among the set of random variables $X = \{X_1, \ldots, X_p\}$ and the set of random variables $Y = \{Y_1, \ldots, Y_q\}$, given the values of a third set of random variables $Z = \{Z_1, \ldots, Z_r\}$, is defined by

$$I(X; Y \mid Z) = H(X \mid Z) - H(X \mid Y, Z) \tag{6.7}$$

1 To avoid problems, we suppose that all probabilities are strictly positive, without fundamental limitation.

$$= H(Y \mid Z) - H(Y \mid X, Z)$$
$$= -\sum_{\mathbf{x}\in\mathcal{X}}\sum_{\mathbf{y}\in\mathcal{Y}}\sum_{\mathbf{z}\in\mathcal{Z}} P_{X,Y,Z}(\mathbf{x},\mathbf{y},\mathbf{z}) \log_2 \frac{P_{X|Z}(\mathbf{x}\mid\mathbf{z})P_{Y|Z}(\mathbf{y}\mid\mathbf{z})}{P_{X,Y|Z}(\mathbf{x},\mathbf{y}\mid\mathbf{z})}$$

We also recall the chaining rule

$$I(X, Z; Y \mid W) = I(X; Y \mid W) + I(Z; Y \mid W, X), \tag{6.8}$$

and the symmetry of the (conditional) mutual information among sets of random variables

$$I(X; Y \mid Z) = I(Y; X \mid Z). \tag{6.9}$$

### 6.2.2 *Asymptotic analysis*

Let us now consider the MDI importance as defined by Equation 6.2, and let us assume a set $V = \{X_1, ..., X_p\}$ of *categorical* input variables and a *categorical* output $Y$. For the sake of simplicity we will use the Shannon entropy as impurity measure, and focus on totally randomized trees; later on we will discuss other impurity measures and tree construction algorithms.

Given a training sample $\mathcal{L}$ of $N$ joint observations of $X_1, ..., X_p, Y$ independently drawn from the joint distribution $P(X_1, ..., X_p, Y)$, let us assume that we infer from it an infinitely large ensemble of *totally randomized and fully developed trees*.

**Definition 6.1.** *A* totally randomized and fully developed tree *is a decision tree in which each node $t$ is partitioned using a variable $X_j$ picked uniformly at random among those not yet used at the parent nodes of $t$, where each node $t$ is split into $|\mathcal{X}_j|$ sub-trees, i.e., one for each possible value of $\mathcal{X}_j$, and where the recursive construction process halts only when all $p$ variables have been used along the current branch.*

In such a tree, leaves are all at the same depth $p$, and the set of leaves of a fully developed tree is in bijection with the set $\mathcal{X}$ of all possible joint configurations of the $p$ input variables. For example, if the input variables are all binary, the resulting tree will have exactly $2^p$ leaves. As a result, totally randomized trees, as considered here, are in fact closer to trees grown with ID3 [Quinlan, 1986] than trees built from the CART procedure (as described throughout Chapters 3 and 4).

**Theorem 6.1.** *The MDI importance of $X_j \in V$ for $Y$ as computed with an infinite ensemble of fully developed totally randomized trees and an infinitely large training sample is:*

$$Imp(X_j) = \sum_{k=0}^{p-1} \frac{1}{C_p^k} \frac{1}{p-k} \sum_{B \in \mathcal{P}_k(V^{-j})} I(X_j; Y|B), \tag{6.10}$$

where $V^{-j}$ denotes the subset $V \setminus \{X_j\}$, $\mathcal{P}_k(V^{-j})$ is the set of subsets of $V^{-j}$ of cardinality $k$, and $I(X_j;Y|B)$ is the conditional mutual information of $X_j$ and $Y$ given the variables in $B$.

*Proof.* By expanding $\Delta i(s,t) = i(t) - p_L i(t_L) - p_R i(t_R)$ into Equation 6.2 and using the entropy $H(Y|t) = -\sum_j p(j|t)\log_2(p(j|t))$ as impurity measure $i(t)$, Equation 6.2 can be rewritten in terms of mutual information:

$$\text{Imp}(X_j) = \frac{1}{M}\sum_{m=1}^{M}\sum_{t\in\varphi_m} 1(j_t = j)p(t)I(Y;X_j|t) \tag{6.11}$$

As the size $N$ of the training sample grows to infinity, $p(t)$ becomes the (exact) probability (according to $P(X_1,\ldots,X_p,Y)$) that an object reaches node $t$, i.e., a probability $P(B(t) = b(t))$ where $B(t) = (X_{i_1},...,X_{i_k})$ is the subset of $k$ variables tested in the branch from the root node to the parent of $t$ and $b(t)$ is the vector of values of these variables. As the the number $M$ of totally randomized trees also grows to infinity, the importance of a variable $X_j$ can then be written:

$$\text{Imp}(X_j) = \sum_{B\subseteq V^{-j}}\sum_{b\in\mathcal{X}_{i_1}\times...\times\mathcal{X}_{i_k}} \alpha(B,b,X_j,p)P(B=b)I(Y;X_j|B=b), \tag{6.12}$$

where $b$ is a set of values for the variables in $B$ and $\alpha(B,b,X_j,p)$ is the probability that a node $t$ (at depth $k$) in a totally randomized tree tests the variable $X_j$ and is such that $B(t) = B$ and $b(t) = b$.

Let us compute $\alpha(B,b,X_j,p)$. First, let us consider the probability that a node $t$ tests the variable $X_j$ and is such that the branch leading to $t$ follows a path defined, in that particular order, by all $k$ variables $X_{i_1},...,X_{i_k} \in B$ and their corresponding values in $b$. The probability of that branch is the probability of picking (uniformly at random) $X_{i_1}$ at the root node times the probability of testing, in that order, the remaining $X_{i_2},...,X_{i_k}$ variables in the sub-tree corresponding to the value $x_{i_1}$ of $X_{i_1}$ defined in $b$. Note that, by construction, it is certain that this particular sub-tree exists since the root node is split into $|\mathcal{X}_{i_1}|$ sub-trees. Then, the probability of testing $X_j$ at the end of this branch is the probability of picking $X_j$ among the remaining $p-k$ variables. By recursion, we thus have:

$$\frac{1}{p}\frac{1}{p-1}\cdots\frac{1}{p-k+1}\frac{1}{p-k} = \frac{(p-k)!}{p!}\frac{1}{p-k} \tag{6.13}$$

Since the order along which the variables appear in the branch is of no importance, $\alpha(B,b,X_j,p)$ actually includes all $k!$ equiprobable ways of building a branch composed of the variables and values in $B$ and $b$. Then, since a tree may at most contain a single such branch, whatever the order of the tests, the probabilities may be added up and it comes:

$$\alpha(B, b, X_j, p) = k! \frac{(p-k)!}{p!} \frac{1}{p-k} = \frac{1}{C_p^k} \frac{1}{p-k} \tag{6.14}$$

From the above expression, it appears that $\alpha(B, b, X_j, p)$ depends only on the size $k$ of $B$ and on the number $p$ of variables. As such, by grouping in the previous equation of $Imp(X_j)$ conditioning variable subsets $B$ according to their sizes and using the definition of conditional mutual information, $\alpha$ can be factored out, hence leading to the form foretold by Theorem 6.1:

$$Imp(X_j) = \sum_{k=0}^{p-1} \frac{1}{C_p^k} \frac{1}{p-k} \sum_{B \in \mathcal{P}_k(V^{-j})} I(X_j; Y|B). \tag{6.15}$$

□

**Theorem 6.2.** *For any ensemble of fully developed trees in asymptotic learning sample size conditions (e.g., in the same conditions as those of Theorem 6.1), we have that*

$$\sum_{j=1}^{p} Imp(X_j) = I(X_1, \dots, X_p; Y). \tag{6.16}$$

*Proof.* For any tree $\varphi$, we have that the sum of all importances estimated by using an infinitely large sample $\mathcal{L}$ (or equivalently, by assuming perfect knowledge of the joint distribution $P(X_1, ..., X_p, Y)$) is equal to $H(Y) - \sum_{t \in \widetilde{\varphi}} p(t) H(Y|b(t))$, where $\widetilde{\varphi}$ denotes the set of all leaves of $\varphi$, and where $b(t)$ denotes the joint configuration of all input variables leading to leaf $t$. This is true because the impurities of all test nodes intervening in the computation of the variable importances, except the impurity $H(Y)$ at the root node of the tree, cancel each other when summing up the importances.

Since, when the tree is fully developed, $\sum_{t \in \widetilde{\varphi}} p(t) H(Y|b(t))$ is obviously equal to the mean conditional entropy $H(Y|X_1, \dots, X_p)$ of $Y$ given all input variables, this implies that for any fully developed tree we have that the sum of variable importances is equal to $I(X_1, \dots, X_p; Y)$, and so this relation also holds when averaging over an infinite ensemble of totally randomized trees. Note that, for the same reasons as in the proof of Theorem 6.1, this result holds in asymptotic conditions only, i.e., when $p(t)$ becomes the exact probability $P(B(t) = b(t))$. □

Together, theorems 6.1 and 6.2 show that variable importances derived from totally randomized trees in asymptotic conditions provide a three-level decomposition of the information $I(X_1, \dots, X_p; Y)$ contained in the set of input variables about the output variable. The first level is a decomposition among input variables (see Equation 6.16 of Theorem 6.2), the second level is a decomposition along the degrees $k$ of interaction terms of a variable with the other ones (see the outer sum in Equation 6.10 of Theorem 6.1), and the third level is a decomposition along the combinations $B$ of interaction terms of fixed size $k$ of possible interacting variables (see the inner sum in Equation 6.10).

We observe that the decomposition includes, for each variable, each and every interaction term of each and every degree weighted in a fashion resulting only from the combinatorics of possible interaction terms. In particular, since all $I(X_j;Y|B)$ terms are at most equal to $H(Y)$, the prior entropy of $Y$, the $p$ terms of the outer sum of Equation 6.10 are each upper bounded by

$$\frac{1}{C_p^k}\frac{1}{p-k}\sum_{B\in\mathcal{P}_k(V^{-j})} H(Y) = \frac{1}{C_p^k}\frac{1}{p-k}C_{p-1}^k H(Y) = \frac{1}{p}H(Y). \quad (6.17)$$

As such, the second level decomposition resulting from totally randomized trees makes the $p$ sub-importance terms

$$\frac{1}{C_p^k}\frac{1}{p-k}\sum_{B\in\mathcal{P}_k(V^{-j})} I(X_j;Y|B) \quad (6.18)$$

to equally contribute (at most) to the total importance, even though they each include a combinatorially different number of terms.

## 6.3 RELEVANCE OF VARIABLES

Following Kohavi and John [1997], let us define relevant and irrelevant variable as follows:

**Definition 6.2.** *A variable* $X_j$ *is* relevant to $Y$ with respect to $V$ *if there exists at least one subset* $B \subseteq V$ *(possibly empty) such that* $I(X_j;Y|B) > 0$*.*

**Definition 6.3.** *A variable* $X_i$ *is* irrelevant to $Y$ with respect to $V$ *if, for all* $B \subseteq V$, $I(X_i;Y|B) = 0$*.*

Remark that if $X_i$ is irrelevant to $Y$ with respect to $V$, then by definition it is also irrelevant to $Y$ with respect to any subset of $V$. However, if $X_j$ is relevant to $Y$ with respect to $V$, then it is not necessarily relevant to $Y$ with respect to all subsets of $V$. Among the relevant variables, we also distinguish the *marginally* relevant ones, for which $I(X_j;Y) > 0$, the *strongly* relevant ones, for which $I(X_j;Y|V^{-j}) > 0$, and the *weakly* relevant variables, which are relevant but not strongly relevant.

**Theorem 6.3.** $X_i \in V$ *is irrelevant to* $Y$ *with respect to* $V$ *if and only if its infinite sample size importance as computed with an infinite ensemble of fully developed totally randomized trees built on* $V$ *for* $Y$ *is* $0$*.*

*Proof.* The proof directly results from the definition of irrelevance. If $X_i$ is irrelevant with respect to $V$, then $I(X_i;Y|B)$ is zero for all $B \subseteq V^{-i} \subset V$ and Equation 6.10 reduces to 0. Also, since $I(X_i;Y|B)$ is non-negative for any $B$, $\text{Imp}(X_i)$ is zero if and only if all its $I(X_i;Y|B)$ terms are zero. Since $\text{Imp}(X_i)$ includes all $I(X_i;Y|B)$ terms for $B \subseteq V^{-i}$, and since all of them are therefore null if $\text{Imp}(X_i) = 0$, $X_i$ is thus, by definition, irrelevant with respect to $V^{-i}$. $X_i$ is then also trivially irrelevant with respect to $V = V^{-i} \cup \{X_i\}$ since $I(X_i;Y|B \cup \{X_i\}) = 0$ for any $B$. □

**Lemma 6.4.** *Let $X_i \notin V$ be an irrelevant variable for $Y$ with respect to $V$. The infinite sample size importance of $X_j \in V$ as computed with an infinite ensemble of fully developed totally randomized trees built on $V$ for $Y$ is the same as the importance derived when using $V \cup \{X_i\}$ to build the ensemble of trees for $Y$.*

*Proof.* Let $X_i \notin V$ be an irrelevant variable with respect to $V$. For $X_j \in V$, $B \subseteq V^{-j}$, using the chain rules of mutual information, we have:

$$I(X_j, X_i; Y|B) = I(X_j;Y|B) + I(X_i;Y|B \cup \{X_j\}) \tag{6.19}$$

$$= I(X_i;Y|B) + I(X_j;Y|B \cup \{X_i\}) \tag{6.20}$$

If $X_i$ is irrelevant with respect to $V$, i.e., such that $I(X_i;Y|B) = 0$ for all $B \subseteq V$, then $I(X_i;Y|B \cup \{X_j\})$ and $I(X_i;Y|B)$ both equal 0, leading to

$$I(X_j;Y|B \cup \{X_i\}) = I(X_j;Y|B) \tag{6.21}$$

Then, from Theorem 6.1, the importance of $X_j$ as computed with an infinite ensemble of totally randomized trees built on $V \cup \{X_i\}$ can be simplified to:

$$\begin{aligned}
\text{Imp}(X_j) &= \sum_{k=0}^{p-1+1} \frac{1}{C_{p+1}^k} \frac{1}{p+1-k} \sum_{B \in \mathcal{P}_k(V^{-j} \cup \{X_i\})} I(X_j;Y|B) \\
&= \sum_{k=0}^{p} \frac{1}{C_{p+1}^k} \frac{1}{p+1-k} \left[ \sum_{B \in \mathcal{P}_k(V^{-j})} I(X_j;Y|B) + \sum_{B \in \mathcal{P}_{k-1}(V^{-j})} I(X_j;Y|B \cup \{X_i\}) \right] \\
&= \sum_{k=0}^{p-1} \frac{1}{C_{p+1}^k} \frac{1}{p+1-k} \sum_{B \in \mathcal{P}_k(V^{-j})} I(X_j;Y|B) + \\
&\hookrightarrow \sum_{k=1}^{p} \frac{1}{C_{p+1}^k} \frac{1}{p+1-k} \sum_{B \in \mathcal{P}_{k-1}(V^{-j})} I(X_j;Y|B) \\
&= \sum_{k=0}^{p-1} \frac{1}{C_{p+1}^k} \frac{1}{p+1-k} \sum_{B \in \mathcal{P}_k(V^{-j})} I(X_j;Y|B) + \\
&\hookrightarrow \sum_{k'=0}^{p-1} \frac{1}{C_{p+1}^{k'+1}} \frac{1}{p+1-k'-1} \sum_{B \in \mathcal{P}_{k'}(V^{-j})} I(X_j;Y|B)
\end{aligned}$$

$$= \sum_{k=0}^{p-1} \left[ \frac{1}{C_{p+1}^{k}} \frac{1}{p+1-k} + \frac{1}{C_{p+1}^{k+1}} \frac{1}{p-k} \right] \sum_{B \in \mathcal{P}_k(V^{-j})} I(X_j;Y|B)$$

$$= \sum_{k=0}^{p-1} \frac{1}{C_p^k} \frac{1}{p-k} \sum_{B \in \mathcal{P}_k(V^{-j})} I(X_j;Y|B) \tag{6.22}$$

The last line above exactly corresponds to the importance of $X_j$ as computed with an infinite ensemble of totally randomized trees built on $V$, which proves Lemma 6.4. □

**Theorem 6.5.** *Let $V_R \subseteq V$ be the subset of all variables in $V$ that are relevant to $Y$ with respect to $V$. The infinite sample size importance of any variable $X_j \in V_R$ as computed with an infinite ensemble of fully developed totally randomized trees built on $V_R$ for $Y$ is the same as its importance computed in the same conditions by using all variables in $V$. That is:*

$$Imp(X_j) = \sum_{k=0}^{p-1} \frac{1}{C_p^k} \frac{1}{p-k} \sum_{B \in \mathcal{P}_k(V^{-j})} I(X_j;Y|B)$$

$$= \sum_{l=0}^{r-1} \frac{1}{C_r^l} \frac{1}{r-l} \sum_{B \in \mathcal{P}_l(V_R^{-j})} I(X_j;Y|B) \tag{6.23}$$

*where $r$ is the number of relevant variables in $V_R$.*

*Proof.* Let us assume that $V_R$ contains $r \leqslant p$ relevant variables. If an infinite ensemble of totally randomized trees were to be built directly on those $r$ variables then, from Theorem 6.1, the importance of a relevant variable $X_j$ would be:

$$\text{Imp}(X_j) = \sum_{l=0}^{r-1} \frac{1}{C_r^l} \frac{1}{r-l} \sum_{B \in \mathcal{P}_l(V_R^{-m})} I(X_j;Y|B) \tag{6.24}$$

Let $X_i \in V \setminus V_R$ be one of the $p-r$ irrelevant variables in $V$ with respect to $V$. Since $X_i$ is also irrelevant with respect to $V_R$, using Lemma 6.4, the importance of $X_j$ when the ensemble is built on $V_R \cup \{X_i\}$ is the same as the one computed on $V_R$ only (i.e., as computed by the equation above). Using the same argument, adding a second irrelevant variable $X_{i'}$ with respect to $V$ – and therefore also with respect to $V_R \cup \{X_i\}$ – and building an ensemble of totally randomized trees on $V_R \cup \{X_i\} \cup \{X_{i'}\}$ will yield importances that are the same as those computed on $V_R \cup \{X_i\}$, which are themselves the same as those computed by an ensemble built on $V_R$. By induction, adding all $p-r$ irrelevant variables has therefore no effect on the importance of $X_j$, which means that:

$$\text{Imp}(X_j) = \sum_{k=0}^{p-1} \frac{1}{C_p^k} \frac{1}{p-k} \sum_{B \in \mathcal{P}_k(V^{-j})} I(X_j;Y|B)$$

$$= \sum_{l=0}^{r-1} \frac{1}{C_r^l} \frac{1}{r-l} \sum_{B \in \mathcal{P}_l(V_R^{-m})} I(X_j; Y|B) \tag{6.25}$$

□

Theorems 6.3 and 6.5 show that the importances computed with an ensemble of totally randomized trees depends only on the relevant variables. Irrelevant variables have a zero importance and do not affect the importance of relevant variables. Practically, we believe that such properties are desirable conditions for a sound criterion assessing the importance of a variable. Indeed, noise should not be credited of any importance and should not make any other variable more (or less) important.

Intuitively, the independence with respect to irrelevant variables can be partly attributed to the fact that splitting at $t$ on some irrelevant variable $X_i$ should only dillute the local importance $p(t)\Delta i(t)$ of a relevant variable $X_j$ into the children $t_L$ and $t_R$, but not affect the total sum. For instance, if $X_j$ was to be used at $t$, then the local importance would be proportional to $p(t)$. By contrast, if $X_i$ was to be used at $t$ and $X_j$ at $t_L$ and $t_R$, then the sum of the local importances for $X_j$ would be proportional to $p(t_L) + p(t_R) = p(t)$, which does not change anything. Similarly, one can recursively invoke the same argument if $X_j$ was to be used deeper in $t_L$ or $t_R$.

A second reason comes from the fact that local importances are collected only in nodes $t$ where $X_j$ is used. By contrast, if local importances were summed over all nodes (e.g., using surrogate splits), then it would necessarily depend on the total number of nodes in a tree, which itself directly depends on $p$ – that is, not on $r$.

Finally, it is also worth noting that this result is consistent with the work of Biau [2012], who proved that rate of convergence of forests of randomized trees also only depends on the relevant variables.

## 6.4 VARIABLE IMPORTANCES IN RANDOM FOREST VARIANTS

In this section, we consider and discuss variable importances as computed with other types of ensembles of randomized trees. We first show how our results extend to any other impurity measure, and then analyze importances computed by depth-pruned ensemble of randomized trees and those computed by randomized trees built on random subspaces of fixed size. Finally, we discuss the case of non-totally randomized trees.

### 6.4.1 *Generalization to other impurity measures*

Although our characterization in sections 6.1 and 6.3 uses Shannon entropy as the impurity measure, theorems 6.1, 6.3 and 6.5 hold for other impurity measures, simply substituting conditional mutual information for conditional impurity reduction in the different formulas and in the definition of irrelevant variables. In particular, our results thus hold for the Gini index in classification and can be extended to regression problems using variance as the impurity measure.

Let us consider a generic impurity measure $i(Y|t)$ and, by mimicking the notation used for conditional mutual information, let us denote by $G(Y;X_j|t)$ the impurity decrease for a split on $X_j$ at node $t$:

$$G(Y;X_j|t) = i(Y|t) - \sum_{x \in \mathcal{X}_j} p(t_x) i(Y|t_x), \tag{6.26}$$

where $t_x$ denotes the successor node of $t$ corresponding to value $x$ of $X_j$. The importance score associated to a variable $X_j$ (see Equation 6.2) is then rewritten:

$$\text{Imp}(X_j) = \frac{1}{M} \sum_{m=1}^{M} \sum_{t \in \varphi_m} 1(j_t = j) p(t) G(Y;X_j|t). \tag{6.27}$$

As explained in the proof of Theorem 6.1, conditioning over a node $t$ is equivalent to conditioning over an event of the form $B(t) = b(t)$, where $B(t)$ and $b(t)$ denote respectively the set of variables tested in the branch from the root to $t$ and their values in this branch. When the learning sample size $N$ grows to infinity, this yields the following population based impurity decrease at node $t$:

$$\begin{aligned} & G(Y;X_j|B(t) = b(t)) \qquad (6.28) \\ & = i(Y|B(t) = b(t)) - \sum_{x \in \mathcal{X}_j} P(X_j = x|B(t) = b(t)) i(Y|B(t) = b(t), X_j = x) \end{aligned}$$

Again by analogy with conditional entropy and mutual information[2], let us define $i(Y|B)$ and $G(Y;X_j|B)$ for some subset of variables $B \subseteq V$ as follows:

$$i(Y|B) = \sum_{b} P(B = b) i(Y|B = b) \tag{6.29}$$

$$\begin{aligned} G(Y;X_j|B) &= \sum_{b} P(B = b) G(Y;X_j|B = b) \qquad (6.30) \\ &= i(Y|B) - i(Y|B, X_j) \end{aligned}$$

where the sums run over all possible combinations $b$ of values for the variables in $B$.

With these notations, the proof of Theorem 6.1 can be easily adapted to lead to the following generalization of Equation 6.10:

$$\text{Imp}(X_j) = \sum_{k=0}^{p-1} \frac{1}{C_p^k} \frac{1}{p-k} \sum_{B \in \mathcal{P}_k(V^{-j})} G(Y;X_j|B). \tag{6.31}$$

2 Note however that $G(Y;X_j|B)$ does not share all properties of conditional mutual information as for example $G(X_j;Y|B)$ might not be equal to $G(Y;X_j|B)$ or even be defined, depending on the impurity measure and the nature of the output $Y$.

Note that this generalization is valid without any further specific constraints on the impurity measure $i(Y|t)$.

Let us now define as *irrelevant to* $Y$ *with respect to* $V$ a variable $X_i$ for which, for all $B \subseteq V$, $G(Y; X_i|B) = 0$ (i.e. a variable that neither affects impurity whatever the conditioning). From this definition, one can deduce the following property of an irrelevant variable $X_i$ (for all $B \subseteq V$ and $X_j \in V$):

$$G(Y; X_j|B \cup \{X_i\}) = G(Y; X_j|B).$$

Indeed, by a simple application of previous definitions, we have:

$$\begin{aligned}
&G(Y; X_j|B) - G(Y; X_j|B \cup \{X_i\}) \\
&= i(Y|B) - i(Y|B \cup \{X_j\}) - i(Y|B \cup \{X_i\}) + i(Y|B \cup \{X_i, X_j\}) \\
&= i(Y|B) - i(Y|B \cup \{X_i\}) - i(Y|B \cup \{X_j\}) + i(Y|B \cup \{X_i, X_j\}) \\
&= G(Y; X_i|B) - G(Y; X_i|B \cup \{X_j\}) \\
&= 0,
\end{aligned} \tag{6.32}$$

where the last step is a consequence of the irrelevance of $X_i$.

Using this property, the proofs of Lemma 6.4 and Theorem 6.5 can be straightforwardly adapted, showing that, in the general case also, the MDI importance of a variable is invariant with respect to the removal or the addition of irrelevant variables.

Given the general definition of irrelevance, all irrelevant variables also get zero MDI importance but, without further constraints on the impurity measure $i$, there is no guarantee that all relevant variables (defined as all variables that are not irrelevant) will get a non zero importance. This property, and in consequence theorem 6.3, will be however satisfied as soon as the impurity measure is such that $G(Y; X_j|B) \geqslant 0$ for all $X_j \in V$ and for all $B \subseteq V$.

Previous developments show that all results presented in this chapter remain valid for any impurity measure leading to non negative impurity decreases, provided that the definition of variable irrelevance is adapted to this impurity measure. The choice of a specific impurity measure should thus be guided by the meaning one wants to associate to irrelevance.

Measuring impurity with Shannon entropy, i.e., taking $i(Y|t) = H(Y|t)$ and $i(Y|B = b) = H(Y|B = b)$, one gets back all previous results. Given the properties of conditional mutual information, irrelevance for this impurity measure strictly coincides with conditional independence: a variable $X_i$ is irrelevant to $Y$ with respect to $V$ if and only if $X_i \perp Y|B$ for all $B \subseteq V$.

A common alternative to Shannon entropy for growing classification trees is Gini index, which, in the finite and infinite sample cases, is written:

$$i(Y|t) = -\sum_j p(j|t)(1 - p(j|t)) \tag{6.33}$$

$$i(Y|B=b) = -\sum_j P(Y=j|B=b)(1-P(Y=j|B=b)). \tag{6.34}$$

Like the Shannon entropy, this measure leads to non negative impurity decreases and the corresponding notion of irrelevance is also directly related to conditional independence.

The most common impurity measure for regression is variance, which, in the finite and infinite sample cases, is written:

$$i(Y|t) = \frac{1}{N_t}\sum_{i\in t}(y_i - \frac{1}{N_t}\sum_{i\in t} y_i)^2 \tag{6.35}$$

$$i(Y|B=b) = E_{Y|B=b}\{(Y - E_{Y|B=b}\{Y\})^2\}. \tag{6.36}$$

Variance can only decrease as a consequence of a split and therefore, Theorem 6.3 is also valid for this impurity measure, meaning that only irrelevant variables will get a zero variance reduction. Note however that with this impurity measure, irrelevance is not directly related to conditional independence, as some variable $X_i$ can be irrelevant in the sense of our definition and still affects the distribution of output values.

### 6.4.2 *Pruning and random subspaces*

In sections 6.1 and 6.3, we considered totally randomized trees that were fully developed, i.e. until all $p$ variables were used within each branch. When totally randomized trees are developed only up to some smaller depth $q \leqslant p$, we show in Proposition 6.6 that the variable importances as computed by these trees is limited to the $q$ first terms of Equation 6.10. We then show in Proposition 6.7 that these latter importances are actually the same as when each tree of the ensemble is fully developed over a random subspace [Ho, 1998] of $q$ variables drawn prior to its construction.

**Proposition 6.6.** *The importance of $X_j \in V$ for $Y$ as computed with an infinite ensemble of pruned totally randomized trees built up to depth $q \leqslant p$ and an infinitely large training sample is:*

$$Imp(X_j) = \sum_{k=0}^{q-1} \frac{1}{C_p^k}\frac{1}{p-k} \sum_{B\in\mathcal{P}_k(V^{-j})} I(X_j;Y|B) \tag{6.37}$$

*Proof.* The proof of Theorem 6.1 can be directly adapted to prove Proposition 6.6. If the recursive procedure is stopped at depth $q$, then it means that $B(t)$ may include up to $q-1$ variables, which is strictly equivalent to summing from $k=0$ to $q-1$ in the outer sum of Equation 6.10. □

**Proposition 6.7.** *The importance of $X_j \in V$ for $Y$ as computed with an infinite ensemble of pruned totally randomized trees built up to depth $q \leqslant p$* *and an infinitely large training sample is identical to the importance as computed for* $Y$ *with an infinite ensemble of fully developed totally randomized trees built on random subspaces of* $q$ *variables drawn from* $V$*.*

*Proof.* Let us define a random subspace of size $q$ as a random subset $V_S \subseteq V$ such that $|V_S| = q$. By replacing $p$ with $q$ in Equation 6.10 (since each tree is built on $q$ variables) and adjusting by the probability

$$\frac{C^{q-k-1}_{p-k-1}}{C^q_p}$$

of having selected $X_j$ and the $k$ variables in the branch when drawing $V_S$ prior to the construction of the tree, it comes:

$$\text{Imp}(X_j) = \sum_{k=0}^{q-1} \frac{C^{q-k-1}_{p-k-1}}{C^q_p} \frac{1}{C^k_q} \frac{1}{q-k} \sum_{B \in \mathcal{P}_k(V^{-j})} I(X_j; Y|B) \tag{6.38}$$

The multiplicative factor in the outer sum can then be simplified as follows:

$$\begin{aligned}
\frac{C^{q-k-1}_{p-k-1}}{C^q_p} \frac{1}{C^k_q} \frac{1}{q-k} &= \frac{\frac{(p-k-1)!}{(p-k)!(q-k-1)!}}{\frac{p!}{(p-q)!q!}} \frac{1}{C^k_q} \frac{1}{q-k} \\
&= \frac{(p-k-1)!q!}{(q-k-1)!p!} \frac{1}{C^k_q} \frac{1}{q-k} \\
&= \frac{q(q-1)...(q-k)}{p(p-1)...(p-k)} \frac{1}{C^k_q} \frac{1}{q-k} \\
&= \frac{q(q-1)...(q-k)}{p(p-1)...(p-k)} \frac{k!(q-k)!}{q!} \frac{1}{q-k} \\
&= \frac{1}{p(p-1)...(p-k)} \frac{k!(q-k)!}{(q-k-1)!} \frac{1}{q-k} \\
&= \frac{k!}{p(p-1)...(p-k)} \\
&= \frac{k!(p-k)!}{p!} \frac{1}{p-k} \\
&= \frac{1}{C^k_p} \frac{1}{p-k}
\end{aligned} \tag{6.39}$$

which yields the same importance as in Proposition 6.6 and proves the proposition. □

As long as $q \geqslant r$ (where $r$ denotes the number of relevant variables in V), it can easily be shown that all relevant variables will still obtain a strictly positive importance, which will however differ in general from the importances computed by fully grown totally randomized trees built over all variables. Also, each irrelevant variable of course keeps an importance equal to zero, which means that, in asymptotic conditions, these pruning and random subspace methods would still allow us identify the relevant variables, as long as we have a good upper bound $q$ on $r$.

### 6.4.3 *Non-totally randomized trees*

In our analysis in the previous sections, trees are built totally at random and hence do not directly relate to those built in Random Forest [Breiman, 2001] or in Extremely Randomized Trees [Geurts et al., 2006a]. To better understand the importances as computed by those algorithms, let us consider a close variant of totally randomized trees: at each node $t$, let us instead draw uniformly at random $1 \leqslant K \leqslant p$ variables and let us choose the one that maximizes $\Delta i(t)$. As previously, $t$ is split into as many subtrees as the cardinality of the chosen variable. Asymptotically, for binary variables, this variant exactly matches Random Forests and Extremely Randomized Trees. For variables with a larger cardinality, the correspondence no longer exactly holds but the trees still closely relate. Notice that, for $K = 1$, this procedure amounts to building ensembles of totally randomized trees as defined before, while, for $K = p$, it amounts to building classical single trees in a deterministic way.

First, the importance of $X_j \in V$ as computed with an infinite ensemble of such randomized trees is not the same as Equation 6.10. For $K > 1$, masking effects indeed appear: at $t$, some variables are never selected because there always is some other variable for which $\Delta i(t)$ is larger. Such effects tend to pull the best variables at the top of the trees and to push the others at the leaves. As a result, the importance of a variable no longer decomposes into a sum including all $I(X_j; Y|B)$ terms. The importance of the best variables decomposes into a sum of their mutual information alone or conditioned only with the best others – but not conditioned with all variables since they no longer ever appear at the bottom of trees. By contrast, the importance of the least promising variables now decomposes into a sum of their mutual information conditioned only with all variables – but not alone or conditioned with a couple of others since they no longer ever appear at the top of trees. In other words, because of the guided structure of the trees, the importance of $X_j$ now takes into account only some of the conditioning sets $B$, which may over- or underestimate its overall relevance.

To make things clearer, let us consider a simple example. Let $X_1$ perfectly explains $Y$ and let $X_2$ be a slightly noisy copy of $X_1$ (i.e., $I(X_1; Y) \approx I(X_2; Y)$, $I(X_1; Y|X_2) = \epsilon$ and $I(X_2; Y|X_1) = 0$). Using totally randomized trees, the importances of $X_1$ and $X_2$ are nearly equal – the importance of $X_1$ being slightly higher than the importance of $X_2$:

$$\text{Imp}(X_1) = \frac{1}{2}I(X_1;Y) + \frac{1}{2}I(X_1;Y|X_2) = \frac{1}{2}I(X_1;Y) + \frac{\epsilon}{2} \tag{6.40}$$

$$\text{Imp}(X_2) = \frac{1}{2}I(X_2;Y) + \frac{1}{2}I(X_2;Y|X_1) = \frac{1}{2}I(X_2;Y) + 0 \tag{6.41}$$

In non-totally randomized trees, for $K = 2$, $X_1$ is always selected at the root node and $X_2$ is always used in its children. Also, since $X_1$ perfectly explains $Y$, all its children are pure and the reduction of entropy when splitting on $X_2$ is null. As a result, $\text{Imp}_{K=2}(X_1) = I(X_1;Y)$ and $\text{Imp}_{K=2}(X_2) = I(X_2;Y|X_1) = 0$. Masking effects are here clearly visible: the true importance of $X_2$ is masked by $X_1$ as if $X_2$ were irrelevant, while it is only a bit less informative than $X_1$.

As a direct consequence of the example above, for $K > 1$, it is no longer true that a variable is irrelevant if and only if its importance is zero. In the same way, it can also be shown that the importances become dependent on the number of irrelevant variables. Let us indeed consider the following counter-example: let us add in the previous example an irrelevant variable $X_i$ with respect to $\{X_1, X_2\}$ and let us keep $K = 2$. The probability of selecting $X_2$ at the root node now becomes positive, which means that $\text{Imp}_{K=2}(X_2)$ now includes $I(X_2;Y) > 0$ and is therefore strictly larger than the importance computed before. For $K$ fixed, adding irrelevant variables dampens masking effects, which thereby makes importances indirectly dependent on the number of irrelevant variables.

In conclusion, the importances as computed with totally randomized trees exhibit properties that do not possess, by extension, neither Random Forests nor Extremely Randomized Trees. With totally randomized trees, the importance of $X_j$ only depends on the relevant variables and is $0$ if and only if $X_j$ is irrelevant. As we have shown, it may no longer be the case for $K > 1$. Asymptotically, the use of totally randomized trees for assessing the importance of a variable may therefore be more appropriate. In a finite setting (i.e., a limited number of samples and a limited number of trees), guiding the choice of the splitting variables remains however a sound strategy. In such a case, $I(X_j;Y|B)$ cannot be measured neither for all $X_j$ nor for all $B$. It is therefore pragmatic to promote those that look the most promising, even if the resulting importances may be biased.

## 6.5 illustration

In this section, we consider the digit recognition problem of [Breiman et al., 1984] for illustrating variable importances as computed with totally randomized trees. We verify that they match with our theoretical developments and that they decompose as foretold. We also compare these importances with those computed by an ensemble of non-totally randomized trees, as discussed in section 6.4.3.

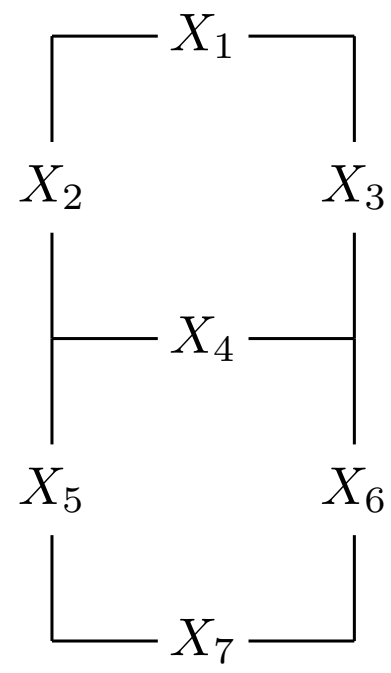

Figure 6.1: 7-segment display

Let us consider a seven-segment indicator displaying numerals using horizontal and vertical lights in on-off combinations, as illustrated in Figure 6.1. Let $Y$ be a random variable taking its value in $\{0, 1, ..., 9\}$ with equal probability and let $X_1, ..., X_7$ be binary variables whose values are each determined univocally given the corresponding value of $Y$ in Table 6.1.

Since Table 6.1 perfectly defines the data distribution, and given the small dimensionality of the problem, it is practicable to directly apply Equation 6.10 to compute variable importances. To verify our theoretical developments, we then compare in Table 6.2 variable importances as computed by Equation 6.10 and those yielded by an ensemble of $10000$ totally randomized trees ($K = 1$). Note that given the known structure of the problem, building trees on a sample of finite size that perfectly follows the data distribution amounts to building them on a sample of infinite size. At best, trees can thus be built on a 10-sample dataset, containing exactly one sample for each of the equiprobable outcomes of $Y$. As the table illustrates, the importances yielded by totally randomized trees match those computed by Equation 6.10, which confirms Theorem 6.1. Small differences stem from the fact that a finite number of trees were built in our simulations, but those discrepancies should disappear as the size of the ensemble grows towards infinity. It also shows that importances indeed add up to $I(X_1, ...X_7; Y)$, which confirms Theorem 6.2. Regarding the actual importances, they indicate that $X_5$ is stronger than all others, followed – in that order – by $X_2$, $X_4$ and $X_3$ which also show large importances. $X_1$, $X_7$ and $X_6$ appear to be the less informative. The table also reports importances for increasing values of $K$. As discussed before, we see that a large value of $K$ yields importances that can be either overestimated (e.g., at $K = 7$, the importances of $X_2$ and $X_5$ are larger than at $K = 1$) or underestimated due to masking effects (e.g., at $K = 7$, the importances of $X_1$, $X_3$, $X_4$ and $X_6$ are smaller than at $K = 1$, as if they were less important). As Figure 6.2 illustrates, it can also be observed that masking effects may even induce changes in the variable rankings (e.g., compare the rankings at $K = 1$ and at $K = 7$), which thus confirms that importances are differently affected.

| y | $x_1$ | $x_2$ | $x_3$ | $x_4$ | $x_5$ | $x_6$ | $x_7$ |
|---|---|---|---|---|---|---|---|
| 0 | 1 | 1 | 1 | 0 | 1 | 1 | 1 |
| 1 | 0 | 0 | 1 | 0 | 0 | 1 | 0 |
| 2 | 1 | 0 | 1 | 1 | 1 | 0 | 1 |
| 3 | 1 | 0 | 1 | 1 | 0 | 1 | 1 |
| 4 | 0 | 1 | 1 | 1 | 0 | 1 | 0 |
| 5 | 1 | 1 | 0 | 1 | 0 | 1 | 1 |
| 6 | 1 | 1 | 0 | 1 | 1 | 1 | 1 |
| 7 | 1 | 0 | 1 | 0 | 0 | 1 | 0 |
| 8 | 1 | 1 | 1 | 1 | 1 | 1 | 1 |
| 9 | 1 | 1 | 1 | 1 | 0 | 1 | 1 |

Table 6.1: Values of $Y, X_1, ..., X_7$

To better understand why a variable is important, it is also insightful to look at its decomposition into its $p$ sub-importances terms, as shown in Figure 6.3. Each row in the plots of the figure corresponds to one the $p$ variables $X_1, \dots, X_7$ and each column to a size $k$ of conditioning sets. As such, the value at row $j$ and column $k$ corresponds the importance of $X_j$ when conditioned with $k$ other variables (e.g., to the term $\frac{1}{C_p^k}\frac{1}{p-k}\sum_{B \in \mathcal{P}_k(V^{-j})} I(X_j;Y|B)$ in Equation 6.10 in the case of totally randomized trees). In the left plot, for $K = 1$, the figure first illustrates how importances yielded by totally randomized trees decomposes along the degrees $k$ of interactions terms. We can observe that they each equally contribute (at most) the total importance of a variable. The plot also illustrates why $X_5$ is important: it is informative either alone or conditioned with any combination of the other variables (all of its terms are significantly larger than 0). By contrast, it also clearly shows why $X_6$ is not important: neither alone nor combined with others $X_6$ seems to be very informative (all of its terms are close to 0). More interestingly, this figure also highlights redundancies: $X_7$ is informative alone or conditioned with a couple of others (the first terms are significantly larger than 0), but becomes uninformative when conditioned with many others (the last terms are closer to 0). The right plot, for $K = 7$, illustrates the decomposition of importances when variables are chosen in a deterministic way. The first thing to notice is masking effects. Some of the $I(X_j;Y|B)$ terms are indeed clearly never encountered and their contribution is therefore reduced to 0 in the total importance. For instance, for $k = 0$, the sub-importances of $X_2$ and $X_5$ are positive, while all others are null, which means that only those two variables are ever selected at the root node, hence masking the others. As a consequence, this also means that the importances of the remaining variables are biased and that they actually only account of their relevance when conditioned to $X_2$ or $X_5$, but not of their relevance in other contexts. At $k = 0$, masking effects also amplify the contribution of $I(X_2; Y)$ (resp. $I(X_5; Y)$) since $X_2$ (resp. $X_5$) appears more frequently at the root node than in totally randomized trees. In addition, because nodes become pure before reaching depth $p$, conditioning sets of size $k \geqslant 4$ are never actually encountered, which means that we can no longer know whether variables are still informative when conditioned to many others. All in all, this figure thus indeed confirms that importances as computed with non-totally randomized trees take into account only some of the conditioning sets $B$, hence biasing the measured importances.

| | Eqn. 6.10 | $K=1$ | $K=2$ | $K=3$ | $K=4$ | $K=5$ | $K=6$ | $K=7$ |
|---|---|---|---|---|---|---|---|---|
| $X_1$ | 0.412 | 0.414 | 0.362 | 0.327 | 0.309 | 0.304 | 0.305 | 0.306 |
| $X_2$ | 0.581 | 0.583 | 0.663 | 0.715 | 0.757 | 0.787 | 0.801 | 0.799 |
| $X_3$ | 0.531 | 0.532 | 0.512 | 0.496 | 0.489 | 0.483 | 0.475 | 0.475 |
| $X_4$ | 0.542 | 0.543 | 0.525 | 0.484 | 0.445 | 0.414 | 0.409 | 0.412 |
| $X_5$ | 0.656 | 0.658 | 0.731 | 0.778 | 0.810 | 0.827 | 0.831 | 0.835 |
| $X_6$ | 0.225 | 0.221 | 0.140 | 0.126 | 0.122 | 0.122 | 0.121 | 0.120 |
| $X_7$ | 0.372 | 0.368 | 0.385 | 0.392 | 0.387 | 0.382 | 0.375 | 0.372 |
| $\sum$ | 3.321 | 3.321 | 3.321 | 3.321 | 3.321 | 3.321 | 3.321 | 3.321 |

Table 6.2: Variable importances as computed with an ensemble of randomized trees, for increasing values of $K$. Importances at $K = 1$ follow their theoretical values, as predicted by Equation 6.10 in Theorem 6.1. However, as $K$ increases, importances diverge due to masking effects. In accordance with Theorem 6.2, their sum is also always equal to $I(X_1, \ldots, X_7; Y) = H(Y) = \log_2(10) = 3.321$ since inputs allow to perfectly predict the output.

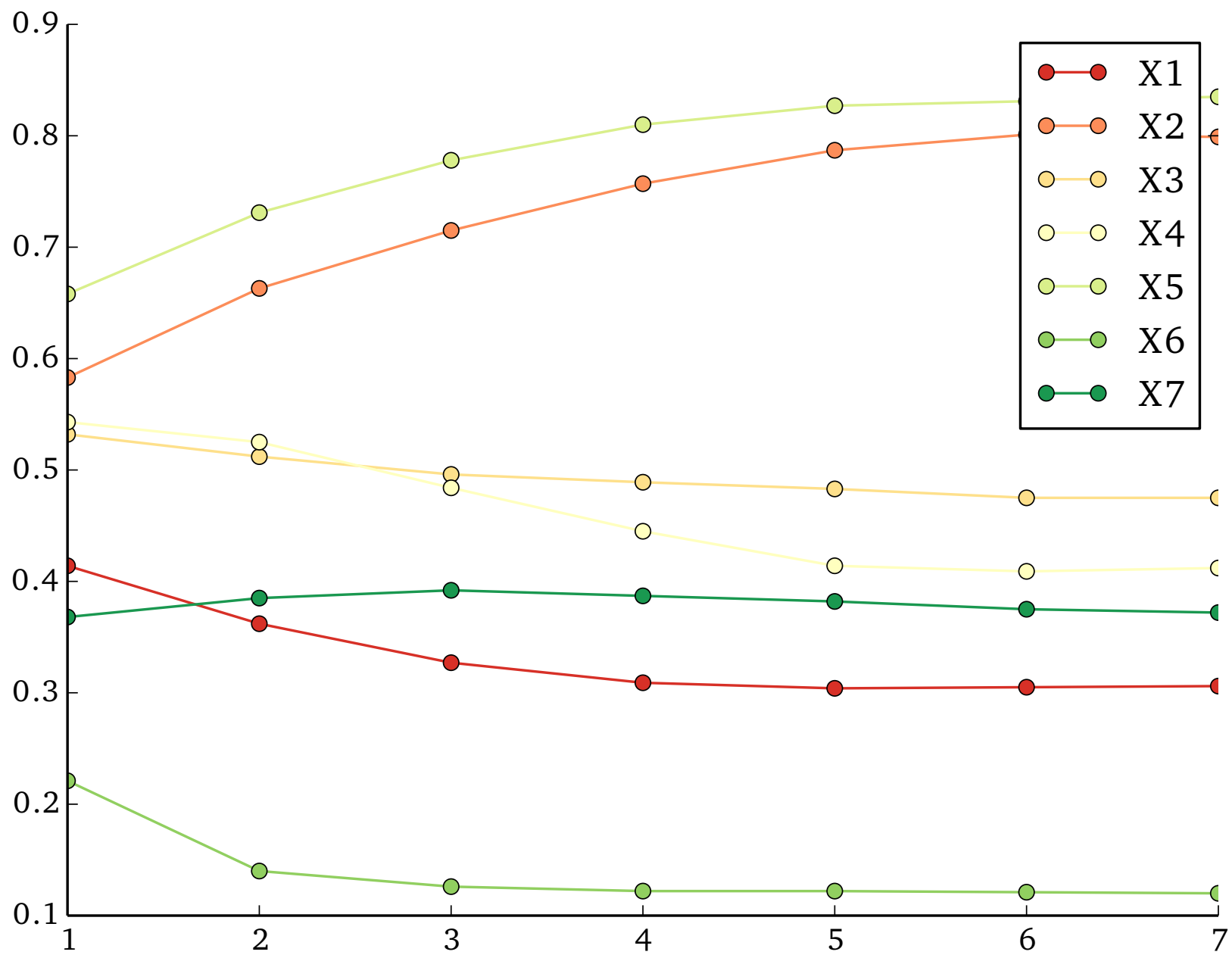

Figure 6.2: Variable importances as computed with an ensemble of randomized trees, for increasing values of $K$. Importances at $K = 1$ follow their theoretical values, as predicted by Equation 6.10 in Theorem 6.1. However, as $K$ increases, importances diverge due to masking effects.

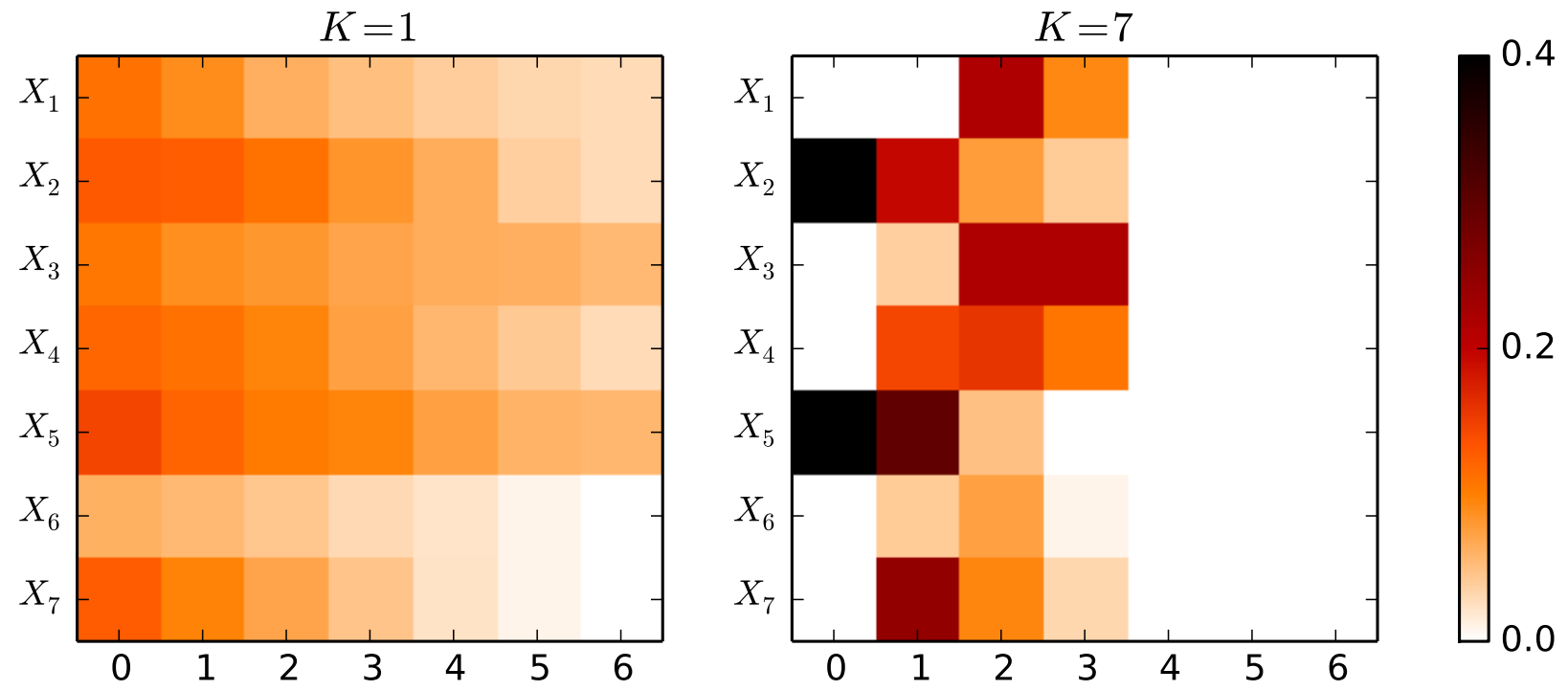

Figure 6.3: Decomposition of variable importances along the degrees $k$ of interactions of one variable with the other ones. At $K = 1$, all $I(X_j; Y|B)$ are accounted for in the total importance, while at $K = 7$ only some of them are taken into account due to masking effects.

## 6.6 CONCLUSIONS

In this chapter, we made a first step towards understanding variable importances as computed with a forest of randomized trees. In particular, we derived a theoretical characterization of the Mean Decrease Impurity importances as computed by totally randomized trees in asymptotic conditions. We showed that they offer a three-level decomposition of the information jointly provided by all input variables about the output (Section 6.2). We then demonstrated (Section 6.3) that MDI importances as computed by totally randomized trees exhibit desirable properties for assessing the relevance of a variable: it is equal to zero if and only if the variable is irrelevant and it depends only on the relevant variables. We discussed the case of Random Forests and Extremely Randomized Trees (Section 6.4) and finally illustrated our developments on an artificial but insightful example (Section 6.5).

There remain several limitations to our framework that we would like address in the future. First, our results should be adapted to binary splits as used within an actual Random Forest-like algorithm. In this setting, any node t is split in only two subsets, which means that any variable may then appear one or several times within a branch, and thus should make variable importances now dependent on the cardinalities of the input variables. In the same direction, our framework should also be extended to the case of continuous variables. Finally, results presented in this work are valid in an asymptotic setting only. An important direction of future work includes the characterization of the distribution of variable importances in a finite setting.

# 7

# FURTHER INSIGHTS FROM IMPORTANCES

**Outline**

In this chapter, we build upon results from Chapter 6 to further study variable importances as computed from random forests. In Section 7.1, we first examine importances for variables that are redundant. In Section 7.2, we revisit variable importances in the context of binary decision trees and ordered variables. In this framework, we highlight various sources of bias that may concurrently happen when importances are computed from actual random forests. Finally, we present in Section 7.3 some successful applications of variable importance measures.

**Caution.** *The work presented in this chapter is exploratory. Conclusions should be considered with a grain of salt, until further empirical verifications.*

## 7.1 REDUNDANT VARIABLES

In most machine learning problems, it is typical for input variables to be correlated, at least to some extent, and to share common bits of information. In image classification for instance, pixels are usually highly correlated and individually represent nearly the same information as their neighbors. In that sense, variables are often *partially redundant*, i.e., some of the variables may share some of the same information about the output variable $Y$. In the extreme case, redundancy is *total* or *complete*, with some of the variables redundantly conveying exactly the same information with respect to the output variable $Y$. In this section, we study redundancy in random forests and show that it may have a significant effect on both the accuracy of the ensemble and variable importance measures.

As a guiding example for our discussion, let us consider a set of input variables and let us discuss the effect of adding redundant variables on the structure of randomized trees. Intuitively, two variables $X_i$ and $X_j$ are redundant if one can perfectly explains the other and vice-versa. Formally, we define redundancy as follows:

**Definition 7.1.** *Two variables* $X_i, X_j$ *are* totally redundant *if no additional information is required for describing* $X_i$ *given* $X_j$ *and vice-versa. I.e., if*

$$H(X_i|X_j) = H(X_j|X_i) = 0. \tag{7.1}$$

In particular, a variable $X_j$ and its copy, denoted $X'_j$, are totally redundant. With respect to random forests, adding copies of variables (e.g., duplicating $X_j$, hence resulting in a new set of $p + 1$ input variables) has no effect when the selection of the split is deterministic (e.g., in RF for $K$ set to the maximum value). No matter the number of totally redundant variables, the best split that is selected is always the same, even if the same splits need to be recomputed multiple times due to redundancy. When the choice of the best split is stochastic however (e.g., for $K$ strictly smaller than the total number of variables), adding multiple copies of a variable $X_j$ results in splits that may be biased towards this variable (or one of its copies), which in turn may have a significant effect on the resulting accuracy of the ensemble. For a fixed value of $K$, it is indeed not difficult to see that adding copies of $X_j$ increases the probability of $X_j$, or of one of its copies, to be in the random subset of $K$ input variables on which to look for splits. As a corollary, it therefore also simultaneously decreases the probability of any of the others to be selected, hence biasing the structure of the generated decision trees. Note that the resulting net effect on accuracy depends on the nature of duplicated variable. If $X_j$ is very informative with respect to the input, then favoring splits on $X_j$ by adding copies may result in an increase of accuracy. By contrast, if $X_j$ is irrelevant, then adding copies increases the risk of overfitting.

With respect to variable importances, the effect of adding redundant variables can be derived both qualitatively and quantitatively using results from Chapter 6. From Theorem 6.5, we already know that adding irrelevant variables does not change the resulting variable importances. Adding copies of a relevant variable however, has an effect on both the importance of the duplicated variable and on the importance of the remaining variables. As in the previous chapter, let us assume a set $V = \{X_1, ..., X_p\}$ of categorical input variables and a categorical output $Y$, for which we derive MDI importances, as computed from totally randomized and fully developed trees built on an infinitely large dataset.

**Lemma 7.1.** *Let $X_i$ and $X_j$ be totally redundant variables. For any conditioning set $B$,*

$$I(X_i; Y|B, X_j) = I(X_j; Y|B, X_i) = 0 \tag{7.2}$$

$$I(X_i; Y|B) = I(X_j; Y|B). \tag{7.3}$$

*Proof.* By symmetry of the mutual information, it comes

$$\begin{aligned} I(X_i; X_j) &= H(X_i) - H(X_i|X_j) \\ &= H(X_j) - H(X_j|X_i), \end{aligned} \tag{7.4}$$

which implies that $H(X_i) = H(X_j)$ since $H(X_i|X_j) = H(X_j|X_i) = 0$ if $X_i$ and $X_j$ are totally redundant. Since $0 \leqslant H(X_i|X_j, B) \leqslant H(X_i|X_j)$ and $H(X_i|X_j) = 0$, we also have $H(X_i|X_j, B) = 0$ for any conditioning set $B$. Likewise, $H(X_j|X_i, B) = 0$. By reusing the same argument for $I(X_i; X_j|B)$ (instead of $I(X_i; X_j)$), equality therefore extends to any conditioning set B, giving $H(X_i|B) = H(X_j|B)$. From these, it follows that,

$$I(X_i; Y|B, X_j) = H(X_i|B, X_j) - H(X_i|B, X_j, Y) = 0 - 0, \tag{7.5}$$
$$I(X_j; Y|B, X_i) = H(X_j|B, X_i) - H(X_j|B, X_i, Y) = 0 - 0, \tag{7.6}$$

which proves Equation 7.2. We also have

$$\begin{aligned} I(X_i; Y|B) &= H(X_i|B) - H(X_i|B, Y) &(7.7)\\ &= H(X_j|B) - H(X_j|B, Y) &(7.8)\\ &= I(X_j; Y|B), &(7.9)\end{aligned}$$

which proves Equation 7.3. □

**Proposition 7.2.** *Let $X_j \in V$ be a relevant variable with respect to $Y$ and $V$ and let $X'_j \notin V$ be a totally redundant variable with respect to $X_j$. The infinite sample size importance of $X_j$ as computed with an infinite ensemble of fully developed totally randomized trees built on $V \cup \{X'_j\}$ is*

$$Imp(X_j) = \sum_{k=0}^{p-1} \frac{p-k}{p+1} \frac{1}{C_p^k} \frac{1}{p-k} \sum_{B \in \mathcal{P}_k(V^{-j})} I(X_j; Y|B) \tag{7.10}$$

*Proof.* From Theorem 6.1, the variable importance of $X_j$ is

$$\begin{aligned} \text{Imp}(X_j) &= \sum_{k=0}^{p-1+1} \frac{1}{C_{p+1}^k} \frac{1}{p+1-k} \sum_{B \in \mathcal{P}_k(V^{-j} \cup \{X'_j\})} I(X_j; Y|B)\\ &= \sum_{k=0}^{p-1} \frac{1}{C_{p+1}^k} \frac{1}{p+1-k} \sum_{B \in \mathcal{P}_k(V^{-j})} I(X_j; Y|B)\\ &= \sum_{k=0}^{p-1} \frac{p-k}{p+1} \frac{1}{C_p^k} \frac{1}{p-k} \sum_{B \in \mathcal{P}_k(V^{-j})} I(X_j; Y|B), &(7.11)\end{aligned}$$

since from Lemma 7.1, $I(X_j; Y|B \cup X'_j) = 0$ for all $B \in \mathcal{P}(V^{-j})$. □

**Lemma 7.3.** *Let $X_i$ and $X_j$ be totally redundant variables. For any conditioning set $B$ and for any variable $X_l$,*

$$I(X_l; Y|B, X_i) = I(X_l; Y|B, X_j) = I(X_l; Y|B, X_i, X_j) \tag{7.12}$$

*Proof.* From the chaining rule of the mutual information, we have

$$\begin{aligned} I(X_i, X_j, X_l; Y|B) &= I(X_l; Y|B) + I(X_i, X_j; Y|B, X_l)\\ &= I(X_l; Y|B) + I(X_i; Y|B, X_l) + I(X_i; Y|B, X_j, X_l)\\ &= I(X_l; Y|B) + I(X_i; Y|B, X_l) \quad \text{(Lemma 7.1)}\\ &= I(X_i, X_l; Y|B)\\ &= I(X_i; Y|B) + I(X_l; Y|B, X_i). &(7.13)\end{aligned}$$

By symmetry,

$$I(X_i, X_j, X_l; Y|B) = I(X_j; Y|B) + I(X_l; Y|B, X_j), \tag{7.14}$$

which proves that $I(X_l; Y|B, X_i) = I(X_l; Y|B, X_j)$, by combining both equations and using the fact that $I(X_i; Y|B) = I(X_j; Y|B)$ (Lemma 7.1).

From the chaining rule, we also have

$$\begin{aligned} I(X_i, X_j, X_l; Y|B) &= I(X_i, X_j; Y|B) + I(X_l; Y|B, X_i, X_j) \\ &= I(X_i; Y|B) + I(X_j; Y|B, X_i) + I(X_l; Y|B, X_i, X_j) \\ &= I(X_i; Y|B) + I(X_l; Y|B, X_i, X_j). \end{aligned} \tag{7.15}$$

By combining this last equation with Equation 7.13, we finally have $I(X_l; Y|B, X_i) = I(X_l; Y|B, X_i, X_j)$, which proves Lemma 7.3. □

**Proposition 7.4.** *Let $X_j \in V$ be a relevant variable with respect to $Y$ and $V$ and let $X'_j \notin V$ be a totally redundant variable with respect to $X_j$. The infinite sample size importance of $X_l \in V^{-j}$ as computed with an infinite ensemble of fully developed totally randomized trees built on $V \cup \{X'_j\}$ is*

$$\begin{aligned} Imp(X_l) = &\sum_{k=0}^{p-2} \frac{p-k}{p+1} \frac{1}{C_p^k} \frac{1}{p-k} \sum_{B \in \mathcal{P}_k(V^{-l} \setminus X_j)} I(X_l; Y|B) + \\ &\hookrightarrow \sum_{k=0}^{p-2} \left[ \sum_{k'=1}^{2} \frac{C_2^{k'}}{C_{p+1}^{k+k'}} \frac{1}{p+1-(k+k')} \right] \sum_{B \in \mathcal{P}_k(V^{-l} \setminus X_j)} I(X_l; Y|B \cup X_j). \end{aligned} \tag{7.16}$$

*Proof.* From Lemma 7.3, conditioning by either $X_j$, $X'_j$ or by both variables yield terms $I(X_l; Y|B, X_j)$, $I(X_l; Y|B, X'_j)$ and $I(X_l; Y|B, X_j, X'_j)$ that are all equal. From Theorem 6.1, the variable importance of $X_l$ can therefore be rewritten as follows:

$$\begin{aligned} Imp(X_l) &= \sum_{k=0}^{p-1+1} \frac{1}{C_{p+1}^k} \frac{1}{p+1-k} \sum_{B \in \mathcal{P}_k(V^{-l} \cup X'_j)} I(X_l; Y|B) \\ &= \sum_{k=0}^{p-2} \sum_{k'=0}^{2} \frac{1}{C_{p+1}^{k+k'}} \frac{1}{p+1-(k+k')} \sum_{\substack{B \in \mathcal{P}_k(V^{-l} \setminus X_j) \\ B' \in \mathcal{P}_{k'}(\{X_j, X'_j\})}} I(X_l; Y|B \cup B') \\ &= \sum_{k=0}^{p-2} \frac{1}{C_{p+1}^k} \frac{1}{p+1-k} \sum_{B \in \mathcal{P}_k(V^{-l} \setminus X_j)} I(X_l; Y|B) + \\ &\hookrightarrow \sum_{k=0}^{p-2} \left[ \sum_{k'=1}^{2} \frac{C_2^{k'}}{C_{p+1}^{k+k'}} \frac{1}{p+1-(k+k')} \right] \sum_{B \in \mathcal{P}_k(V^{-l} \setminus X_j)} I(X_l; Y|B \cup X_j) \\ &= \sum_{k=0}^{p-2} \frac{p-k}{p+1} \frac{1}{C_p^k} \frac{1}{p-k} \sum_{B \in \mathcal{P}_k(V^{-l} \setminus X_j)} I(X_l; Y|B) + \end{aligned}$$

$$\hookrightarrow \sum_{k=0}^{p-2} \left[ \sum_{k'=1}^{2} \frac{C_2^{k'}}{C_{p+1}^{k+k'}} \frac{1}{p+1-(k+k')} \right] \sum_{B \in \mathcal{P}_k(V^{-l} \setminus X_j)} I(X_l; Y | B \cup X_j).$$

□

Proposition 7.2 shows that the importance of $X_j$ decreases when a redundant variable $X_j'$ is added to the set of input variables, since all mutual information terms are multiplied by a factor $\frac{p-k}{p+1} < 1$. Intuitively, this result is in fact expected since the same information is then conveyed within two variables (i.e., in $X_j$ and its copy $X_j'$). It also shows that the terms in the total importance are not all modified in the same way. The weight of the terms corresponding to small conditioning sets remains nearly unchanged (i.e., for a large number $p$ of variables and small values of $k$, $\frac{p-k}{p+1}$ is close to 1), while the weight of the terms of large conditioning sets is greatly impacted (i.e., for large values of $k$, $\frac{p-k}{p+1}$ tends to 0).

As shown by Proposition 7.4, the effect of adding a redundant variable $X_j'$ on the importance of the other variables $X_l$ (for $l \neq j$) is twofold. The first part of Equation 7.16 shows that the weight of all terms that do not include $X_j$ (or its copy) strictly decreases by a factor $\frac{p-k}{p+1}$. The second part of Equation 7.16 shows that the the weight of all terms that include $X_j$ (or its copy) increases, since several equivalent conditioning sets (i.e., $B \cup \{X_j\}$, $B \cup \{X_j\}'$ and $B \cup \{X_j, X_j'\}$) can now appear within the branches of a tree. Like for Proposition 7.2, impurity terms are not all modified in the same way and changes depend on the size $k$ of the conditioning set $B$. Overall, the net effect on the total importance of $X_l$ is therefore a compromise between these opposing changes. If the weighted sum of the $I(X_l; Y|B, X_j)$ terms is small with respect to the sum of the terms that do not include $X_j$ (or its copy), then the decrease effect dominates and the importance of $X_l$ should be smaller. By contrast, if the $I(X_l; Y|B, X_j)$ terms are larger, then the increase effect dominates and the resulting importance is larger. As shown later in Figure 7.2, redundant variables therefore increase the importance of all variables that *interact* with the duplicated variable.

Propositions 7.2 and 7.4 can be extended to the case where $N_c$ redundant variables $X_j^c$ (for $c = 1, \dots, N_c$) are added to the input variables instead of one, leading to the general Proposition 7.5. From this result, the same qualitative conclusions can be drawn, except that the decrease or increase effects discussed above are even stronger as more redundant variables are added.

**Proposition 7.5.** *Let $X_j \in V$ be a relevant variable with respect to $Y$ and $V$ and let $X_j^c \notin V$ (for $c = 1, \dots, N_c$) be $N_c$ totally redundant variables with respect to $X_j$. The infinite sample size importances of $X_j$ and $X_l \in V$* *as computed with an infinite ensemble of fully developed totally randomized trees built on* $V \cup \{X_j^1, \dots, X_j^{N_c}\}$ *are*

$$Imp(X_j) = \sum_{k=0}^{p-1} \left[ \frac{C_p^k (p-k)}{C_{p+N_c}^k (p+N_c-k)} \right] \frac{1}{C_p^k} \frac{1}{p-k} \sum_{B \in \mathcal{P}_k(V^{-j})} I(X_j; Y|B),$$

$$Imp(X_l) = \sum_{k=0}^{p-2} \left[ \frac{C_p^k (p-k)}{C_{p+N_c}^k (p+N_c-k)} \right] \frac{1}{C_p^k} \frac{1}{p-k} \sum_{B \in \mathcal{P}_k(V^{-l} \setminus X_j)} I(X_l; Y|B) +$$

$$\hookrightarrow \sum_{k=0}^{p-2} \left[ \sum_{k'=1}^{N_c+1} \frac{C_{N_c+1}^{k'}}{C_{p+N_c}^{k+k'}} \frac{1}{p+N_c-(k+k')} \right] \sum_{B \in \mathcal{P}_k(V^{-l} \setminus X_j)} I(X_l; Y|B \cup X_j).$$

*Proof.* Omitted here, but Proposition 7.5 can be proved by generalizing for $N_c$ the proofs of Propositions 7.2 and 7.4. □

Finally, let us note that Propositions 7.2, 7.4 and 7.5 are in fact valid as soon as variables $X_i$ and $X_j$ satisfy conditions of Lemma 7.1, even if they are not totally redundant. Accordingly, we call two variables satisfying these conditions *totally redundant with respect to the output* $Y$.

As an illustrative example, let us reconsider the LED classification problem from Section 6.5, for which $X_5$ was found to be the most important variable. As shown in Figure 7.1, adding variables that are redundant with $X_5$ makes its importance decrease, as predicted by our theoretical result from Propositions 7.2 and 7.5. When 5 or more copies of $X_5$ are added, the importance of $X_5$ is the smallest of all, as if $X_5$ had become less informative. Similarly, we observe that the importance of the other variables remains about the same or slightly decreases, as if they all had become a bit less informative. With regards to our previous results in Propositions 7.4 and 7.5, this indicates that the importance due to the $I(X_l; Y|B)$ terms prevails from the importance due to the $I(X_l; Y|B, X_5^c)$ terms. As a matter of fact, this example highlights a fundamental property of variable importances as computed in a random forest: *importance measures are computed not only with respect to the output* $Y$*, but also with respect to all the other input variables that define the problem.* In particular, a variable which is not important is not necessarily uninformative, as the example illustrates. A variable may be considered as less important because the information it conveys is also redundantly conveyed and diluted in other variables, and not necessarily because it has no information about the output.

As a second example, Figure 7.2 illustrates redundancy effects for a XOR classification problem defined on two variables $X_1$ and $X_2$. Again, the importance of the duplicated variable $X_1$ decreases as redundant variables are added, which confirms our results from Propositions 7.2 and 7.5. More interestingly, we now observe that the importance of the other variable, $X_2$, increases as copies of $X_1$ are added. For this problem, the $I(X_2; Y|B, X_1^c)$ terms are prevalent with respect to the $I(X_2; Y|B)$ terms (which is in fact unique and equal to 0), thereby artificially increasing the overall importance of $X_2$ as redundancy augments, as expected from Propositions 7.4 and 7.5.

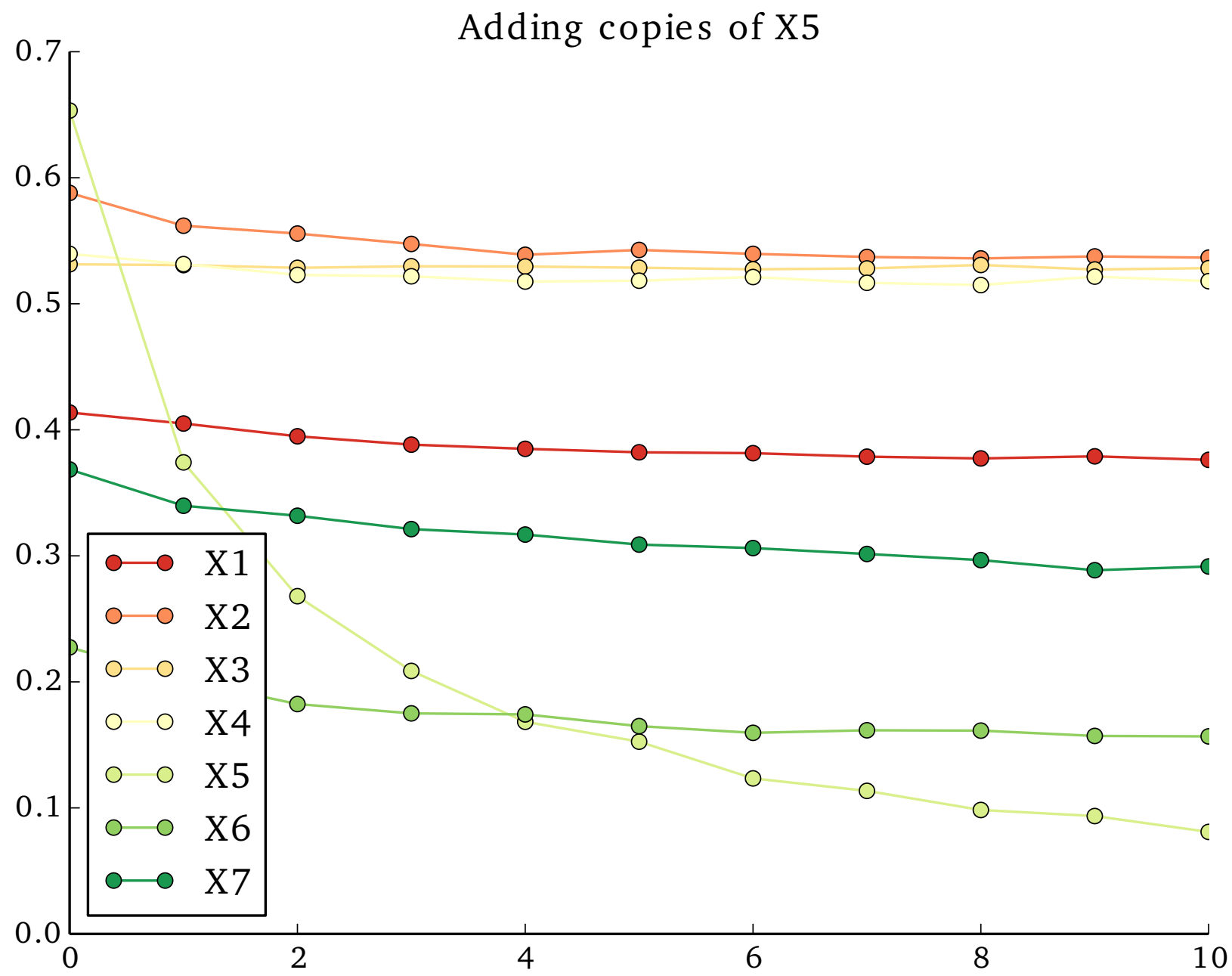

Figure 7.1: Adding copies of $X_5$ on the LED classification task. The more redundant variables are added, the lesser the importance of $X_5$.

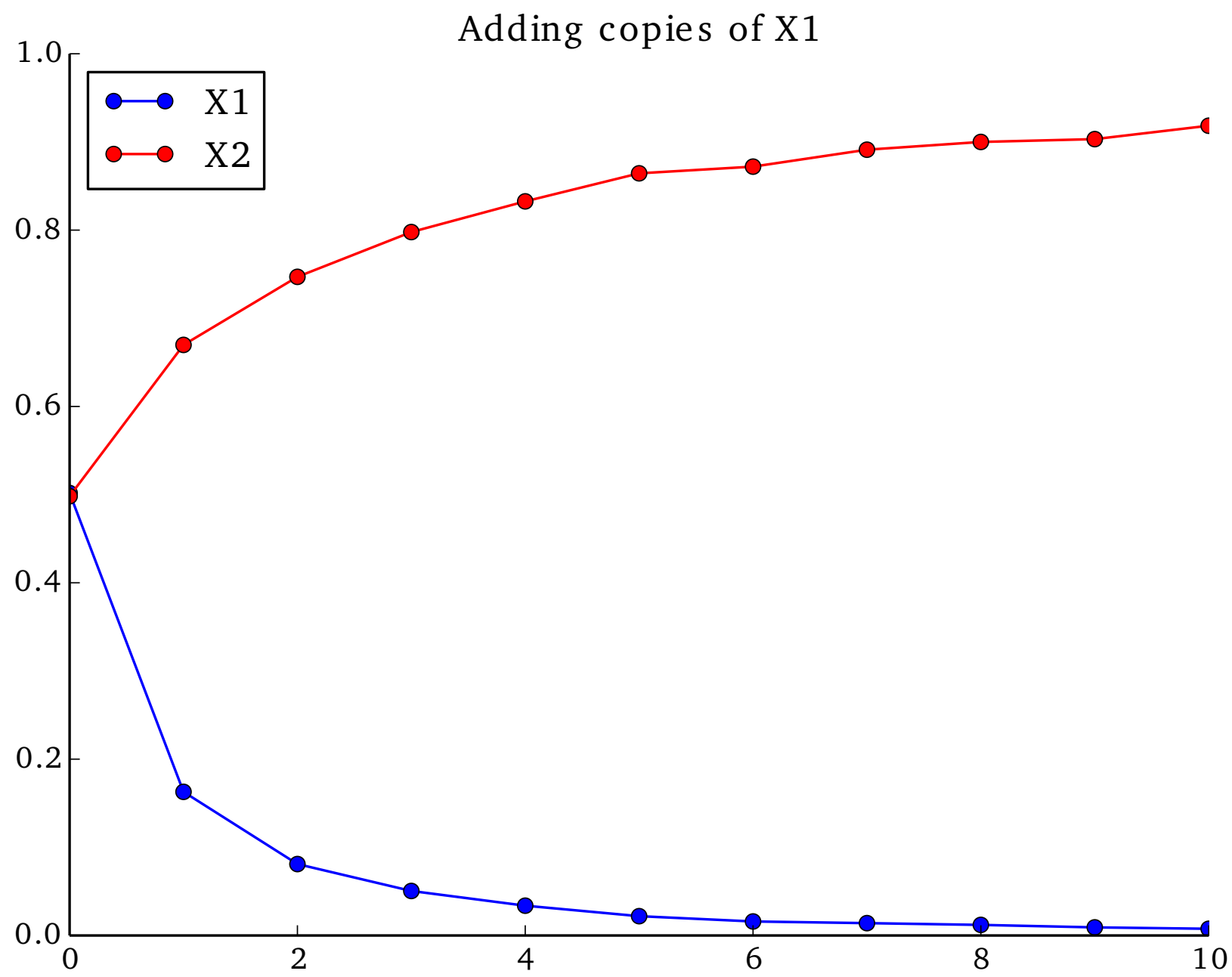

Figure 7.2: Adding copies of $X_1$ on a XOR classification task. The more redundant variables are added, the lesser the importance of $X_1$, but the larger the importance of $X_2$.

Overall, results presented in this section call for caution when interpreting variable importance scores. Due to redundancy effects – either total, as studied here, or partial as it would often arise in practice – it may happen that the total importance of a given variable is either misleadingly low or deceptively high because the same information is spread within several redundant variables and therefore taken into account several times within the total importances. As such, we advise to complement the interpretation with a systematic decomposition of variable importance scores, e.g., as previously done in Figure 6.3, in order to better understand why a variable is in fact important and to possibly detect redundancy.

## 7.2 BIAS IN VARIABLE IMPORTANCES

In this section, we study sources of bias in variable importances and show that variable selection (as previously discussed in Section 6.4.3) is not the only cause of bias. In practice, complementary forces due to masking effects, impurity misestimations and the structure of the trees make variable importances deviate from the theoretical results found in asymptotic conditions for totally randomized trees.

### 7.2.1 *Bias due to masking effects*

As shown in the previous chapters, the guided selection of the split variable (i.e., for $K > 1$) is necessary for balancing bias and variance in randomized decision trees and to produce accurate ensembles. In particular, we studied that decision trees built with too much randomization usually lead to an increase of bias (with respect to the generalization error) which cannot be compensated by a reciprocal decrease of variance, making it necessary to adjust the value of $K$ to find the appropriate trade-off. By contrast, we also showed in Section 6.4.3 that when variable selection is not totally random (i.e., as soon as $K > 1$), masking effects induce a bias with respect to variable importances, since it forces some of the branches to never be built, and therefore some of the conditioning sets $B \in \mathcal{P}(V)$ to never be taken into account. As a consequence, random forests whose parameter $K$ has been tuned to maximize accuracy may yield variable importances that are biased (either over- or underestimated). More specifically, it can be shown (see Section 6.4.3) that a relevant variable may be null with regards to its importance, thereby making it indistinguishable from irrelevant variables, and that the importance of relevant variables becomes dependent on the number of irrelevant variables.

| | | |
|---|---|---|
| | $X_1$ | $\sim \mathcal{N}(0,1)$ |
| | $X_2$ | $\sim \mathcal{M}(2)$ |
| | $X_3$ | $\sim \mathcal{M}(4)$ |
| | $X_4$ | $\sim \mathcal{M}(10)$ |
| | $X_5$ | $\sim \mathcal{M}(20)$ |
| *null case* | $Y$ | $\sim \mathcal{B}(0.5)$ |
| *power case* | $Y \vert X_2 = 0$ | $\sim \mathcal{B}(0.5 - \text{relevance})$ |
| | $Y \vert X_2 = 1$ | $\sim \mathcal{B}(0.5 + \text{relevance})$ |

Table 7.1: Input variables are independent random variables as defined from the table. $\mathcal{N}(0,1)$ is the standard normal distribution. $\mathcal{M}(k)$ is the multinomial distribution with values in $\{0, \ldots, k-1\}$ and equal probabilities. $\mathcal{B}(p)$ is the binomial distribution. In the null case, $Y$ is independent from $X_1, \ldots, X_5$. In the power case, $Y$ depends on the value $X_2$ while other input variables remain irrelevant.

### 7.2.2 *Bias due to empirical impurity estimations*

The analysis of variable importances carried out so far has considered asymptotic conditions for which the true node impurity $i(t)$ is assumed to be known. In practice however, due to the finite size of the learning set, impurity measurements suffer from an empirical misestimation bias. In this section, we study this effect in the context of heterogeneous variables[1], with respect to their scale or their number of categories, and show that the misestimation of node impurity is directly proportional to the cardinality of the split variable and inversely proportional to the number $N_t$ of samples used for the evaluation. As a result, impurity reductions become overestimated as we go deeper in the tree and/or as the number of values of the variable is large. In consequence, variable importances also suffer from bias, making variables of higher cardinality wrongly appear as more important.

To guide our discussion, let us revisit the simulation studies from [Strobl et al., 2007b] which consider a binary output variable $Y$ and five input variables $X_1, \ldots, X_5$, as defined in Table 7.1.

Let us first analyze the case where $Y$ is independent from $X_1, \ldots, X_5$, i.e., such that none of the input variables are informative with respect to the output. For a randomly sampled dataset $\mathcal{L}$ of $N = 120$ samples, Figure 7.3 plots variable importances for different kinds of random forests. TRT corresponds to totally randomized trees, as defined in Section 6.2, RF corresponds to standard Random Forest with bootstrap sampling and ETs corresponds to Extremely Randomized Trees.

1 As an example, in the case of meteorological problems, variables often comprise mixed environmental measurements of different nature and scale, like speed of wind, temperature, humidity, pressure, rainfall or solar radiation.

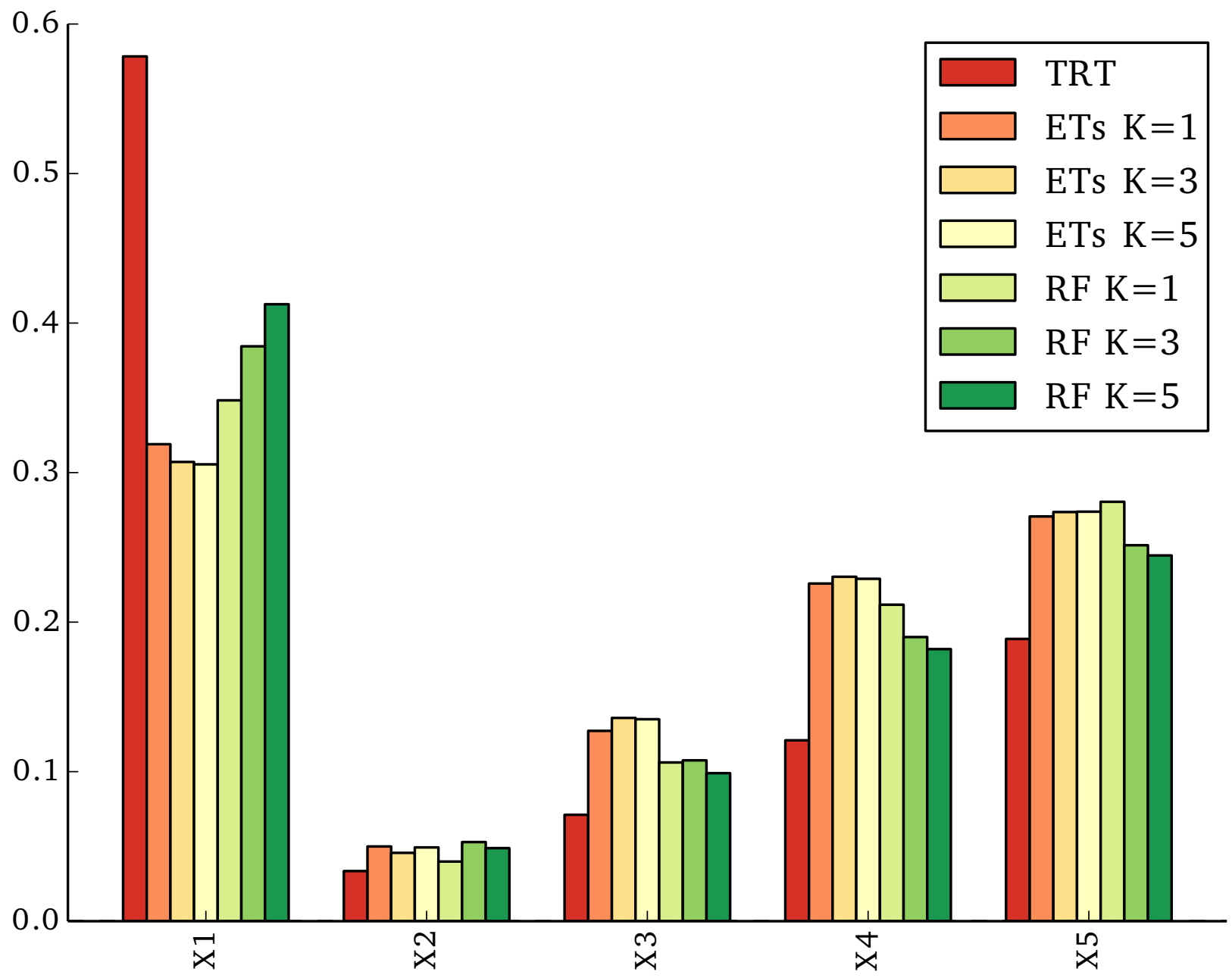

Figure 7.3: Variable importances for Y independent of $X_1, \dots, X_5$. ($N = 120$, $M = 500$)

Both RF and ETs use binary splits while TRT rely on multiway exhaustive splits[2]. In asymptotic conditions, we proved in Theorem 6.3 that the importance of irrelevant variables is strictly equal to 0. For a finite value of N however, this result does not hold, as Figure 7.3 indeed confirms. In contrast with what would be expected, we observe that the importance of none of the variables is in fact nowhere close to 0. In light of Theorem 6.2 however, this result is not that surprising: as long as decision trees can be fully developed, the sum of variable importances is equal to the (empirically estimated) mutual information between $X_1, \dots, X_p$ and the output Y, which is itself upper bounded by the (empirically estimated) entropy $H(Y)$ of the output variable. In this case, $H(Y) = \log_2(2) = 1$, which indeed corresponds to the sum of variable importances for all methods compared in the figure. More importantly, we also observe that the larger the cardinality of the variable (i.e., the larger its number of unique values), the larger its importance. For example, the importance of $X_1$ (for which samples all have unique values) or of $X_5$ (which counts up to 40 unique values) appears nearly 3 times larger as the importance of $X_2$ (which is binary).

In their study, Strobl et al. [2007b] argue that this bias is due to variable selection: *variables with more potential cut-points are more likely to produce a good criterion value by chance, as in a multiple testing situation*. As a result, variables of higher cardinality are more likely to be chosen for splitting than those of lower cardinality. While this bias has

2 Thereby creating as many branches as the number of values of the split variable, even if this variable is continuous and count unique values only.

been known for long in decision trees [Kononenko, 1995; Kim and Loh, 2001; Hothorn et al., 2006], we argue that it is not the primary cause for the bias observed here. Indeed, this argument does not directly explain why similar trends are also observed when no variable selection is performed (e.g., for TRT or for $K = 1$) nor why it similarly happens when cut-points are chosen at random, independently of the cardinality of the variable (e.g., in ETs).

For multiway splits like in TRT, bias in variable importances can be traced to the misestimations of the mutual information terms

$$\Delta i(s,t) \approx I(X_j; Y|t) \tag{7.17}$$

due to the finite size $N_t$ of the node samples. As shown in [Goebel et al., 2005], when $X_j$ and $Y$ are independent random variables (i.e., when $X_j$ is irrelevant), the distribution of finite sample size estimates of their mutual information follows approximately a gamma distribution

$$\widehat{I}(X_j; Y) \sim \Gamma\Big(\frac{1}{2}(|\mathcal{X}_j| - 1)(|\mathcal{Y}| - 1), \frac{1}{N_t \log 2}\Big) \tag{7.18}$$

whose mean is linearly proportional to the cardinalities $|\mathcal{X}_j|$ and $|\mathcal{Y}|$ of $X_j$ and $Y$ and inversely proportional to $N_t$, that is

$$\mathbb{E}\{\widehat{I}(X_j; Y)\} = \frac{(|\mathcal{X}_j| - 1)(|\mathcal{Y}| - 1)}{2N_t \log 2}. \tag{7.19}$$

As a result, estimates get larger as $X_j$ counts many unique values, and become even larger as nodes are deep in the tree (since $N_t$ gets smaller). Consequently, the weighted mean of all such estimated impurity terms $I(X_j; Y|t)$, for all nodes $t$ where $X_j$ is the split variable, and resulting in the total importance $\text{Imp}(X_j)$, is also linearly dependent on the cardinality of $X_j$. For TRT, this result explains why variables of high cardinality in Figure 7.3 appear as more important than those of lower cardinality. Intuitively, the closer the number of unique values with respect to the total number of samples, the larger the impurity decrease when splitting exhaustively on this variable. In the extreme case, when values for $X_j$ are all unique, splitting on the variable perfectly memorizes the values of $Y$, resulting in child nodes that are all pure, therefore maximizing the estimated mutual information. As such, this explains why $X_1$, whose values are all unique, appears as the most important variable.

For binary splits (i.e., for RF and ETs) the mutual information $I(X_j; Y|t)$ is not directly estimated at each node. Rather, in the case of ordered variables, $\Delta i(s,t)$ corresponds to an estimate of the mutual information $I(X_j \leqslant v; Y|t)$ of the binary split $X_j \leqslant v$. Under the simplifying assumption that binary splits on the same variable are all directly consecutive in the decision tree, it is easy to see that binary trees are equivalent to multiway trees [Knuth, 1968], as illustrated in Figure 7.4.

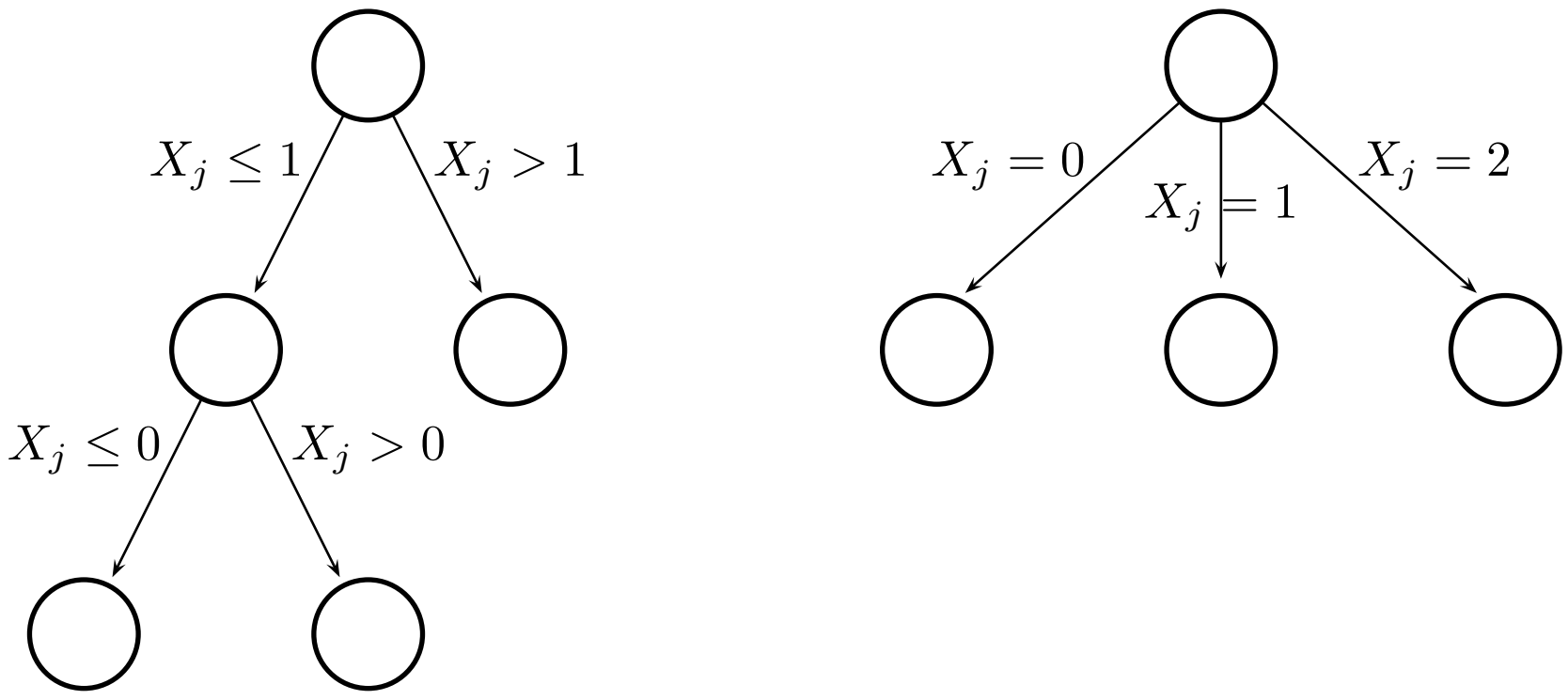

Figure 7.4: Consecutive binary splits on the same variable are equivalent to direct multiway splits.

Using an argument similar to the proof of Theorem 6.2, intermediate impurity terms between the first split and the last of those splits cancel each other when summing up the importances, which finally amounts to collect an actual estimate of $I(X_j; Y|t)$ from the sequence of binary splits. For the same reasons as before, variables of high cardinality are therefore biased in the same way. (As we will study in Section 7.2.3, this consecutiveness assumption does not hold in practice, making the importances from binary decision trees strictly different from those of multiway decision trees. Yet, the qualitative conclusions are still valid since variables of high cardinality can be reused far more often than variables of low cardinality, before all possible splits are exhausted.)

In both situations, the origin of the problem stems from the fact that node impurity is misestimated when the size $N_t$ of the node samples is too small. To a large extent, the issue is aggravated by the fact that trees are fully developed by default, making impurity terms collected near the leaves usually unreliable. As a precaution, a safe and effective solution for the bias problem therefore simply consists in collecting impurity terms only for those nodes where the impurity estimates can be considered as reliable. Equivalently, the construction of the tree can also be stopped early, when impurity estimates become unreliable, e.g., by limiting the depth of the tree, controlling for the minimum number of samples in internal nodes or using any of the other stopping criteria defined in Section 3.5. Among all alternatives, conditional inference trees [Hothorn et al., 2006] and earlier methods [Quinlan, 1986; Wehenkel, 1998] make use of statistical tests for assessing the independence of $X_j$ and $Y$ at a pre-specified level of confidence $\alpha$. If the null hypothesis cannot be rejected, then recursive partitioning halts. In particular, variable importances collected from conditional inference trees were shown experimentally by Strobl et al. [2007b] not to suffer from bias. The authors argue that it is due to the unbiased variable selection mechanisms also implemented in conditional inference trees. By contrast, we argue that the absence of bias in importances from such trees is mostly due to the early stopping criterion, and not to variable selection. Although variable selection plays an important and exacerbating role, it is not the true cause of the observed bias. Indeed, since the impurity reduction of variables of high cardinality is overestimated, searching for a split among $K > 1$ randomly drawn variables increases their probability of being selected when compared to others of lower cardinality, therefore masking and reducing the importances of these latter variables. Yet, bias stems from the fact that impurity reductions are misestimated in the first place.

As an illustrative example, Figure 7.5 reconsiders the simulation study of Table 7.1, when varying both the maximum depth of the trees (from `max_depth` $= 1$ to $9$) and the relevance of $X_2$ with respect to $Y$. Let us first consider ETs built with no variable selection ($K = 1$), as shown in four top plots of the figure. For the null case, when relevance $= 0$, limiting the depth of the trees correctly fixes the bias that was observed before. For shallow trees, the importances before normalization of all five input variables are close to $0$, as expected from irrelevant variables. The normalized importances, as shown in the figure, are also all close to $\frac{1}{p} = 0.2$, confirming that no variable is detected as more relevant than the others. However, when the depth of the decision trees increases, importances deviate and bias proportional to the cardinality of the variables appears as discussed before. When $X_2$ is a relevant variable (i.e., for relevance $> 0$), its importance is expected to be strictly positive and at least as large as the importances of the irrelevant variables. For relevance $= 0.1$ and `max_depth=1`, the importance of $X_2$ appears nearly 6 times larger than the importances of the other variables, confirming that $X_2$ is correctly identified as a relevant variable. For deeper trees however, noise dominates and the importance of the irrelevant variables is larger due to misestimations of the impurity terms. As relevance increases, $X_2$ can more clearly be identified as a relevant variable. In particular, the more relevant $X_2$, that is the stronger the signal, the deeper trees can be built until $X_2$ is made unrecognizable from the irrelevant variables.

By comparison, the four bottom plots of Figure 7.5 illustrate variable importances for RF built with variable selection ($K = 5$). Again, we observe that limiting the depth of the trees helps reduce the misestimation bias. However, the supplementary effect due variable selection is also clearly visible: variables of larger cardinality appear as significantly more important than the other variables. Consequently, this makes the detection of $X_2$ as a relevant variable more difficult when trees are grown deeper. When comparing the relative importance of $X_2$ for a given depth and relevance, we indeed observe that $X_2$ consistently appears as less important in RF than in ETs. It is only for very shallow trees (`max_depth=1` or 2) and high relevance that $X_2$ is identified with higher confidence than in ETs.

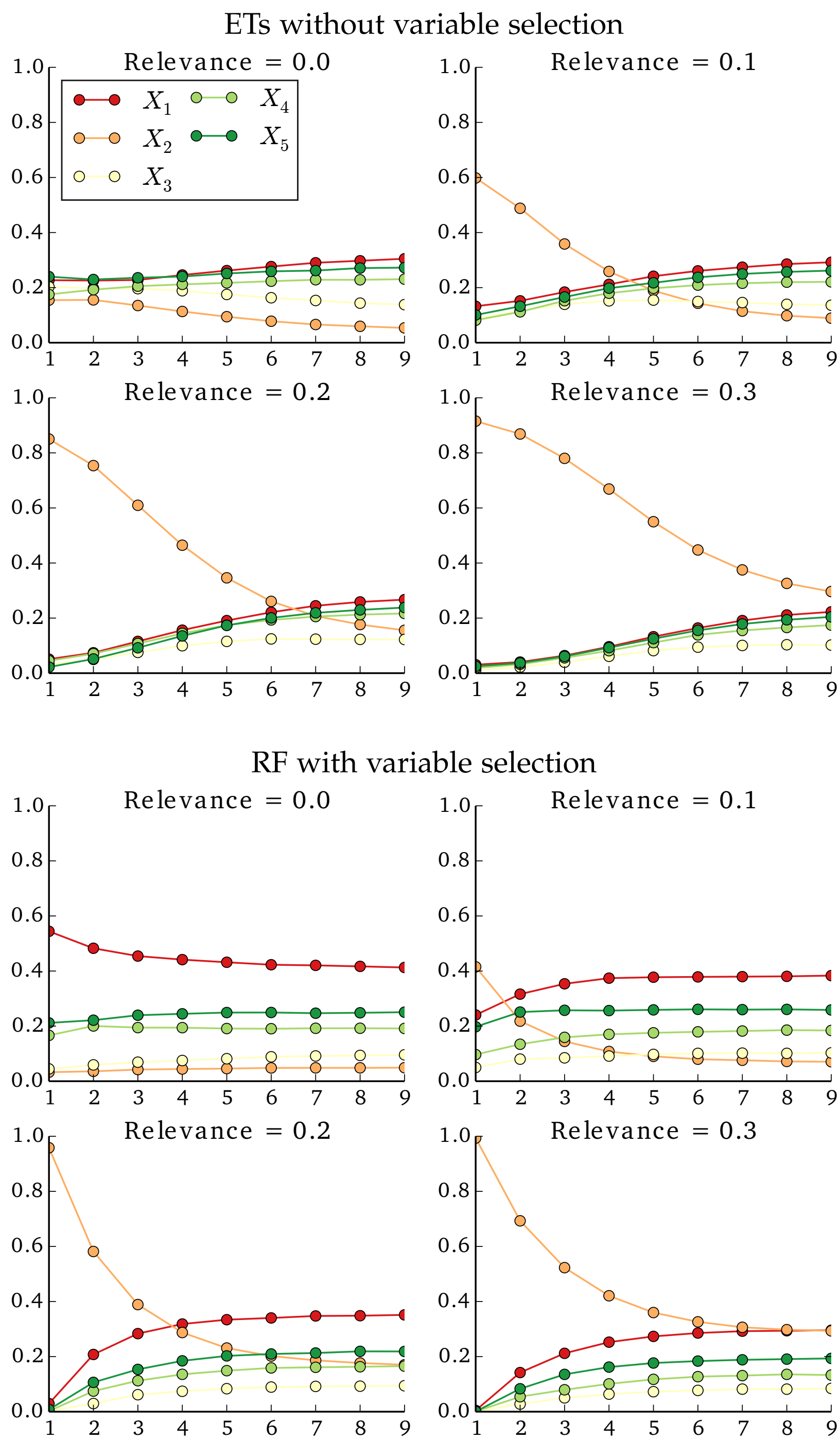

Figure 7.5: Variable importances of $X_1, \ldots, X_5$ when varying both the maximum depth of the trees and the degree of relevance of $X_2$. Importance scores are normalized by the sum of importances. (Top) ETs with no variable selection, $K = 1$, $N = 120$, $M = 500$. (Bottom) Random Forest with variable selection, $K = 5$, $N = 120$, $M = 500$.

In conclusion, evaluating impurity on small samples lead to over-estimations of the mutual information, resulting in biased variable importances. In particular, the higher the cardinality of the variable, the larger the misestimations. To minimize this effect, caution should be taken by only considering impurity terms that were computed from a large enough sample. This can be guaranteed, e.g., by stopping the construction of the tree early or making leaves grow more slowly than the size $N$ of the learning set. Additionally, we have also shown that variable selection may increase the bias due to over-estimations. In this case, a simple remedy consists in not using variable selection when assessing the relevance of variables. Finally, let us also note the connection with Propositions 6.6 and 6.7: as long as trees are built to a maximum depth which is larger than the number $r$ of relevant variables, early stopping the construction of the trees does not prevent us from detecting the relevant variables.

### 7.2.3 *Bias due to binary trees and threshold selection*

Previous developments from Chapter 6 studied variable importances for fully developed totally randomized trees and multiway exhaustive splits. In practice however, random forests usually rely on binary splits rather than on multiway splits. In terms of impurity, this results in additional and distinct information terms that were not previously accounted for, because i) a same variable can be reused several times along the same branch and ii) binary splits discretize the information contained in a variable, making variable importances dependent on the split threshold selection strategy. In lack of a rigorous theoretical framework, we give in this section preliminary insights on variable importances for binary decision trees, hopefully shedding some light on their interpretation.

As a simplistic but illustrative example, let us consider a toy classification problem composed of a ternary input variable $X_1$ and of a binary input variable $X_2$, both of them being ordered. Let us further assume that input samples are uniformly drawn from Table 7.2, which defines the output as $Y = X_1 < 1$ and $X_2$ as a copy of $Y$. With respect to $Y$, both variables are as informative and one would expect their importances to be the same. With totally randomized trees and exhaustive splits, two equiprobable decision trees can be built, as represented in Figure 7.6. In both of them, splitting either on $X_1$ or on $X_2$ at the root node results in child nodes that are pure, hence halting the construction process. As expected, the measured variable importances are the same:

$$\text{Imp}(X_1) = \frac{1}{2} I(X_1; Y) = \frac{1}{2} H(Y) = 0.459,$$

$$\text{Imp}(X_2) = \frac{1}{2} I(X_2; Y) = \frac{1}{2} H(Y) = 0.459.$$

| y | $x_1$ | $x_2$ |
|---|---|---|
| 0 | 0 | 0 |
| 1 | 1 | 1 |
| 1 | 2 | 1 |

Table 7.2: Toy problem. All 3 possible samples are equiprobable.

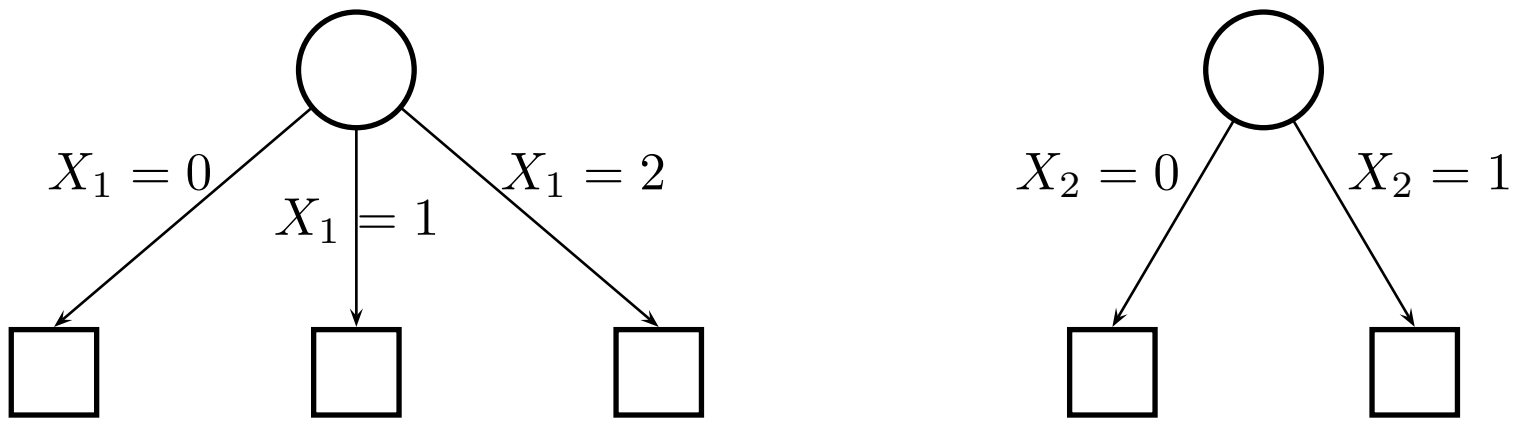

Figure 7.6: Totally randomized trees built from Table 7.2. Both decision trees are equiprobable. The resulting variable importances indicate that both $X_1$ and $X_2$ are as important.

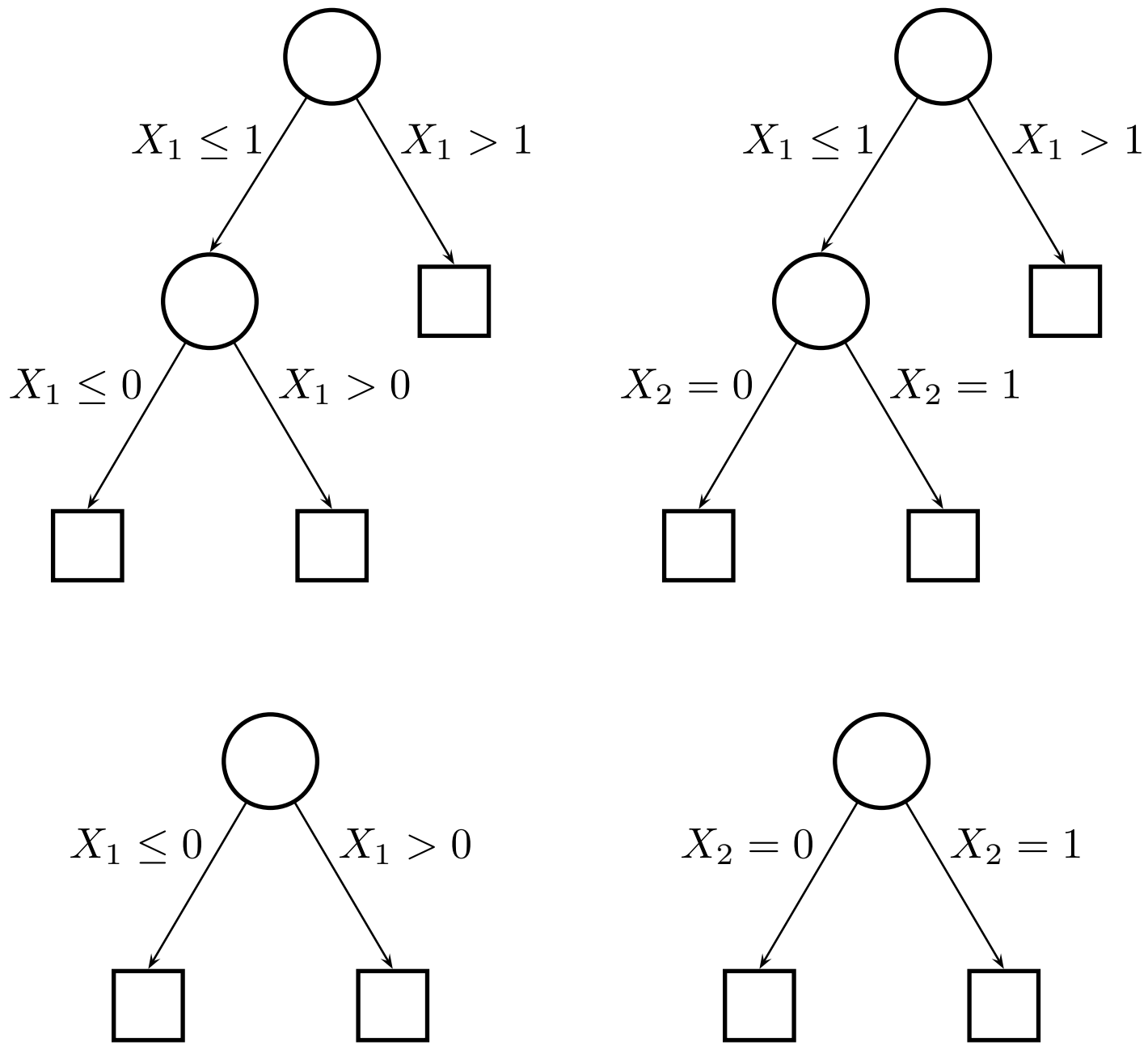

Figure 7.7: Extremely randomized trees ($K = 1$) built from Table 7.2. From left to right, top to bottom, decision trees are respectively generated with probability $\frac{1}{8}$, $\frac{1}{8}$, $\frac{1}{4}$ and $\frac{1}{2}$. The resulting variable importances indicate that $X_2$ is now more important than $X_1$.

By contrast, using binary splits and ETs (for $K = 1$) results in 4 different decision trees, as represented in Figure 7.7. When splitting on $X_1$ at the root node, two binary splits are now possible with equal probability: $t_L = X_1 \leqslant 0, t_R = X_1 > 0$ or $t_L = X_1 \leqslant 1, t_R = X_1 > 1$. In the former case, the resulting child nodes are pure, hence halting the construction process. In the latter case, the right child (corresponding to $X_1 > 1$) is pure, while the left child is not. For this node, recursive partitioning proceeds and a second binary split can be made either on $X_1$ or on $X_2$. Overall, the measured variable importances in these binary trees, in asymptotic conditions, are

$$\begin{aligned}
\text{Imp}(X_1) &= \frac{2}{8} I(X_1 \leqslant 1; Y) + \frac{1}{8} P(X_1 \leqslant 1) I(X_1 \leqslant 0; Y | X_1 \leqslant 1) + \frac{1}{4} I(X_1 \leqslant 0; Y) \\
&= 0.375 \\
\text{Imp}(X_2) &= \frac{1}{2} I(X_2; Y) + \frac{1}{8} P(X_1 \leqslant 1) I(X_2; Y | X_1 \leqslant 1) \\
&= 0.541,
\end{aligned}$$

which makes them strictly different from the importances collected from multiway totally randomized trees. In particular, due to the binarization of the split variables, importances now account for conditioning sets that may include several values of a same variable. For instance, the importance of $X_2$ includes $I(X_2; Y | X_1 \leqslant 1)$, which measures the mutual information between $X_2$ and $Y$ when $X_1$ is either equal to 0 or 1. With multiway exhaustive splits, these conditioning sets are not considered because branches correspond to single values only. Similarly, importances also account for binarized mutual information terms such as $I(X_1 \leqslant 1; Y)$. Again, these are not evaluated in totally randomized trees because of the multiway splits.

Accordingly, threshold selection in binary splits has a dominant effect on variable importances since it controls how the original variables are binarized. In ETs, all intermediate thresholds $v$ are possible, resulting in a combinatorial number of additional impurity terms $I(X_j \leqslant v; Y|\cdot)$ and $I(\cdot; Y|X_j \leqslant v, \cdot)$. In RF, only the best threshold is selected locally at each node, resulting in fewer additional impurity terms $I(X_j < v^*; Y|\cdot)$ and $I(\cdot; Y|X_j < v^*, \cdot)$ (and masking all others).

As a last example, Figure 7.8 illustrates variable importances for $X_1$ and $X_2$ when increasing the cardinality $L = |\mathcal{X}_1|$ of $X_1$ on the previous toy problem. That is, let us redefine the output as $Y = X_1 < \frac{L}{2}$, for $\mathcal{X}_1 = \{0, \dots, L-1\}$, while keeping $X_2$ as a binary variable defined as a copy of $Y$. Assuming that all input samples are equiprobable, the importances yielded by totally randomized trees with multiway splits remain the same as before. For ETs however, increasing $L$ induces even more new impurity terms that are now accounted for in the importances. As the figure shows, increasing the cardinality of $X_1$ makes its importance decrease, but also simultaneously makes the importance of $X_2$ increase. Indeed, splitting on $X_2$ always yield child nodes that are pure, which is unlikely to be the case when splitting randomly on $X_1$. For RF, only the best thresholds are selected, yielding in this case child nodes that are always pure, as for totally randomized trees. As such, their importances respectively appear as equal as the figure confirms.

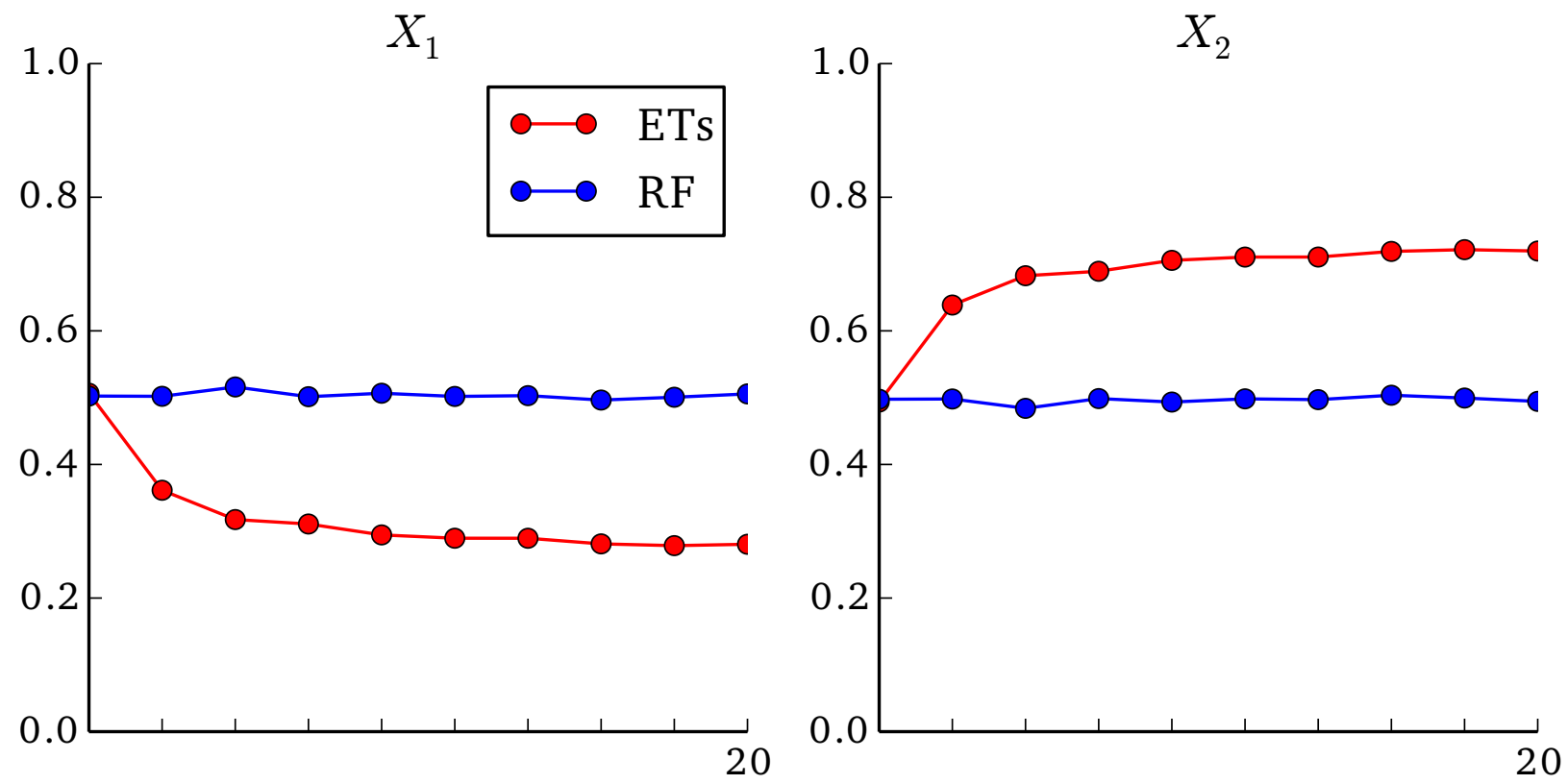

Figure 7.8: Variable importances of $X_1$ and $X_2$ when increasing the cardinality of $X_1$.

Overall, these additional effects due to binary splits make variable importances computed from classical random forests very difficult to interpret and understand, as soon as data include many variables of different number of categories. While they can still be used to identify the most relevant variables, caution should be taken when interpreting the amplitude of importances. As our last example illustrates, they may be misleadingly low or high because of combinatorial effects, solely due to the possible ways variables are binarized through the implemented threshold selection mechanisms.

### Encoding L-ary variables into binary variables

Partitioning node samples with binary splits on L-ary input variables is equivalent to individually transforming each input variable $X_j$ into a set of $L_j - 1$ binary input variables $\{X_j^l | l = 1, \dots, L_j - 1\}$, each encoding for one the possible splits, and then partitioning node samples using of one these new variables. If we consider the binarized learning set $\mathcal{L}'$, for which all p the input variables have been transformed into $\sum_j (L_j - 1)$ binary variables, then our theoretical framework for totally randomized trees and exhaustive splits can be applied and Theorem 6.1 could possibly be adapted. The only difference lies in the way binary variables are drawn at random: instead of splitting randomly on one of the $\sum_j (L_j - 1)$ variables, first, one of the p original variables is drawn uniformly at random; second, one of its $L_j - 1$ binary variables is selected for splitting the current node. In this setting, the importance of $X_j$ therefore amounts to the sum of importances $\sum_l \text{Imp}(X_j^l)$ of its binary variables.

## 7.3 APPLICATIONS

In spite of the various concurrent effects discussed earlier, variable importance measures have been used with success in a wide range of scientific applications. While they have proven to be a good proxy for assessing the relevance of input variables, providing helpful insights, too often variable importances are used as a black-box metric, hence under-exploiting the information they offer. As such, examples presented in this section all demonstrate that any progress towards a better theoretical understanding of variable importances may help to further advance a wide range of research domains.

### 7.3.1 *Feature selection*

Within the past 10 or 20 years, typical machine learning problems have grown from a few tens of input variables to domains exploring hundreds of thousands of variables, often comprising noisy, irrelevant or redundant predictors, mixing both numerical and categorical variables and involving complex interaction effects. In this context, the feature selection problem consists in identifying a subset of the original input variables that are useful for building a good model [Guyon and Elisseeff, 2003; Liu and Yu, 2005]. The advantages and benefits of reducing the dimensionality of the problem include: speeding up machine learning algorithms, reducing measurement and storage requirements, improving the accuracy of the models or facilitating data visualization and understanding.

Because of the properties of random forests (good prediction performance, robustness to noisy variables, support of numerical and categorical variables and ability to model complex interactions), variable importances often provide an effective solution for the feature selection problem. The most straightforward solution consists in ranking variables according to their importance scores and to only keep the most important ones. Depending on the objective, the best subset of variables can be identified in several ways:

- When the goal is simply to reduce dimensionality because of speed and storage requirements, the simplest solution is to keep only those variables whose importances $\text{Imp}(X_j)$ is greater than some manually defined threshold $\alpha$.

- If the goal is to improve accuracy, a good subset of variables can typically be found by tuning the threshold $\alpha$ so as to minimize some user-defined criterion (e.g., the zero-one loss in classification or the squared error loss in regression) for the model built on the subset $\{X_j | \text{Imp}(X_j) > \alpha\}$.

  At the price of more computational efforts, even better performance can usually be reached by embedding variable im-portances into a dedicated iterative feature selection procedure, such as those described in [Guyon et al., 2002; Tuv et al., 2009].

- In some other applications, the objective is to identify variables that are relevant, in order to better understand the underlying relations with the output $Y$. In asymptotic conditions, this could be done by discarding all variables whose importances is null, as shown by Theorem 6.5. In a finite setting however, bias due to masking effects or impurity misestimations (as previously discussed in Section 7.2) makes it more difficult to identify variables that are truly relevant since their importances might appear to be lower than those of irrelevant variables. Yet, several options are available for controlling and limiting false positives, such as stopping the construction process when impurity estimations become statistically unreliable (Section 7.2.2), comparing the importances of the original input variables to artificial contrasts [Tuv et al., 2006] or robustly controlling the conditional error rate through permutation tests [Huynh-Thu et al., 2012].

### 7.3.2 *Biomarker identification*

With the rise of *-omics* data, random forests have become one of the most popular tools in life sciences, providing practitioners with both high-prediction accuracy and helpful insights about the importances of variables. In many cases, variable importance measures (either MDI or the permutation importance) is exploited to better understand complex interaction effects between the inputs and the output. Examples of successful applications include the identification of disease-associated SNPs in genome-wide association studies [Lunetta et al., 2004; Meng et al., 2009; Botta et al., 2014], the discovery of important genes and pathways from micro-array gene-expression data [Pang et al., 2006; Chang et al., 2008] or the identification of factors for predicting protein-protein interactions [Qi et al., 2006]. These examples are not isolated and tens of further studies based on random forests could in fact be cited from the fields of genomics, metabolomics, proteomics or transcriptomics. Some of them are reviewed in [Geurts et al., 2009; Touw et al., 2013; Boulesteix et al., 2012].

In light of our study and discussion of variable importances, recommendations for biomarker identification depend on the exact objective of the application. If the goal is to identify all relevant variables, then totally randomized trees with multiway splits should be preferred. They indeed constitute the only method for which variable importances are unbiased, taking into account all possible interaction terms in a fair and exhaustive way. The only caveat is that a (very) large number of trees may be required before variable importances converge, making them computationally intensive to compute, even if the full randomization of the induction procedure actually make individual decision trees to be quite fast to generate. By contrast, if the objective is only to identify a subset of good predictors for predicting the output (hence possibly omitting relevant but redundant variables), then non-totally randomized trees (e.g., standard RF or ETs) with K chosen to maximize accuracy should be preferred. Even if some informative variables may be masked because of variable selection, a good subset of input variables should still be identifiable from importance scores. From a computational point of view, convergence should also be faster, making this approach more appealing when resources are limited. In all cases, impurity misestimations should be controlled by stopping the construction process early or collecting importances only for those nodes where impurity reductions can be considered as reliable. Finally, whenever practical, variable importances should also be decomposed in order to better understand why some variables appear as more important than others. In particular, studying the decomposition might help identify redundancy effects.

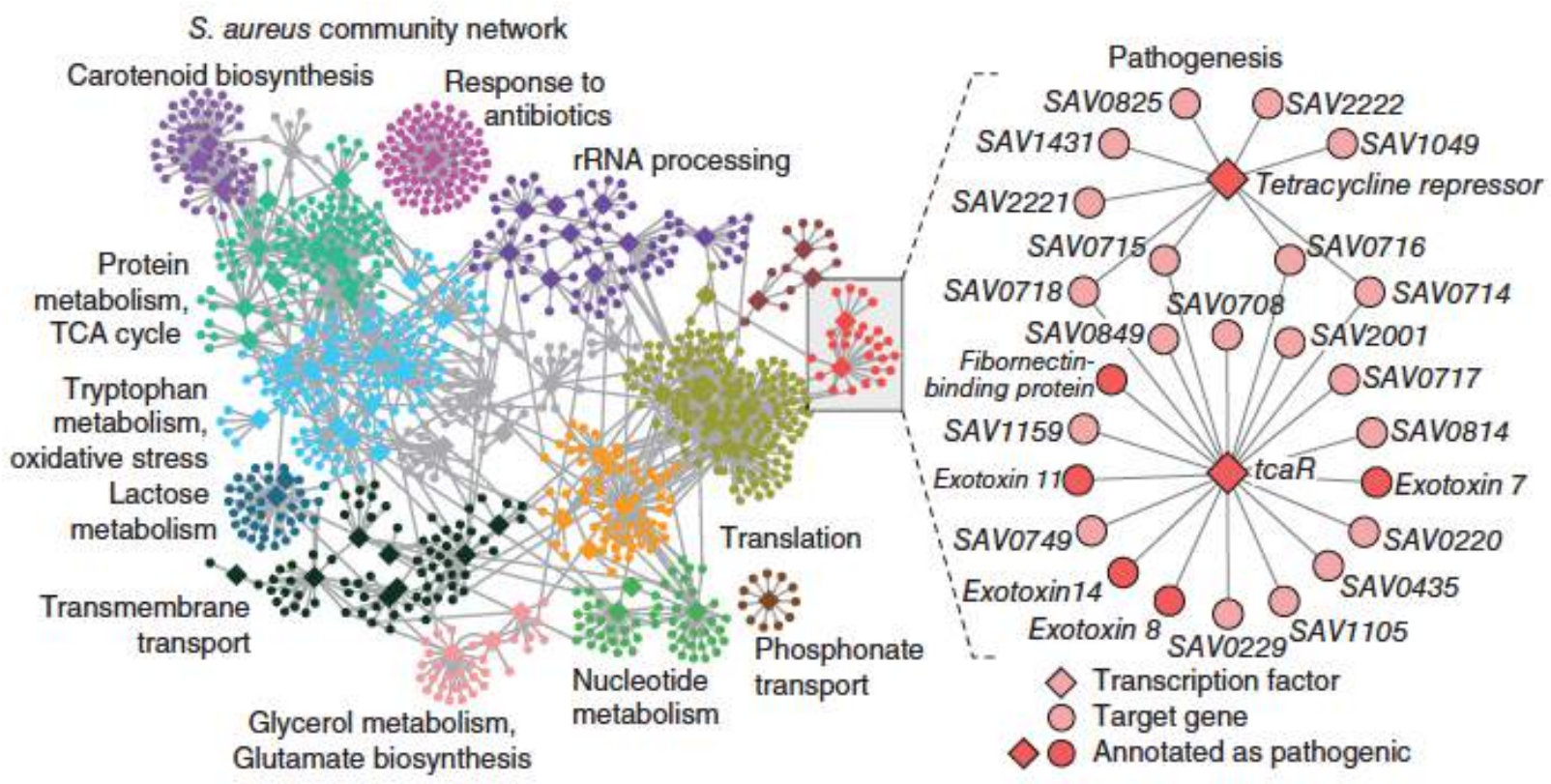

Figure 7.9: Gene regulatory network in Staphylococcus aureus. Image from [Marbach et al., 2012].

### 7.3.3 *Network inference*

Given a set of input variables $V = \{X_1, \dots, X_p\}$, the *network inference* problem consists in the identification of conditional dependencies between variables. In genomics for example, regulatory network inference consists in the identification of interactions between genes or transcription factors on the basis of their expression level, in order to reconstruct a global network of interactions (as illustrated in Figure 7.9 for Staphylococcus aureus).

As proposed in the GENIE3 method [Huynh-Thu et al., 2010], the network inference problem can be solved generically by remarking that it can decomposed into p independent supervised learning problems. Formally, for each input variable $X_j$ (for $j = 1, \dots, p$), GENIE3 considers the supervised sub-problem consisting in the prediction of the target variable $X_j$ from the remaining $p-1$ variables $V^{-j}$. Using random forests for solving each of these $p$ problems, variable importances can be derived and used as an indication of the (directed) putative link between the predictor variables $X_i$ (for $i \neq j$) and the target $X_j$. Intuitively, the larger the importance, the more likely the conditional dependency with $X_j$. Once all $p$ problems are solved, putative links are aggregated over all $p$ variables to provide a ranking of interactions from which the network can finally be reconstructed.

Again, our previous theoretical analysis calls for caution when plugging variable importances into network inference procedures. Indeed, since the goal is to retrieve all *direct* interactions, importances might in fact not be as appropriate as desired since they also account for indirect or combined effects. In this case, a good heuristic is to intentionally induce masking effects (i.e., by setting $K > 1$) in order to recover only the strongest (and assumingly direct) interactions. Alternatively, a promising strategy might be to directly look for *strongly* relevant variables, that is for variables $X_i$ such that $I(X_i; X_j | B) > 0$ for *all* $B$. In both cases, impurity misestimations effects should be mitigated by using an adequate stopping criterion. When input variables are of different scale of measurements or vary in the number of categories, care should finally be taken when aggregating variable importances into a single ranking of interactions. As shown by Theorem 6.2, variable importances are in this case upper bounded by the entropy $H(X_j)$ of the target variable, which may greatly varies from one target to another. As such, variable importances are not directly comparable and should preferably be normalized (e.g., by $H(X_j)$) before their aggregation.

# Part III

# SUBSAMPLING DATA

# 8 ENSEMBLES ON RANDOM PATCHES

**Outline**

In this chapter, we consider supervised learning under the assumption that the available memory is small compared to the size of the dataset. This general framework is relevant in the context of big data, distributed databases and embedded systems. In Section 8.1, we propose a very simple, yet effective, ensemble framework that builds each individual model of the ensemble from a random patch of data obtained by drawing random subsets of *both* samples and input variables from the whole dataset. In sections 8.2 and 8.3, we carry out an extensive and systematic evaluation of this method on 29 datasets, using decision trees as base models. With respect to popular ensemble methods, these experiments show that the proposed method provides on par performance in terms of accuracy while simultaneously lowering the memory needs, and attains significantly better performance when memory is severely constrained. We conclude and discuss future work directions in Section 8.4. *This chapter is based on previous work published in [Louppe and Geurts, 2012].*

Within the past few years, big data has become a popular trend among many scientific fields. In life sciences, computer vision, Internet search or finance, to cite a few, quantities of data have grown so large that it is increasingly difficult to process, analyze or visualize. In many cases, single computers are no longer fit for big data and distributed environments need to be considered to handle it. Although research is very active in this area, machine learning is no exception to this new paradigm. Much still needs to be done and methods and algorithms have to be reinvented to take this constraint into account.

In this context, we consider supervised learning problems for which the dataset is so large that it cannot be loaded into memory. Breiman [1999] proposed the Pasting method to tackle this problem by learning an ensemble of models individually built on random subsets of the training examples, hence alleviating the memory requirements since the base models would be built on only small parts of the whole dataset. Earlier, Ho [1998] proposed to learn an ensemble of models individually built on random subspaces, i.e., on random subsets of the input variables (or *features*). While the first motivation of the Random Subspace method was to increase the diversity within the models of the ensemble, it can actually also be seen as way to reduce the memory requirements of building individual models. In this work, we propose to combine and leverage both approaches at the same time: learn an ensemble of models on *random patches*, i.e., on random subsets of the samples *and* of the input variables. Through an extensive empirical study, we show that this approach (1) improves or preserves comparable accuracy with respect to other ensemble approaches which build base models on the whole dataset while (2) drastically lowering the memory requirements and hence allowing an equivalent reduction of the global computing time. In other words, this analysis shows that there is usually no need to build individual models on the whole dataset. For the same accuracy, models can be learned independently on small portions of the data, within significantly lower computational requirements.

## 8.1 RANDOM PATCHES

The Random Patches algorithm proposed in this work (further referred to as RP) is a wrapper ensemble method that can be described in the following terms. Let $\mathcal{R}(\alpha_s, \alpha_f, \mathcal{L})$ be the set of all random patches of size $\alpha_s N \times \alpha_f p$ than can be drawn from the dataset $\mathcal{L}$, where $N$ (resp., $p$) is the number of samples in $\mathcal{L}$ (resp., the number of input variables in $\mathcal{L}$) and where $\alpha_s \in [0, 1]$ (resp. $\alpha_f$) is an hyper-parameter that controls the number of samples in a patch (resp., the number of variables). That is, $\mathcal{R}(\alpha_s, \alpha_f, \mathcal{L})$ is the set of all possible subsets containing $\alpha_s N$ samples (among $N$) with $\alpha_f p$ variables (among $p$). The method then works as follows:

**Algorithm 8.1.** *Random Patches algorithm.*

1: **for** $m = 1, \dots, M$ **do**
2: Draw a patch $r \sim U(\mathcal{R}(\alpha_s, \alpha_f, \mathcal{L}))$ uniformly at random
3: Build a model on the selected patch $r$
4: **end for**
5: Aggregate the predictions of the $M$ models in an ensemble

While the RP algorithm can exploit any kind of base estimators, we consider in this work only tree-based estimators. In particular, we evaluate the RP algorithm using standard classification trees (as described in Chapter 3) and (single) extremely randomized trees [Geurts et al., 2006a]. Unless stated otherwise, trees are unpruned and grown using Gini index as impurity criterion for node splitting. The parameter $K$ of extremely randomized trees within RP is set to its maximum value $K = \alpha_f p$ (i.e., corresponding to no further random selection of variables).

The first benefit of RP is that it generalizes both the Pasting Rvotes (P) method [Breiman, 1999] (and its extensions [Chawla et al., 2004; Basilico et al., 2011]) and the Random Subspace (RS) algorithm [Ho, 1998]. Both are indeed merely particular cases of RP: setting $\alpha_s = 1.0$ yields RS while setting $\alpha_f = 1.0$ yields P. As such, it is expected that when both hyper-parameters $\alpha_s$ and $\alpha_f$ are tuned simultaneously, RP should be at least as good as the best of the two methods, provided there is no overfitting associated with this tuning.

When the base estimators are standard decision trees (resp., extremely randomized trees with $K = \alpha_f p$), interesting parallels can also be drawn between RP and the RF algorithm (resp., ET). For $\alpha_s = 1.0$, the value of $\alpha_f p$ is indeed nearly equivalent to the number $K$ of features randomly considered when splitting a node. A major difference remains though. In RP, the subset of features is selected globally once and for all, prior to the construction of each tree. By contrast, in RF (resp., in ET) subsets of features are drawn locally at each node. Clearly, the former approach already appears to be more attractive when dealing with large databases. Non-selected features indeed do not need to be considered at all, hence lowering the memory requirements for building a single tree. Another interesting parallel can be made when bootstrap samples are used like in RF: it nearly amounts to set $\alpha_s = 0.632$, i.e. the average proportion of unique samples in a bootstrap sample. Differences are that in a bootstrap sample, the number of unique training samples varies from one to another (while it would be fixed to $0.632N$ in RP), and that samples are not all equally weighted.

In addition, RP also closely relates to the SubBag algorithm which combines Bagging and RS for constructing ensembles. Using $N$ bootstrapped samples (i.e., nearly equivalent to $\alpha_s = 0.632$) and setting $\alpha_f = 0.75$, Panov and Džeroski [2007] showed that SubBag has comparable performance to that of RF. An added advantage of SubBag, and hence of RP, is that it is applicable to any base estimator without the need to randomize the latter.

## 8.2 ON ACCURACY

Our validation of the RP algorithm is carried out in two steps. In this section, we first investigate how RP compares with other popular tree-based ensemble methods in terms of accuracy. In the next section, we then focus on its memory requirements for achieving optimal accuracy and its capability to handle strong memory constraints, again in comparison with other ensemble methods.

Considering accuracy only, our main objective is to investigate whether the additional degrees of freedom brought by $\alpha_s$ and $\alpha_f$ significantly improve, or degrade, the performance of RP. At first glance, one might indeed think that since the base estimators are (intentionally) built on parts of the data, the accuracy of the ensemble will be lower than if they were all built on the whole set. Additionally, our goal is also to see whether sampling features once globally, instead of locally at each node, impairs performance, as this is the main difference between RP and state-of-the-art methods such as RF or ET.

### 8.2.1 *Protocol*

We compare our method with P and RS, as well as with RF and ET. For RP, P and RS, two variants have been considered, one using standard decision trees (suffixed below with '-DT') as base estimators, and the other using extremely randomized trees (suffixed below with '-ET') as base estimators. Overall, 8 methods are compared: RP-DT, RP-ET, P-DT, P-ET, RS-DT, RS-ET, RF and ET.

We evaluate the accuracy of the methods on an extensive list of both artificial and real classification problems. For each dataset, three random partitions were drawn: the first and larger (50% of the original dataset) to be used as the training set, the second (25%) as validation set and the third (25%) as test set. For all methods, the hyper-parameters $\alpha_s$ and $\alpha_f$ were tuned on the validation set with a grid-search procedure, using the grid {0.01, 0.1, ..., 0.9, 1.0} for both $\alpha_s$ and $\alpha_f$. All other hyper-parameters were set to default values. In RF and ET, the number $K$ of features randomly selected at each node was tuned using the grid $\alpha_f p$. For all ensembles, 250 fully developed trees were generated and the generalization accuracy was estimated on the test set. Unless otherwise mentioned, for all methods and for all datasets, that procedure was repeated 50 times, using the same 50 random partitions between all methods, and all scores reported below are averages over those 50 runs. All algorithms and experiments have been implemented in Python, using Scikit-Learn [Pedregosa et al., 2011] as base framework.

### 8.2.2 *Small datasets*

Before diving into heavily computational experiments, we first wanted to validate our approach on small to medium datasets. To that end, experiments were carried out on a sample of 16 well-known and publicly available datasets (see Table 8.1) from the UCI machine learning repository [Frank and Asuncion, 2010], all chosen a priori and independently of the results obtained. Overall, these datasets cover a wide range of conditions, with the sample sizes ranging from 208 to 20000 and the number of features varying from 6 to 168. Detailed average performances of the 8 methods for all 16 datasets using the protocol described above are reported in Table 8.2. Below, we analyze general trends by performing various statistical tests.

Following recommendations in [Demsar, 2006], we first performed a Friedman test that rejected the hypothesis that all algorithms are equivalent at a significance level $\alpha = 0.05$. We then proceeded with a post-hoc Nemenyi test for a pairwise comparison of the average ranks of all 8 methods. According to this test, the performance of two classifiers is significantly different (at $\alpha = 0.05$) if their average ranks differ by at least the critical difference $CD = 2.6249$ (See [Demsar,

| Dataset | N | p |
|---|---|---|
| DIABETES | 768 | 8 |
| DIG44 | 18000 | 16 |
| IONOSPHERE | 351 | 34 |
| PENDIGITS | 10992 | 16 |
| LETTER | 20000 | 16 |
| LIVER | 345 | 6 |
| MUSK2 | 6598 | 168 |
| RING-NORM | 10000 | 20 |
| SATELLITE | 6435 | 36 |
| SEGMENT | 2310 | 19 |
| SONAR | 208 | 60 |
| SPAMBASE | 4601 | 57 |
| TWO-NORM | 9999 | 20 |
| VEHICLE | 1692 | 18 |
| VOWEL | 990 | 10 |
| WAVEFORM | 5000 | 21 |

Table 8.1: Small datasets.

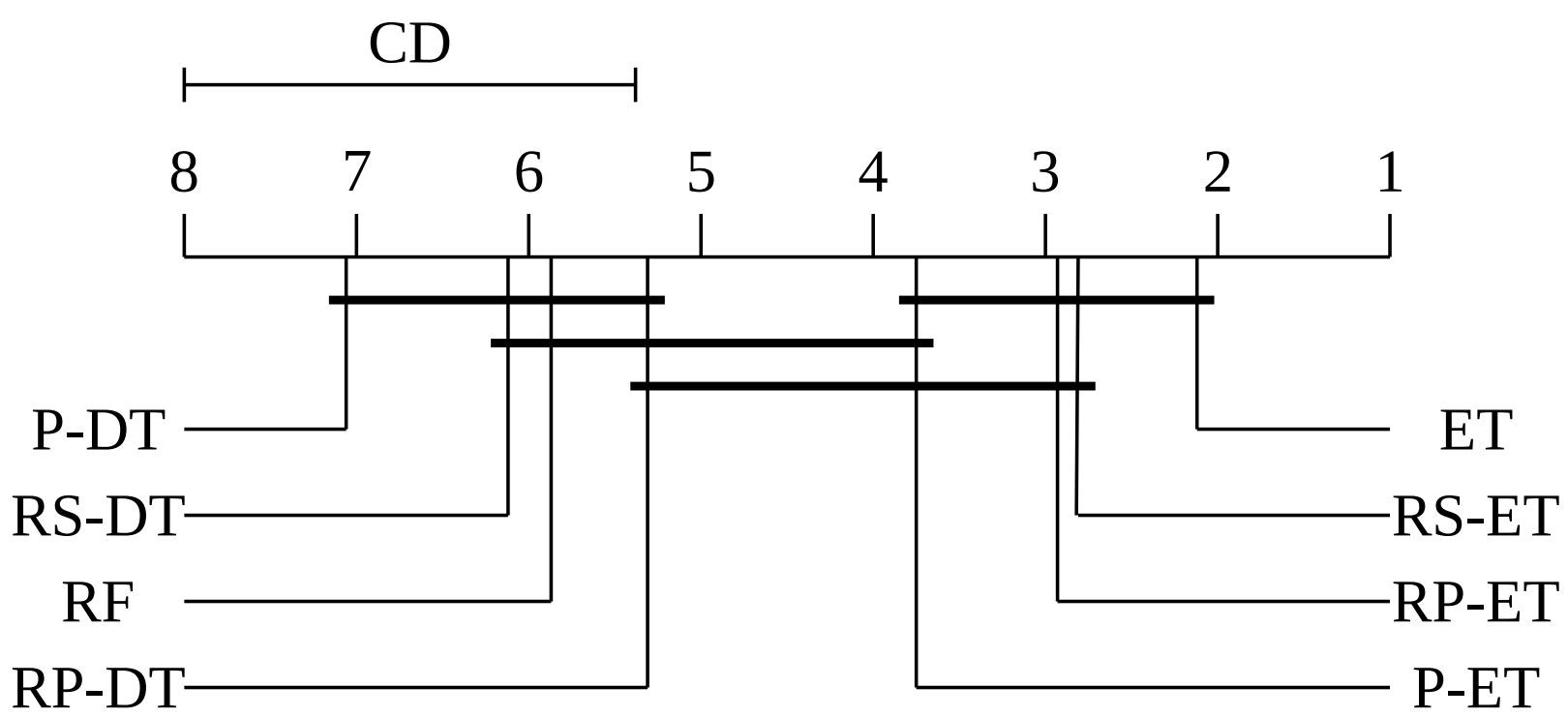

Figure 8.1: Average ranks of all methods on small datasets.

2006] for further details). The diagram of Figure 8.1 summarizes these comparisons. The top line in the diagram is the axis along which the average rank $R_m$ of each method $m$ is plotted, from the highest ranks (worst methods) on the left to the lowest ranks (best methods) on the right. Groups of methods that are not statistically different from each other are connected. The critical difference CD is shown above the graph. To further support these rank comparisons, we also compare the 50 accuracy values obtained over each dataset split for each pair of methods by using a paired t-test (with $\alpha = 0.01$). The results of these comparisons are summarized in Table 8.3 in terms of "Win-Draw-Loss" statuses of all pairs of methods; the three values at the intersection of row $i$ and column $j$ of this table respectively indicate on how many datasets method $i$ is significantly better/not significantly different/significantly worse than method $j$.

| **Validation** | RF | ET | P-DT | P-ET | RS-DT | RS-ET | RP-DT | RP-ET |
|---|---|---|---|---|---|---|---|---|
| DIABETES | 77.12 (6) | 77.25 (5) | 77.67 (4) | 78.01 (3) | 75.11 (8) | 76.77 (7) | 78.82 (2) | **79.07** (1) |
| DIG44 | 94.99 (7) | **95.78** (1) | 91.86 (8) | 95.46 (4) | 95.07 (6) | 95.69 (3) | 95.13 (5) | 95.72 (2) |
| IONOSPHERE | 94.40 (6) | 95.15 (3) | 93.86 (8) | 94.75 (5) | 94.11 (7) | 94.90 (4) | 95.20 (2) | **95.36** (1) |
| PENDIGITS | 98.94 (7) | **99.33** (1) | 98.09 (8) | 99.28 (4) | 99.02 (6) | 99.31 (3) | 99.07 (5) | 99.32 (2) |
| LETTER | 95.36 (7) | **96.38** (1) | 92.72 (8) | 95.87 (4) | 95.68 (6) | 96.08 (3) | 95.74 (5) | 96.10 (2) |
| LIVER | 72.37 (5) | 71.90 (6) | 72.55 (4) | 72.88 (3) | 68.06 (8) | 70.88 (7) | **74.53** (1) | 74.37 (2) |
| MUSK2 | 97.18 (7) | **97.73** (1) | 96.89 (8) | 97.60 (4) | 97.58 (6) | 97.72 (3) | 97.60 (5) | 97.73 (2) |
| RING-NORM | 97.44 (6) | 98.10 (5) | 96.41 (8) | 97.28 (7) | 98.25 (4) | 98.41 (3) | 98.50 (2) | **98.54** (1) |
| SATELLITE | 90.97 (7) | **91.56** (1) | 90.01 (8) | 91.40 (5) | 91.31 (6) | 91.50 (3) | 91.41 (4) | 91.54 (2) |
| SEGMENT | 97.46 (6) | 98.17 (2) | 96.78 (8) | 98.10 (4) | 97.33 (7) | 98.14 (3) | 97.52 (5) | **98.21** (1) |
| SONAR | 82.92 (7) | 86.92 (3) | 80.03 (8) | 84.73 (5) | 83.07 (6) | 87.07 (2) | 85.42 (4) | **88.15** (1) |
| SPAMBASE | 94.80 (7) | 95.36 (3) | 93.69 (8) | 95.01 (6) | 95.01 (5) | 95.50 (2) | 95.11 (4) | **95.57** (1) |
| TWO-NORM | 97.54 (6) | 97.77 (2) | 97.52 (7) | 97.59 (5) | 97.46 (8) | 97.63 (4) | 97.76 (3) | **97.82** (1) |
| VEHICLE | 88.67 (5) | 88.68 (4) | 88.26 (8) | 88.74 (3) | 88.41 (7) | 88.60 (6) | **89.22** (1) | 89.21 (2) |
| VOWEL | 92.04 (5) | **95.12** (1) | 85.19 (8) | 93.49 (4) | 89.76 (7) | 94.34 (3) | 91.10 (6) | 94.48 (2) |
| WAVEFORM | 85.45 (6) | 85.96 (2) | 84.89 (8) | 85.68 (5) | 84.91 (7) | 85.69 (4) | 85.85 (3) | **86.21** (1) |
| *Average rank* | 6.25 | 2.5625 | 7.4375 | 4.4375 | 6.5 | 3.75 | 3.5626 | 1.5 |
| **Test** | RF | ET | P-DT | P-ET | RS-DT | RS-ET | RP-DT | RP-ET |
| DIABETES | 75.62 (4) | 75.38 (5) | 75.67 (3) | **76.34** (1) | 73.03 (8) | 74.63 (7) | 75.32 (6) | 75.82 (2) |
| DIG44 | 94.96 (6) | **95.67** (1) | 91.79 (8) | 95.39 (4) | 94.98 (5) | 95.58 (2) | 94.95 (7) | 95.55 (3) |
| IONOSPHERE | 92.20 (6) | **93.22** (1) | 92.09 (7) | 92.40 (4) | 92.02 (8) | 93.22 (2) | 92.34 (5) | 92.68 (3) |
| PENDIGITS | 98.84 (7) | **99.23** (1) | 97.97 (8) | 99.21 (3) | 98.95 (5) | 99.21 (2) | 98.93 (6) | 99.20 (4) |
| LETTER | 95.27 (7) | **96.29** (1) | 92.57 (8) | 95.89 (4) | 95.61 (5) | 96.03 (2) | 95.61 (6) | 95.99 (3) |
| LIVER | 69.95 (3) | 68.22 (6) | **70.43** (1) | 69.58 (5) | 63.17 (8) | 67.35 (7) | 70.20 (2) | 69.67 (4) |
| MUSK2 | 97.08 (7) | **97.61** (1) | 96.69 (8) | 97.54 (4) | 97.47 (5) | 97.58 (2) | 97.42 (6) | 97.56 (3) |
| RING-NORM | 97.48 (6) | 98.07 (5) | 96.42 (8) | 97.25 (7) | 98.16 (4) | **98.31** (1) | 98.22 (3) | 98.30 (2) |
| SATELLITE | 90.67 (7) | 91.22 (2) | 89.66 (8) | 91.20 (3) | 91.15 (5) | **91.28** (1) | 91.04 (6) | 91.20 (4) |
| SEGMENT | 97.02 (5) | **97.93** (1) | 96.44 (8) | 97.86 (3) | 96.86 (7) | 97.90 (2) | 96.88 (6) | 97.84 (4) |
| SONAR | 79.53 (5) | **82.76** (1) | 75.15 (8) | 80.07 (4) | 78.50 (6) | 82.19 (2) | 78.26 (7) | 81.92 (3) |
| SPAMBASE | 94.76 (7) | 95.17 (3) | 93.51 (8) | 94.84 (5) | 94.88 (4) | 95.22 (2) | 94.80 (6) | **95.22** (1) |
| TWO-NORM | 97.22 (7) | **97.50** (1) | 97.26 (6) | 97.29 (4) | 97.20 (8) | 97.33 (2) | 97.33 (3) | 97.28 (5) |
| VEHICLE | 87.47 (7) | 87.85 (3) | 87.01 (8) | 87.98 (2) | 87.50 (6) | 87.68 (5) | 87.73 (4) | **88.08** (1) |
| VOWEL | 91.51 (5) | **93.95** (1) | 84.17 (8) | 92.93 (4) | 89.51 (7) | 93.60 (2) | 89.89 (6) | 93.09 (3) |
| WAVEFORM | 85.23 (5) | **85.77** (1) | 84.68 (8) | 85.40 (3) | 84.74 (7) | 85.38 (4) | 85.16 (6) | 85.56 (2) |
| *Average rank* | 5.875 | 2.125 | 7.0625 | 3.75 | 6.125 | 2.8125 | 5.3125 | 2.9375 |

Table 8.2: Accuracy on small datasets (in %).

| | RF | ET | P-DT | P-ET | RS-DT | RS-ET | RP-DT | RP-ET |
|---|---|---|---|---|---|---|---|---|
| RF | — | 1/2/13 | 12/4/0 | 1/7/8 | 4/7/5 | 2/2/12 | 1/10/5 | 0/4/12 |
| ET | 13/2/1 | — | 14/1/1 | 10/5/1 | 13/3/0 | 4/11/1 | 12/2/2 | 5/10/1 |
| P-DT | 0/4/12 | 1/1/14 | — | 0/4/12 | 2/3/11 | 2/1/13 | 0/4/12 | 0/4/12 |
| P-ET | 8/7/1 | 1/5/10 | 12/4/0 | — | 9/6/1 | 2/6/8 | 9/6/1 | 0/11/5 |
| RS-DT | 5/7/4 | 0/3/13 | 11/3/2 | 1/6/9 | — | 0/2/14 | 1/11/4 | 0/4/12 |
| RS-ET | 12/2/2 | 1/11/4 | 13/1/2 | 8/6/2 | 14/2/0 | — | 11/4/1 | 1/13/2 |
| RP-DT | 5/10/1 | 2/2/12 | 12/4/0 | 1/6/9 | 4/11/1 | 1/4/11 | — | 0/6/10 |
| RP-ET | 12/4/0 | 1/10/5 | 12/4/0 | 5/11/0 | 12/4/0 | 2/13/1 | 10/6/0 | — |

Table 8.3: Pairwise t-test comparisons on small datasets.

Since all methods are variants of ensembles of decision trees, average accuracies are not strikingly different from one method to another (see Table 1 of the supplementary materials). Yet, significant trends appear when looking at Figure 8.1 and Table 8.3. First, all ET-based methods are ranked before DT-based methods, including the popular Random Forest algorithm. Overall, the original ET algorithm is ranked first ($R_{ET} = 2.125$), then come RS-ET and RP-ET at close positions ($R_{RS-ET} = 2.8125$ and $R_{RP-ET} = 2.9375$) while P-ET is a bit behind ($R_{P-ET} = 3.75$). According to Figure 8.1, only ET is ranked significantly higher than all DT-based method but looking at Table 8.3, the worse ET-based variant (P-ET) is still 9 times significantly better (w.r.t. the 50 runs over each set) and only 1 time significantly worse than the best DT-based variant (RP-DT). The separation between these two families of algorithm thus appears quite significant. This observation clearly suggests that using random split thresholds, instead of optimized ones like in decision trees, pays off in terms of generalization.

Among ET-based methods, RP-ET is better than P-ET but it is superseded by ET and RS-ET in terms of average rank. Since RS-ET is a particular case of RP-ET, this suggests that we are slightly overfitting when tuning the additional parameter $\alpha_s$. And indeed RP-ET is better ranked than RS-ET in average on the validation set (results not shown). Table 8.3 however indicates otherwise and makes RP-ET appear as slightly better than RS-ET (2/13/1). Regarding ET over RP-ET, the better performance of the former (5/10/1) is probably due to the fact that in ET subsets of features are redrawn locally at each node when building trees and not once and for all prior to their construction. This gives less chances to generate improper trees because of a bad initial choice of features and thus leads to a lower bias and a better accuracy.

Among DT-based methods, RP-DT now comes first (mean rank of 5.3125), then RF ($R_{RF} = 5.875$), RS-DT ($R_{RS-DT} = 6.125$) and then P-DT in last ($R_{P-DT} = 7.0625$). RP is only significantly worse than another DT-based variant on 1 dataset. The extra-randomization brought by the random choices of both samples and features seems to be beneficial with decision trees that do not benefit from the randomization of discretization thresholds. The fact that RF samples features locally does not appear here anymore as an advantage over RP (RF is significantly worse on 5 problems and better on only one), probably because the decrease of bias that it provides does not exceed the increase of variance with respect to global feature selection.

### 8.2.3 *Larger datasets*

While the former experiments revealed promising results, it is fair to ask whether the conclusions that have been drawn would hold on and generalize to larger problems, for example when dealing with a few relevant features buried into hundreds or thousands of not important features (e.g., in genomic data), or when dealing with many correlated features (e.g., in images). To investigate this question, a second bench of experiments was carried out on 13 larger datasets (see Table 8.4). All but MADELON are real data. In terms of dimensions, these datasets are far bigger, ranging from a few hundreds of samples and thousands of features, to thousands of samples but hundreds of features. As such, the complexity of the problems is expected to be greater. We adopted the exact same protocol as for smaller datasets. However, to lower computing times, for datasets marked with $*$, the methods were run using 100 trees instead of 250 and the minimum number of samples required in an internal node was set to 10 in order to control complexity. Detailed results are provided in Table 8.5 and are summarized in Figure 8.2 and Table 8.6, respectively in terms of average rank (the critical difference at $\alpha = 0.05$ is now 2.9120) and Win/Draw/Loss statuses obtained with paired t-tests. A Friedman test (at $\alpha = 0.05$) still indicates that some methods are significantly different from the others.

As it may be observed from Figure 8.2, the average ranks of the methods are closer to each other than in the previous experiments, now ranging from 2.38 to 6.61, while they were previously ranging from 2.12 to 7. Methods are more connected by critical difference bars. This suggests that overall they behave more similarly to each other than before. General trends are nevertheless comparable to what we observed earlier. ET-based methods still seem to be the front-runners. From Figure 8.2, RS-ET, ET and RP-ET are in the top 4, while P-DT, RF and RP-DT remain in the second half of the ranking. Surprisingly however, RS-DT now comes right after RS-ET and ET and just before RP-ET whereas it ranked penultimate on the smaller datasets. Table 8.6 however suggests that RS-DT performs actually a little worse against RP-ET (1/10/2). All in all, it thus still seems beneficial to randomize split thresholds on the larger datasets.

| Dataset | N | p |
|---|---|---|
| CIFAR10* | 60000 | 3072 |
| MNIST3VS8 | 13966 | 784 |
| MNIST4VS9 | 13782 | 784 |
| MNIST* | 70000 | 784 |
| ISOLET | 7797 | 617 |
| ARCENE | 900 | 10000 |
| BREAST2 | 295 | 24496 |
| MADELON | 4400 | 500 |
| MARTI0 | 500 | 1024 |
| REGED0 | 500 | 999 |
| SECOM | 1567 | 591 |
| TIS | 13375 | 927 |
| SIDO0* | 12678 | 4932 |

Table 8.4: Large datasets.

Comparing ET-based variants, ET is no longer the best method on average, but RS-ET is (with 4/9/0 for RS-ET versus ET). This suggests than on larger datasets, picking features globally at random prior to the construction of the trees is as good, or even beat picking them locally at each node. Due to the quantitatively larger number of samples in a patch, and also to the larger number of redundant features expected in some large datasets (e.g., in CIFAR10 or MNIST), it is indeed less likely to build improper trees with strong biases. As a result, variance can be further decreased by sampling globally. In support of this claim, on a few problems such as ARCENE, BREAST2, or MADELON that contain many irrelevant features, ET remains the best method. In that case, it is indeed more likely to sample globally improper random patches, and hence to build improper trees. The average rank of RP-ET suggests that it performs worse than RS-ET and thus that there is some potential overfitting when tuning $\alpha_s$ in addition to $\alpha_f$. This difference is however not confirmed in Table 8.6 where the accuracies of these two methods are shown to be never significantly different (0/13/0). RP-ET is also on a perfect par with ET (1/11/1). Among DT-based variants, RP-DT, which was the best performer on small datasets, is still ranked above RF and P-DT, but it is now ranked below RS-DT with a win/draw/loss of 0/11/2. This is again due to some overfitting.

While less conclusive than before, the results on larger datasets are consistent with what we observed earlier. In particular, they indicate that the Random Patches method (with ET) remains competitive with the best performers.

| **Validation** | RF | ET | P-DT | P-ET | RS-DT | RS-ET | RP-DT | RP-ET |
|---|---|---|---|---|---|---|---|---|
| CIFAR10 | 45.16 (6) | **46.17** (1) | 44.92 (7) | 44.88 (8) | 45.83 (5) | 46.09 (3) | 45.89 (4) | 46.13 (2) |
| MNIST3V8 | 98.42 (7) | 98.77 (5) | 97.63 (8) | 98.57 (6) | 98.79 (4) | 98.86 (2) | 98.81 (3) | **98.87** (1) |
| MNIST4V9 | 98.47 (7) | 98.82 (3) | 97.39 (8) | 98.47 (6) | 98.77 (5) | 98.89 (2) | 98.80 (4) | **98.91** (1) |
| MNIST | 96.14 (7) | 96.56 (3) | 94.35 (8) | 96.19 (6) | 96.52 (5) | 96.57 (2) | 96.53 (4) | **96.57** (1) |
| ISOLET | 93.96 (7) | 95.07 (3) | 90.90 (8) | 94.23 (6) | 94.38 (5) | 95.15 (2) | 94.47 (4) | **95.20** (1) |
| ARCENE | 73.84 (7) | 77.76 (3) | 73.76 (8) | 75.12 (6) | 76.40 (5) | 77.60 (4) | 79.44 (2) | **80.00** (1) |
| BREAST2 | 69.64 (6) | 69.91 (4) | 70.54 (3) | 69.83 (5) | 69.59 (7) | 69.54 (8) | **72.59** (1) | 71.62 (2) |
| MADELON | 76.12 (8) | **81.41** (1) | 78.48 (7) | 81.06 (5) | 80.68 (6) | 81.29 (3) | 81.18 (4) | 81.36 (2) |
| MARTI | 87.66 (8) | 87.92 (6) | 88.08 (4) | 88.09 (3) | 87.77 (7) | 88.00 (5) | 88.19 (2) | **88.20** (1) |
| REGED | 98.00 (6) | 98.46 (2) | 97.08 (8) | 98.24 (5) | 98.00 (7) | 98.41 (3) | 98.40 (4) | **98.57** (1) |
| SECOM | 93.37 (5) | 93.36 (6) | 93.34 (8) | 93.42 (2) | 93.35 (7) | 93.40 (4) | 93.41 (3) | **93.45** (1) |
| TIS | 91.81 (5) | 91.53 (7) | 91.59 (6) | 91.50 (8) | 92.04 (3) | 91.97 (4) | **92.26** (1) | 92.06 (2) |
| SIDO | 97.46 (5) | 97.44 (6) | 97.36 (8) | 97.36 (7) | 97.47 (3) | 97.46 (4) | **97.52** (1) | 97.52 (2) |
| *Avg. rank* | 6.4615 | 3.8461 | 7.0 | 5.6153 | 5.3076 | 3.5384 | 2.8461 | 1.3846 |

| **Test** | RF | ET | P-DT | P-ET | RS-DT | RS-ET | RP-DT | RP-ET |
|---|---|---|---|---|---|---|---|---|
| CIFAR10 | 44.87 (7) | 45.95 (2) | 44.75 (8) | 44.95 (6) | 45.86 (4) | **46.02** (1) | 45.74 (5) | 45.93 (3) |
| MNIST3V8 | 98.31 (7) | 98.64 (5) | 97.47 (8) | 98.48 (6) | 98.70 (2) | **98.71** (1) | 98.68 (4) | 98.68 (3) |
| MNIST4V9 | 98.33 (6) | 98.68 (2) | 97.16 (8) | 98.32 (7) | 98.59 (4) | **98.69** (1) | 98.59 (5) | 98.67 (3) |
| MNIST | 96.10 (7) | 96.47 (3) | 94.30 (8) | 96.18 (6) | 96.47 (4) | **96.49** (1) | 96.46 (5) | 96.48 (2) |
| ISOLET | 93.87 (7) | **94.95** (1) | 90.71 (8) | 94.15 (6) | 94.33 (4) | 94.94 (2) | 94.29 (5) | 94.88 (3) |
| ARCENE | 71.04 (4) | **72.56** (1) | 66.16 (8) | 68.56 (7) | 71.52 (3) | 72.24 (2) | 69.44 (6) | 70.00 (5) |
| BREAST2 | 65.86 (6) | **66.16** (1) | 65.54 (8) | 65.59 (7) | 66.13 (2) | 65.91 (5) | 65.94 (4) | 66.08 (3) |
| MADELON | 75.33 (8) | **80.83** (1) | 77.86 (7) | 80.78 (2) | 80.06 (6) | 80.59 (3) | 80.06 (5) | 80.42 (4) |
| MARTI | 87.74 (8) | 87.85 (6) | **88.24** (1) | 88.17 (2) | 87.82 (7) | 88.06 (4) | 88.08 (3) | 88.01 (5) |
| REGED | 97.60 (5) | **98.20** (1) | 96.40 (8) | 97.92 (3) | 97.39 (6) | 98.00 (2) | 97.32 (7) | 97.82 (4) |
| SECOM | 93.30 (7) | 93.22 (8) | **93.38** (1) | 93.30 (6) | 93.37 (2) | 93.31 (4) | 93.33 (3) | 93.31 (5) |
| TIS | 91.68 (5) | 91.42 (7) | 91.49 (6) | 91.40 (8) | **92.05** (1) | 91.91 (3) | 92.05 (2) | 91.90 (4) |
| SIDO | 97.33 (4) | 97.33 (3) | 97.26 (7) | 97.26 (8) | **97.35** (1) | 97.34 (2) | 97.33 (5) | 97.32 (6) |
| *Avg. rank* | 6.2307 | 3.1538 | 6.6153 | 5.6923 | 3.5384 | 2.3846 | 4.5384 | 3.8461 |

Table 8.5: Accuracy on larger datasets (in %).

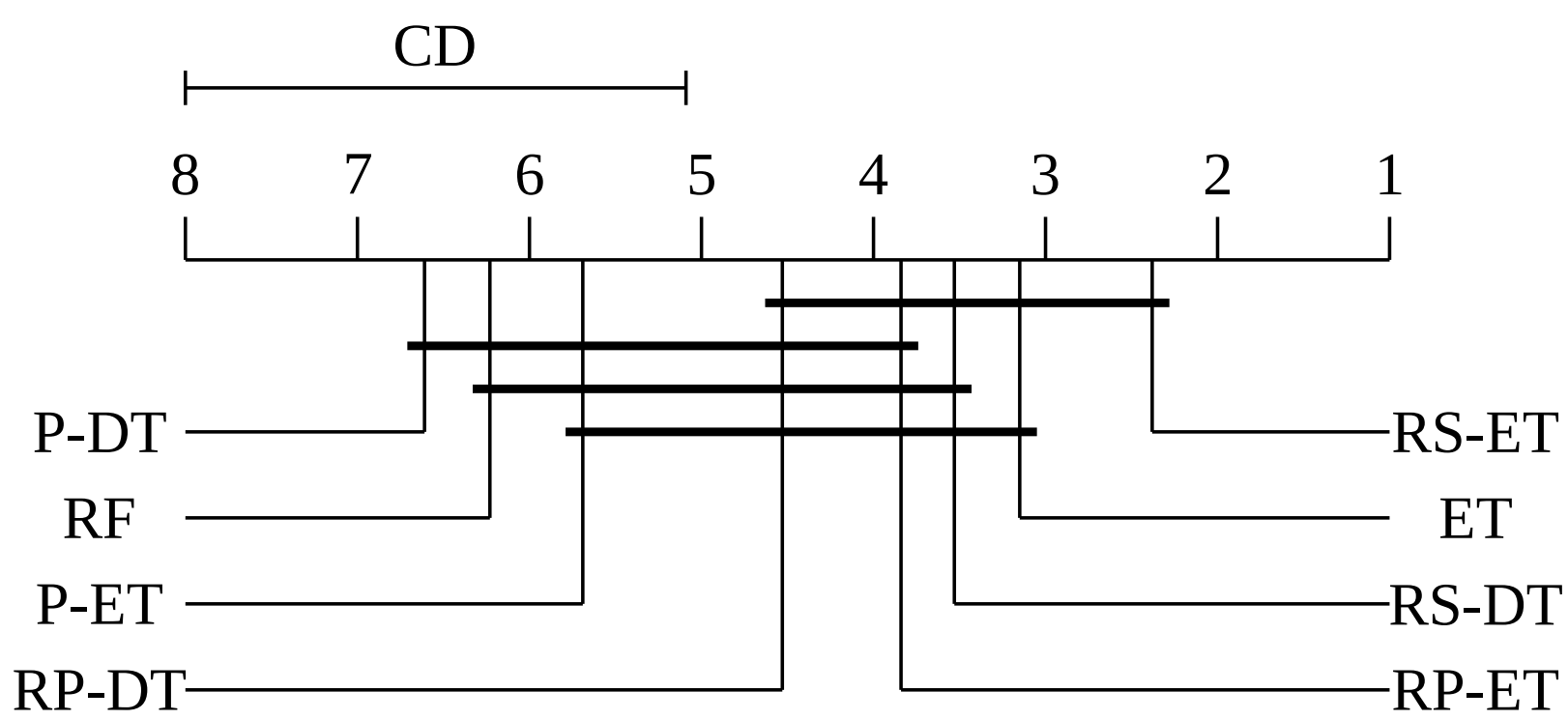

Figure 8.2: Average ranks of all methods on larger datasets.

| | RF | ET | P-DT | P-ET | RS-DT | RS-ET | RP-DT | RP-ET |
|---|---|---|---|---|---|---|---|---|
| RF | — | 1/5/7 | 8/3/2 | 2/6/5 | 0/6/7 | 0/5/8 | 0/6/7 | 0/6/7 |
| ET | 7/5/1 | — | 9/2/2 | 7/6/0 | 3/7/3 | 0/9/4 | 5/6/2 | 1/11/1 |
| P-DT | 2/3/8 | 2/2/9 | — | 1/5/7 | 0/3/10 | 0/3/10 | 1/3/9 | 0/4/9 |
| P-ET | 5/6/2 | 0/6/7 | 7/5/1 | — | 0/6/7 | 0/5/8 | 2/5/6 | 1/5/7 |
| RS-DT | 7/6/0 | 3/7/3 | 10/3/0 | 7/6/0 | — | 1/8/4 | 2/11/0 | 1/10/2 |
| RS-ET | 8/5/0 | 4/9/0 | 10/3/0 | 8/5/0 | 4/8/1 | — | 4/8/1 | 0/13/0 |
| RP-DT | 7/6/0 | 2/6/5 | 9/3/1 | 6/5/2 | 0/11/2 | 1/8/4 | — | 1/9/3 |
| RP-ET | 7/6/0 | 1/11/1 | 9/4/0 | 7/5/1 | 2/10/1 | 0/13/0 | 3/9/1 | — |

Table 8.6: Pairwise t-test comparisons on larger datasets.

### 8.2.4 *Conclusions*

Overall, this extensive experimental study reveals many interesting results. The first and foremost result is that ensembles of randomized trees nearly always beat ensembles of standard decision trees. As off-the-shelf methods, we advocate that ensembles of such trees should be preferred to ensembles of decision trees. In particular, these results show that the well-known Random Forest algorithm does not compete with the best performers. Far more important to our concern though, this study validates our RP approach. Building ensembles (of ET) on random patches of data is competitive in terms of accuracy. Overall, there is no strong statistical evidence that the method performs less well, but there is also no conclusive evidence that it significantly improves performance. Yet, results show that RP is often as good as the very best methods. Regarding the shape of the random patches, the strategy behind Pasting (i.e., $\alpha_s$ free and $\alpha_f = 1.0$) proved to be (very) ineffective on many datasets while the Random Subspace algorithm (i.e., $\alpha_s = 1.0$ and $\alpha_f$ free) always ranked among the very best performers. On average, RS indeed came in second on the small datasets and in first on the larger datasets, which tends to indicate that sampling features is crucial in terms of accuracy. As for patches of freely adjustable size (i.e., using RP), they showed to be slightly sensitive to overfitting but proved to remain closely competitive with the very best methods. In addition, these results also suggest that sampling features globally, once and for all, prior to the construction of a (randomized) decision tree, does not actually impair performance. For instance, RS-ET or RP-ET are indeed not strongly worse, nor better, than ET, in which candidates features are re-sampled locally at each node.

## 8.3 ON MEMORY

Section 8.2 reveals that building an ensemble of base estimators on random patches, instead of the whole data, is a competitive strategy.

In the context of big data, that is when the size of the dataset is far bigger than the available memory, this suggests that using random parts of the data of the appropriate size to build each base estimator would likely result in an ensemble which is actually as good as if the whole data could have been loaded and used.

Formally, we assume a general framework where the number of data units that can be loaded at once into memory is constrained to be lower than a given threshold $\mu_{max}$. Not considering on-line algorithms within the scope of this study, $\mu_{max}$ can hence be viewed as the total units of data allowed to be used to build a single base estimator. In the context of our sampling methods, the amount of memory required for a patch is given by $(\alpha_s N)(\alpha_f p)$ and thus constraining memory by $\mu_{max}$ is equivalent to constraining the relative patch size $\alpha_s \alpha_f$ to be lower than $\mu'_{max} = \mu_{max}/(Np)$. While simplistic[1], this framework has the advantage of clearly addressing one of the main difficulties behind big data, that is the lack of fast memory. Yet, it is also relevant in other contexts, for example when data is costly to access (e.g., on remote locations) or when algorithms are run on embedded systems with strong memory constraints.

In Section 8.3.1, we first study the effects of $\alpha_s$ and $\alpha_f$ on the accuracy of the resulting ensemble and show that it is problem and base estimator dependent. Second, we show that the memory requirements, i.e., the relative size $\alpha_s \alpha_f$ of the random patches, can often be drastically reduced without significantly degrading the performance of the ensemble (Section 8.3.2). Third, because the sensitivity of the ensemble to $\alpha_s$ and $\alpha_f$ is problem and base estimator specific, we show that under very strong memory constraints adjusting both parameters at the same time, as RP does, is no longer merely as good but actually significantly better than other ensemble methods (Section 8.3.3).

### 8.3.1 *Sensitivity to $\alpha_s$ and $\alpha_f$*

Let us first consider and analyze the triplets

$$\{(\alpha_s, \alpha_f, \text{Acc}_{\mathcal{L}}(\alpha_s, \alpha_f)) | \forall \alpha_s, \alpha_f\} \tag{8.1}$$

for various problems, where $\text{Acc}_{\mathcal{L}}(\alpha_s, \alpha_f)$ is the average test accuracy of an ensemble built on random patches of size $\alpha_s \alpha_f$ (using the same protocol as previously) on the dataset $\mathcal{L}$.

As Figure 8.3 illustrates for six datasets, the surfaces defined by these points vary significantly from one problem to another. We observed four main trends. In Figures 8.3a, and 8.3b (resp., 8.3c), accuracy increases with $\alpha_s$ (resp., $\alpha_f$) while adjusting $\alpha_f$ (resp., $\alpha_s$) has no or limited impact. In Figure 8.3d, the best strategy is to increase both $\alpha_s$ and $\alpha_f$. Finally, in Figures 8.3e and 8.3f, the surface features

1 e.g., the quantity of memory used by the estimator itself is not taken into account.

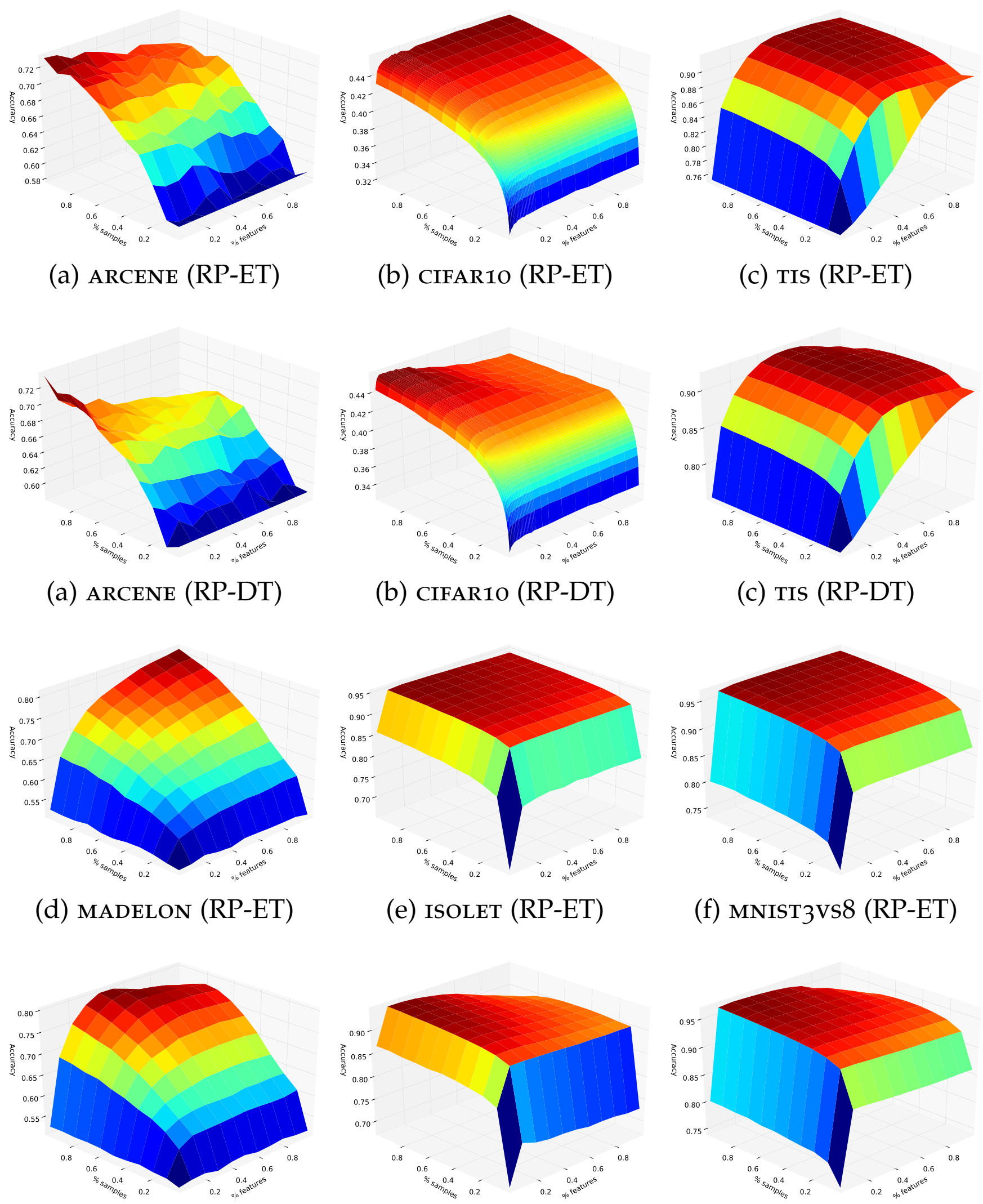

(a) ARCENE (RP-ET) (b) CIFAR10 (RP-ET) (c) TIS (RP-ET)

(a) ARCENE (RP-DT) (b) CIFAR10 (RP-DT) (c) TIS (RP-DT)

(d) MADELON (RP-ET) (e) ISOLET (RP-ET) (f) MNIST3VS8 (RP-ET)

(d) MADELON (RP-DT) (e) ISOLET (RP-DT) (f) MNIST3VS8 (RP-DT)

Figure 8.3: Learning surfaces.

plateaus, which means that beyond some threshold, increasing $\alpha_s$ or $\alpha_f$ does not yield any significant improvement. Interestingly, in most of the cases, the optimum corresponds to a value $\alpha_s \alpha_f$ much smaller than 1.

The choice of the base estimators does not have a strong impact on the aspect of the curves (compare the 1$^{st}$ and 3$^{rd}$ rows of sub-figures in Figure 8.3 with those in the 2$^{nd}$ and 4$^{th}$ rows). The only difference is the decrease of the accuracy of RP-DT when $\alpha_s$ and $\alpha_f$ grow towards 1.0. Indeed, since the only source of randomization in RP-DT is patch selection, it yields in this case ensembles of identical trees and therefore amounts to building a single tree on the whole dataset. By contrast, because of the extra-randomization of the split thresholds in ET, there is typically no drop of accuracy for RP-ET when $\alpha_s$ and $\alpha_f$ grow to 1.0.

Overall, this analysis suggests that not only the best pair $\alpha_s \alpha_f$ depends on the problem, but also that the sensitivity of the ensemble to changes to the size of a random patch is both problem and base estimator specific. As a result, these observations advocate for a method that could favor $\alpha_s$, $\alpha_f$ or both, and do so appropriately given the base estimator.

### 8.3.2 *Memory reduction, without significant loss*

We proceed to study in this section the actual size of the random patches when the values of $\alpha_s$ and $\alpha_f$ are tuned using an independent validation set. Our results are summarized in Figure 8.4a. Each ellipse corresponds to one of the 29 datasets of our benchmark, whose center is located at $(\overline{\alpha_s}, \overline{\alpha_f})$ (i.e., the average parameter values over the 50 runs) and whose semi-axes correspond to the standard deviations of $\alpha_s$ and $\alpha_f$. Any point in the plot corresponds to a pair $(\alpha_s, \alpha_f)$ and thus to a relative consumption $\mu' = \alpha_s \alpha_f$ of memory. To ease readability, level curves are plotted for $\mu' = 0.01, 0.1, ..., 0.9$. In the right part of the figure, the histogram counts the number of datasets such that $\overline{\alpha_s} \cdot \overline{\alpha_f}$ falls in the corresponding level set.

Figure 8.4a corroborates our previous discussion. On some datasets, it is better to favor $\alpha_s$ while on some other increasing $\alpha_f$ is a better strategy. The various sizes of the ellipses also confirm that the sensitivity to variations of $\alpha_s$ and $\alpha_f$ is indeed problem-specific. The figure also clearly highlights the fact that, even under no memory constraint, the optimal patches rarely consume the whole memory. A majority of ellipses indeed lie below the level set $\mu' = 0.5$ and only a couple of them are above $\mu' = 0.75$. With respect to ET or RF for which the base estimators are all built on the whole dataset, this means that ensembles of patches are not only as competitive but also less memory greedy. In addition, the figure also points out the difference between RP-ET and RP-DT as discussed in the previous section. To ensure diversity, RP-DT is constrained to use smaller patches than RP-ET, hence explaining why the ellipses in red are on average below those in blue. While RP-DT proved to be a bit less competitive in terms of accuracy, this indicates on the other hand that RP-DT may actually be more interesting from a memory consumption point of view.

In Section 8.3.1, we observed plateaus or very gentle slopes around the optimal pair $(\alpha_s, \alpha_f)$. From a memory point of view, this suggests that the random patches are likely to be reducible without actually degrading the accuracy of the resulting ensemble. Put otherwise, our interest is to find the smallest size $\alpha_s \alpha_f$ such that the accuracy of the resulting ensemble is not significantly worse than an ensemble built without such constraint. To that end, we study the extent at which the constraint $\alpha_s \alpha_f < \mu'_{max}$ can be strengthened without any significant drop in accuracy. If $\mu'_{max}$ can be reduced significantly then it would indeed mean that even when only small parts of the data are actually used to build single base estimators, competitive performance can still be achieved.

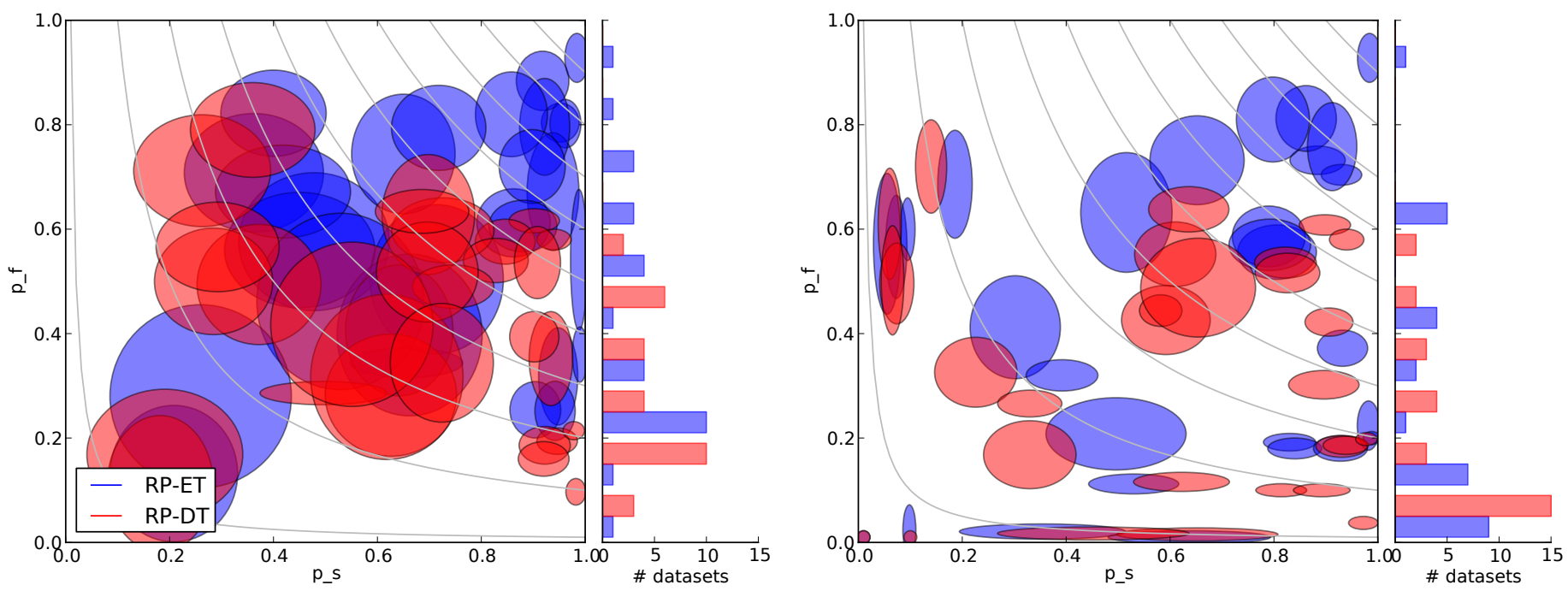

(a) Original patches (b) Reduced patches

Figure 8.4: Optimal sizes of the random patches on our benchmark.

Figure 8.4b summarizes our results. For all datasets, $\mu'_{max}$ was set to the lowest value such that it cannot be statistically detected that the average accuracy of the resulting ensemble is different from the average accuracy of an ensemble built with no memory constraint (at $\alpha = 0.05$). With regard to Figure 8.4a, the shift of most ellipses to lower memory level sets confirm our first intuition. In many cases, the size of the random patches can indeed be reduced, often drastically, without significant decrease of accuracy. For more than half of the datasets, memory can indeed be decreased to $\mu' = 0.1$ or $\mu' = 0.2$. In other words, building trees on small parts of the data (i.e., 10% or 20% of the original dataset) is, for more than half of the datasets, enough to reach competitive accuracy. Also, the sensitivity to $\alpha_s$ and $\alpha_f$ is now even more patent. Some ensembles use very few samples ($\alpha_s < 0.1$) but with many features, while other uses many samples with few features ($\alpha_f < 0.1$). Again, from a memory point of view, RP-DT appears to be more interesting than RP-ET. The memory reduction is larger, as the histogram indicates. Optimized splits in the decision trees may indeed lead to a better exploitation of the data, hence to a potentially larger reduction of memory. In conclusion, while not detecting significant differences in accuracy does not allow to conclude that the performances are truly similar, these figures at least illustrate that memory requirements can be drastically reduced without apparent loss in accuracy.

### 8.3.3 *Memory reduction, with loss*

The previous section has shown that the memory consumption can be reduced up to some threshold $\mu'_{max}$ with no significant loss in accuracy. In this section we now look at the accuracy of the resulting ensemble when $\mu'_{max}$ is further decreased. We argue that with severe constraints, and because datasets have all a different sensitivity, it is even more crucial to better exploit data and thus to find the right trade-off between both $\alpha_s$ and $\alpha_f$, as only RP can.

To illustrate our point, Figure 8.5 compares for 6 representative datasets the accuracy of the methods with respect to the memory constraint $\alpha_s \alpha_f < \mu'_{max}$. A plain line indicates that the generalization error of the best resulting ensemble under memory constraint $\mu'_{max}$ is significantly (at $\alpha = 0.05$) worse on the test sets than when there is no constraint (i.e., $\mu'_{max} = 1$). A dotted line indicates that on average, on the test set, the ensemble is not significantly less accurate.

As the figure shows, when $\mu'_{max}$ is low, RP-based ensembles often achieve the best accuracy. Only on ARCENE (Figure 8.5a), RS seems to be a better strategy, suggesting some overfitting in setting $\alpha_s$ in RP. On all 5 other example datasets, RP is equivalent or better than RS and P for low values of $\mu'_{max}$, with the largest gaps appearing on ISOLET (Figure 8.5e) and MNIST3VS8 (Figure 8.5f). As already observed in the previous section, although RP-DT is not the best strategy when memory is unconstrained, its curve dominates the curve of RP-ET for small values of $\mu'_{max}$ in Figures 8.5b, 8.5c, and 8.5d. Because split thresholds are not randomized in RP-DT, this method is more resistant than RP-ET to the strong randomization induced by a very low $\mu'_{max}$ threshold.

For comparison, Figure 8.5 also features the learning curves of both ET and RF (with K optimized on the validation set), in which the trees have all been built on the *same* training sample of $\mu'_{max}N$ instances, with all features. These results are representative of the use of a straightforward sub-sampling of the instances to handle the memory constraint. On all datasets, this setting yields very poor performance when $\mu'_{max}$ is low. Building base estimators on re-sampled random patches thus brings a clear advantage to RP, RS and P and hence confirms the conclusions of Basilico et al. [2011] who showed that using more data indeed produces more accurate models than learning from a single sub-sample. This latter experiment furthermore shows that the good performances of RP cannot be trivially attributed to the fact that our datasets contain so many instances that only processing a sub-sample of them would be enough. On most problems, the slopes of the learning curves of RF and ET indeed suggest that convergence has not yet been reached on these datasets. Yet, important improvement are gained by sub-sampling random patches. Contrary to intuition, it also does not appear that the bigger the datasets, the

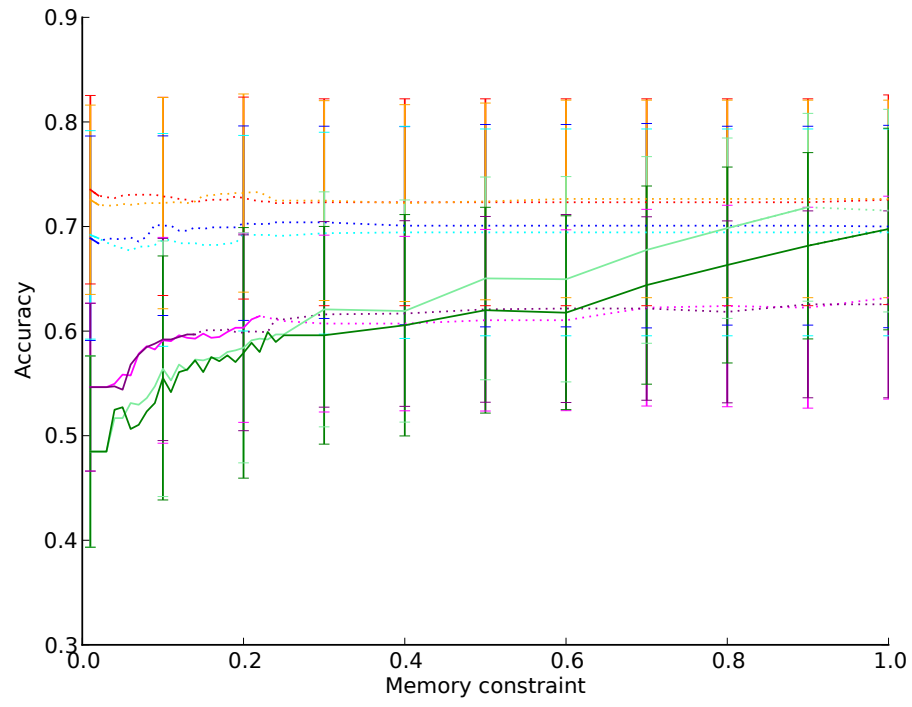

(a) ARCENE

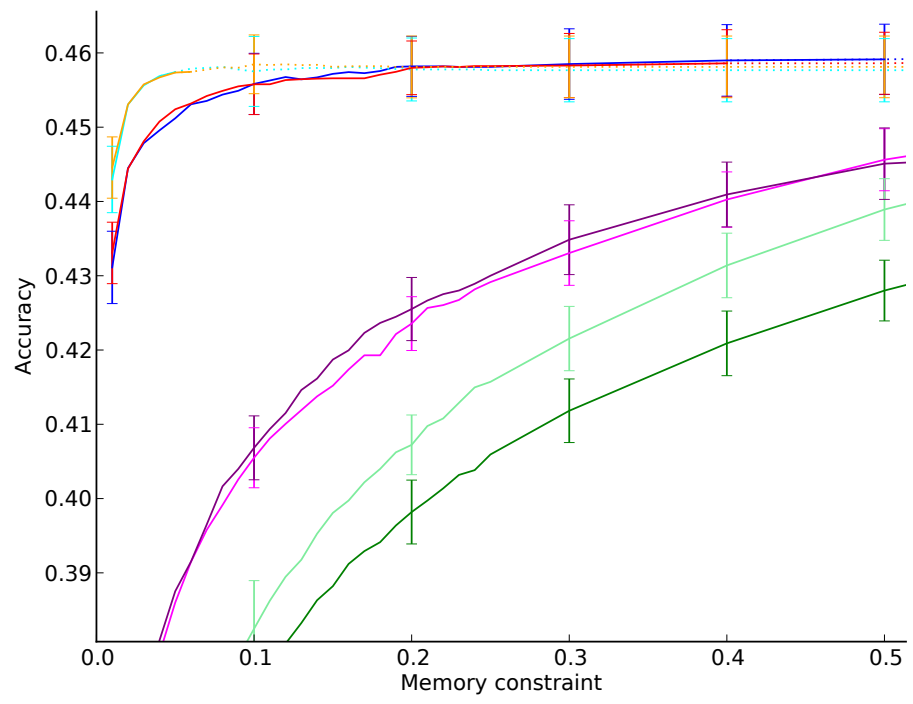

(b) CIFAR10

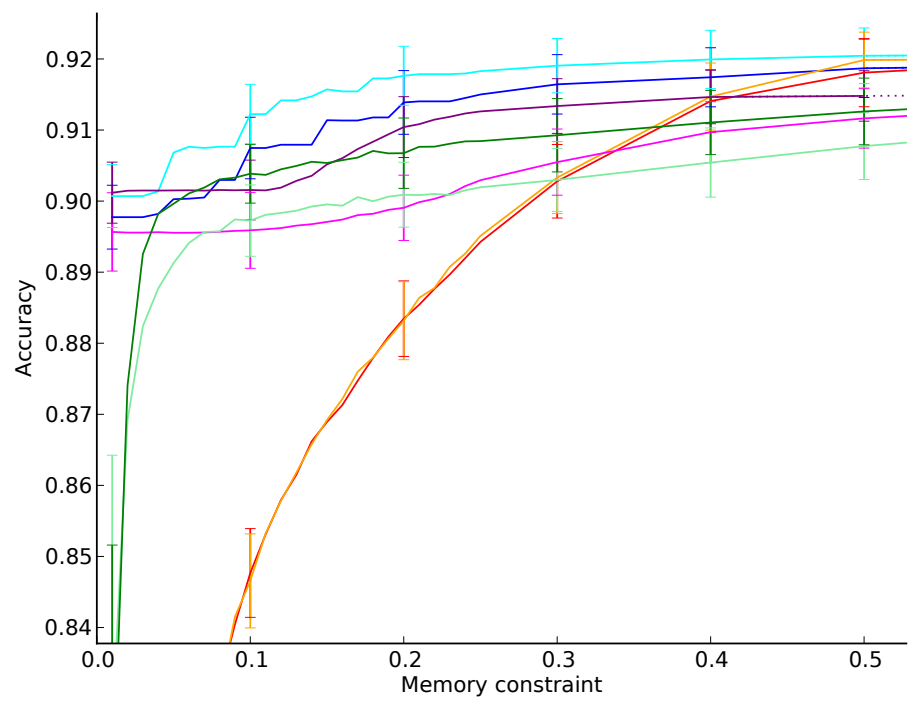

(c) TIS

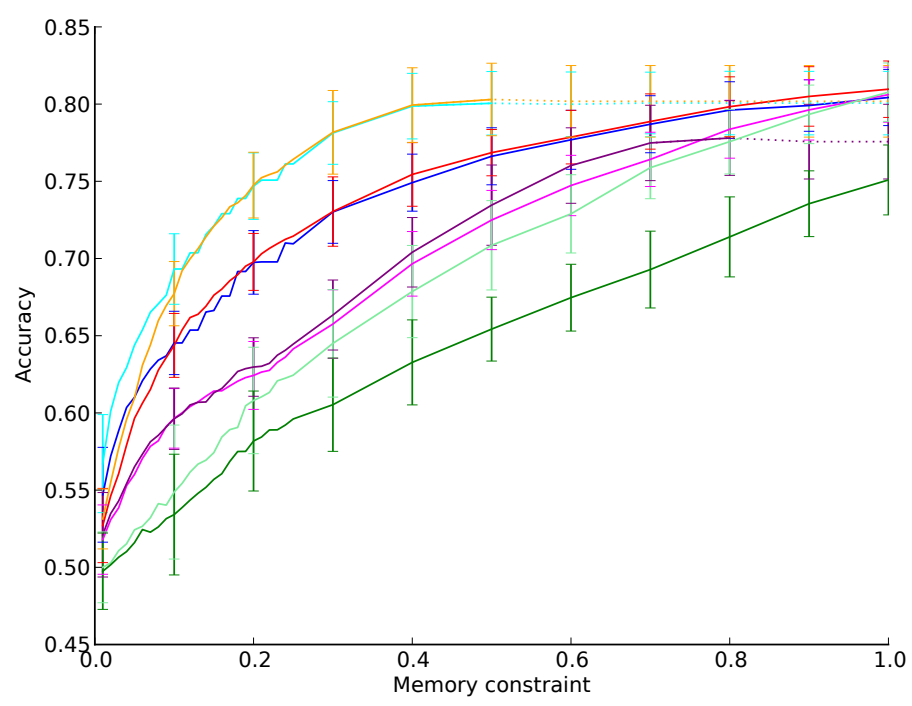

(d) MADELON

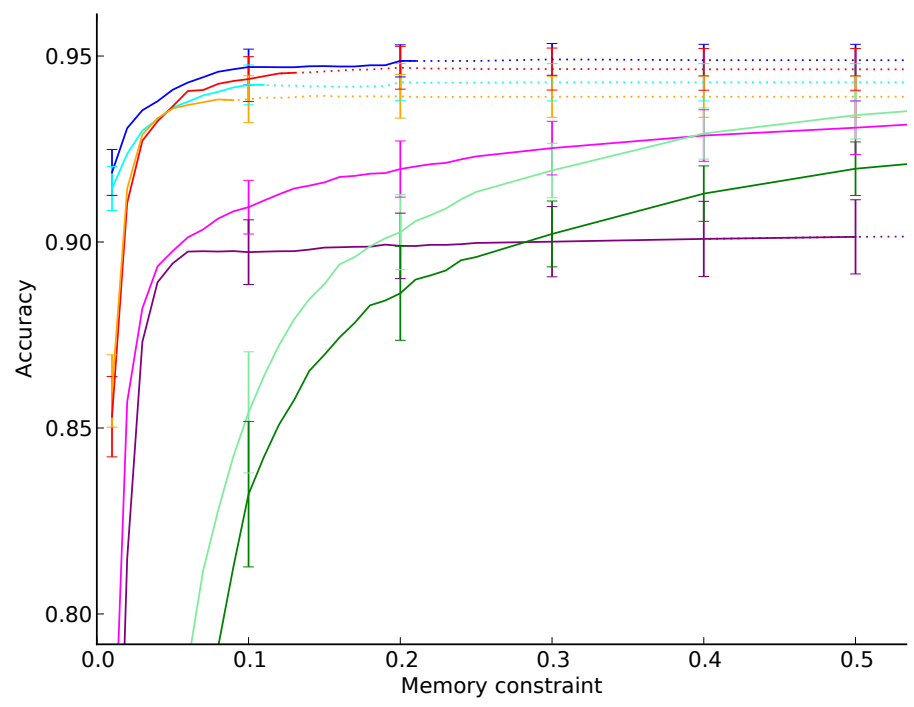

(e) ISOLET

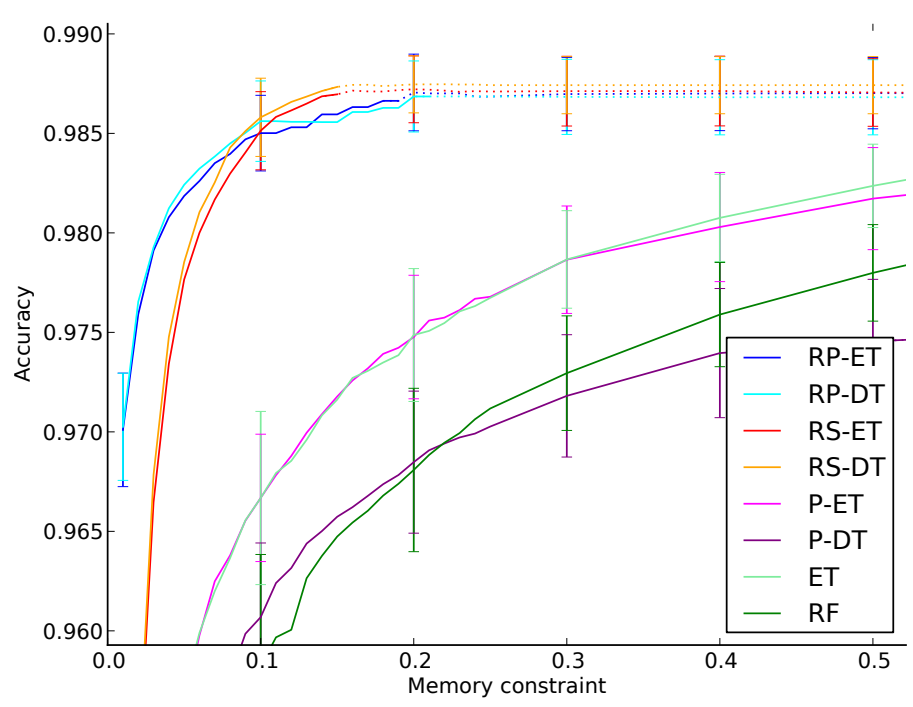

(f) MNIST3VS8

Figure 8.5: Accuracy under memory constraint.

lower $\mu'_{max}$ can be reduced without loss of accuracy. Indeed, we found no conclusive correlation between the size $Np$ of the dataset and the minimum value of $\mu'_{max}$ to reach good performance[2]. Overall, these results thus indicate that building an ensemble on random patches is not only a good strategy when data is abundant and redundant but also that it works even for scarce datasets with limited information regarding the problem.

### 8.3.4 *Conclusions*

We have shown in this section that the memory requirements of sampling based ensembles are intrinsically low. Better, we have shown that they can often be drastically decreased without significant loss in accuracy. When the size of the dataset is far bigger than the available memory, we have also demonstrated that sampling data along both samples and features, as RP does, not only competes with other ensemble algorithms but also significantly improves the accuracy of the resulting ensemble. It also brings a significant improvement over a straightforward sub-sampling of the instances.

## 8.4 OVERALL CONCLUSIONS

The main contribution of this work is to explore a new framework for supervised learning in the context of very strong memory constraints or, equivalently, very large datasets. To address such problems, we proposed the Random Patches ensemble method that builds each individual model of the ensemble from a random patch of the dataset obtained by drawing random subsets of both samples and features from the whole dataset. Through extensive experiments with tree-based estimators, we have shown that this strategy works as well as other popular randomization schemes in terms of accuracy (Section 8.2), at the same time reduces very significantly the memory requirements to build each individual model (Section 8.3.2), and, given its flexibility, attains significantly better accuracy than other methods when memory is severely constrained (Section 8.3.3). Since all models are built independently of each other, the approach is furthermore trivial to parallelize. All in all, we believe that the paradigm of our method highlights a very promising direction of research to address supervised learning on big data.

There remain several open questions and limitations to our approach that we would like to address in the future. First, this study motivates our interest in experimenting with truly large-scale problems (of giga-scale and higher). Since RP already appears advantageous for small to medium datasets, the potential benefits on very large-scale data indeed look very promising.

2 Spearman's rank correlation coefficient is -0.0354 ($p-\text{value} = 0.8550$).

Second, the conclusions drawn in sections 8.2 and 8.3 are all based on the optimal values of the parameters $\alpha_s$ and $\alpha_f$ tuned through an exhaustive grid search on the validation set. Our analysis did not account for the memory and time required for tuning these two parameters. In practice, hyper-parameter tuning can not be avoided as we have shown that the optimal trade-off between $\alpha_f$ and $\alpha_s$ was problem dependent. It would therefore be interesting to design an efficient strategy to automatically find and adjust the values of $\alpha_s$ and $\alpha_f$, taking into account the global memory constraint. Our simplistic framework also only accounts for the memory required to store the training set in memory and not for the total memory required to actually build the ensemble.

We have only explored uniform sampling of patches of fixed size in our experiments. In the context of the Pasting approach, Breiman proposed an iterative instance weighting scheme that proved to be more efficient than uniform sampling [Breiman, 1999]. It would be interesting to extend this approach when sampling both instances and features. Yet, parallelization would not be trivial anymore, although probably still possible in the line of the work in [Chawla et al., 2004].

Finally, our analysis of RP is mostly empirical. In the future, we would like to strengthen these results with a more theoretical analysis. A starting point could be the work in [Zinkevich et al., 2010] that studies a scheme similar to the Pasting method applied to linear models trained through parallel stochastic gradient descent. The extension of this work to non parametric tree-based estimators does not appear trivial however, since these latter are not well characterized theoretically. In retrospect, latest results [Scornet et al., 2014] regarding the consistency of random forests could certainly also be used to motivate the use of random subsets of samples. Similarly, theoretical results from Chapter 6 (Section 6.4.2) also indicate that the use of random subsets of features does not systematically degrade performance.

# 9

# CONCLUSIONS

By and large, machine learning remains an open field of research for which many questions are still left unanswered, even regarding well-established methods. In this dissertation, we have revisited decision trees and random forests, consistently calling into question each and every part of these algorithms, in order to shed new light on their learning capabilities, inner workings and interpretability.

In Part I of this work, we laid out the decision trees and random forests methodology in the context of classification and regression tasks. Our treatment first considered the induction of individual decision trees and put them into a unified and composable framework. In particular, our analysis reviewed assignment rules, stopping criteria and splitting rules, theoretically motivating their design and purpose whenever possible. We then proceeded with a systematic study of randomized ensemble methods within the bias-variance framework. We established that variance depends on the correlation between individual tree predictions, thereby showing why randomization acts as a mechanism for reducing the generalization error of an ensemble. Random forest and its variants were then presented within the framework previously introduced, and their properties and features discussed and reviewed. Our contributions followed with an original time and space complexity analysis of random forests, hence showing their good computational performance and scalability to larger problems. Finally, the first part of this work concluded with an in-depth discussion of implementation details of random forests, highlighting and discussing considerations that are critical, yet easily overlooked, for guaranteeing good computational performance. While not directly apparent within this manuscript, this discussion also underlined our contributions in terms of software, within the open source Sckit-Learn library. As open science and reproducibility concerns are gaining momentum, we indeed believe that good quality software should be an integrative part, acknowledged for its own value and impact, of any modern scientific research activity.

Part II of this dissertation analyzed and discussed the interpretability of random forests in the eyes of variable importance measures. The core of our contributions rests in the theoretical characterization of the Mean Decrease of Impurity variable importance measure, from which we have then proved and derived some of its properties in the case of multiway totally randomized trees and in asymptotic conditions. In particular, we have shown that variable importances offer a three-level decomposition of the information jointly provided by the input variables about the output, accounting for all possible interaction terms in a fair and exhaustive way. More interestingly, we have also shown that variable importances only depend on relevant variables and that the importance of irrelevant variables is strictly equal to zero, thereby making importances a sound and appropriate criterion for assessing the usefulness of variables. In consequence of this work, our analysis then demonstrated that variable importances as computed from non-totally randomized trees (e.g., standard Random Forest or Extremely Randomized Trees) suffer from a combination of defects, due to masking effects, misestimations of node impurity or due to the binary structure of decision trees. Overall, we believe that our analysis should bring helpful insights in a wide range of applications, by shedding new light on variable importances. In particular, we advise to complement their interpretation and analysis with a systematic decomposition of their terms, in order to better understand why variables are (or are not) important.

This preliminary work unveils various directions of future work, both from a theoretical and practical point of view. To our belief, the most interesting theoretical open question would be the characterization of the distribution of variable importances in the finite setting. Such a characterization would indeed allow to more reliably distinguish irrelevant variables (whose importances are positive in the finite case) from relevant variables. Another interesting direction of future work would be to derive a proper characterization of variable importances in the case of binary trees – even if we believe, as pointed out earlier, that variable importances derived from such ensembles may in fact not be as appropriate as desired. From a more practical point of view, this study also calls for a re-analysis of previous empirical studies. We indeed believe that variable importances along with their decomposition should yield new insights in many cases, providing a better understanding of the interactions between the input variables and the output, but also between the input variables themselves. Again, we recommend multiway totally randomized trees to mitigate sources of bias as much as possible.

Finally, Part III addressed limitations of random forests in the context of large datasets. Through extensive experiments, we have shown that subsampling either samples, features or both simultaneously provides on par performance while lowering at the same time the memory requirements. Overall this paradigm highlights an intriguing practical fact: there is often no need to build single models over immensely large datasets. Good performance can often more simply be achieved by building models on small random parts of the data and then combining them all in an ensemble, thereby avoiding all practical and computational burdens of making large data fit into memory. Again, this work raises interesting questions of further work. From a theoretical point of view, one would be to identify the statistical properties in the learning problem that are necessary for guaranteeing subsampling strategies to work. In particular, in which cases is it better to subsample examples rather than features? From a more practical point of view, other directions of research also include the study of smarter sampling strategies or the empirical verification that conclusions extend to non tree-based methods.

Overall, this thesis calls for a permanent re-assessment of machine learning methods and algorithms. It is only through a better understanding of their mechanisms that algorithms will advance in a consistent and reliable way. Always seek for the what and why. In conclusion, machine learning should not be considered as a black-box tool, but as a methodology, with a rational thought process that is entirely dependent on the problem we are trying to solve.

# Part IV

# APPENDIX

# A

# NOTATIONS

| | | |
|---|---|---|
| $\mathcal{A}$ | A supervised learning algorithm | 12 |
| $\mathcal{A}(\theta, \mathcal{L})$ | The model $\varphi_{\mathcal{L}}$ produced by algorithm $\mathcal{A}$ over $\mathcal{L}$ and hyper-parameters $\theta$ | 17 |
| $\alpha_s$ | The proportion of samples in a random patch | 170 |
| $\alpha_f$ | The proportion of features in a random patch | 170 |
| $b_l$ | The $l$-th value of a categorical variable | 39 |
| $B$ | A subset $B \subseteq V$ of variables | 128 |
| $c_k$ | The $k$-th class | 11 |
| $C_p^k$ | The number of $k$-combinations from a set of $p$ elements | 129 |
| $C(N)$ | The time complexity for splitting $N$ samples | 88 |
| $\mathbb{E}$ | Expectation | |
| $\overline{E}(\varphi_{\mathcal{L}}, \mathcal{L}')$ | The average prediction error of $\varphi_{\mathcal{L}}$ over $\mathcal{L}'$ | 13 |
| $Err(\varphi_{\mathcal{L}})$ | The generalization error of $\varphi_{\mathcal{L}}$ | 12, 55 |
| $H(X)$ | The Shannon entropy of $X$ | 126 |
| $H(X\|Y)$ | The Shannon entropy of $X$ conditional to $Y$ | 126 |
| $\mathcal{H}$ | The space of candidate models | 16 |
| $i(t)$ | The impurity of node $t$ | 30, 40 |
| $i_R(t)$ | The impurity of node $t$ based on the local resubstitution estimate | 41, 46 |
| $i_H(t)$ | The entropy impurity of node $t$ | 45 |
| $i_G(t)$ | The Gini impurity of node $t$ | 45 |
| $\Delta i(s, t)$ | The impurity decrease of the split $s$ at node $t$ | 30 |
| $I(X; Y)$ | The mutual information between $X$ and $Y$ | 126 |
| $Imp(X_j)$ | The variable importance of $X_j$ | 124, 125 |
| $J$ | The number of classes | 11 |
| $K$ | The number of folds in cross-validation | 14 |
| | The number of input variables drawn at each node for finding a split | 71 |
| $K(\mathbf{x}_i, \mathbf{x}_j)$ | The kernel of $\mathbf{x}_i$ and $\mathbf{x}_j$ | 22, 79 |
| $L$ | A loss function | 12 |
| | The number of values of a categorical variable | 39 |
| $\mathcal{L}$ | A learning set $(\mathbf{X}, \mathbf{y})$ | 10 |
| $\mathcal{L}^m$ | The $m$-th bootstrap replicate of $\mathcal{L}$ | 70 |
| $\mathcal{L}_t$ | The subset of node samples falling into node $t$ | 30 |
| $M$ | The number of base models in an ensemble | 62 |
| $\mu_{\mathcal{L},\theta_m}(\mathbf{x})$ | The mean prediction at $X = \mathbf{x}$ of $\varphi_{\mathcal{L},\theta_m}$ | 64 |
| $N$ | The number of input samples | 10 |
| $N_t$ | The number of node samples in node $t$ | 30 |
| $N_{ct}$ | The number of node samples of class $c$ in node $t$ | 32 |
| $\Omega$ | The universe, or population, from which cases are sampled | 9 |
| $p$ | The number of input variables | 9 |
| $p_L$ | The proportion of node samples going to $t_L$ | 30 |
| $p_R$ | The proportion of node samples going to $t_R$ | 30 |
| $p(t)$ | The estimated probability $p(X \in \mathcal{X}_t) = \frac{N_t}{N}$ | 33 |
| $p(c\|t)$ | The empirical probability estimate $p(Y = c\|X \in \mathcal{X}_t) = \frac{N_{ct}}{N_t}$ of class $c$ at node $t$ | 33 |
| $\widehat{p}_{\mathcal{L}}$ | An empirical probability estimate computed from the learning set $\mathcal{L}$ | 59 |
| $P(X, Y)$ | The joint probability distribution of the input variables $X = (X_1, \dots, X_p)$ and the output variable $Y$ | 11 |
| $\mathcal{P}_k(V)$ | The set of subsets of $V$ of size $k$ | 128 |
| $\varphi$ | A model or function $\mathcal{X} \mapsto \mathcal{Y}$ | 10 |
| | A single decision tree | 27 |
| $\widetilde{\varphi}$ | The set of terminal nodes in $\varphi$ | 32 |
| $\varphi(\mathbf{x})$ | The prediction of $\varphi$ for the sample $\mathbf{x}$ | 10 |
| $\varphi_{\mathcal{L}}$ | A model built from $\mathcal{L}$ | 12 |
| $\varphi_{\mathcal{L},\theta}$ | A model built from $\mathcal{L}$ with random seed $\theta$ | 61 |
| $\varphi_B$ | A Bayes model | 15 |
| $\psi_{\mathcal{L},\theta_1,\dots,\theta_M}$ | An ensemble of $M$ models built from $\mathcal{L}$ and random seeds $\theta_1, \dots, \theta_M$ | 62 |
| $\mathcal{Q}$ | A set $\mathcal{Q} \subseteq \mathcal{S}$ of splits of restricted structure | 27, 38 |
| $\mathcal{Q}(X_j)$ | The set $\mathcal{Q}(X_j) \subseteq \mathcal{Q}$ of univariate binary splits that can be defined on variable $X_j$ | 38, 39 |
| $\rho(\mathbf{x})$ | The correlation coefficient between the predictions at $X = \mathbf{x}$ of two randomized models | 65 |
| $s$ | A split | 27, 37 |
| $s^*$ | The best split | 30, 39 |
| $s_j^*$ | The best binary split defined on variable $X_j$ | 30, 39 |
| $s_j^v$ | The binary split $(\{\mathbf{x}\|x_j \leqslant v\}, \{\mathbf{x} > v\})$ defined on variable $X_j$ with discretization threshold $v$ | 38 |
| $s_t$ | The split labeling node $t$ | 27 |
| $\tilde{s}_t^j$ | The best surrogate split for $s_t$ defined from $X_j$ | 124 |

| | |
|---|---|
| $\mathcal{S}$ | The set of all possible splits s . . . 38 |
| $\sigma^2_{\mathcal{L},\theta_m}(\mathbf{x})$ | The prediction variance at $X = \mathbf{x}$ of $\varphi_{\mathcal{L},\theta_m}$ . . . 64 |
| $t$ | A node in a decision tree . . . 27 |
| $t_L$ | The left child of node t . . . 30, 37 |
| $t_R$ | The right child of node t . . . 30, 37 |
| $\theta$ | A vector of hyper-parameter values . . . 17<br>A random seed . . . 62 |
| $\theta^*$ | The optimal hyper-parameters . . . 17 |
| $\widehat{\theta}^*$ | The approximately optimal hyper-parameters . . . 17 |
| $\theta_m$ | The seed of the m-th model in an ensemble . . . 62 |
| $v$ | A discretization threshold in a binary split . . . 38 |
| $v_k$ | The k-th value of an ordered variable, when node samples are in sorted order . . . 48 |
| $v'_k$ | The mid-cut point between $v_k$ and $v_{k+1}$ . . . 48 |
| $V$ | The set $\{X_1, \dots, X_p\}$ of input variables . . . 127 |
| $V^{-j}$ | $V \setminus \{X_j\}$ . . . 128 |
| $\mathbb{V}$ | Variance . . . |
| $\mathbf{x}$ | A case, sample or input vector $(x_1, \dots, x_p)$ . . . 9 |
| $\mathbf{x}_i$ | The i-th input sample in $\mathcal{L}$ . . . 10 |
| $x_j$ | The value of variable $X_j$ for the sample $\mathbf{x}$ . . . 9 |
| $\mathbf{X}$ | The $N \times p$ matrix representing the values of all N samples for all p input variables . . . 10 |
| $X_j$ | The j-th input variable or feature . . . 9, 10 |
| $X$ | The random vector $(X_1, \dots, X_p)$ . . . 11 |
| $\mathcal{X}_j$ | The domain or space of variable $X_j$ . . . 9 |
| $\mathcal{X}$ | The input space $\mathcal{X}_1 \times \dots \times \mathcal{X}_p$ . . . 9 |
| $\mathcal{X}_t$ | The subspace $\mathcal{X}_t \subseteq \mathcal{X}$ represented by node t . . . 27 |
| $y$ | A value of the output variable Y . . . 9 |
| $\widehat{y}_t$ | The value labelling node t . . . 28 |
| $\widehat{y}^*_t$ | The optimal value labelling node t . . . 32 |
| $\mathbf{y}$ | The output values $(y_1, \dots, y_N)$ . . . 10 |
| $Y$ | The output or response variable Y . . . 9 |
| $\mathcal{Y}$ | The domain or space of variable Y . . . 9 |

# B REFERENCES